\documentclass[11pt, oneside]{Thesis} % The default font size and one-sided printing (no margin offsets)

\usepackage[english]{babel}

\usepackage{subfigure}
\usepackage{verbatim}
\usepackage{multirow}
\usepackage{graphicx, epstopdf}
\usepackage{xspace,epsfig,url}
\usepackage{amssymb,amsmath}
\usepackage[noadjust]{cite}
\usepackage{enumerate}
\usepackage{float}
\usepackage{amsmath}
\usepackage{epsf}
\usepackage{psfrag}
\usepackage{verbatim}
\usepackage[all]{xy}
\usepackage{color}
\usepackage{comment}
\usepackage{subfigure}
\usepackage{cases}
\usepackage{algorithm}
\usepackage[noend]{algpseudocode}
\usepackage{array}
\usepackage{multicol}
\usepackage[table]{xcolor}
\usepackage{rotating}

\newcolumntype{P}[1]{>{\centering\arraybackslash}p{#1}}

\newcommand{\vect}[1]{\mathbf{#1}}
\newcommand{\vects}[1]{\boldsymbol{#1}}
\newcolumntype{?}{!{\vrule width 2.5pt}}

\graphicspath{{./Figures/}}

\title{\ttitle} % Defines the thesis title - don't touch this

\begin{document}

\frontmatter % Use roman page numbering style (i, ii, iii, iv...) for the pre-content pages

\setstretch{1.3} % Line spacing of 1.3

% Define the page headers using the FancyHdr package and set up for one-sided printing
\fancyhead{} % Clears all page headers and footers
\rhead{\thepage} % Sets the right side header to show the page number
\lhead{} % Clears the left side page header

\pagestyle{fancy} % Finally, use the "fancy" page style to implement the FancyHdr headers

\newcommand{\HRule}{\rule{\linewidth}{0.5mm}} % New command to make the lines in the title page

% PDF meta-data
\hypersetup{pdftitle={\ttitle}}
\hypersetup{pdfsubject=\subjectname}
\hypersetup{pdfauthor=\authornames}
\hypersetup{pdfkeywords=\keywordnames}

%----------------------------------------------------------------------------------------
%	TITLE PAGE,
%----------------------------------------------------------------------------------------

\begin{titlepage}
\begin{center}

\textsc{\LARGE Scuola Superiore Sant'Anna}\\[1.5cm] % University name
\textsc{\Large Doctoral Thesis}\\[0.5cm] % Thesis type

\HRule \\[0.4cm] % Horizontal line
{\huge \bfseries Design, Implementation and Control of an Underactuated Hand Exoskeleton}\\[0.4cm] % Thesis title
\HRule \\[1.5cm] % Horizontal line

\begin{minipage}{0.4\textwidth}
\begin{flushleft} \large
\emph{Author:} \\Mine SARAÇ STROPPA
\end{flushleft}
\end{minipage}
\begin{minipage}{0.4\textwidth}
\begin{flushright} \large
\emph{Supervisor:} \\ Antonio FRISOLI,  Professor
\end{flushright}
\end{minipage}\\[3cm]

\large \textit{A thesis submitted in fulfillment of the requirements\\ for the degree of Doctor of Philosophy}\\[0.3cm] % University requirement text
\textit{in the}\\[0.4cm]
Perceptual Robotics Laboratory\\ [0.4cm]
Institute of Communication, Information and Perception Technologies \\ [0.4cm]
Scuola Superiore Sant'Anna\\ [0.4cm]
Pisa - Italy\\ [0.4cm]

{\large April, 2017}\\[4cm]
%{\large \today}\\[4cm] % Date
%\includegraphics{Logo} % University/department logo - uncomment to place it

\vfill
\end{center}

\end{titlepage}

%\clearpage % Start a new page

%----------------------------------------------------------------------------------------
%	QUOTATION PAGE
%----------------------------------------------------------------------------------------

%\pagestyle{empty} % No headers or footers for the following pages
%
%\null\vfill % Add some space to move the quote down the page a bit
%
%\clearpage % Start a new page

%----------------------------------------------------------------------------------------
%	ABSTRACT PAGE
%----------------------------------------------------------------------------------------

\begin{center}
\huge
\begin{singlespace}
Design, Implementation and Control of an Underactuated Hand Exoskeleton\end{singlespace}
\end{center}
\vspace{-9mm}

\begin{singlespace}
\begin{center} \large
Mine Saraç Stroppa

Perceptual Robotics, Doctor of Philosophy (PhD), 2017. \\ [0.4cm]

Thesis Supervisor: Assoc. Prof. Dr. Antonio Frisoli
\normalsize
\end{center}
\end{singlespace}
\vspace{-3mm}

%\begin{center}
\begin{singlespace}
\end{singlespace}
%\end{center}
\vspace{-3mm}

\begin{center}
\textbf{Abstract} \vspace{-5mm}
\end{center}
%\small
%\begin{singlespace}

%{\huge \bfseries Abstract}\\[0.4cm] % Thesis title

I present the design, implementation and control of a novel, linkage-based, underactuated hand exoskeleton aimed to assist patients with hand disabilities during grasping tasks for robot assisted physical rehabilitation or robot assisted activities of daily living. Even though the proposed exoskeleton was designed with the purpose of assisting users during interactions with real objects around them, its use can be extended for haptic applications as well.

The design requirements of a generic hand exoskeleton, which can be used for all types of applications, were listed after an intensive literature search. These requirements lead us to have a novel kinematics selection. In particular, a generic hand exoskeleton should be portable, lightweight and easily wearable for allowing patients with hand disabilities to use the device. The hand exoskeleton should assist all fingers of the user independently. Using linkage-based kinematics with intentional misalignment between mechanical and anatomical finger joints allows the device to adopt its operation for different hand sizes automatically. The device assists only $2$ finger joints of each finger for flexion/extension using a single actuator. Such an assistance can be achieved by adopting underactuation concept, which can adjust force transmission for finger joints based on physical interaction forces. Doing so, the exoskeleton can assist users grasping objects with different sizes and shapes automatically, with no prior information. Finally, the connection mechanics of the device is designed to exert only perpendicular forces to the finger phalanges to increase the realism of natural interaction forces between the user and objects. Overall, the easiness of the attachment to user's fingers, better comfort and improved security are guaranteed.

I performed pose analysis, differential kinematics analysis, statics analysis and stability of grasp analysis for the proposed kinematics. The lengths of mechanical links for each finger component are optimized to increase range of motion for finger joints and efficacy of force transmission. An additional potentiometer attached on the system allows the finger pose to be predicted during operation. For the electronic components, a DSP board is selected to run all sensory measurements, control algorithms and motor driver connections.

The proposed underactuated hand exoskeleton can be controlled in various ways. First of all, a strict position control is performed based on a simple position controller. The performance of the PI control is enhanced by temperature filters and backdrive force support. The position controller can be used to perform grasping tasks for assistive or rehabilitation applications, thanks to the automatic adjustment of operation. The same position control can be used with a EMG based trajectory, instead of pre-defined passive trajectory. In particular, patients with disabilities are asked to move their unimpaired hand, on which muscular activity can be measured with EMG sensors, and the exoskeleton can assist their impaired hand to perform the same movement. Involving users in the execution of the tasks to define the trajectory turns the task into an active exercise. The EMG activities can move user's impaired hand in a coupled or independent manner. Finally, a force control algorithm is proposed by equipping an additional force sensor for each finger component to let the user open/close his fingers by applying external forces to it. Thanks to the active backdriveability, the exoskeleton can detect user's intentions and follow his intentions or amplify the movements for assistance.

The force control algorithm was extended for stiffness rendering algorithm to provide force feedback to user based on virtual interactions during a haptic task. Since the underactuation property suffers from the lack of controllability of joints, additional rendering strategies are proposed, which can be generalized for any underactuated device in the literature. Feasibility studies show the efficacy of the proposed strategies to determine the transmitted forces along finger joints, while ensuring the safety of these strategies compared to conventional rendering algorithms.

\clearpage % Start a new page

%----------------------------------------------------------------------------------------
%	ACKNOWLEDGEMENTS
%----------------------------------------------------------------------------------------
\pagestyle{fancy}

\setstretch{1.3} % Reset the line-spacing to 1.3 for body text (if it has changed)
\lhead{\emph{Acknowledgement}}
%\acknowledgements {\addtocontents{toc}}%{\vspace{1em}}

%\vspace*{-10mm}

\begin{acknowledgements}

It is a great pleasure to extend my gratitude to my thesis advisor Assoc. Prof. Dr. Antonio Frisoli and Assoc. Prof. Dr. Miguel Angel Otaduy with whom I worked with during my period abroad for their guidance and support. I would also like to thank Dr. Massimilano Solazzi for his advices and support throughout my doctoral studies, and Assist. Prof. Dr. Eray Baran and Assoc. Prof. Dr. Rocco Vertechy for their time to review my work and help me to improve it.

I am grateful for having the decision of moving to Italy and before anyone else, I would like to thank my dearest husband Fabio Stroppa, who gave me a new family. Thank you for all the coffee you made for me, all your love and your understanding. Without your support, I could never have completed the obligatory credits and my doctoral degree.

I would also like to extend my appreciation to my parents, my dearest aunt, and my sweetest cousins for always supporting and loving me despite of all the distance between us. I would also like to thank my new Italian family for welcoming me with open arms.

It was a pleasure to work with Massimiliano Solazzi and Daniele Leonardis and I am very thankful to them for their help and advices during my studies. I am also grateful to Daniel Lobo for all his help during my survival in Spain and his collaboration with Mickeal Verschoor. Many thanks to my group friends in PERCRO who I spent all my working days with for making the laboratory enjoyable. I would also like to thank all the members of Multimodal Simulation Lab in Universidad Rey Juan Carlos in Madrid, Spain for their friendship and brain storming lunch talks. Furthermore, I would like to thank Bengi Deniz Kırındı Demirkıran and many other friends from Turkey, who gave me all their support online in sadness and happiness.

I would like to express my gratitude for Elisa Zanobini, who work in the International Office of Scuola Superiore Sant'Anna, for all her time and effort during any bureaucratic chaos I was passing through. %She was always there with all her help and kindness every time I needed help.

This research was funded within the project ''WEARHAP – WEARable HAPtics for humans and robots" of the European Union Seventh Framework Programme FP7/2007-2013, grant agreement n. 601165 and project ''CENTAURO - Robust Mobility and Dexterous Manipulation in Disaster Response by Fullbody Telepresence in a Centaur-like Robot"  of the   the European Union's Horizon 2020 Programme, Grant Agreement  n. 644839.

\end{acknowledgements}

\newpage

%\acknowledgements{\addtocontents{toc}{\vspace{1em}} % Add a gap in the Contents, for aesthetics

%The acknowledgements and the people to thank go here, don't forget to include your project advisor\ldots}
%\clearpage % Start a new page

%----------------------------------------------------------------------------------------
%	LIST OF CONTENTS/FIGURES/TABLES PAGES
%----------------------------------------------------------------------------------------

\pagestyle{fancy} % The page style headers have been "empty" all this time, now use the "fancy" headers as defined before to bring them back

\lhead{\emph{Contents}} % Set the left side page header to "Contents"
\tableofcontents % Write out the Table of Contents

\lhead{\emph{List of Figures}} % Set the left side page header to "List of Figures"
\listoffigures % Write out the List of Figures

\lhead{\emph{List of Tables}} % Set the left side page header to "List of Tables"
\listoftables % Write out the List of Tables

%----------------------------------------------------------------------------------------
%	ABBREVIATIONS
%----------------------------------------------------------------------------------------

%\clearpage % Start a new page
%\setstretch{1.5} % Set the line spacing to 1.5, this makes the following tables easier to read
%\lhead{\emph{Abbreviations}} % Set the left side page header to "Abbreviations"
%\listofsymbols{ll} % Include a list of Abbreviations (a table of two columns)
%{
%\textbf{LAH} & \textbf{L}ist \textbf{A}bbreviations \textbf{H}ere \\
%\textbf{Acronym} & \textbf{W}hat (it) \textbf{S}tands \textbf{F}or \\
%}

%----------------------------------------------------------------------------------------
%	THESIS CONTENT - CHAPTERS
%----------------------------------------------------------------------------------------

\mainmatter % Begin numeric (1,2,3...) page numbering

\pagestyle{fancy} % Return the page headers back to the "fancy" style

% Include the chapters of the thesis as separate files from the Chapters folder
% Uncomment the lines as you write the chapters

\chapter{Introduction} % Main chapter title

\label{sec:Chapter1} % For referencing the chapter elsewhere, use \ref{Chapter1}

\lhead{Chapter 1. \emph{Introduction}} % This is for the header on each page - perhaps a shortened title

As humans, we can sense the physical environment around us, discriminate the properties of objects we encounter with or perform tasks with high dexterity through our hands. Therefore, the functionality of our hands is highly crucial for our daily lives. Unfortunately, hand injuries are very common, and the complexity of such an important organ makes it harder to be treated in the event of these injuries. This study mainly focuses on developing a portable, efficient, and wearable hand exoskeleton. Besides our main focus, we want to create an exoskeleton that can be used for assistive applications, where the exoskeleton assists patients to complete various tasks during their daily lives, or for haptic applications, where the exoskeleton renders kinesthetic feedback based on teleoperation tasks or user's activity in the virtual environment. 

This chapter focuses on the properties and the requirements of physical rehabilitation. The importance and the positive impact of utilizing robotic devices to assist rehabilitation exercises are stated in general and specifically for the hand. Finally, the objectives and contributions of this study, which presents the design and implementation of a novel hand exoskeleton, are listed briefly.

%% HAND INJURIES

\newpage
\section{Physical Rehabilitation}

The nervous system can be explained as an incredibly complex communication system, that controls and regulates the functions of body. This system is composed of the brain, the spinal cord and an intricate network of nerves. Unfortunately, this extraordinary system is highly vulnerable to diseases or injury~\cite{NI}. Neurological injuries are the leading cause of serious, long-term disabilities that restrict the daily functions of millions of patients. Stroke is one of the most serious neurological injuries and approximately 9 million people have had a stroke in 2008.

30 million people have previously had a stroke and are still alive~\cite{GBD}. $75\%$ of stroke survivors suffer from the disabilities that affect their daily living routines physically, mentally, and emotionally~\cite{Stroke}. When the stroke affects the upper limbs, having hand injuries is inevitable. After these injuries, stroke patients not only experience discomfort and pain, but also get affected significantly during the daily living activities in physical, psychological, social, and financial aspects of their lives~\cite{Daud2016}. Survivors mostly experience pain and stiffness for up to four years after their injury.

Especially during the initial stage of the injury, patients with hand disabilities cannot perform many activities of daily living (ADLs) without any help or assistance from other people. Not being able to perform even the simplest task during their daily lives might cause them to go through a psychological burden, such as frustration, discouragement and loss of confidence. In the meantime, they might face with stress, the loss of hope, and the fear of getting injured again. All of these emotions might lead to avoidance of many activities. In addition, sustaining hand injuries might have a negative impact on social and financial security, as the individuals will be absent from their works or may lose their employment following the injury. %Physical rehabilitation helps these survivors to regain their hand functions, and to perform as many ADLs as possible with less restrictions and external assistance. 

Physical rehabilitation is an indispensable solution to treat patients, who are dealing with disabilities of neurological injuries, and to help them regain their functional abilities. There are two main approaches of physical therapy to treat hand injuries. The first one focuses on increasing the effective Range of Motion (RoM) for the impaired finger joints. These therapy exercises consist of repetitive implicit joint rotations, without considering any possible reflections in the real life. The second approach simulates tasks and scenarios from real life, and provides a platform for survivors to practice these given scenarios. By engaging with daily activities repeatedly in a clinical setting, these survivors might regain the functionality of their hand. %With both approaches, 

Recovering from the injuries in many instances requires a long treatment period and often results in a variable but persistent disability. Such a recovery is challenging even if the treatment procedure is initiated in an early stage of disability. The treatment of hand injuries might take even longer time compared to other regions of the body. The rehabilitation therapies were found to be more effective when they are repetitive \cite{Butefisch1995}, intense \cite{Kwakkel1999}, long term \cite{Sunderland1992}, and task specific \cite{Bayona2005}.

Conventionally, the physical rehabilitation sessions are designed and performed by a therapist~\cite{Levanon2013}. During the sessions, the therapist decides the needs and physical limitations of the patient, and designs a unique path of rehabilitation based on the physical examination. For every session, the patient meets the therapist in the clinic, as the therapist physically leads the impaired limbs of the patient. The motivation of these sessions is mostly to increase the motion, dexterity and/or strength of the impair hand. To achieve these goals, the rehabilitation exercises to be performed should be motivations, repetitious, interesting, challenging and graded.

%% ROBOT-ASSISTED REHABILITATION

The technological developments created robot-assisted rehabilitation devices that can be used during therapy exercises instead of conventional, manual assistance/guidance techniques. Such robotic rehabilitation devices have been shown to be improving the functional independence of patients when integrated to the clinical therapy exercises~\cite{Kwakkel2008,Mehrholz2009,Prange2006,Nykanen2010}. During the robot-assisted rehabilitation therapies, the patient still needs the therapist to design the therapy pathway, but they can use these devices instead of manually guiding all the tasks for the patients. As a result, they eliminate the physical burden of repetitive physical therapy for the therapists. Furthermore, a clinic can operate with less number of therapists but more health staff, and still treat more patients at a given time. Such a practicality might actually decrease the overall treatment cost, despite the initial cost of purchasing these devices. Robot assisted rehabilitation also increases the reliability and accuracy of desired tasks, ensuring that the same task can be repeated by patients. Providing quantitative measurements might allow the therapist to track patient's progress over time. Finally, various control algorithms can be implemented on the same robotic device to address different therapy scenarios that might improve the treatment for patients with various levels of impairments, while motivating them to endure long, intense therapy sessions.

%As the robotic devices become a part of treatment, therapists must adapt to these technological changes and need to use relevant and familiar tools with the patients in the future. 
The changes in the technology and current trends in the rehabilitation create the need of developing better treatment scenarios. Portable, low-cost and user friendly hand devices might even carry the rehabilitation exercises from hospitals to homes. %However, the use of technology has social and therapeutic justifications. The absence of human contact, high costs, and sensitivity of a system that needs technical support must be taken into consideration before buying and using these systems. It is important to note that in order to promote the use of technology in rehabilitation, failures in technology must be avoided, therapeutic principles within games should be maintained, and feedback that is easily and readily associated with success must be provided. In summary, the system should be user-friendly, realistic yet challenging.

\newpage
\section{Hand Exoskeletons}

A hand exoskeleton is a wearable, haptic device that provides realistic kinesthetic feedback to user's fingers through active force transmission over a series of mechanical components. Such a device applies forces to the fingers in order to move and to assist them completing a pre-defined task to imitate ADLs in a natural manner. The interaction with users, especially with patients suffering from hand disabilities, the stability of the device, ergonomic and comfortable mechanical design, and efficacy of force transmission become highly crucial. A hand exoskeleton can be used for physical rehabilitation as much as haptic applications based on their mechanical properties.

The human hand has a very complex kinematics. Many researchers tried to analyze and imitate the natural behavior of a human hand in robotic devices to be used for industrial grippers~\cite{Qiao2015, Hashimoto2015, Cheung2015}, prosthesis~\cite{Wolf2014, Matrone2011, Bennett2015} or humanoid robotics~\cite{Mnyusiwalla2016, Fukaya2013, Mouri2015, Kamogawa2014}. Despite of these efforts, we are still nowhere near getting close to copying the whole behavior, so reaching realistic devices is still an up to date challenge. Creating hand exoskeletons that can allow the natural and realistic behavior of a human hand is even more challenging due to the safety measures and ergonomic and easy wearability of the mechanism that should be considered for any mechanical system with human interaction.

A hand exoskeleton should be designed to satisfy generic requirements, such as user's safety, ergonomy and compliance with the natural behavior of the user, and specific requirements, which might depend on the target application, such as rehabilitation, assistance, teletoperation or haptics. Most of the hand exoskeletons existing in the literature are designed for a specific purpose, therefore with limited requirements. However, the diversity of the design requirements of each application type and the complexity of the hand model forces the hand exoskeletons in the literature to be defined mostly for one type of application in an efficient manner. Therefore, the literature has an increased interest of generic hand exoskeletons that can be used for all types of applications in an easy and efficient manner.

%Rehabilitative hand exoskeletons~\cite{Wang2009, Chiri2012, Cempini2013, Agarwal2013, Yamaura2009, Weiss2013, Wei2013, Tang2013, Shields1997} %Tong2015, } aim to help patients with hand disabilities performing repetitive finger or hand exercises in order to recover their functional disabilities. Assistive hand exoskeletons~\cite{Burton2011, Tadano2010, Iqbal2011} assist patients with disabilities to perform difficult or impossible daily activities by helping them to survive with minimum dependency on other people. Assistive and rehabilitative devices should be lightweight, easy to be worn, efficient to transmit forces and comfortable. In addition to these requirements, assistive devices should be portable and designed in a specific way to perform real activities of everyday living with no prior definition or constraint set on the performing trajectory. Haptic devices aim to track user's movements~\cite{Xing2014, Fahn2005, Rahmat2009, Heurner2007, Wang2015} and/or provide force feedback to user's hand based on his activities in the virtual environment~\cite{Taheri2014, Stergiopoulos, Gosselin2005, Fontana2009, Brenosa2011, Bouzit2002}. Haptic devices can collect user's movements to control an avatar in a virtual environment or another robotic device through teleoperation. Furthermore, they might give feedback to user in order to let him perceive the state of virtual or teleoperation task. Therefore, these devices should be easily manipulated by user and effectively stimulate forces acting on fingers while rendering daily tasks.

\newpage
\section{Objectives}
\label{ch1:objectives}

Our motivation is to develop a hand exoskeleton to be used mainly for rehabilitation purposes, but also for assistive and haptic tasks. Such an exoskeleton must satisfy a few conditions for each application. In the first one, the hand exoskeleton should be assisting patient's fingers along desired tasks repetitively. It is important to create an easily wearable device with no pose requirement since different disability levels of patients might create difficulties and pain during the preparation time. A generic device that can be useful for different disability definitions and levels should be providing passive tasks, where the device leads the user's fingers to a predefined trajectory, or active tasks, where the device measures the user's intentions through force, pressure or muscular activity sensors and allow them to contribute to the tasks based on their performances and abilities. Meanwhile, the exoskeletons provide realistic feedback to patients during serious game scenarios.

%This work presents the development of a hand exoskeleton that can be used all types of applications, even though is focused for two contexts. The first application of the proposed hand exoskeleton is providing rehabilitation exercises. With this motivation, the hand exoskeleton should be assisting patient's fingers along desired tasks repetitively. It is important to create an easily wearable device with no pose requirement since different disability levels of patients might create difficulties and pain during the preparation time. A generic device that can be useful for different disability definitions and levels should be providing passive tasks, where the device leads the user's fingers to a predefined trajectory, or active tasks, where the device measures the user's intentions through force, pressure or muscular activity sensors and allow them to contribute to the tasks based on their performances and abilities. Meanwhile, the exoskeletons provide realistic feedback to patients during serious game scenarios.

For the second application, the hand exoskeleton should be used for is to assist patients with hand disabilities during ADLs. One of the most common daily activities is to grasp different objects around the environment. Grasping real objects under the assistance of the exoskeleton is possible only when the internal part of the fingers is free. With the same motivation, the fingertip should be left free in case the user needs to press objects. The device should be portable in order to allow the user to explore the environment with minimum attachment to a stationary place. Such a hand exoskeleton should assist the user to grasp various objects with any shape and any size in an automatic manner.

%Even though it is not the first design focus, the proposed hand exoskeleton can be used for haptic applications, 

For the final application, user's finger movements should be estimated and represented in a virtual environment. Furthermore, the exoskeleton can give kinesthetic feedback to the user based on his interactions in the virtual environment with other objects. Even though the proposed exoskeleton suffers from the lack of controllability, which was chosen for the sake of simplicity and task adjustability during grasping tasks for objects with different sizes and shapes, during virtual tasks, utilizing further control strategies can provide sufficient performance regarding the perception of virtual grasping. 

Besides these specific properties, the proposed hand exoskeleton should have a light-weight mechanical design to minimize the fatigue during all types of operations and an efficient force transmission in a natural manner.

\newpage
\section{Contributions}
\label{ch1:contributions}

The contributions of this study to the literature can be summarized as follows: 

%In this study, the development of the hand exoskeleton is presented to achieve the objectives above. The contributions of this work can be summarized as follows.

\begin{itemize}

\item A novel kinematics design was developed in the form of a single finger component of the hand exoskeleton, for which the international patent application is pending~\cite{sarac_patent}. The properties of this kinematics can be listed as :

\begin{itemize}
\item achieving $2~DoFs$ mobility to rotate the finger joints thanks to the linkage based kinematic selection,
    \item performing independent finger control by actuating each finger component by a single motor,
    \item automatically adapting the operation for different hand sizes by completing the kinematics loop only when worn by the user,
    \item automatically adapting the operation for different object sizes and shapes thanks to the underactuation concept based on the contact forces, and
    \item constraining the applied forces to be always perpendicular to the finger phalanges.
    \end{itemize}

\item The initial feasibility study~\cite{Sarac2016} presented various performance analyses and link length optimization. In this work,

    \begin{itemize}
    \item The inverse kinematics was analyzed for the given mechanism design,
    \item The differential kinematics was derived with the assumption of an additional sensory tool, and
    \item The link length optimization was performed to find a set of link lengths with the best force transmission performance satisfying the following constraints:

            \begin{itemize}
            \item the required actuator displacement should be less than the stroke of the chosen actuator,
            \item the displacement of the passive sliders should be less than the length of finger phalanges,
            \item the ratio between the joint torques should be in a reasonable range for the safety and comfort of the user, and
            \item all possible combinations of finger pose in the natural range of motion should be reached.
            \end{itemize}

    \item The interaction forces between the user and the object while grasping various objects were presented to show the adaptability of the device for various grasping tasks.
    \end{itemize}

\item The undersensing property of the device was overcome to reach the pose estimation in an online manner as the user moves attached to the exoskeleton~\cite{Sarac2017}. The content of this work can be summarized as:

    \begin{itemize}
    \item utilizing an additional potentiometer to measure the rotation of one of the of passive joints,
    \item reaching a unique solution for the forward kinematics numerically thanks to the additional measurements,
    \item presenting the pose estimation results for various tasks, and
    \item creating a calibration process to estimate the length of the first finger phalange at a certain pose.
    \end{itemize}

\item The proposed hand exoskeleton was also used to perform a haptic grasping task in a virtual environment simulation by proposing an optimized solution to overcome the issues of the underactuation~\cite{Lobo2017}. With this study, we won the Best Student Presentation awards in IEEE World Haptics Conference, 2017. We are recently invited to submit an extended version of this work as a journal for IEEE Transactions on Haptics. In particular,

    \begin{itemize}
    \item the literature based on the underactuated devices in haptics was investigated,
    \item the existing methods for haptic rendering were detailed,
    \item a new method was proposed to implement haptic rendering tasks for underactuated devices, and
    \item a set of experiment results were used to show the feasibility of this method in which the hand exoskeleton was being worn by a user to perform a simple virtual grasping task in Unreal simulation engine.
    \end{itemize}

\item A journal is being prepared for submission regarding the design of specific control algorithms for underactuated systems stating:

    \begin{itemize}
    \item a force control algorithm to be implemented for the underactuated hand exoskeleton,
    \item an experimental stage for stiffness rendering task, where the underactuated device is limited by the task to perform only $1-DoF$ rotation,
    \item two different optimization algorithms to implement stiffness rendering task to the underactuated exoskeleton with no limitations on the task, by

        \begin{itemize}
        \item optimizing the desired torques to satisfy the underactuation constraint, or
        \item optimizing the desired finger pose that is predicted to be reachable based on the previous behavior of the user.
        \end{itemize}

    \end{itemize}

\end{itemize}

The rest of this study goes as follows: Chapter~\ref{sec:Chapter2} will list the hand exoskeletons existing in the literature for rehabilitative, assistive, or haptic applications. These devices in the literature will be categorized in terms of their: (i) mobility, (ii) number of contact points, (iii) mechanism type, (iv) actuation and (v) control. Chapter~\ref{sec:Chapter3} will describe the design requirements for a hand exoskeleton to be used for assistive and rehabilitative exercises specifically using the previous categorizations studied for the devices in the literature. These requirements will define the kinematics of a novel mechanism for an exoskeleton. Once the device kinematics is defined, the link lengths of each finger component will be optimized to maximize the Range of Motion (RoM) of finger joints and the efficiency of the force transmission to finger phalanges. Chapter~\ref{sec:Chapter4} will focus on the implementation of mechanical design and electronic components. Later on, various control algorithms will be designed and tested on a single finger component based on (i) position control to follow a strict reference, (ii) position control to follow an EMG based reference set by the healthy hand of the user and (iii) force control to follow a force reference set by the therapist or user's interactions in the virtual environment. Chapter~\ref{sec:Chapter5} will detail the implementation of various haptic rendering strategies to improve the performance of underactuated exoskeleton by (i) modeling the user's hand as $2 DoFs$ with stiff bone structure and (ii) modeling the user's avatar as countless $DoFs$ due to the deformative model definition of user's hand. Finally, Chapter~\ref{sec:Chapter6} will conclude this study by summarizing the overall findings, achievements and future works that are left unfinished.

\clearpage
\pagebreak

\newpage

\chapter{Background and Literature}\label{sec:Chapter2} % Main chapter title

\lhead{Chapter 2. \emph{Background and Literature}} % This is for the header on each page - perhaps a shortened title

This chapter will first focus on the anatomical varieties and average properties of human hands. Then, it will reveal how hand exoskeletons in the literature, in terms of active mobility, actuation, number of connection points, and control while detailing the advantages and disadvantages for each application individually. In fact, the literature consists of a wide range of hand exoskeletons with many different properties, characteristics and purposes of use. The existing literature surveys~\cite{Sarakoglou2007, Foumashi2011, Heo2012} provide a good beginning. However, further analysis might be useful to create a generic hand exoskeleton, and to keep up with the latest technological development in the literature.

Despite of common characteristics to be fulfilled, the restrictions and requirements of a hand exoskeleton are shaped by the purpose of use. There is not a unique, straightforward design kinematics for a hand exoskeletons due to the variations of these design properties. Therefore, a systematic classification of the existing robotic devices in the literature is useful as a prior study in order to understand the consequences of all these decisions. Investigating the design aspects such as the active and passive degrees of freedom for each finger, portability, wearability, the kinematic structure, actuation and control strategies can be useful before starting the new design of a hand exoskeleton.

%In this chapter, anatomical properties of an average hand will be stated, some of the most important studies in the literature will be detailed and categorized in terms of the kinematic architecture. In the classification section, the devices will be compared to each other 

\newpage
\section{Anatomical Properties of Human Hand} \label{sec:anatomy}

\begin{figure}[t!]
\centering
%\vspace*{4\baselineskip}
\includegraphics[width=1.1\textwidth]{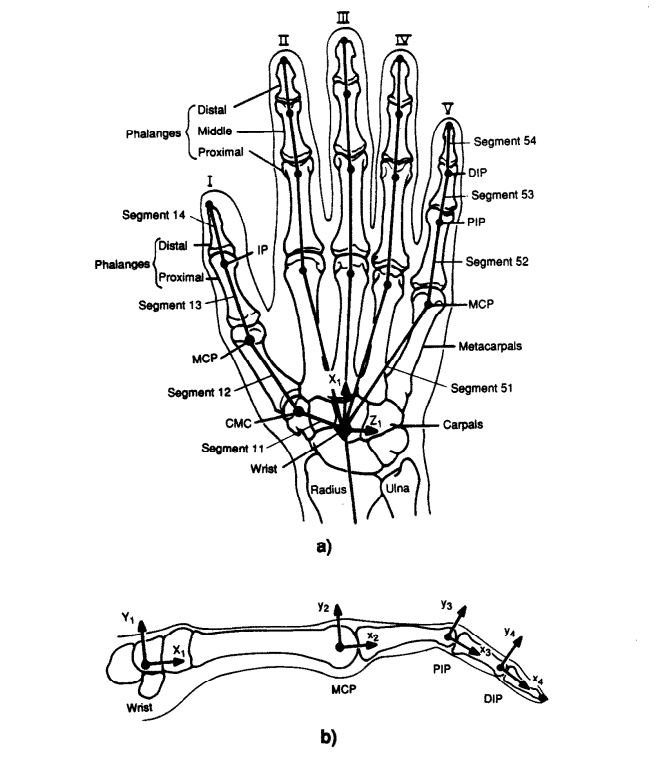}
\caption{Kinematic model of the human hand~\cite{Buchholz1992}. (a) dorsal view of the hand shows the definitions for the thumb (I), the index (II), the middle (III), the ring (IV), and the little fingers (V). The thumb has two phalanges (proximal, and distal), while other fingers have three (proximal, middle, and distal) from the palm to the fingertip. (b) ursal view of the index finger as an example for finger kinematics. MCP, PIP and DIP joints are shown from the palm to the fingertip.}
%\vspace*{4\baselineskip}
\label{fig:hand}
\end{figure}

A human hand has five fingers; thumb, index, middle, ring and little fingers as in Figure~\ref{fig:hand}. From the kinematic point of view, index, middle, ring and little fingers are similar with different phalange lengths, while the thumb is different than the rest of the fingers. Overall, anatomy of a human hand can be modelled with $20~DoFs$ with $15$ joints.

The index, middle, ring, and little fingers have $3$ phalanges named as proximal, middle and distal from the palm to the fingertip. Each phalange is connected to each other with joints named as metacarpophalangeal (MCP), proximal interphalangeal (PIP) and distalinterphalangeal (DIP). With these $3$ joints, a finger can reach $4$ degrees of freedom (DoFs) mobility~\cite{Wang2009}, such that  PIP and DIP joints perform only flexion/extension, while MCP joint performs both flexion/extension and abduction/adduction movements.

On the other hand, the thumb has $2$ phalanges named as proximal and distal from the palm to the fingertip, and $3$ joints named as Carpometacarpal (CMC), metacarpophalangeal (MCP) and interphalangeal (IP). With these $3$ joints, a finger can reach $4~DoFs$ mobility, such that IP and MCP joints have flexion/extension, whereas CMC joint has both flexion/extension and abduction/adduction $DoFs$. The thumb is crucial to perform meaningful tasks during virtual or real tasks. However, we handle the design of the thumb component as a separate research topic than the rest of the fingers due to the differences in their kinematics. 

The literature reports a wide variety of hand sizes within the society~\cite{Buryanov2010}. In particular, Buryanov \textit{et al.} analyzed the anatomical variations using right and left hands of $66$ adult patients between the ages of $19$ and $78$ from anterior-posterior X-ray images. As a result, they reported an average size for bone lengths of proximal, medial and distal phalanges as well as metacarpal bone, soft tissue at fingertip and web height from metacarpophalangeal joint. Table~\ref{tab:fingersize} shows the average size of human fingers for index, middle, ring and little fingers. Since modeling a finger should consider distances between MCP and PIP joints as the length of the first phalange; the size of proximal phalange should be added to web height. Table~\ref{tab:fingerratio} also shows the ratio between the proximal and the middle phalanges over distal phalanges using the data gathered by Buryanov \textit{et al.}.

\newpage

\begin{table}[!htb]
\vspace*{8\baselineskip}
\caption{Average finger phalange sizes for index, middle, ring and little fingers~\cite{Buryanov2010}.}
\label{tab:fingersize}
\begin{center}
\begin{tabular}{c||c c c c}
\hline \hline \\
 Finger & Web Height & Proximal P. & Middle P. & Distal P. \\ [1ex] \hline \\
 Index Finger & $15.52$ & $39.78 \pm 4.94$ & $22.38 \pm 2.51$ & $15.82 \pm 2.26$ \\ [1ex]
 Middle Finger & $15.33$ & $44.63 \pm 3.81$ & $26.33 \pm 3.00$ & $17.40 \pm 1.85$ \\ [1ex]
 Ring Finger & $18.49$ & $41.37 \pm 3.87$ & $25.65 \pm 3.29$ & $17.30 \pm 2.22$ \\ [1ex]
 Little Finger & $24.72$ & $32.74 \pm 2.77$ & $18.11 \pm 2.54$ & $15.96 \pm 2.45$ \\ [1ex]
 \hline \hline
\end{tabular}
\end{center}
\vspace*{2\baselineskip}
\end{table}

\begin{table}[!htb]
\vspace*{2\baselineskip}
\caption{Average ratios between proximal and middle phalanges over distal phalange for the index, middle, ring, and little fingers~\cite{Buryanov2010}.}
\label{tab:fingerratio}
\begin{center}
\begin{tabular}{c||c c}
\hline \hline \\
 Finger & Middle/Distal Phalanges & Proximal/Distal Phalanges \\ [1ex] \hline \\
 Index Finger & $1.4$ & $2.5$ \\ [1ex]
 Middle Finger & $1.5$ & $2.6$ \\ [1ex]
 Ring Finger & $1.5$ & $2.4$ \\ [1ex]
 Little Finger & $1.1$ & $2.1$ \\ [1ex]
 \hline \hline
\end{tabular}
\end{center}
% \vspace*{2\baselineskip}
\end{table}

\newpage

Similarly, Becker \textit{et al.}~\cite{Becker1988} studied natural range of motion (RoM) for the flexion/extension movements of MCP, PIP and DIP joints using a video image analyzer. The findings of Becker \textit{et al.} has been summarized in Table~\ref{tab:fingerrom}. 

\begin{table}[!htb]
\vspace*{3\baselineskip}
\caption{Ranges of motion (RoM) for finger joints during flexion/extension: means (standard deviations) in degrees~\cite{Becker1988}.}
\label{tab:fingerrom}
\begin{center}
\begin{tabular}{c||c c c}
\hline \hline \\
 Finger & MCP & PIP & DIP \\ [1ex] \hline \\
 Index Finger & 70.83 (11.09) & 103.87 (7.79) & 61.17 (12.71) \\ [1ex]
 Middle Finger & 85.30 (9.87) & 103.98 (8.98) & 73.64 (16.30) \\ [1ex]
 Ring Finger & 85.09 (14.46) & 107.15 (13.49) & 66.96 (15.77) \\ [1ex]
 Little Finger & 85.58 (18.09) & 98.95 (11.20) & 70.79 (15.84) \\ [1ex]
 \hline \hline
\end{tabular}
\end{center}
\vspace*{3\baselineskip}
\end{table}

Even though anatomical studies regarding hand sizes and the RoMs were stated independently, Buchholz \textit{et al.}~\cite{Buchholz1992} presented a simulation study where relation between finger sizes and required joint angles while grasping cylindrical object, which shows that required RoM to perform the single task might vary based on hand sizes.

%  Another elimination for finger mobility was performed based on abduction/adduction movement of each finger. MCP joints of $4$ fingers act as spherical joints, combining abduction/adduction movement with flexion/extension movement. However, most of hand functions during ADLs can be performed using flexion/extension of finger joints only, by restricting the abduction/adduction movements of MCP joint, or leave it passive to user. Such assumption allows human finger to be modelled much simpler while constraining the overall hand movement to a planar scene. %, as can be seen in Figure ~\ref{fig:underact}-(b).

\newpage
\section{Possible Applications for Hand Exoskeletons} \label{sec:appl}

\begin{figure*}[h!]
\centering
%\vspace*{-.5\baselineskip}
  \subfigure[Rehabilitation use~\cite{Tong2010} \label{fig:rehab}]%
	{\includegraphics[width=0.31\textwidth]{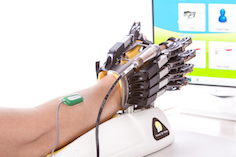}} \hspace{0.03\textwidth}
  \subfigure[Assistive use~\cite{Gasser2015} \label{fig:assist}]%
	{\includegraphics[width=0.24\textwidth]{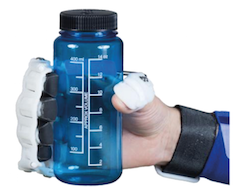}} \hspace{0.03\textwidth}
	\subfigure[Haptic use~\cite{Choi2016} \label{fig:hapt}]
	{\includegraphics[width=0.31\textwidth]{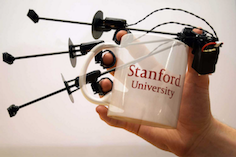}} \hspace{0.03\textwidth}
% \vspace*{-.75\baselineskip}
  \caption{Hand exoskeletons can be used to lead, assist or resist user's hand movements for: (a) rehabilitation use, (b) assistive use and (c) haptic use.}
	\label{fig:applications}
% \vspace*{-1\baselineskip}
\end{figure*}

As briefly mentioned previously, a hand exoskeleton can be used for specific desired tasks, user profiles and external factors during different applications, such as rehabilitation, assistive and haptic use, as shown in Figure~\ref{fig:applications}. %Each application has , worth to focus before investigating the exoskeletons in the literature. 

% \vspace*{0.5\baselineskip}
\textbf{Rehabilitation exoskeletons} are designed to treat disabilities of patients in a clinical setting (see Figure~\ref{fig:rehab}). They repeat mimicking the most common activities of daily living (ADL) by opening/closing the fingers. They must be easily wearable not to cause discomfort or pain for the patients during preparation. They should apply high output forces, and monitor finger movements for performance evaluation. They might allow patients to interact with real objects to increase the realism of therapy exercises. Instant adjustability for different tasks also is favorable, even though patients with severe disabilities would not take advantage of the task variety due to the loss of isolated individual finger movement after injury~\cite{Welmer2008}. Their portability is not mandatory especially for clinical devices, but still preferable.

% \vspace*{0.5\baselineskip}
\textbf{Assistive exoskeletons} are designed to assist patients with hand disabilities in performing ADLs, such as grasping a cup while drinking coffee, or holding a key while opening the door (see Figure~\ref{fig:assist}). Instant adjustability for different tasks, easy wearability and portability are highly important for assistive exoskeletons. They must allow patients to interact with real objects, and apply high output forces. Finger tracking is neither mandatory, nor favorable.

% \vspace*{0.5\baselineskip}
\textbf{Haptic exoskeletons} are designed for healthy subjects to interact with objects in a virtual environment (see Figure~\ref{fig:hapt}). Instant adjustability for different tasks, portability and efficient finger tracking are highly important for haptic exoskeletons. Since the target user profile is assumed to be healthy, the wearability or high output forces are not mandatory, but favorable.

\newpage

\begin{sidewaystable}[ph!]
\scriptsize
\caption{\textbf{5 fingered hand exoskeletons}: Main application (rehabilitation (R) / assistive (A) / haptic (H)), number of independent fingers, number of assisted and independent DoF for each finger, mechanism type, device placement (dorsal (DOR)/ palmar (PAL)/ lateral (LAT)), actuator, control modes (position (POS)/ velocity (VEL)/ backdriveable (BAC)/ admittance (ADM)/ impedance (IMP)/ EMG triggered (EMG)), pose estimation method (encoder (ENC)/ flex sensor (FLE)/ motion tracking (MT)/ additional sensor (SEN))}
\vspace*{-1\baselineskip}
\label{tab:5fin}
\begin{center}
{\rowcolors{2}{red!20!red!10}{red!5!red!2}
 \renewcommand\arraystretch{1.2}
 \begin{tabular}{p{3.5cm}| p{1cm} p{1.5cm} p{2.2cm} p{2.2cm} p{1.6cm} p{1.6cm} p{1.7cm} p{1.4cm} }
 \hline
Device & Applic. & Ind. Fing. & Fing. DoF (act.) & Mechanism & Placement & Actuation & Control & Fing. pose \\ \hline %& Device & Indep. fingers & Placement & Control modes \\ \hline \hline
\textbf{Tong et al.}~\cite{Tong2010, Tong2013, Tong2015} & R & 5 & 2(1) & Link (coupled) & DOR & DC (Lin) & BAC, EMG & ENC \\ % PAS, ACT\ K-ENC \\ % Y Y \cite{Tong2013}
\textbf{HEXOSYS}~\cite{Iqbal2011, Iqbal2015} & R & 5 & 4(1) & Link (fingertip) & DOR & DC (Rot) & POS & SEN \\ % PAS \ K-SEN \\ % N Y
\textbf{SAFE}~\cite{BenTzvi2015, Ma2016} & H,R & 5 & 3(1) & Link (fingertip) & DOR & DC (Rot) & ADM & SEN \\%PAS/ D-SEN \\%
\textbf{Fu et al.}~\cite{Fu2007} & R & 5 & 4(1) & Link (coupled) & DOR & DC (Rot) & POS & SEN \\%PAS/ D-SEN \\
\textbf{HANDEXOS}~\cite{Chiri2009, Chiri2012} & R & 5 & 4(1) & Link (coupled) & DOR & DC (Rot) & BAC & SEN,MT \\% PAS, BAC/ MT, D-SEN \\%
\textbf{Rahman et al.}~\cite{Rahman2012} & R & 5 & 3(1) & Link (coupled) & DOR & DC (Lin) & POS & ENC \\ %PAS, ACT / K-ENC \\ %&
\textbf{Jo et al.}~\cite{Jo2014, Jo2015} & H & 5 & 3(1) & Link-glove & DOR & DC (Lin) & VEL & FLE \\ % PAS, ACT /
\textbf{PneuGlove}~\cite{Connelly2010} & H,R & 5 & 3(1) & Glove & DOR, PAL & Pneumatic & POS, BAC & FLE \\ %PAS, ACT &
%\textbf{Kawasaki et al.}~\cite{Kawasaki2007} & R & 5 & 2(2) & Grounded & DOR & Servo & POS & ENC \\%ACT / D-ENC\\
\textbf{Fang et al.}~\cite{Fang2009} & H & 5 & 3(1) & Link (fingertip) & DOR & DC (Rot) & ADM & ENC \\ %ACT / K-ENC \\ %&
\textbf{Delph et al.}~\cite{Delph2013} & R & 5 & 3(1) & Glove & DOR, PAL & Servo & POS, EMG & - \\ %PAS, ACT &
\textbf{Polygerinos et al.}~\cite{Polygerinos2015} & R & 5 & 3(1) & Link (compliant) & DOR & Pneumatic & POS & - \\ % PAS/
\textbf{Cui et al.}~\cite{Cui2015} & R & 5 & 3(1) & Link (coupled) & DOR & DC (Lin) & POS & - \\ %PAS&
\textbf{Kim et al.}~\cite{Kim2017} & R & 5 & 3(2) & Link (coupled) & DOR & DC (Rot) & POS & ENC \\%PAS / K-ENC\
\textbf{Decker et al.}~\cite{Decker2017} & R & 5 & 2(1) & Link-glove & DOR & DC (Rot) & BAC & FLE \\ %PAS, BAC &
\textbf{Jo et al.}~\cite{Jo2017} & R & 5 & 2(1) & Link-glove & DOR & DC (Lin) & POS & ENC \\ %PAS / K-ENC
\textbf{Yap et al.}~\cite{Yap2015, Yap2017} & A,R & 5 & 3(1) & Compliant & DOR & Pneumatic & POS & - \\ %PAS &
\textbf{Lu et al.}~\cite{Lu2016} & R & 5 & 2(1) & Link (coupled) & DOR & DC (Rot) & - & ENC \\ % K-ENC \\
%\textbf{Ueki et al.}~\cite{Ueki2012} & R & 5 & 3(3) & Grounded & DOR & Servo & POS & ENC \\ % ACT, PAS /D-ENC \\
\textbf{Sarac et al.}~\cite{Sarac2016, Sarac2017, Sarac2018, Gabardi2018} & R,A,H & 5 & 2(1) & Link (underact.) & DOR & DC (Lin) & POS, ADM & SEN \\ %ACT, PAS / K-SEN
\textbf{Hasegawa et al.}~\cite{Hasegawa2008, Hasegawa2011} & A & 3 & 3(3) & Link (indep) & DOR, LAT & DC (Rot) & POS, ADM & ENC \\ %PAS, ACT / D-ENC \\ %
\textbf{Burton et al.}~\cite{Burton2011} & R & 3 & 3(2) & Link (coupled) & DOR & Pneumatic & - & ENC,MT \\ %K-ENC, MT \\ %Y Y
\textbf{Ferguson et al.}~\cite{Ferguson2018} & R & 3 & 4(2) & Link (coupled) & DOR & DC (Rot) & POS, ADM & - \\ %K-ENC, MT \\ %Y Y
\textbf{BiomHED}~\cite{Lee2013, Lee2014} & R & 2 & 4(3) & Link-glove & DOR, PAL & DC (Rot) & ADM & MT \\ % PAS &
%\textbf{HEXORR}~\cite{Schabowsky2010} & R & 2 & 2(1) & Grounded & PAL, LAT & DC (Rot) & BAC & ENC \\ % BAC / K-ENC\\
\textbf{BRAVO Hand}~\cite{Troncossi2012, leonardis2015emg} & R & 2 & 3(1) & Link (mitten) & DOR & DC (Rot) & POS, EMG & ENC \\ % ACT / K-ENC \\
\textbf{Mulas et al.}~\cite{Mulas2005} & R & 2 & 3(1) & Link-glove & DOR, LAT & Servo & EMG & ENC \\ %PAS, ACT / K-ENC
%\textbf{HWARD}~\cite{Takahashi2005} & R & 2 & 2(1) & Grounded & DOR, LAT & Pneumatic & BAC & SEN \\ % PAS, BAC / K-SEN
\textbf{HANDSOME}~\cite{Brokaw2011} & R & 2 & 1 & Link (indep) & DOR & Spring & - & ENC \\ % y y
%\textbf{Loureiro et al.}~\cite{Loureiro2007} & R & 2 & 2(2) & Stationary & DOR, LAT & DC (Rot) & ADM & ENC \\ %PAS, BAS / D-ENC
\textbf{Exophalanx-2}~\cite{Kobayashi2013} & H & 2 & 3(2) & Link-glove & DOR, LAT & SMA & ADM & ENC \\ %ACT / K-ENC
%\textbf{Butler et al.}~\cite{Butler2017} & R & 2 & 2(0) & Grounded & LAT & Spring & - & - \\
%\textbf{HandCARE}~\cite{Dovat2008} & R & 1 & 4(1) & Grounded & FIN & DC (Rot) & POS, BAC& - \\ %PAS, BAC
\textbf{Li et al.}~\cite{Li2017} & R & 1 & 3(1) & Glove & DOR & Pneumatic & POS & FLE \\ %PAS
%\textbf{Chang et al.} \cite{Chang2014} & R & 1 & 3(1) & Link (station) & DOR & DC (Rot) & POS & - % PAS
\end{tabular} }
% \vspace*{-1.5\baselineskip}
\end{center}
\end{sidewaystable}

\begin{sidewaystable}[ph!]
\scriptsize
\caption{\textbf{4, and 3 fingered hand exoskeletons}: Main application (rehabilitation (R) / haptic (H) / assistive (A)), number of assisted and independent fingers, number of assisted and independent DoF for each finger, mechanism type, device placement (dorsal (DOR)/ palmar (PAL)/ lateral (LAT)), actuator, control modes (position (POS)/ velocity (VEL)/ backdriveable (BAC)/ admittance (ADM)/ impedance (IMP)/ EMG triggered (EMG)), pose estimation method (encoder (ENC)/ flex sensor (FLE)/ motion tracking (MT)/ additional sensor (SEN))}
\vspace*{-2\baselineskip}
\label{tab:43fin}
\begin{center}
 \renewcommand\arraystretch{1.2}
{\rowcolors{2}{red!20!red!10}{red!5!red!2}
 \begin{tabular}{p{3.5cm}| p{1cm} p{1.5cm} p{2.2cm} p{2.2cm} p{1.6cm} p{1.6cm} p{1.7cm} p{1.4cm} }
 \hline
Device & Applic. & Ind. Fing. & Fing. DoF (act.) & Mechanism & Placement & Actuation & Control & Fing. pose \\ \hline %& Device & Indep. fingers & Placement & Control modes \\ \hline \hline
\textbf{Rutgers Master}~\cite{Bouzit2002} & H & 4(4) & 4(1) & Link (fingertip) & PAL & Pneumatic & POS & SEN \\ %PAS /K-SEN \\
\textbf{Popov et al.}~\cite{Popov2017} & A & 4(4) & 3(1) & Glove & DOR & Pneumatic & POS & FLE \\ % PAS, ACT
\textbf{Allotta et al.}~\cite{Allotta2015} & A & 4(4) & 3(1) & Link (coupled) & DOR & Servo & POS & ENC \\ %PAS / K-ENC
%\textbf{Choi et al.}~\cite{Choi2016} & H & 4(4) & 3(0) & Grounded & FIN & Spring & - & - \\
\textbf{Wu et al.}~\cite{Wu2010} & R & 4(1) & 2(2) & Link (mitten) & DOR & Pneumatic & POS & SEN \\ %PAS / K-SEN \\
\textbf{Weiss et al.}~\cite{Weiss2013} & R & 4(1) & 4(1) & Link (coupled) & DOR,LAT & DC (Rot) & POS & SEN \\%PAS/ D-SEN
\textbf{Wei et al.}~\cite{Wei2013} & R & 4(1) & 2(1) & Link (mitten) & DOR,LAT & DC (Rot) & - & FLE \\
\textbf{Arata et al.}~\cite{Arata2013, Arata2016} & A & 4(1) & 3(1) & Link (compliant) & DOR & DC (Lin) & POS, BAC & ENC,MT \\ %PAS, ACT / K-ENC, MT \\
\textbf{Gasser et al.}~\cite{Gasser2015, Gasser2017} & R,A & 4(1) & 2(1) & Link (mitten) & DOR,LAT & DC (Rot) & POS & ENC \\ %K-ENC \\
\textbf{Lince et al.}~\cite{Lince2017} & R,A & 4(1) & 3(1) & Link (mitten) & DOR & DC (Rot) & EMG & ENC \\ \hline \hline
%\textbf{Brenosa et al.}~\cite{Brenosa2011} & H & 3(3) & 3(3) & Grounded & FIN & DC (Rot) & POS, BAC & ENC \\ %D-ENC \\
\textbf{Wei et al.}~\cite{Wei2017} & R,A & 3(3) & 3(2) & Link (coupled) & DOR & DC (Rot) & IMP & ENC \\ %K-ENC \\
\textbf{Exophalanx}~\cite{Kobayashi2012} & H & 3(3) & 3(1) & Link-glove & DOR & SMA & ADM & FLE,MT \\
\textbf{Ryu et al.}~\cite{Ryu2008} & H & 3(3) & 3(1) & Link (compliant) & DOR & Pneumatic & POS, BAC & FLE \\ % y y
\textbf{Sarakoglou et al.}~\cite{Sarakoglou2016} & H & 3(3) & 4(1) & Link (fingertip) & DOR & DC (Rot) & ADM & SEN \\ %K-SEN\\
\textbf{In et al.}~\cite{In2010, In2015} & A & 3(1) & 3(1) & Glove & DOR & DC (Rot) & ADM & ENC \\ %K-ENC \\ % y y
\textbf{Geo et al.}~\cite{Geo2016} & R & 3(1) & 3(1) & Link (coupled) & DOR & DC (Rot) & - & ENC 
\end{tabular} }
\vspace*{-1.5\baselineskip}
\end{center}
%\vspace*{-1.5\baselineskip}
\end{sidewaystable}

\begin{sidewaystable}[ph!]
\scriptsize
\caption{\textbf{2, and 1 fingered hand exoskeletons}: Main application (rehabilitation (R) / haptic (H) / assistive (A)), number of assisted and independent fingers, number of assisted and independent DoF for each finger, mechanism type, device placement (dorsal (DOR)/ palmar (PAL)/ lateral (LAT)), actuator, control modes (position (POS)/ velocity (VEL)/ backdriveable (BAC)/ admittance (ADM)/ impedance (IMP)/ EMG triggered (EMG)), pose estimation method (encoder (ENC)/ flex sensor (FLE)/ motion tracking (MT)/ additional sensor (SEN))}
\vspace*{-2\baselineskip}
\label{tab:21fin}
\begin{center}
 \renewcommand\arraystretch{1.2}
{\rowcolors{2}{red!20!red!10}{red!5!red!2}
 \begin{tabular}{p{3.5cm}| p{1cm} p{1.5cm} p{2.2cm} p{2.2cm} p{1.6cm} p{1.6cm} p{1.7cm} p{1.4cm} }
 \hline
Device & Applic. & Ind. Fing. & Fing. DoF (act.) & Mechanism & Placement & Actuation & Control & Fing. pose \\ \hline %& Device & Indep. fingers & Placement & Control modes \\ \hline \hline
\textbf{Cempini et al.}~\cite{Cempini2013, Cortese2015} & R & 2(2) & 4(1) & Link (f.coupled) & DOR,LAT & DC (Rot) & VEL & ENC \\ %K-ENC \\
\textbf{FINGER}~\cite{Taheri2014} & R & 2(2) & 2(2) & Link (indep) & DOR & DC (Lin) & POS, BAC & ENC \\ %PAS, BAC / D-ENC
\textbf{WHIPFI}~\cite{Gosselin2005} & H & 2(2) & 3(1) & Link (fingertip) & DOR & DC (Rot) & ADM & SEN \\ %PAS / K-SEN \\
\textbf{Fontana et al.}~\cite{Fontana2009} & H & 2(2) & 4(3) & Link (coupled) & DOR & DC (Rot) & - & ENC \\ %K-ENC \\
\textbf{iHandRehab}~\cite{Li2011} & R & 2(2) & 4(4) & Link (indep) & DOR & DC (Rot) & POS, BAC & ENC \\ %PAS, BAC / D-ENC
\textbf{Fiorilla et al.}~\cite{Fiorilla2008} & R & 2(2) & 1 & Link (MCP) & DOR & DC (Rot) & - & SEN \\ \hline \hline%D-SEN \\ % y y
\textbf{Wang et al.}~\cite{Wang2009} & R & 1(1) & 4(4) & Link (indep) & DOR & DC (Rot) & POS, BAC & ENC \\%ACT, BAC/ D-ENC
\textbf{Yamaura et al.}~\cite{Yamaura2009} & R & 1(1) & 3(2) & Link (coupled) & DOR & Servo & POS & ENC \\%PAS,ACT/ K-ENC
\textbf{Jones et al.}~\cite{Jones2010, Jones2012} & R & 1(1) & 3(3) & Link (indep) & DOR,LAT & DC (Rot) & POS, BAC & ENC \\ %D-ENC \\
\textbf{Polotto et al.}~\cite{Polotto2012} & R, A & 1(1) & 4(4) & Link (indep) & DOR,LAT & DC (Rot) & POS, BAC & ENC \\ %D-ENC \\
\textbf{Tang et al.}~\cite{Tang2011, Tang2013} & R & 1(1) & 2(1) & Link (coupled) & DOR & Ultrasonic & POS, BAC & ENC \\ %K-ENC \\
\textbf{DiCicco et al.}~\cite{DiCicco2004} & A & 1(1) & 3(1) & Link (coupled) & DOR & Pneumatic & EMG & ENC \\ %K-ENC \\
\textbf{Wege et al.}~\cite{Wege2005, Wege2006, Wege2007} & R & 1(1) & 4(1) & Link (coupled) & DOR & DC (Rot) & POS, EMG & SEN \\ %K-SEN \\ \cite{}
\textbf{Sun et al.}~\cite{Sun2009} & R & 1(1) & 4(1) & Link (fingertip) & DOR & Pneumatic & POS, ADM & MT \\
\textbf{Agarwal et al.}~\cite{Agarwal2013} & R & 1(1) & 4(3) & Link (indep) & DOR & Pneumatic & - & SEN \\ %K-SEN \\
\textbf{Agarwal2 et al.}~\cite{Agarwal2015} & R & 1(1) & 3(2) & Link (coupled) & DOR & Elastic & IMP & ENC\\ %D-ENC \\
\textbf{AssistOnFinger}~\cite{Hocaoglu2009, Ertas2014} & R & 1(1) & 3(1) & Link (underac.) & DOR & DC (Rot) & POS,BAC,EMG & - \\
\textbf{Aubin et al.}~\cite{Aubin2013} & R & 1(1) & 2(2) & Link-glove & DOR & Servo & POS, BAC & ENC \\ %D-ENC\\
\textbf{Lee et al.}~\cite{Lee2015} & R & 1(1) & 3(1) & Link (coupled) & DOR & DC (Rot)& - & SEN \\ %D-SEN \\
\textbf{Maeder York et al.}~\cite{Maeder-York2014} & R & 1(1) & 3(1) & Link (compliant) & DOR & Pneumatic & - & ENC \\ %D-ENC
\end{tabular} }
\vspace*{-1.5\baselineskip}
\end{center}
%\vspace*{-1.5\baselineskip}
\end{sidewaystable}

\newpage
\section{Hand Exoskeleton Literature} \label{sec:classification}

\begin{figure}[h!]
  \centering
  \vspace*{1\baselineskip}
  \resizebox{4.5in}{!}{\includegraphics{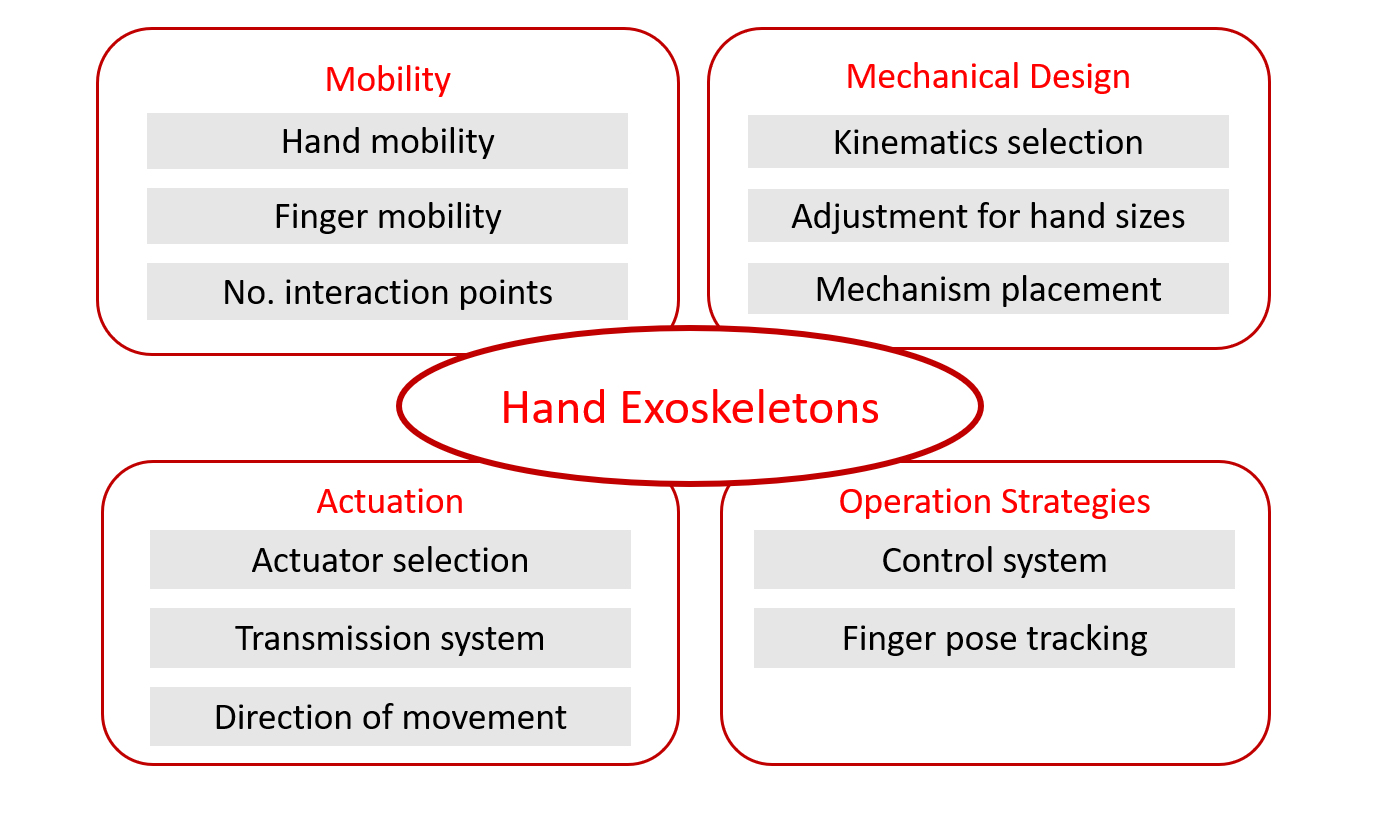}}
%   \vspace*{-.75\baselineskip}
  \caption{Hand exoskeletons should be categorized by design selections that can be categorized under design aspects: mobility, mechanical design, actuation and operational strategies.}
  \label{fig:structure}
  \vspace*{1\baselineskip}
\end{figure}

The complex anatomy of the human hand and specific requirements for each applications have previously leaded designers to implicitly select a single use and desired task, and to simplify the mechanical design of the exoskeleton. Such differences among the possible target applications increase the need to create a categorization factor while investigating the devices in the literature. State-of-the-art exoskeletons (see Table~\ref{tab:5fin}, Table~\ref{tab:43fin} and Table~\ref{tab:21fin}) will be investigated based on other properties, such as the number of actuated fingers, number of active and passive joints for each finger, number of connection points, the type of finger pose estimation, the actuation type, the mechanism placement type and the control algorithms, as summarized in Figure~\ref{fig:structure}. %We will then discuss whether each possible design choice is suitable for a generic hand exoskeleton. 

The very first step of designing a hand exoskeleton from scratch should be listing the design criteria for the focus of application. In particular, these design criteria can be formed only by investigating the advantages and disadvantages of each classification category and find out the most appealing option for the future device. With this motivation, this section combines the hand exoskeleton detailed in the previous section under all possible classification category to study all the possible alternatives.

\subsection{Mobility}

\begin{figure}[h!]
  \centering
  \vspace*{1\baselineskip}
  \resizebox{4.5in}{!}{\includegraphics{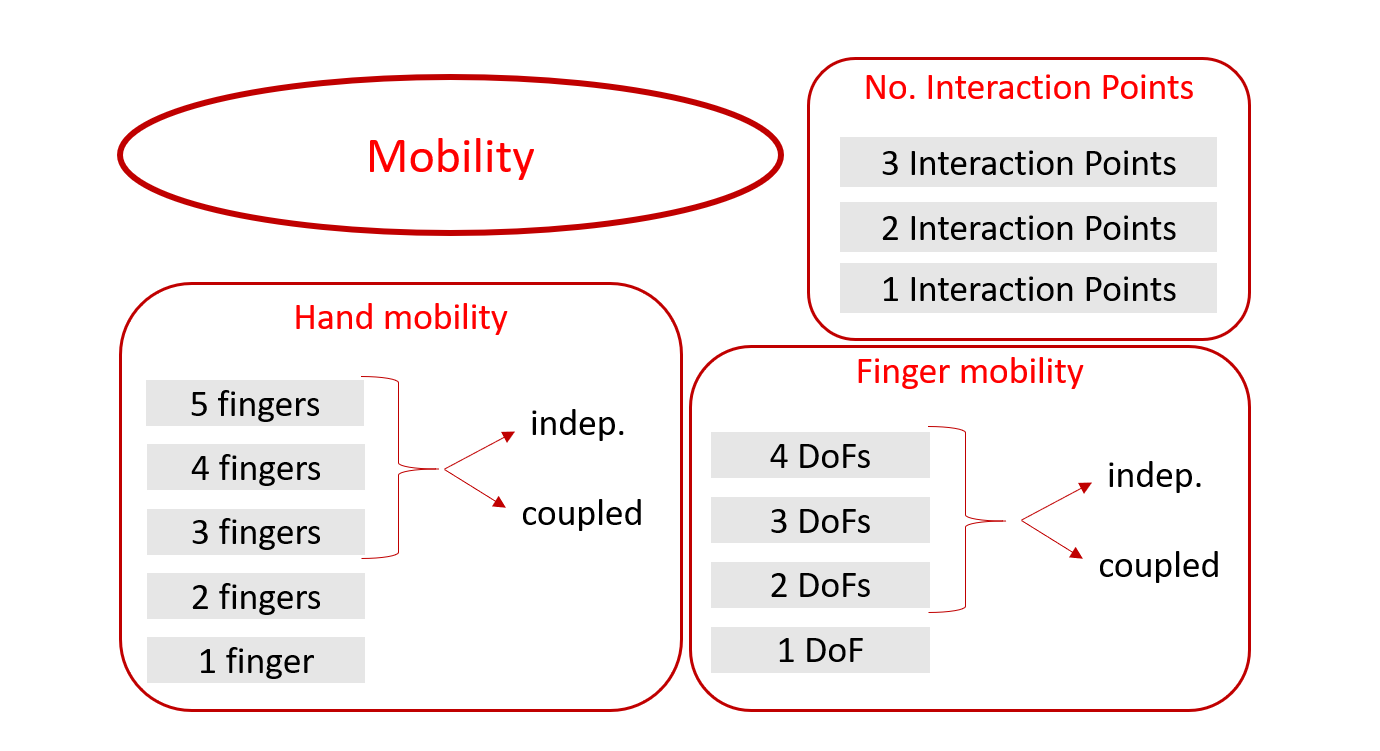}}
%   \vspace*{-.75\baselineskip}
  \caption{Possible design choices for mobility based on hand mobility, finger mobility and number of interaction points.}
  \label{fig:mobility}
  %\vspace*{-.5\baselineskip}
\end{figure}

Mobility assisted by an exoskeleton can be handled in terms of hand mobility, finger mobility and the number of interaction points between the mechanism and user's finger (see Figure~\ref{fig:mobility}). Both hand and finger mobility can be categorized further based on the number of assisted and independently controlled mobility.

% \vspace*{1\baselineskip}

\subsubsection{Hand mobility} \label{sec:q1q2}

A human hand has $5$ fingers, and an exoskeleton can be designed to assist and control various numbers of fingers. Finger exoskeletons~\cite{Aubin2013, Maeder-York2014, Hocaoglu2009, Sun2009, Lee2015, Jones2010, Polotto2012, Wang2009, Agarwal2013, Tang2011, DiCicco2004, Wege2005, Yamaura2009} are designed mostly for the index finger, and are mostly stated as an initial study for a multi-finger exoskeleton. $2$-finger exoskeletons control thumb and index finger independently, and support only specific hand movements for rehabilitation or haptics, such as finger tapping or pick-and-place tasks~\cite{Cempini2013, Cempini2015, Taheri2014, Gosselin2005, Fontana2009, Li2011, Fiorilla2008, Cortese2015, Hasegawa2011}.

Even though exoskeletons with $1$ or $2$ fingers are simpler to implement, most of ADLs require at least $3$ fingers to be assisted. One approach to design multi-finger exoskeletons is to control each finger component individually. $5$-finger exoskeletons control each finger independently, and can be used for all applications with minimum constraints~\cite{Burton2011, Fu2007, Tong2010, Chiri2009, Rahman2012, Cui2015, Kim2017, Connelly2010, Delph2013, Decker2017, Jo2017, Yap2015, Polygerinos2015, BenTzvi2015, Fang2009, Lu2016, Jo2014, Iqbal2011, Sarac2016}. Since the middle, ring and little fingers of a healthy person are highly coupled, $4$-finger exoskeletons, which control thumb, index, middle and ring fingers~\cite{Popov2017, Bouzit2002}, or $3$-finger exoskeletons, which control thumb, index and middle fingers~\cite{Kobayashi2012, Ryu2008, Sarakoglou2016, Wei2017}, can be used all applications. Even though these devices can assist users during all ADLs, the perception of realism would drop as the number of assisted fingers decrease. On the other hand, $4$-finger exoskeletons, which control index, middle, ring and little fingers, cannot be effective for grasping or picking tasks during assistive or haptic applications due to the lack of resistive forces acting on the objects through the thumb~\cite{Allotta2015}.

Increasing the number of assisted fingers improves the overall mobility while complicating the design. The second approach to design multi-finger exoskeletons is to couple finger movements through mechanical~\cite{Hasegawa2008, Ferguson2018, Burton2011, Mulas2005, Lee2013, Brokaw2011, Troncossi2012, Kobayashi2013, Wu2010, Wei2013, Gasser2015, Lince2017} or differential~\cite{Li2017, Arata2013, Weiss2013, In2010, Geo2016, Gasser2017} systems.

Even though we cannot claim that moving finger components together prevents the exoskeleton to be used for certain applications, it limits certain tasks. For instance, a $5$-finger exoskeleton with coupled index, middle, ring and little fingers can assist users grasping objects only in certain shapes (e.g. a water bottle) during assistive or haptic applications, but not a key. This is why a generic hand exoskeleton should control $4$ or $5$ fingers independently.

%together might be useful for certain physical rehabilitation tasks, where repetitive muscle activity is the key, but can be extended for assistive or haptic applications only partially, where the interaction with objects are predefined. a $4$ fingers are coupled mechanically  For instance, a $5$ fingered exoskeleton that is controlled with a single actuator can be used for all applications, but only to perform the same movement repeatedly. This is why a generic device should control each finger independently, and an exoskeleton with $5$ finger provide the full mobility. %Exoskeletons with $5$ independent fingers is the most natural and generic option.%, while also ones with $4$ independent fingers but little finger is natural enough, since ring and little fingers are highly coupled to each other anatomically.

%\vspace*{-0.5\baselineskip}

A generic exoskeleton should assist user's natural finger movements. The designer can choose to independently control $5$ fingers or $4$ fingers, while the little finger is either left free or coupled with the ring finger. In particular, the anatomic coupling between the ring and little fingers would allow designers to simplify the mechanical system without sacrificing the natural hand movements.

\subsubsection{Finger mobility} \label{sec:q3q4}

A human finger has $4~DoF$ mobility, and an exoskeleton can be designed to assist and control various numbers of finger joints for each finger. $1~DoF$ mechanisms~\cite{Brokaw2011, Fiorilla2008} only flex/extend MCP joint for repetitive rehabilitation exercises and enhanced motor learning. Even though finger components with $1~DoF$ mobility are simpler to implement and easier to be worn, most of ADLs require at least $2$ DoF to be assisted for each finger. One approach to design multi-DoF mechanisms is to control each finger joint individually with $2~DoF$~\cite{Aubin2013, Taheri2014, Wu2010}, $3~DoF$~\cite{Jones2010, Hasegawa2008} or $4~DoF$~\cite{Wang2009, Li2011, Polotto2012} mobility.

\begin{figure*}[h!]
\centering
  \subfigure[$3~Points$ ($3~DoF$)~\cite{Wege2007} \label{fig:3cp}]%
	{\includegraphics[width=0.27\textwidth]{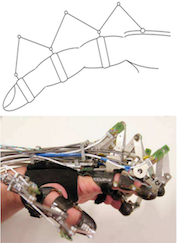}} \hspace{0.1\textwidth}
  \subfigure[$2~Points ($2~DoF$)$~\cite{Tong2013} \label{fig:2cp}]%
	{\includegraphics[width=0.27\textwidth]{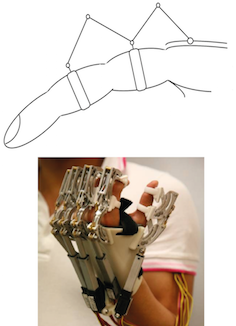}} \\ %\hspace{0.01\textwidth} \
	\subfigure[$1~Point$ ($1~DoF$)~\cite{Fiorilla2008} \label{fig:1cp_1}]%
	{\includegraphics[width=0.27\textwidth]{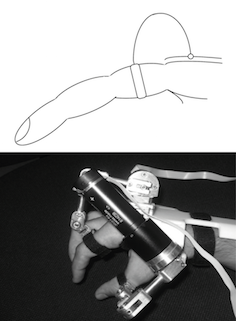}}\hspace{0.1\textwidth}
  \subfigure[$1~Point$ ($3~DoF$)~\cite{Chang2014} \label{fig:1cp_3}]%
	{\includegraphics[width=0.32\textwidth]{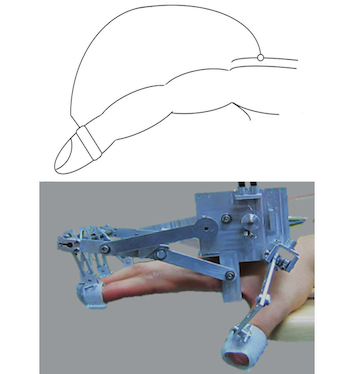}} %\hspace{0.01\textwidth}
%   \vspace*{.5\baselineskip}
	\caption{Hand exoskeletons can be designed with different numbers of interaction points between the device and user's fingers. Multiple interaction points improve grasping stability, user's safety and perception of touch but are harder to be worn.}
	\label{fig:cp}
  \vspace*{.5\baselineskip}
\end{figure*}

Increasing the number of assisted joints improves the overall mobility while complicating the design. The second approach to design multi-DoF mechanisms is to couple finger joints through mechanical or differential systems. Towards simplifying the finger components, the first step can be leaving the abduction/adduction of MCP joint passive~\cite{Agarwal2013}, or neglected completely, since most of the ADLs focus on finger opening/closing. Even then, controlling $3~DoF$ flexion/extension independently can be challenging. As the second simplification step, DIP and PIP joints can be coupled with a mechanically adjustable ratio, while MCP joint is controlled independently~\cite{Fontana2009, Ferguson2018, Yamaura2009, Burton2011, Wei2017, Cortese2015, Kim2017, Agarwal2015, Kobayashi2012, Kobayashi2013}. Since DIP and PIP joints are anatomically coupled, this simplification does not affect the perception significantly, but coupling them with a constant ratio might limit certain finger synergies.

Finally, a mechanism can be designed with a single actuator to control finger opening/closing through $4~DoF$~\cite{Bouzit2002, Iqbal2011, Sun2009, Sarakoglou2016, Fu2007, Wege2005, Chiri2009, Weiss2013, Cempini2013, Lee2013}, $3~DoF$~\cite{In2010, Connelly2010, Delph2013, Li2017, Popov2017, Mulas2005, Kobayashi2012, Jo2014, Arata2013, Yap2015, Polygerinos2015, Ryu2008, Gosselin2005, BenTzvi2015, Fang2009, Lince2017, Troncossi2012, Hocaoglu2009, DiCicco2004, Rahman2012, Allotta2015, Cui2015, Lee2015, Geo2016, Maeder-York2014} or $2~DoF$~\cite{Decker2017, Jo2017, Tong2010, Tang2011, Lu2016, Wei2013, Gasser2015, Sarac2016} mobility. Such coupling can set by a constant ratio through mechanical linkages or differential systems, or by adjusting the transmitted forces automatically based on contact forces~\cite{Gosselin2003}.

Even though we cannot claim that moving finger joints together prevents the exoskeleton to be used for certain applications, it limits certain tasks. For instance, a $3~DoF$ mechanism with constant ratio can assist users grasping objects in certain shapes (e.g. a water bottle) during assistive or haptic
applications, but not a phone without having mechanical adjustments. This is why a generic hand exoskeleton should flex/extend $2$ or $3$ finger joints independently, or coupled based on contact forces. Compared to fully controlled mechanisms, underactuated systems based on contact forces are mechanically simpler and cheaper, but require more complicated operational strategies.

A generic exoskeleton should allow finger joints to flex/extend in different synergies, based on different tasks. The designer can passively abduct/adduct MCP joint, since it does not significantly change the task performance during ADLs. Furthermore, the designer might focus on flexion/extension of MCP and PIP joints only, since the natural coupling between DIP and PIP joints would cause the DIP joints to move accordingly even without assistance. The designer can choose to achieve $2~DoF$ or $3~DoF$ mobility for each finger either by controlling them independently, or by coupling them using strategies to adjust for different tasks.

\subsubsection{Number of interactions} \label{sec:q5}

A human finger has $3$ phalanges, and an exoskeleton can be designed to interact with various numbers of phalanges to transmit actuator forces and to rotate finger joints. The number of interactions mostly depends on finger mobility. One approach to design finger components is to choose the same number of interaction points as the number of DoF. In other words, an exoskeleton can be designed with $4~DoF$ and $3$ interaction points~\cite{Fu2007, Wege2005, Lee2013, Weiss2013, Cempini2013, Chiri2009, Li2011, Wang2009, Polotto2012, Agarwal2013}, $3~DoF$ and $3$ interaction points (Figure~\ref{fig:3cp})~\cite{In2010, Connelly2010, Delph2013, Li2017, Popov2017, Mulas2005, Kobayashi2012, Kobayashi2013, Jo2014, Arata2013, Yap2015, Ryu2008, Jones2010, Hasegawa2008, DiCicco2004, Rahman2012, Allotta2015, Geo2016, Lee2015, Yamaura2009, Burton2011, Wei2017, Kim2017, Agarwal2015, Troncossi2012, Lince2017, Hocaoglu2009, Maeder-York2014}, $2~DoF$ and $2$ interaction points (Figure~\ref{fig:2cp})~\cite{Aubin2013, Decker2017, Jo2017, Taheri2014, Tong2010, Tang2011, Lu2016, Wu2010, Wei2013, Sarac2016} or $1~DoF$ and $1$ interaction point (Figure~\ref{fig:1cp_1})~\cite{Brokaw2011, Fiorilla2008}.

Devices with multiple interactions enhance the grasping stability during assistive and rehabilitation, and improve the haptic perception. Furthermore, they improve patients' safety by strictly limiting the spasticity. However, they might suffer from the design complexity of choosing high finger mobility. Mechanisms with $2$ interaction points can achieve $3~DoF$~\cite{Cui2015, Cortese2015} or $4~DoF$~\cite{Fontana2009} finger mobility. Alternatively, fingertip devices can achieve $2~DoF$~\cite{Gasser2015}, $3~DoF$ (Figure~\ref{fig:1cp_3})~\cite{Polygerinos2015, Gosselin2005, BenTzvi2015, Fang2009} or $4~DoF$~\cite{Bouzit2002, Iqbal2011, Sun2009, Sarakoglou2016, Ferguson2018} mobility. Even though having less number of interaction points simplifies the device mechanically, they might fail to reflect realistic interactions for certain haptic or assistive applications. For instance, a fingertip device can allow users to interact with objects and apply event-based forces, but cannot apply grasping forces on finger phalanges realistically.

Even though having less number of interaction points have simpler design and are easier to be worn, a generic exoskeleton should be designed with the same number of interaction points as the number of DoF. The number of interaction points between a generic exoskeleton and user's fingers should be decided according to the number of finger mobility.

\vspace*{-0.3\baselineskip}

\subsection{Mechanical Design}

\begin{figure}[h!]
  \centering
  \vspace*{1\baselineskip}
  \resizebox{4.2in}{!}{\includegraphics{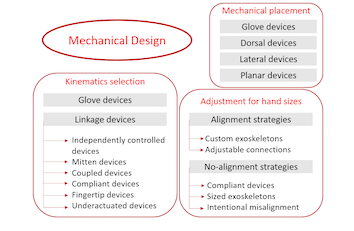}}
%   \vspace*{-1\baselineskip}
  \caption{Possible design choices for mechanical design based on kinematics selection, mechanical placement and adjustment strategies for hand sizes.}
  \label{fig:mechanical}
%   \vspace*{-1.2\baselineskip}
\end{figure}

Towards creating a hand exoskeleton, the next step of the designer should be how to achieve the mobility decisions through mechanical design. The mechanical design aspect can be handled based on kinematics selection, mechanical placement, and adjustment strategies for different hand sizes (see Figure~\ref{fig:mechanical}).

\newpage
\subsubsection{Kinematics selection} \label{sec:q6}

\begin{figure*}[h!]
\centering
  \vspace*{1\baselineskip}
\subfigure[Glove~\cite{Connelly2010} \label{fig:glove}]%
{\includegraphics[width=0.25\textwidth]{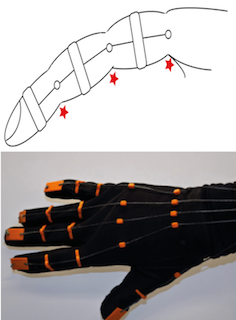}} \hspace{0.07\textwidth}
\subfigure[Glove with links~\cite{Jo2014} \label{fig:halfglove}]%
{\includegraphics[width=0.25\textwidth]{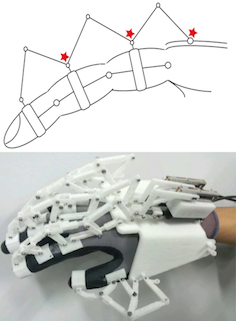}} \hspace{0.07\textwidth}
\subfigure[Independent control~\cite{Agarwal2013} \label{fig:indep}]%
{\includegraphics[width=0.25\textwidth]{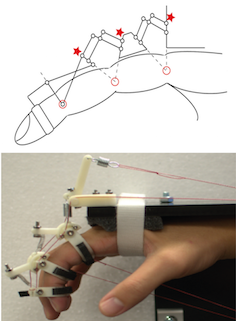}} \\ [1em]
\subfigure[Fingertip device~\cite{Sun2009} \label{fig:fingertip_coupled}]%
{\includegraphics[width=0.25\textwidth]{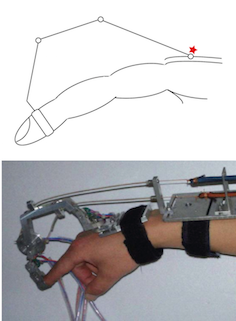}} \hspace{0.07\textwidth}
\subfigure[Coupled device~\cite{Troncossi2012} \label{fig:coupled}]%
{\includegraphics[width=0.25\textwidth]{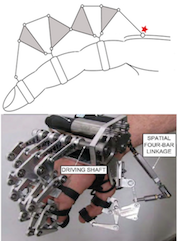}} \hspace{0.07\textwidth}
\subfigure[Underactuation~\cite{Ertas2014} \label{fig:underactuated}]%
{\includegraphics[width=0.25\textwidth]{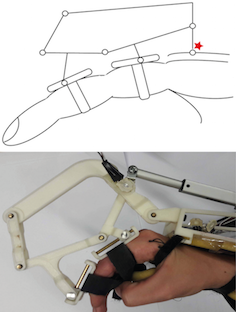}}
\caption{Types of kinematics selections as hand exoskeletons: the black circles show the mechanical joints, while the red stars represent the actuated ones. Glove-based devices can track finger pose easily and efficiently but are hard to be worn. Linkage-based devices are lightweight, portable and easily wearable. Linkage-based devices can be categorized based on the finger mobility choices detailed in Section~\ref{sec:q3q4}.}
	\label{fig:kinematics}
	  \vspace*{1\baselineskip}
\end{figure*}

The kinematics structure of a hand exoskeleton can be handled as glove-based or linkage-based devices. Figure~\ref{fig:kinematics} shows examples of exoskeletons with different kinematics selections. 

\textbf{Glove-based devices} require the user to wear a flexible glove equipped with sensors for motion tracking, and are perfect for haptic applications. They can assist/resist user's activity through cable transmission (Figure~\ref{fig:glove})~\cite{In2010, Connelly2010, Delph2013, Li2017, Popov2017}, or linkage transmission (Figure~\ref{fig:halfglove})~\cite{Mulas2005, Kobayashi2012, Kobayashi2013, Jo2014, Lee2013, Aubin2013, Decker2017, Jo2017}. %However, the glove interface significantly affects the wearability.
Even though their wearability can be improved using Velcro connections in the palm~\cite{Decker2017} or half gloves~\cite{Mulas2005, Jo2017}, patients still have to reach an initial pose to wear the glove.

\textbf{Linkage-based devices} use mechanical links to form the finger components, and can be further categorized with independent joint control, MCP rotation only, full coupling, partial coupling, mitten style, fingertip connection, compliance and contact based underactuation. Devices with independent control have an individual actuator for each assisted finger joint (Figure~\ref{fig:indep})~\cite{Wang2009, Jones2010, Li2011, Hasegawa2008, Polotto2012, Taheri2014, Agarwal2013, Fiorilla2008, Brokaw2011}. These actuators are mostly placed remotely and their forces are transmitted through cables. Even though they can achieve full mobility, increasing the number of actuators significantly affects their cost and portability.

Linkage-based devices can be simplified in terms of the number of actuators with different kinematical structures. Mitten devices open/close the hand in a unique, repetitive way by coupling index, middle, ring and little fingers physically~\cite{Wei2013, Wu2010, Gasser2015, Lince2017, Troncossi2012}. Controlling the hand with $1$ or $2$ actuators simplifies the design and decreases the overall cost, but limit the mobility and task adjustability.

Coupled devices interact with user's finger from multiple points and move finger joints together with a ratio adjusted by mechanical links or differential system (Figure~\ref{fig:coupled}). Such mechanisms can control finger movements with $1$ actuator\cite{Cempini2013, Weiss2013, DiCicco2004, Fu2007, Wege2005, Tong2010, Chiri2009, Rahman2012, Tang2011, Allotta2015, Cui2015, Geo2016, Lee2015, Lu2016} or $2$ actuators~\cite{Yamaura2009, Burton2011, Fontana2009, Wei2017, Cortese2015, Kim2017, Agarwal2015, Ferguson2018}. Compliant devices couple finger joints through compliant elements~\cite{Arata2013}, artificial muscles~\cite{Ryu2008} or soft actuators~\cite{Polygerinos2015, Yap2015, Maeder-York2014} instead of rigid links. Their coupling ratio is set by the mechanical stiffness of these soft elements. They are low-cost, but suffer from mandatory mechanical adjustments to change the finger synergies.

Unlike coupled devices, fingertip devices interact with user's finger from a single point and control the fingertip position regardless how finger joints move (Figure~\ref{fig:fingertip_coupled})~\cite{Iqbal2011, Gosselin2005, Bouzit2002, BenTzvi2015, Fang2009, Sun2009, Sarakoglou2016}. Each finger component is controlled using a single actuator, so they are low-cost, easily wearable and portable. Not having strict mechanical connections around every finger phalange allows users to adjust tasks within the limits of their abilities. However, they cannot impose strict finger synergies, limit spastic movements for patients with disabilities or convey realistic information about virtual interactions.

Finally, underactuated devices based on contact forces control multiple finger joints with a single actuator by adjusting forces acting on finger phalanges automatically based on interaction forces, thanks to passive elements along the mechanism (Figure~\ref{fig:underactuated})~\cite{Hocaoglu2009, Sarac2016}. Each finger component is controlled using a single actuator, so they are low-cost, lightweight and portable. Passive elements along the mechanism ensures the device to be worn easily. Even though the actuator does not control the joints implicitly, alternative control strategies can improve the trajectory following tasks because they have multiple interactions for each finger (see Section~\ref{sec:operation}).

The kinematics of a generic exoskeleton should be consistent with the desired finger mobility. Full finger mobility can be achieved with linkage-based devices with independent control. Alternatively, finger joints can be coupled with underactuated linkage-based devices based on contact forces. Doing so, a single actuator controls a single finger component while adjusting the operation for different tasks automatically.

%While developing a generic exoskeleton, designers should eliminate kinematics strategies that couple finger and finger joint movements in a strict way to allow users adjust their movements. A generic device can be designed to control each assisted joint individually, but the actuators should be carefully chosen to be simpler and cheaper. Alternatively, underactuation based on interaction forces is promising thanks to their automatic adjustability, and reduced number of actuators, which improves the overall weight and cost, while complex control strategies are required for strict joint control tasks

Kinematics selection should be made based on mobility. The designer can adopt glove-based or linkage-based exoskeletons for independently controlled finger components. Despite their bulky and expensive design, they will achieve high performance for strict trajectory following tasks. Furthermore, mechanical and finger joints must be aligned carefully to ensure user's safety and efficacy of applied forces.

% \vspace*{-0.5\baselineskip}
\subsubsection{Strategies for adjusting to different hand sizes} \label{sec:q7}

The society has a wide range of hand sizes~\cite{Buryanov2010}, and a hand exoskeleton should operate correctly and comfortably for all users~\cite{Morel2012}. Exoskeletons with a single interaction point~\cite{Gosselin2005, Sarakoglou2016, Fiorilla2008} can neglect such variety, since they control the fingertip pose without imposing strict trajectory for finger joints. For exoskeletons with multiple interaction points, several adjustment strategies can be found in the literature:

% \vspace*{0.3\baselineskip}
\textbf{Alignment strategies} require mechanical and finger joints to be aligned, such that the exoskeleton can fit on user's hand accurately, and actuator forces can be mapped into perceived ones directly. The first alignment strategy is to manufacture a custom exoskeleton for each user individually~\cite{Delph2013, Hasegawa2008, Burton2011, Weiss2013, Cui2015}. A custom exoskeleton must be designed with variable link lengths corresponding to user's hand size. Such an exoskeleton must be manufactured individually, so the user must agree to purchase it for personal use. Due to the lack of mass production, the overall cost of the device is expected to be high. Even though this strategy might be suitable for assistive or haptic applications, it is not applicable for clinical use, where a single device is expected to serve for multiple patients in a day.

Alternatively, an exoskeleton can align mechanical and finger joints through adjusted mechanical connections and links~\cite{Popov2017, Aubin2013, Decker2017, Brokaw2011, Bouzit2002, Iqbal2011, DiCicco2004, Fang2009, Wang2009, Jones2010, Li2011, Polotto2012, Taheri2014, Fu2007, Wege2005, Tong2010, Geo2016, Agarwal2013, Wei2017, Agarwal2015, Wei2013}. The user wears the device before operation and a technician fixes a slider-screw system for fitting. Even though it requires a crucial preparation process, the exoskeleton can fit all users in the end. The constant need for a technician's presence might make such an exoskeleton suitable for clinical settings more than home therapy.

% \vspace*{0.3\baselineskip}
\textbf{No-Alignment Strategies} accept the misalignment between mechanical and finger joints, and address the issue of hand sizes in other ways. Increasing the compliance of the actuator~\cite{In2010, Connelly2010, Li2017, Polygerinos2015, Yap2015, Ryu2008, Wu2010} or the mechanical links~\cite{Jo2014} transmits lower interaction forces, hence minimizes the after effects of misalignment. However, the output forces might be insufficient for certain rehabilitation or assistive applications.

A hand exoskeleton can be designed in small, medium and large sizes, such that the misalignment between mechanical and finger joints can be limited~\cite{Mulas2005, Kobayashi2012, Lee2013, Jo2017, Arata2013, Sun2009, Gasser2015, Rahman2012, Lu2016}. Even though misalignment are not prevented, they are ensured not to harm users. Finally, a designer can place passive joints along the mechanical structure to turn additional loads, which are caused by misalignment, into motion~\cite{BenTzvi2015, Sarac2016, Ertas2014, Chiri2009, Cempini2013, Tang2011, Allotta2015, Lee2015, Fontana2009, Yamaura2009, Cortese2015, Kim2017, Troncossi2012, Lince2017, Ferguson2018}. Such an exoskeleton adapts its behavior for different hand sizes automatically. Designing sized exoskeletons and passive joints are the best practices for generic exoskeletons, thanks to their usability and preparation time. Furthermore, since they can be mass produced, they can be low-cost.

The designer can couple finger joints using contact based underactuation, such that a single actuator moves multiple finger joints while passive elements adjust the operation based on interaction forces acting on finger phalanges. Thanks to the automatic adjustability, the mechanism can be simplified significantly. Furthermore, passive elements ensure users' safety during operation. However, they require complex control strategies to achieve high tracking performance.

%\vspace*{-0.5\baselineskip}
\subsubsection{Mechanism Placement} \label{sec:q8}

\begin{figure*}[h!]
\centering
% \vspace*{1\baselineskip}
\subfigure[Palmar device~\cite{Bouzit2002} \label{fig:palmar}]%
{\includegraphics[width=0.25\textwidth]{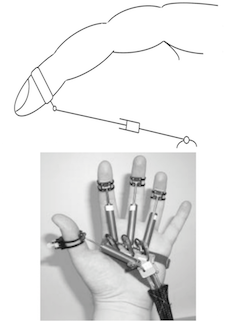}} \hspace{0.04\textwidth}
\subfigure[Lateral device~\cite{Hasegawa2011} \label{fig:lateral}]
{\includegraphics[width=0.25\textwidth]{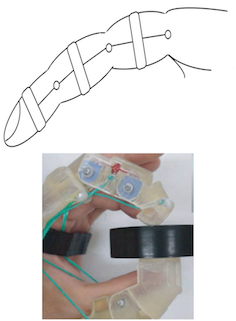}} \hspace{0.04\textwidth}
\subfigure[Dorsal device~\cite{Iqbal2015} \label{fig:dorsal}]%
{\includegraphics[width=0.25\textwidth]{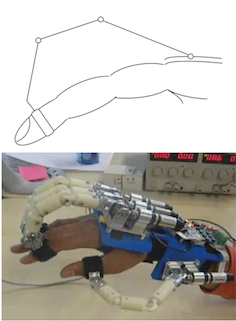}}
% \vspace*{-.5\baselineskip}
\caption{Hand exoskeletons can be designed to placed on different locations with respect of fingers. }
\label{fig:placement}
\vspace*{1\baselineskip}
\end{figure*}

Finally, the designer should device where to place the finger components with respect to the fingers. This design selection is especially important for linkage-based exoskeletons, such that transmission units can be placed on dorsal, lateral or palmar side of fingers (see Figure~\ref{fig:placement}).

% \vspace*{0.3\baselineskip}
\textbf{Palmar devices} consist of mechanical or transmission components placed inside the palm of the hand (see Figure~\ref{fig:palmar})~\cite{Bouzit2002, Connelly2010, Delph2013, Lee2013}. Unfortunately, they prevent users to get in touch with real objects for assistive use.
%do not suffer from possible collisions between finger components, but

% \vspace*{0.3\baselineskip}
\textbf{Lateral devices} consist of mechanical or transmission components placed on both sides of finger phalanges (see Figure~\ref{fig:lateral})~\cite{Jones2010, Hasegawa2008, Polotto2012, Weiss2013, Cempini2013, Cortese2015, Wei2013, Gasser2015, Mulas2005}. Finger joints can be rotated independently through cable transmission or remote center of motion (RCM). These devices free the palm of the hand for future interactions in the real environment. However, they might suffer from possible collisions for multi-finger implementations, especially when abduction/adduction of MCP is allowed. Compared to other options, lateral devices might be harder to be worn by patients with disabilities, so their use for rehabilitation or assistive should be reconsidered.

% \vspace*{0.3\baselineskip}
\textbf{Dorsal devices} consist of mechanical or transmission components placed on top of the finger phalanges (Figure~\ref{fig:dorsal})~\cite{Sarac2016, Wu2010, Troncossi2012, Lince2017, Fontana2009, Yamaura2009, Burton2011, Agarwal2013, Wei2017, Kim2017, Agarwal2015, DiCicco2004, Fu2007, Wege2005, Tong2010, Chiri2009, Rahman2012, Tang2011, Allotta2015, Cui2015, Geo2016, Lee2015, Lu2016, Wang2009, Li2011, Taheri2014, Gosselin2005, Iqbal2011, BenTzvi2015, Fang2009, Sun2009, Sarakoglou2016, Arata2013, Yap2015, Brokaw2011, Fiorilla2008, Li2017, Kobayashi2012, Kobayashi2013, Jo2014, Aubin2013, Decker2017, Jo2017, Cempini2013, Weiss2013, Lee2013, Connelly2010, Delph2013, Polygerinos2015, Hasegawa2008, Mulas2005, Popov2017, Gasser2015, In2010, Jones2010, Polotto2012, Maeder-York2014, Ferguson2018}. Doing so, the collision between multiple finger components can be minimized while user's palm is free for future interactions with real objects. They do not possess any strong limitation regarding the number of finger components to be manufactured or the performance, and can be used for all possible target applications.

Regardless the placement of finger components, linkage-based exoskeletons are attached to user's fingers through rings or flexible attachments. Since there is no recorded impact of mechanism placement on perception during finger opening/closing, we can assume that actuator forced can be distributed around finger phalanges naturally.

\subsection{Actuation}

\begin{figure}[h!]
  \centering
  %\vspace*{-1\baselineskip}
  \resizebox{4.5in}{!}{\includegraphics{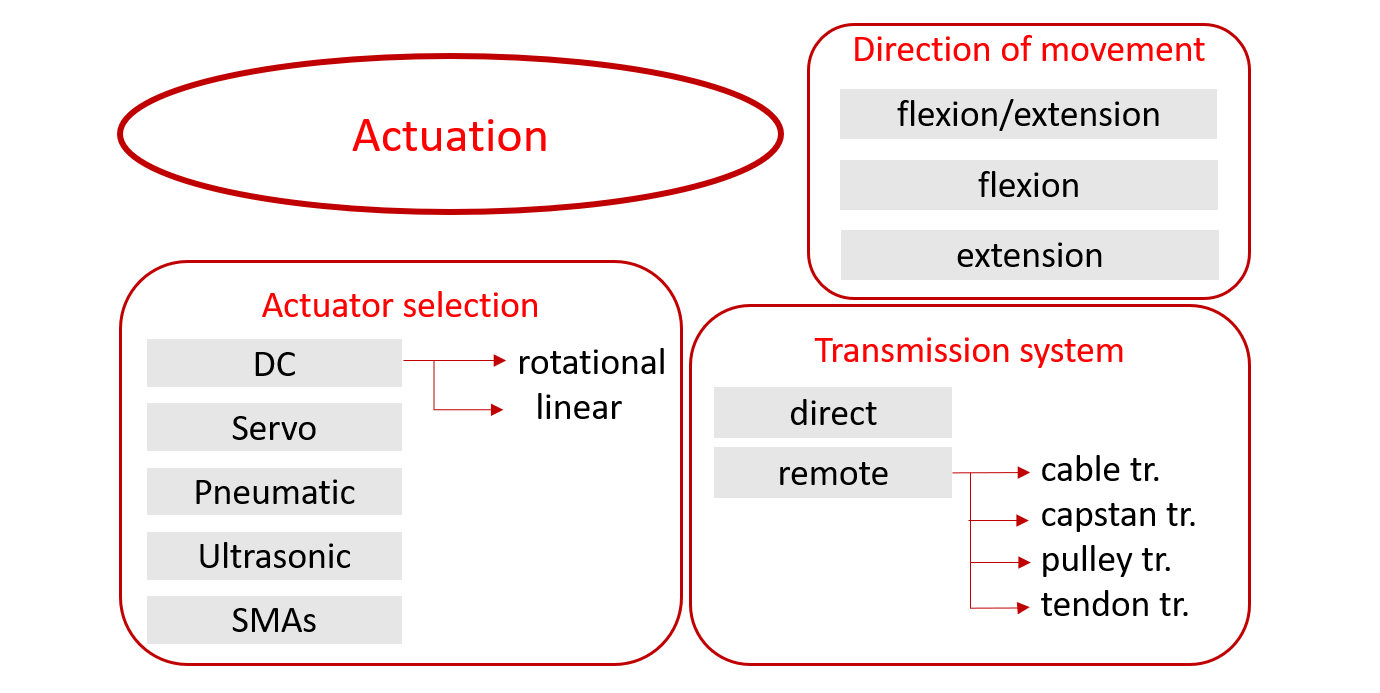}}
%   \vspace*{-.5\baselineskip}
  \caption{Possible design choices for actuation technologies based on actuator selection, transmission system and direction of movement.}
%   \vspace*{1\baselineskip}
  \label{fig:actuation}
\end{figure}

An exoskeleton can assist/resist user's fingers through actuator and transmission technologies. In this section, we will investigate the exoskeletons in the literature based on actuator selection, direction of movement and transmission system from the perspective of achieving generic exoskeletons (see Figure~\ref{fig:actuation}).
%~\cite{Heo2012, Gopura2009}.

%\vspace*{-0.5\baselineskip}
\subsubsection{Actuator type} \label{sec:q10}

Even though there are some exceptional studies that apply assistance based on wrist activity~\cite{Bortoletto2017} or resistance using springs~\cite{Brokaw2011}, most of the exoskeletons move user's fingers through active manipulation. Such a manipulation can be achieved through different actuator types.

\vspace*{0.3\baselineskip}
\textbf{DC motors} are the most popular technology since they are highly available in the market, reliable and easily controllable. Linear movement can be achieved using linear DC motors~\cite{Jo2017, Arata2013, Taheri2014, Tong2010, Rahman2012, Cui2015, Sarac2016, Jo2014} or rotational DC motors with linear sliders~\cite{Chiri2009, Cortese2015}. Then, rotational movement can be achieved using brushed motors~\cite{Popov2017, Decker2017, Lee2013, Fiorilla2008, Gosselin2005, Iqbal2011, BenTzvi2015, Sarakoglou2016, Wang2009, Jones2010, Li2011, Polotto2012, Fu2007, Wege2005, Weiss2013, Cempini2013, Geo2016, Lee2015, Lu2016, Fontana2009, Wei2017, Troncossi2012, Kim2017, Hasegawa2008, Hocaoglu2009, Popov2017, Ferguson2018} or brushless motors~\cite{Fang2009, Wei2013, Gasser2015, Lince2017, In2010}. Linear motors are simpler to be placed on top of the hand for coupled finger opening/closing, while rotational motors are mostly backdriveable and provide unlimited movement. Furthermore, brushed motors have low-cost, simple wiring, compact design and easy control but require maintenance, cause vibration and lose torque in high speeds due to friction.

\vspace*{0.3\baselineskip}
\textbf{Servo motors} can be defined as rotational DC motors with a limited workspace~\cite{Delph2013, Mulas2005, Aubin2013, Yamaura2009, Allotta2015}. They are fast, and can achieve high output torque and accurate position control; but require a special driving circuit for control and have higher cost compared to DC motors. 

\textbf{Ultrasonic motors (USMs)} can also be defined as rotational DC motors powered by ultrasonic vibration~\cite{Tang2011}. They are silent, light weight and efficient in terms of output force, but they suffer from hysteresis and temperature increase over time.

\vspace*{0.3\baselineskip}
\textbf{Pneumatic actuators} control the hydraulic or air flow through compressors, using pneumatic cylinders~\cite{Bouzit2002, DiCicco2004, Burton2011}, air balloons~\cite{Li2017}, hydraulic pump~\cite{Ryu2008}, air bladder~\cite{Connelly2010}, flexible thermoplastic fabrics~\cite{Yap2015}, soft actuation~\cite{Polygerinos2015} or pneumatic artificial muscles~\cite{Sun2009, Agarwal2013, Wu2010, Maeder-York2014}. They can achieve high, adjustable force and speed at low-cost. The size of the compressor and its storage lead the exoskeletons to be controlled remotely. Even though pneumatic actuators are not necessarily compliant, they consequently increase the overall compliance as mentioned in Section~\ref{sec:q7}.

\vspace*{0.3\baselineskip}
\textbf{Shape memory alloy actuators (SMAs) }use deformation of materials upon heating and cooling at critical temperatures~\cite{Kobayashi2012, Kobayashi2013}. Even though they have high power-to-weight ratio, their output motion is hysteresis, highly nonlinear and saturated. As a result, their control is challenging~\cite{Kumar2008}.

Actuation types do not possess strong limitation about applications or tasks. Therefore, any actuator type can be selected for a generic exoskeleton as long as they are low-cost, easily controllable and effective in terms of output forces.
%Since mechanically attaching actuators to the finger components is highly preferable to improve the portability, they should also be small and lightweight.

%\vspace*{-0.5\baselineskip}
\subsubsection{Transmission units} \label{sec:q9}

The actuators should be connected to the mechanical structure through alternative transmission strategies. The simplest transmission scenario is designing a direct-drive system, such that the actuators are placed on top of the hand or along the mechanism, while the actuator shafts are attached to mechanical components directly~\cite{Sarac2016, Hocaoglu2009, Iqbal2011, Troncossi2012, Arata2013}. Even though direct-drive is preferable to improve the portability, the chosen actuators should be highly miniaturized and lightweight.

If the chosen actuators are big and heavy, they should be located away from the exoskeleton and their forces should be transmitted remotely through cables~\cite{Cempini2013, Wege2007, Delph2013, Arata2016, Chiri2009, Agarwal2015, Li2011, Wang2009, Fu2007, Kobayashi2012, Hasegawa2008, Lee2013, Li2011, Aubin2013, Jones2010, Polotto2012, Taheri2014, DiCicco2004, Wege2005, Weiss2013, Allotta2015, Lu2016, Yamaura2009, Agarwal2013, Ferguson2018}, capstan systems~\cite{Fontana2009, Ertas2014, Decker2017}, tendons~\cite{In2010, Popov2017, Cortese2015, Wei2013, Gasser2015}, or pulleys~\cite{Geo2016}. Even though choosing big actuators can create high output forces for the exoskeleton, remote transmission limit the workspace of users.

Both transmission strategies can be implemented for all application types. The designers should make the selection based on the actuator decision.

%\vspace*{-0.5\baselineskip}
\subsubsection{Direction of movement} \label{sec:q11}

Even though the majority of actuators are bidirectional, certain rotational DC motors and pneumatic motors are not. If the chosen actuator is unidirectional, then the designer should decide how to use them for finger movements. One approach is to assist user's fingers in one direction actively, and to leave the other direction passive. The active assistance can be used either to open the finger for rehabilitation~\cite{Chiri2012, Brokaw2011, Agarwal2013, Connelly2010}, or to close the finger for assistive use~\cite{In2010, Lince2017, DiCicco2004}. Devices with active flexion cannot be used for haptic use due to the lack of resistive forces, while devices with active extension cannot be used for assistive use due to the lack of assistive forces. This is why leaving one direction passive cannot be chosen for a generic exoskeleton, even though they provide simple and effective solutions for specific tasks.

The second approach is to achieve bidirectional movement using multiple actuators and transmission units~\cite{Fu2007, Weiss2013, Cempini2013, Yamaura2009, Gasser2015, Wang2009, Jones2010, Li2011, Sun2009}. Bi-directional movements can be adopted for all target applications with no specific limitations. Achieving bi-directional movements might make exoskeletons bulkier and more expensive due to the increased number of actuators. Even though choosing bidirectional actuators is the best choice for generic exoskeletons, the designer should equip multiple actuators if unidirectional actuators are chosen for a specific purpose.

%Obviously, choosing a bi-directional actuator would be more advantageous and simpler, but if the designer needs to choose a single-directional actuator, bi-directional movement can still be achieved.

\vspace*{-0.3\baselineskip}

\subsection{Operation strategies} \label{sec:operation}

\begin{figure}[t!]
  \centering
  %\vspace*{-.5\baselineskip}
  \resizebox{4.5in}{!}{\includegraphics{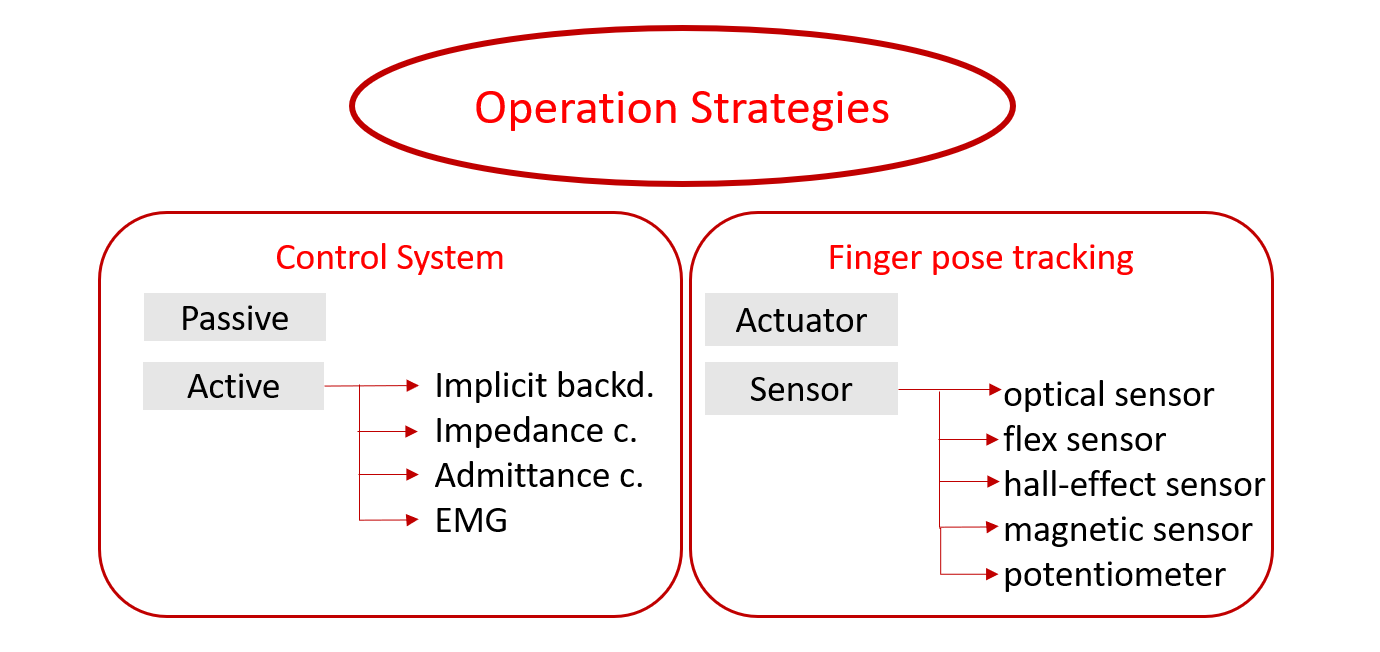}}
%   \vspace*{-.5\baselineskip}
  \caption{Possible design choices for operational strategies based on control and finger pose tracking strategies.}
  \label{fig:operational}
%   \vspace*{-1.25\baselineskip}
\end{figure}

The design of a hand exoskeleton is completed once the mechanical structure is equipped with actuators and transmission units. Then the designer should decide how to control and track user's fingers during operation (see Figure~\ref{fig:operational}).

%\vspace*{-0.5\baselineskip}
\subsubsection{Control} \label{sec:q12}

Control strategies for existing hand exoskeletons can be categorized mainly as active and passive, based on how much users participate to the task~\cite{Jeong2013}.

\vspace*{0.3\baselineskip}
\textbf{Passive control strategies} control the exoskeleton to follow a strict trajectory or to reach a specific target. As the device leads their fingers, the user is asked to obey the movement. The control strategies can be designed based on position~\cite{Iqbal2011, Fu2007, Rahman2012, Connelly2010, Cui2015, Kim2017, Jo2017, Troncossi2012, Popov2017, Allotta2015, Weiss2013, Arata2013, Gasser2015, Wang2009, Sarac2016, Yamaura2009, Wege2005, Polygerinos2015, Yap2015, Burton2011, Li2017, Bouzit2002, Wu2010, Ryu2008, Cempini2015, Sun2009, Ferguson2018} or velocity~\cite{Jo2014, Cempini2015}. Even though passive exercises can be used to treat disabilities of patients during rehabilitation, they might cause patients to lose interest during long, intense therapy sessions. They can be used for assistive applications as long as they are triggered by an external state, such as an additional sensor or a condition satisfied by an arm exoskeleton. However, they are impractical for haptic use.

\vspace*{0.3\baselineskip}
\textbf{Active control strategies} control the exoskeleton to assist/resist user's fingers based on user's performance as they are in charge of following a trajectory or reaching a target. One way to achieve active control is to adopt implicit backdriveability, which requires actuation, transmission and mechanical units to be chosen accordingly. With implicit backdriveability, the user can move their fingers freely even if the exoskeleton is attached to their fingers with no control~\cite{Hocaoglu2009, Wang2009, Jones2010, Li2011, Polotto2012, Taheri2014, Chiri2009, Aubin2013, Tong2013, Decker2017, Connelly2010, Ryu2008, Arata2013}. The backdriveable devices can be controlled with passive strategies when the user fails to keep their performance within a predefined range.

Implicit backdriveability cannot be achieved if the mechanical and actuator components of the designed exoskeleton require high backdrive forces or cause high friction. If so, backdriveability can be achieved actively using force control techniques based on impedance~\cite{Wei2017, Agarwal2015} or admittance~\cite{BenTzvi2015, Sarac2018, Kobayashi2012, Kobayashi2013, Sarakoglou2016, In2015, Gosselin2005, Sun2009, Ferguson2018, Jo2014}. These techniques require additional force sensors to be included for the exoskeleton, such that user's intention to move can be measured and be used as a control reference for the exoskeleton. In either case, backdriveability can easily be used by all target applications and improve user's safety during operation.

Furthermore, user's intention can be detected through additional sensors, such as electromyography (EMG) sensors~\cite{Mulas2005, Delph2013, Arata2013, DiCicco2004, Wege2007, Tong2013, Troncossi2012, Lince2017} or active bioelectric potential electrodes~\cite{Hasegawa2011}. Then, these measurements are used to create a control reference for passive control strategies in an online manner. Similarly, bilateral teleoperation tasks can be developed by controlling the device passively to follow the reference set by user's other hand~\cite{Yamaura2009, Fu2007}. These assist-as-needed or bilateral control strategies are useful for rehabilitation or assistive applications but their use for haptics is out of context.

A generic hand exoskeleton should be backdriveable with or without control, depending on the actuator selection. Additional control strategies might be used for different target applications.

%\vspace*{-0.5\baselineskip}
\subsubsection{Finger pose estimation}

A generic hand exoskeleton must track user's finger movements efficiently during operation. The exoskeletons in the literature adopt various strategies to track finger movements, mostly depending on mechanical and actuation choices.

\vspace*{0.3\baselineskip}
\textbf{Actuator displacements} reveal the finger pose directly with high quality for the exoskeletons with independent finger control~\cite{Brokaw2011, Wang2009, Li2011, Jones2010, Hasegawa2011, Polotto2012, Taheri2014, Aubin2013, Agarwal2013, Fiorilla2008}. Similarly, coupled exoskeletons with constant joint ratio between joint rotations track finger movements using the actuator displacements and this ratio~\cite{In2010, Mulas2005, Jo2017, Arata2013, Fang2009, DiCicco2004, Tong2010, Rahman2012, Tang2011, Cempini2013, Allotta2015, Geo2016, Lu2016, Fontana2009, Yamaura2009, Burton2011, Wei2017, Cortese2015, Kim2017, Troncossi2012, Gasser2015, Lince2017}. Using actuator displacements directly result in simple operational strategies and high quality tracking performance.

\vspace*{0.3\baselineskip}
\textbf{Additional sensors} are needed for exoskeletons with other kinematics selections, when the actuated joints are not directly mapped into finger joints. For glove-based exoskeletons, flex sensors are placed along user's finger joints to measure the finger pose directly~\cite{Wei2013, Popov2017, Connelly2010, Li2017, Kobayashi2012, Jo2014, Decker2017, Ryu2008}. Such flex sensors are low-cost, lightweight and of high quality. Since these flex sensors should be grounded along finger joints to measure the finger pose directly, they require a texture-based interface.

Furthermore, additional sensors can track user's finger movements when inserted along mechanical joints, which are aligned with finger joints. These sensors can be chosen among hall-effect sensors~\cite{Fu2007, Wege2005, Weiss2013, Fiorilla2008} or potentiometers~\cite{Lee2015}. The alignment between mechanical and finger joints measure the finger pose directly, so the measurements are quick and of high quality, while the sensors are mostly low-cost and lightweight. However, such direct pose tracking can be implemented only for exoskeletons with RCM mechanisms.

Non-contact optical~\cite{Lee2013, Kobayashi2012, Arata2013, Burton2011, Chiri2012} or magnetic sensors~\cite{Sun2009} require markers to be placed on finger phalanges or finger joints.  They can be applied only if the exoskeleton allows for these markers to be placed on user's fingers without optical interface. It is important to note that in case of interference, the continuity of finger pose might be disturbed.

The sensor implementations discussed above require certain kinematics decisions. If a hand exoskeleton does not satisfy any of these properties, forward kinematics computation can be used to estimate the finger pose using additional sensors attached along random mechanical joints. Hall-effect sensors~\cite{Bouzit2002, Wu2010, Wege2005}, optical encoders~\cite{Gosselin2005, BenTzvi2015}, magnetic encoders~\cite{Iqbal2011, Sarakoglou2016, Agarwal2015} or potentiometers~\cite{Sarac2017} can be used for such measurements. Even though the speed and efficacy of finger pose tracking depend on the quality of sensors and capabilities of control board, they can be implemented basically for all kinematics choices and target applications.

 For devices with independent joint control, the actuator displacements measure the finger pose directly. For underactuated devices, additional sensors and forward kinematics are needed to estimate the finger pose.

\vspace*{-.5\baselineskip}

\newpage

\chapter{Design of the Hand Exoskeleton} % Main chapter title

\label{sec:Chapter3} % For referencing the chapter elsewhere, use \ref{Chapter1}

\lhead{Chapter 3. \emph{Design}} % This is for the header on each page - perhaps a shortened title

Detailing the state-of-the-art exoskeletons based on their mobility, mechanical design, actuation and operation strategies helps a designer to understand the advantages and the disadvantages of possible design properties. Designing a new hand exoskeleton should start with highlighting the target application, listing the design requirements to be achieved, and analyzing the advantages/disadvantages of each possible design selection detailed in Section \ref{sec:Chapter2}.

This chapter presents the design decisions for the proposed generic hand exoskeleton. Once the design requirements of a new device are revealed, the first kinematic sketch for a finger component will be presented. Furthermore, some of the presented design properties, such as the adjustability of the operational performance with respect to different object shapes and sizes are presented through CAD simulations. 

After the first kinematics design, various pose analyses are needed to be performed in order to study the mechanism performance and operation. First, the inverse kinematics analysis will be performed to study whether the mechanical limits are satisfied for natural RoM of finger joints. Then, forward kinematics analysis will be presented using numerical and analytical approaches to obtain finger pose estimation thanks to an additional measurement tool. These analyses can be transformed to differential kinematics to form a relation between the finger and actuator velocities, as much as actuator forces and finger joint torques. Finally, an optimization algorithm will be performed to obtain the mechanical link lengths, which allows the user to reach their natural finger RoM, maximizing force transmission from actuator to finger joints.

\newpage

\newpage
\section{Design Requirements}

In Chapter~\ref{sec:Chapter2}, hand exoskeletons in the literature were detailed to understand common and individual properties based on their actuation, mechanics and working principles. The wide literature search proved that each choice regarding the previously specified classification items ensure different advantages and drawbacks of the design, considering different application scenarios. Yet, the search for the best hand exoskeleton that can be used for all types of applications is not over. 

The complexity of hand anatomy, discussed in Section~\ref{sec:anatomy} prevents a unique list of design features to be achieved for an exoskeleton design. Furthermore, recent findings of clinical studies, technological developments about sensors, actuators and electronic equipment or recent developments of hand exoskeletons in general might affect main requirements and motivations of creating a new device.

First of all, a hand exoskeleton can be worn by users with impaired or unimpaired hand. Such an interaction introduces a set of generic requirements for a hand exoskeleton, without specifying the target application:

\begin{itemize}
\item \textbf{Effective force transmission:} As all human-interacted devices, the most important and mandatory specification for a hand exoskeleton system is to ensure user's safety throughout the operation. Such safety measures should be defined especially for patients with disabilities since they are more sensitive and vulnerable than healthy users. The overall system should be stable at all times, which might require additional precautions by control algorithms. Furthermore, forces between actuators and user's finger joints should be transmitted in a safe and efficient manner. The mechanism should be designed by strictly respecting the natural RoM of finger joints, such that the mechanism does not impose further reference to finger joints. Rotations along finger joints can be performed by imposing forces to finger phalanges, which should be natural at all times, not to cause any discomfort to users. Finally, while controlling multiple finger joints, torques being applied to finger joints should always have a balanced ratio not to cause finger pain at any orientation during operation.

\item \textbf{Adjustability for the finger sizes:} Section~\ref{sec:anatomy} showed that the society has a wide range of variety regarding finger sizes. It is mandatory for a hand exoskeleton to adjust its operation respectively for different hand sizes. Devices, which align mechanical actuated joints with finger joints, have to create adjustable link lengths to cope with size variations, such that they do not cause pain or unnatural movements to finger joints. Alternatively, such adjustment can be made automatically through misalignments between mechanical and finger joints intentionally and utilizing passive joints along the mechanism to improve the adjustability for different finger sizes.

\item \textbf{Complexity of the hand movements \& diversity of the tasks:} A generic hand exoskeleton should assist user's fingers to perform ADLs as much as basic joint movements. Certain rehabilitation tasks might require simple, repetitive flexion/extention movements of finger joints in order to improve finger mobility and RoM or to strengthen finger muscles. Conversely, other types of rehabilitation exercises might require users to perform ADLs in order to improve the cognitive abilities of patients with disabilities. A hand exoskeleton should be able to apply both assistance or assistance to patients with various needs to accelerate treatment process. Meanwhile, a hand exoskeleton should be able to assist users to perform various tasks in a real environment and to resist users to mimic an interaction during virtual environment tasks. Furthermore, a hand exoskeleton should not be designed only for grasping a certain object with a certain diameter, but the device's operation should be adjustable based on required motion. Such adjustability might involve both the independent mobility of fingers and finger joints for each task.

\item \textbf{Wearability of the device:} A generic hand exoskeleton should be worn by users with impaired and unimpaired hand, which leads the designer to consider the wearability of such a device based on the capabilities of users with least free mobility. Post-stroke patients might exhibit residual contraction forces to close their fingers and difficulties to open their hands. In order to simplify the wearing and preparation process for those users, the device fitting would require to attach the exoskeleton to subject's fingers with no certain initial pose requirement. Avoiding stiff connections could make the fitting easier and less painful while improving the flexibility. With this motivation, the hand exoskeleton should be designed to be attached along the user's phalanges at any pose of the human hand. Creating an easily wearable device with no restriction for the initial pose not only simplifies the preparation time for patients with disabilities, but also for healthy users.

 \item \textbf{Portability:} The mechanical device should be designed with its actuators and power transmission components in a lightweight and efficient manner, without causing fatigue. There are some stationary devices in the literature to address rehabilitation exercises, but they might suffer from unnatural behavior of therapy exercises. While designing a hand exoskeleton, it is crucial to find a solution to increase the portability of the device without increasing the fatigue and overall weight of the device.
\end{itemize}

A hand exoskeleton should satisfy these requirements without defining the purpose of use. Nevertheless, narrowing the purpose of use might define more specific requirements for a new hand exoskeleton. For instance, a hand exoskeleton to be used for physical rehabilitation should provide repetitive and useful physical tasks to patients with physical disabilities. The output forces of the device should be sufficient to overcome the disability level of patients, while the transmission of forces to finger phalanges possess less significance. Another important design criteria for a rehabilitation device is the easy wearability of device not to cause any trouble for users. However, a hand exoskeleton to be used for haptic purposes does not have limitations regarding wearability, but portability and light-weight of the device becomes more crucial so that users can move freely. Since the haptic device is used for force feedback, the transmission of forces to finger phalanges becomes more significant.

All additional properties can be achieved by modifying high-level choices for the kinematic architecture of device. These choices can be organized using classification items that were covered in Chapter~\ref{sec:Chapter2}. The motivation of the hand exoskeleton to be proposed in this thesis can be defined as assistive and rehabilitative applications, without restricting the possibility for haptic implementations. The characteristics of such a hand exoskeleton can be listed based on the previous classification items as following:

\begin{itemize}
\item \textbf{Hand Mobility:} The proposed hand exoskeleton should provide mobility to all fingers of user in order to complete various kinds of ADLs for haptics and assistance, and physical exercises for rehabilitation. In fact, assisting all fingers provide a natural operation for the user to open/close fingers for all application types. Such a mobility can be given by coupling finger joints together, or by transmitting forces to finger joints using a single actuator. Even though such opening/closing the fingers together might be useful for grasping a certain type of objects, some ADLs require the fingers to be moving independently, such as fingertip grasping or using a key. Allowing independent control of fingers require individual finger components mechanically, which utilize independent actuators, in order to increase the variety of different tasks to be performed by the device. Since index, middle, ring and little fingers are kinematically same, the same mechanical structure can be extended to all these fingers, optimizing link lengths of finger components based on finger lengths depicted in Table~\ref{tab:fingersize}. Unfortunately, the joint capabilities of thumb for abduction/adduction require a different mechanism to be designed specifically for it. Due to the complexity of thumb component, this study focuses on the development of other four finger components.

\item \textbf{Finger Mobility:} As mentioned before, a human finger has $4~DoF$ mobility, $3$ flexion/extension and $1$ abduction/adduction. Since most of the ADLs do not require fingers to perform abduction/adduction movements, a hand exoskeleton can be designed to ignore such mobility or allow users to move them passively. A healthy user cannot move his DIP joint independently from PIP joint due to an anatomical coupling between these two. Such natural coupling allows a hand exoskeleton to be simplified by leaving DIP joint free. Assisting $2~DoF$ flexion/extension, rotating MCP and PIP joints only, improves simplicity and wearability of the mechanism without sacrificing operational performance. The assisted joints need to be consistent with natural RoM stated previously in Table~\ref{tab:fingerrom} to ensure user's safety at all times. Even though each finger component aims to achieve $2~DoF$ mobility, actuating both joints independently might cause the device to be costly, bulky and heavy. Instead of controlling finger joints independently, they can be controlled by a single actuator in order to improve design simplicity, portability and affordability.

\item \textbf{Number of Connection Points:} A hand exoskeleton can control MCP and PIP joints for flexion/extension rotations in various ways. Even though embracing a $1~CP$ design, where the device is connected to middle phalange of user, can rotate both joints in a coupled manner using a single actuator, it cannot give sufficient support for both finger joints. Alternatively, a $2~CPs$ mechanism can be designed to connect the device from proximal and middle finger phalanges of users. Doing so, the mobility of finger phalanges are being limited by the mechanical device, even though they are not controlled independently. Leaving fingertip free allows the device to be possible merged with a tactile device while improving the perception of touch during virtual tasks. $2~DoF$ finger mobility by an exoskeleton with $2~CPs$ can be controlled either by utilizing differential mechanisms, which define a rotation ratio between finger joints, or further underactuation methods, which adjust finger movements based on interaction forces. Differential mechanisms to transmit actuator forces to finger phalanges define a strict rotational ratio between finger joints, which can be adjusted only by changing the link lengths of the mechanism or transmission system manually. Therefore, the task adjustment cannot be achieved automatically. For the proposed device, underactuation concept~\cite{Gosselin2003} based on contact forces was adopted by relating forces acting on finger phalanges to each other through linkage mechanism and passive joints. This property allows the device to apply stable forces to finger phalanges during grasping tasks using objects with any shape and size. Kinematics selection to achieve underactuation defines mechanical joints implicitly misaligned with finger joints, thanks to linkage mechanism and passive mechanical joints. Such a misalignment considers finger phalanges as a part of kinematic chain~\cite{leonardis2015emg}, which allows the device to adjust for various finger sizes automatically.

\item \textbf{Finger Pose Estimation:} The choice of embracing underactuation allows $2$ finger joints to be controlled using $1$ actuator without sacrificing from adjustability of achievable tasks, while suffering from the lack of sufficient sensory measurements. Such a lack of measurements prevent finger joints to be predicted during operation only using measurements about actuator activity. Since the proposed underactuation concept is based on contact forces instead of joint ratio, there is no constant relation between MCP and PIP joints to predict instantaneous finger pose. The lack of glove-based interaction between the user and the device abolishes the possibility to implement bending sensors. Alternatively, an additional encoder or a potentiometer can be inserted on a mechanical passive joint to provide extra sensory measurements to calculate finger joints through pose analysis.

\item \textbf{Actuation:} Controlling each underactuated finger component requires independent actuators to be used. Even considering a future thumb component, $5$ required actuators should be inserted on top of the hand in order to connect the actuator directly to finger component, removing the need for a transmission unit. Placing actuators on top of the hand improves the control accuracy and portability, while limiting possible actuators to be used, such that fitting all actuators in a compact manner requires the use of low-cost, light-weight linear DC motors. There are many compact linear DC motors in the market; some of them reach a linear displacement using the magnetic field within closed case. Despite of their miniature design, their magnetic based working principle requires a large gap between actuators not to cause magnetic interference. Alternatively, linear motors can be supported by mechanical gear boxes to amplify output forces, without affecting the environment. Even though utilizing gear boxes prevents backdriveability of actuators, they can apply sufficient force to open/close user's fingers.

\item \textbf{Mechanism Placement:} The finger components of the hand exoskeleton was proposed to be placed above fingers in order to keep the hand palm free for real grasping tasks and to minimize the width of mechanism for multi-finger implementations with no mechanical interference. Placing finger components above fingers allow actuators to be placed on top of the hand and the wrist, simplifying force transmission problem and increasing portability. Since the dorsal side of the hand has no physical limitation for the mechanical device, the height of the device can be extended in order to optimize its performance and reach higher RoM of finger joints, as long as it does not cause fatigue to users during operation.

\item \textbf{Control:} The hand exoskeleton can be controlled passively by a position control, where the actuator displacement is moved to follow its reference. Such control allows fingers to be opened/closed, while the underactuated mechanism flexes/extends finger joints in various modes. The hand exoskeleton can easily by controlled to open/close the fingers through muscular activity of the other hand, through teleoperation. Finally, the device can be controlled by the user in an active manner through force sensors and force control techniques. The mechanical gear boxes of selected actuators limit passive backdriveability, but the overall backdriveability of finger components can be ensured by introducing an individual force sensor for each finger component. The force sensor measures user's intention to open/close the finger and moves the finger component accordingly using force control methods.
\end{itemize}

The hand exoskeleton presented in this work ensures forces between user's phalanges and device to have a perpendicular direction at all times. This property significantly improves the design and functionality of the fasteners, which attach fingers to the exoskeleton. Since the fasteners are crucially used to transmit the tangential shear forces, there is no need to excessively tighten the finger for transmitting either torques or longitudinal forces to the skin. Doing so, the longitudinal forces and torques are prevented to be applied on finger phalanges by the mechanism. The absence of longitudinal forces allows the finger to be connected to the exoskeleton by simple straps around finger phalanges.

\newpage

\section{Proposed Hand Exoskeleton}

\begin{figure}[htb]
\centering
%\vspace*{-1.5\baselineskip}
\includegraphics[width=0.8\textwidth]{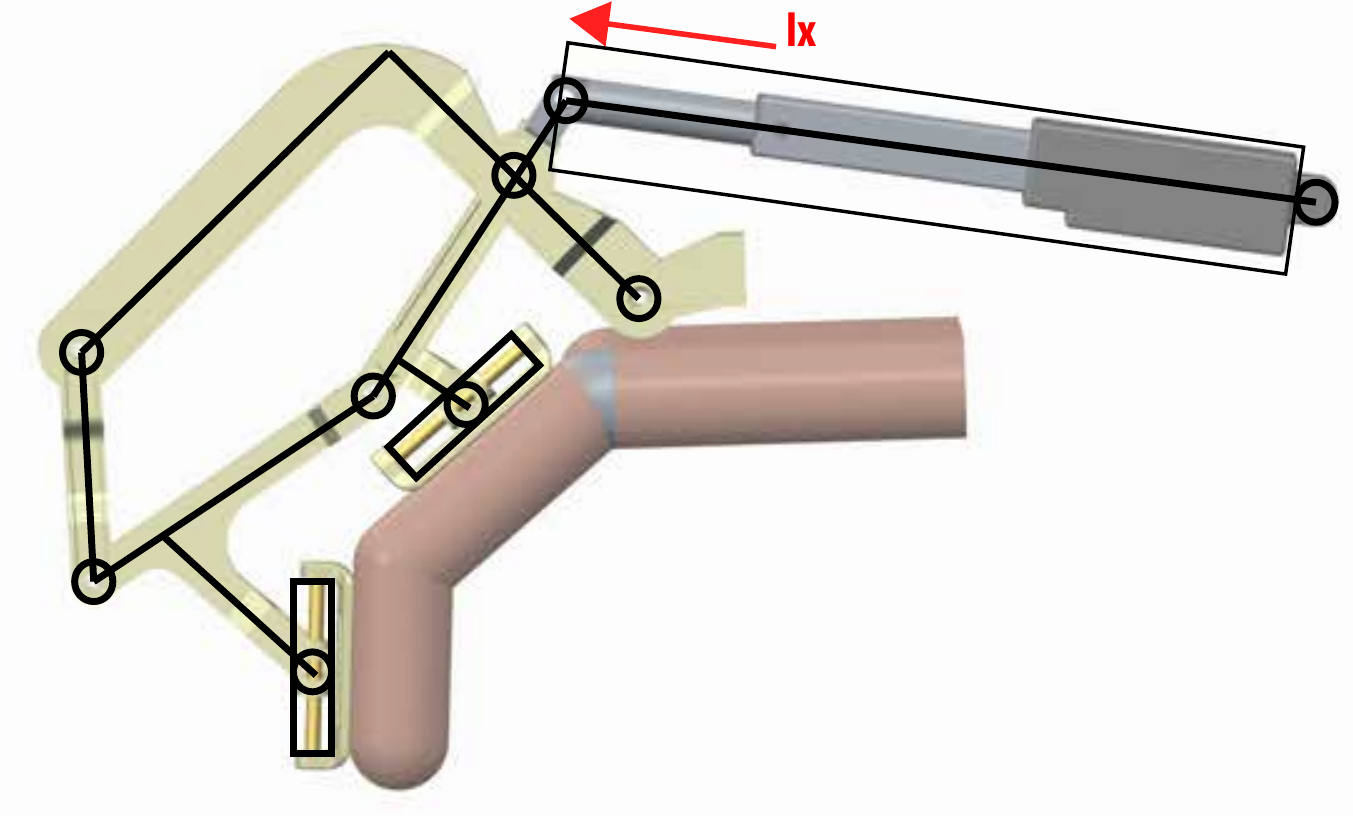}
%\vspace*{-0.5\baselineskip}
\caption{Kinematic selection for a finger component of the proposed hand exoskeleton under given assumptions drawn in CAD.}
\label{fig:model}
%\vspace*{-1\baselineskip}
\end{figure}

Each one of these properties either answers to a design issue or gets us closer to define the kinematics for a finger component of the proposed hand exoskeleton. Figure~\ref{fig:model} shows the sketch of the final kinematics. 

The kinematics structure is designed as an underactuation based linkage structure to ensure easy wearability and high adjustability for different users. Let's focus on the details of the underactuation concept in the next subsection. A linear DC motor allows the device to close/open the finger in a direct manner. The chosen actuators are small and compact in size, so that it can be placed on top of the hand, along with the mechanism itself. 

Placing the mechanical parts on the dorsal side of the finger allows user to contact with real objects. The whole structure is connected to the user's finger from the proximal and the middle phalanges to control the MCP and PIP joints. Mechanical connection between the device and the finger phalanges are performed through combining a cylindrical joint and a rotational joint with perpendicular axis, such that the lateral forces can be transmitted into motion and the perpendicular forces can be used to move the MCP and PIP joints.  

\newpage
\subsection{Underactuation}

Underactuation concept has been implemented on robotic gripper hands to allow a smaller number of actuators without decreasing the number of DoFs during grasping tasks. For mechanical grippers, passive spring-like elements~\cite{Rakic1989} or soft structure~\cite{Hirose1978} can transmit forces to the upcoming joints based on the interaction forces acting on the gripper links. Such a transmission makes the robotic hand to adjust its behavior to the object's size and shape automatically~\cite{chiang1998, Birglen2004}.

The idea behind underactuation is simple: the actuation is turned into motion at the first output joint, until the first mechanical link gets in contact with an object. After the first link reaches an object and starts feeling the interaction forces, the spring-like joint coupling between output joints allows the mechanism to transmit the same actuation for the second output joint. Actuator forces are transmitted to output joints until the grasping is completed. Using a single actuator to move multiple joints allows the device to achieve smaller size, less weight and lower cost, since actuators are typically the largest, heaviest and most expensive components of robotic devices.

\begin{figure}[b!]
\centering
\includegraphics[width=0.8\textwidth]{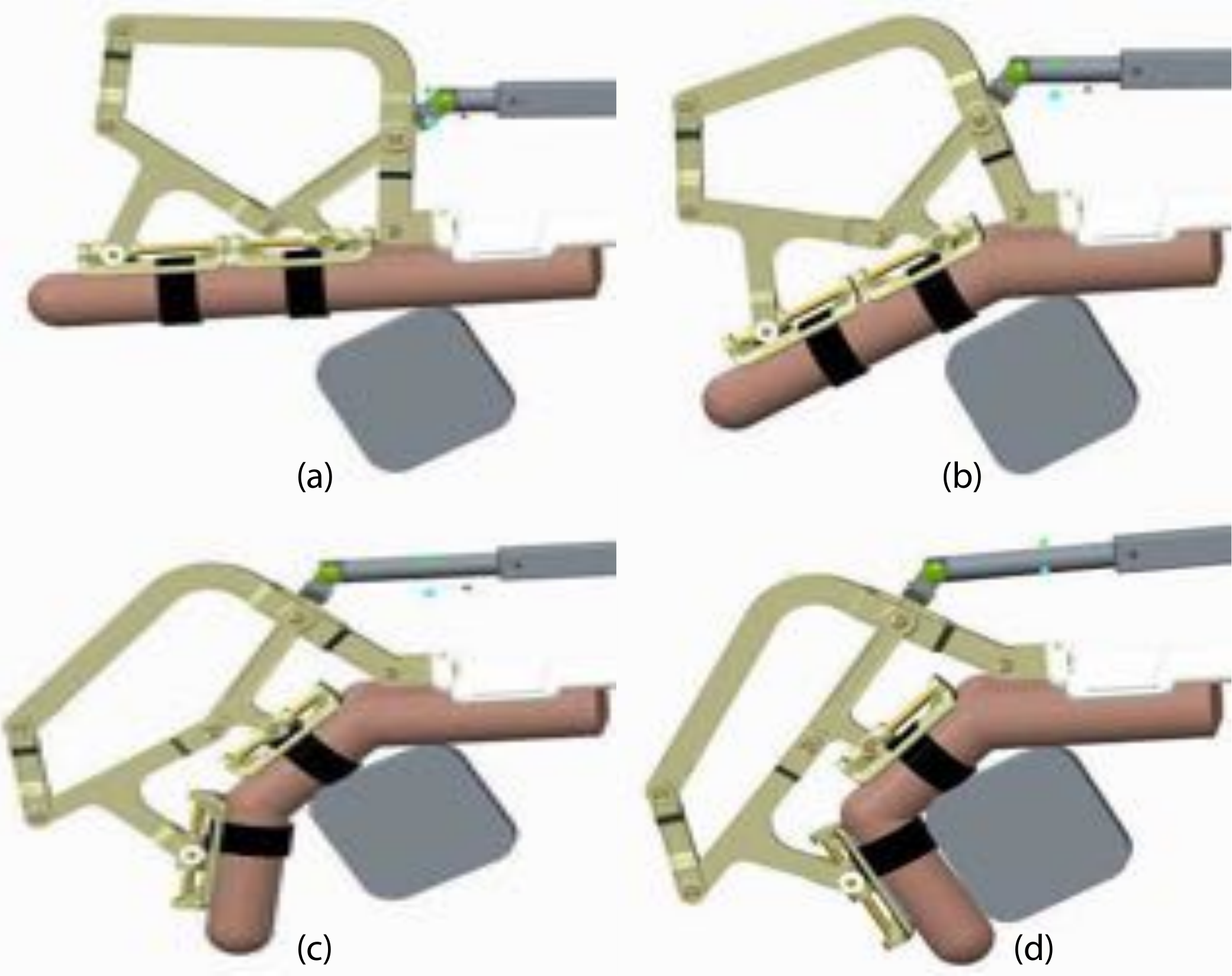}
\caption{Underactuation concept of the proposed hand exoskeleton during a grasping task: (a) initial pose with all finger joints extended, (b) actuator force moves MCP joint until the first finger phalange gets in touch with the object, (c) actuator force is transmitted to move PIP joint when the first finger phalange touches the object and (d) finally grasping task is completed when both phalanges touch the object.}
\label{fig:underact}
\end{figure}

The same underactuation concept can be extended to a hand exoskeleton, by considering user's fingers as a part of the mechanism. Anatomical coupling between finger joints define the required spring-like behavior for the underactuation. Then, the efficient force transmission requires the use of additional passive joints, which can assist motion for finger joints without aligning mechanical joints to the finger joints. Such an inherent passive compensation also improves the robustness of the device.

Figure~\ref{fig:underact} represents the operation flow of the underactuated hand exoskeleton. The finger component is connected to the user's finger, and assists the finger to flex until both finger joints are fully extended and the actuator stroke is at its minimum limit. In the beginning, as the actuator starts to move, the exoskeleton transmits forces to proximal phalange and rotates MCP joint. Encountering external forces along the proximal phalange caused by the physical interaction finalizes the rotation along MCP joint. These forces transmit actuation forces to intermediate phalange through the linkage based mechanism with passive joints, rotating PIP joint. Having $2 DoF$ to control, grasping task is completed when PIP joint is rotated enough for intermediate phalange to reach the object. Similarly, the extension of the finger starts by rotating PIP joint first and MCP joint later until the finger is totally extended and finger joints reach their physical limits. As a safety measure, the displacement is stroke is designed to reach its minimum limit when both finger joints are extended to their natural limits.

\begin{figure}[b!]
\centering
\subfigure[Pose 1]{\includegraphics[width=0.4\textwidth]{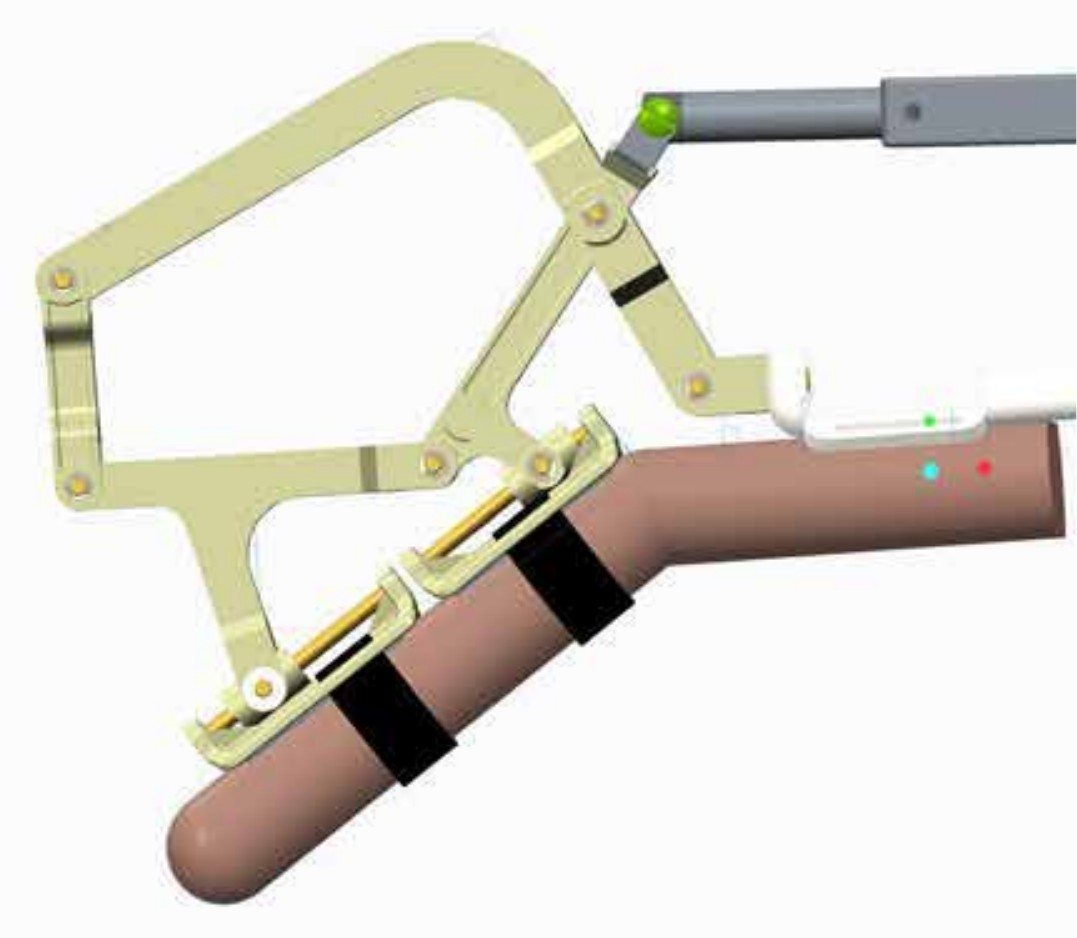}}
\subfigure[Pose 2]{\includegraphics[width=0.4\textwidth]{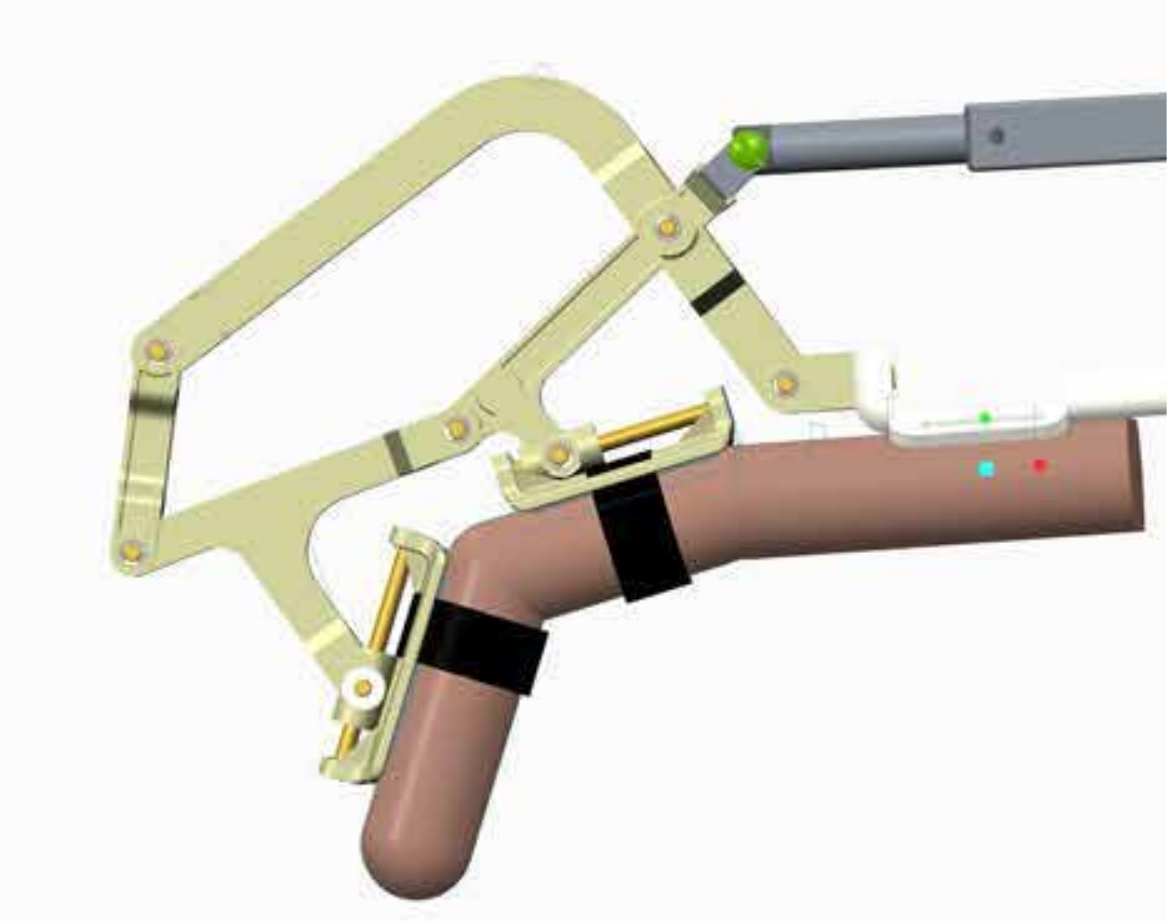}}
\caption{Different poses of finger joints that can be achieved when the actuator displacement is locked due to the extra mobility introduced by the underactuation concept.}
\label{fig:constraint}
\end{figure}

The underactuation concept does not control each finger joint individually, but controls the relation between them based on interaction forces. Fig~\ref{fig:constraint} shows that finger joints can perform a constrained movement even when the actuator is fixed. This constraint is achieved by introducing additional passive joints to the system in order to increase the mobility of the overall device. 

\begin{figure}[htb]
\centering
\includegraphics[width=0.8\textwidth]{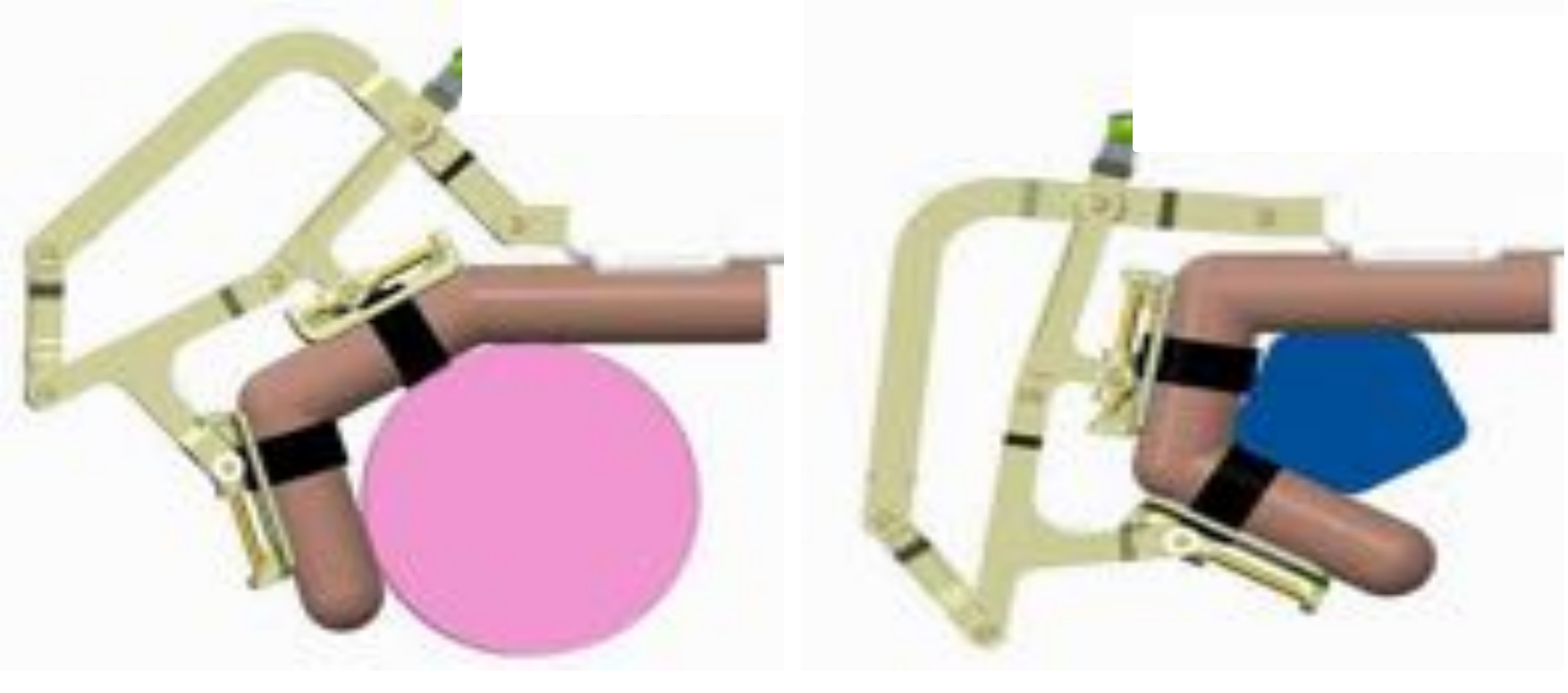}
\caption{Adaptability of underactuated hand exoskeleton for objects with different sizes and shapes due to the extra mobility.}
\label{fig:underact2}
\end{figure}

Thanks to this extra mobility, underactuation concept can adjust the operation of a mechanism based on the shape and the size of the grasping object with no prior information about the object or the orientation of the hand respect to the object. Figure ~\ref{fig:underact2} shows the device grasping objects with different shapes and sizes. 

\newpage
\section{Mechanical Pose Analysis} \label{sec:kinematics}

After designing the first sketch of the proposed exoskeleton as Figure~\ref{fig:model}, we need to analyze the kinematics for the following purposes: 

\begin{itemize}
\item to analyze the behavior of the proposed device during operation,
\item to set limits along mechanical joints and transmitted forces to finger phalanges,
\item to avoid physical limitations of the mechanism, such that the finger lengths, the stroke limits of the actuator, etc., and
\item to optimize the force transmission and RoM for finger joints.
\end{itemize}

The extra mobility introduced by the underactuation prevents a unique solution for finger joints knowing only actuator displacement. The coupled movement between finger joints for a certain actuator displacement cannot predict finger joints with with no further equipment. This is why, the initial mechanical pose analysis will be performed using inverse kinematics in Section \ref{sec:inverse}, such that the unique configuration for actuator displacement and mechanism configuration will be calculated for the given pose of finger joints.

Inverse kinematics is sufficient to study joint limitations and to optimize the link lengths in order to improve joint RoM and force transmission over finger phalanges. However, using inverse kinematics during real time implementations is almost useless. %On the other hand, forward kinematics will be crucial to estimate the finger pose simultaneously during operation, using mechanical pose. 
As discussed before, a unique solution with forward kinematics cannot be reached unless an additional sensory measurement is used within the system. With this motivation, one of the passive joints of the mechanical system is measured with a rotational potentiometer to solve the forward kinematics. We will approach forward kinematics problem numerically in Section \ref{sec:forward}, and analytically in Section \ref{sec:forward_analytical}. 

Having $2$ sensory measurements and $2$ finger joints as output results a square, invertible Jacobian matrix through differential kinematics for velocities (see Section \ref{sec:jacobian}) and for statics (see Section \ref{sec:statics}).

\begin{figure}[htb]
\centering
\subfigure[Point definitions depicted on CAD model of proposed finger exoskeleton component.]{\includegraphics[width=0.75\textwidth]{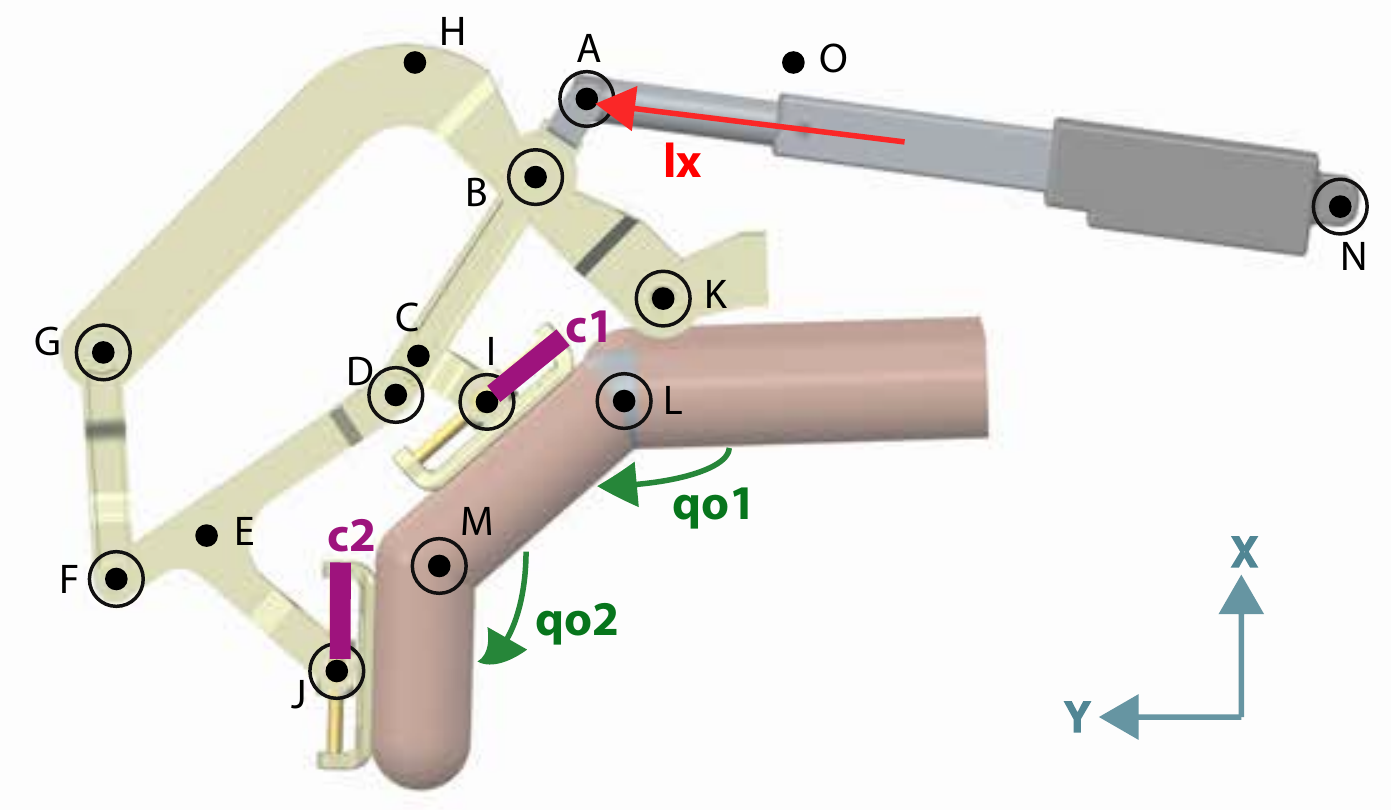}} \\
\subfigure[Joint definitions depicted on CAD model of proposed finger exoskeleton component.]{\includegraphics[width=0.75\textwidth]{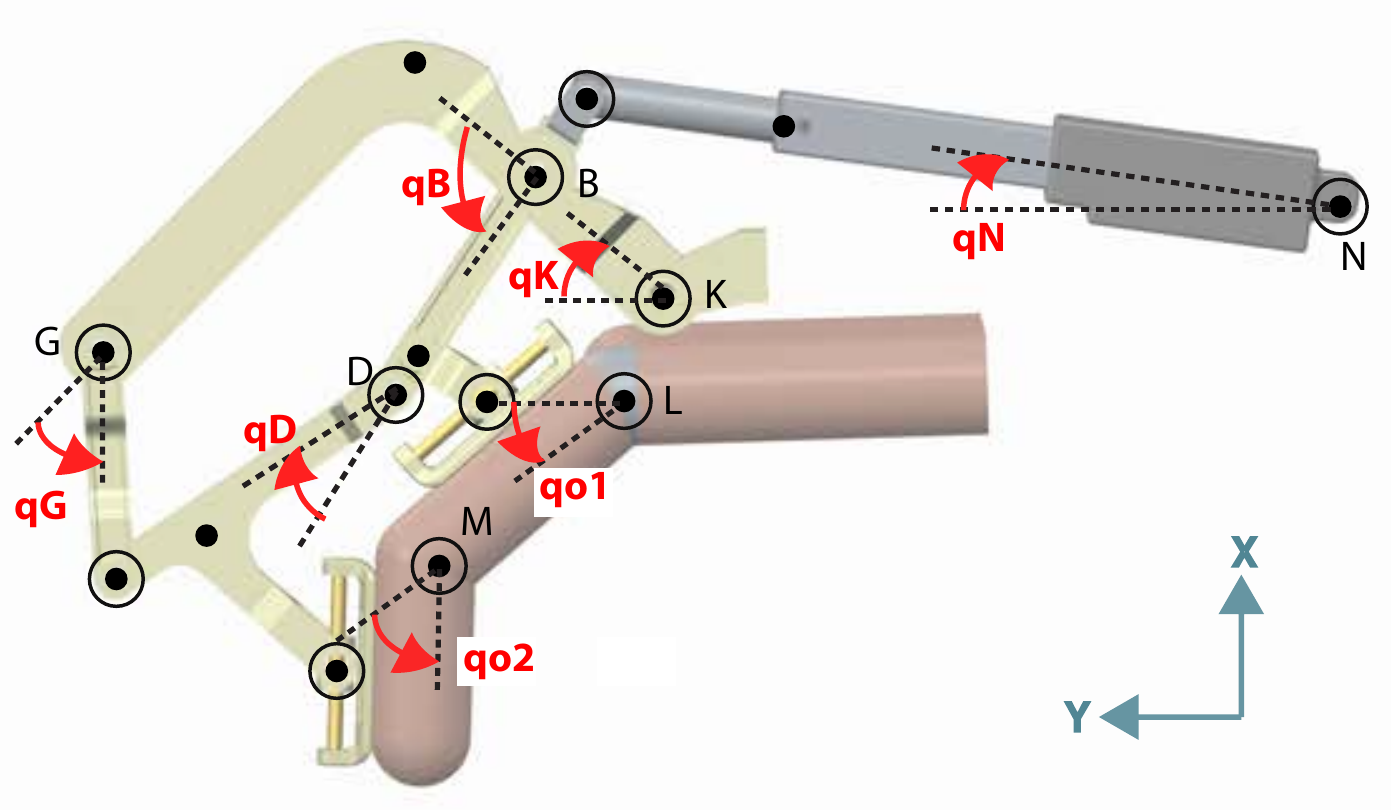}}
\caption{Kinematic scheme with point and joint definitions of a finger component for the proposed underactuated hand exoskeleton.}
\label{fig:definitions}
\end{figure}

Figure~\ref{fig:definitions} defines and shows both active and passive joints to solve kinematics. A linear actuator is attached to the point $A$ with a displacement of $l_x$. MCP and PIP joints of user's finger are defined as points $L$ and $M$ respectively, with the rotations as $q_{o1}$ and $q_{o2}$. The mechanism consists of $9$ passive revolute joints at points $A$, $B$, $D$, $F$, $G$, $I$, $J$, $K$, and $N$ while their rotations are represented as $q_i$, $i$ being the point name. The passive sliders are shown with the points $I$ and $J$, with the displacement of their passive linear joints $c_1$ and $c_2$. Finger phalanges are connected to the device at these points. The points $H$, $C$ and $E$ are additional points that are needed for the calculations, even though they are located on rigid links and do not have any mobility. The point $N$ shows the point between the actuator and the base. Finally, the point $O$ shows the initial position of point $A$ when the actuation stroke $l_x$ is zero.

\newpage
\subsection{Inverse Kinematics} \label{sec:inverse}

Inverse kinematics uses finger joint rotations ([$q_{o1}, q_{o2}$]) and finds the desired mechanical joints and actuation displacement ([$q_B, q_D, q_K, q_G, q_N, c_1, c_2, l_x$]) to achieve these desired joints. Since these rotational joints can be used only in the trigonometric relations, the system becomes nonlinear. Such nonlinearity leads the inverse kinematics to be performed through mechanical closed loops of system with a set of vector equations. Table~\ref{tab:variables} describes all unknown parameters, constants and finger pose parameters that will be used to define vectors between two points stated in Figure~\ref{fig:definitions}. In this table, the notation $\vect{r}^{i}_{j}$ is used to indicate vectors connecting point $i$ to point $j$, and $l_{ij}$ shows a constant length between point $i$ and point $j$. Inverse kinematics problem defines a given finger pose as known parameters, they are specified separately in Table~\ref{tab:variables}.

\begin{table}[!htb]
\caption{Variable and constants for the vectors to solve inverse kinematics}
\label{tab:variables}
\begin{center}
\begin{tabular}{P{0.5in}|| P{0.8in}| P{0.8in}| P{0.8in}}
\hline\hline
Vector & Unknown Parameter & Constants & Finger Pose Parameter \\ [2ex] \hline
$\vect{r}^{O}_{A}$ & \multirow{2}{*}{$l_x, q_N$}& & \multirow{6}{*}{} \\ [2ex] \cline{1-1} \cline{3-3}
$\vect{r}^{O}_{N}$ & & $l_{act}$ & \\ [2ex] \cline{1-1} \cline{2-3}
$\vect{r}^{A}_{D}$ &\multirow{2}{*}{$q_B$} & $l_{AD}$ & \\ [2ex] \cline{1-1} \cline{3-3}
$\vect{r}^{C}_{I}$ & & $l_{CI}$ &\\ [2ex] \cline{1-1} \cline{2-3}
$\vect{r}^{G}_{K}$ & $q_K$ & $l_{GK}$ & \\ [2ex] \cline{1-1} \cline{2-3}
$\vect{r}^{K}_{N}$ & & $l_{KN}$, $q_{KN}$ & \\ [2ex] \cline{1-1} \cline{2-4}
$\vect{r}^{I}_{L}$ & \multirow{2}{*}{$c_1$} & &\multirow{2}{*}{$q_{o1}$}\\ [2ex] \cline{1-1} \cline{3-3}
$\vect{r}^{M}_{I}$ & & $l_{ML}$ & \\ [2ex] \cline{1-1} \cline{2-4}
$\vect{r}^{L}_{K}$ & & $l_{LK}$, $q_{LK}$ &\\[2ex] \cline{1-1} \cline{2-3}
$\vect{r}^{D}_{F}$ & \multirow{2}{*}{$q_D$} & $l_{DF}$ & \\ [2ex] \cline{1-1} \cline{3-4}
$\vect{r}^{E}_{J}$ & & $l_{EJ}$ & \\ [2ex] \cline{1-1} \cline{2-4}
$\vect{r}^{J}_{M}$ & $c_2$ & & $q_{o2}$\\ [2ex] \cline{1-1} \cline{2-4}
$\vect{r}^{G}_{F}$ & $q_G$ & $l_{GF}$ & \\ [2ex]
\hline \hline
\end{tabular}
\end{center}
%\label{table_parameters}
\end{table}

\begin{figure}[htb]
\centering
\includegraphics[width=1\textwidth]{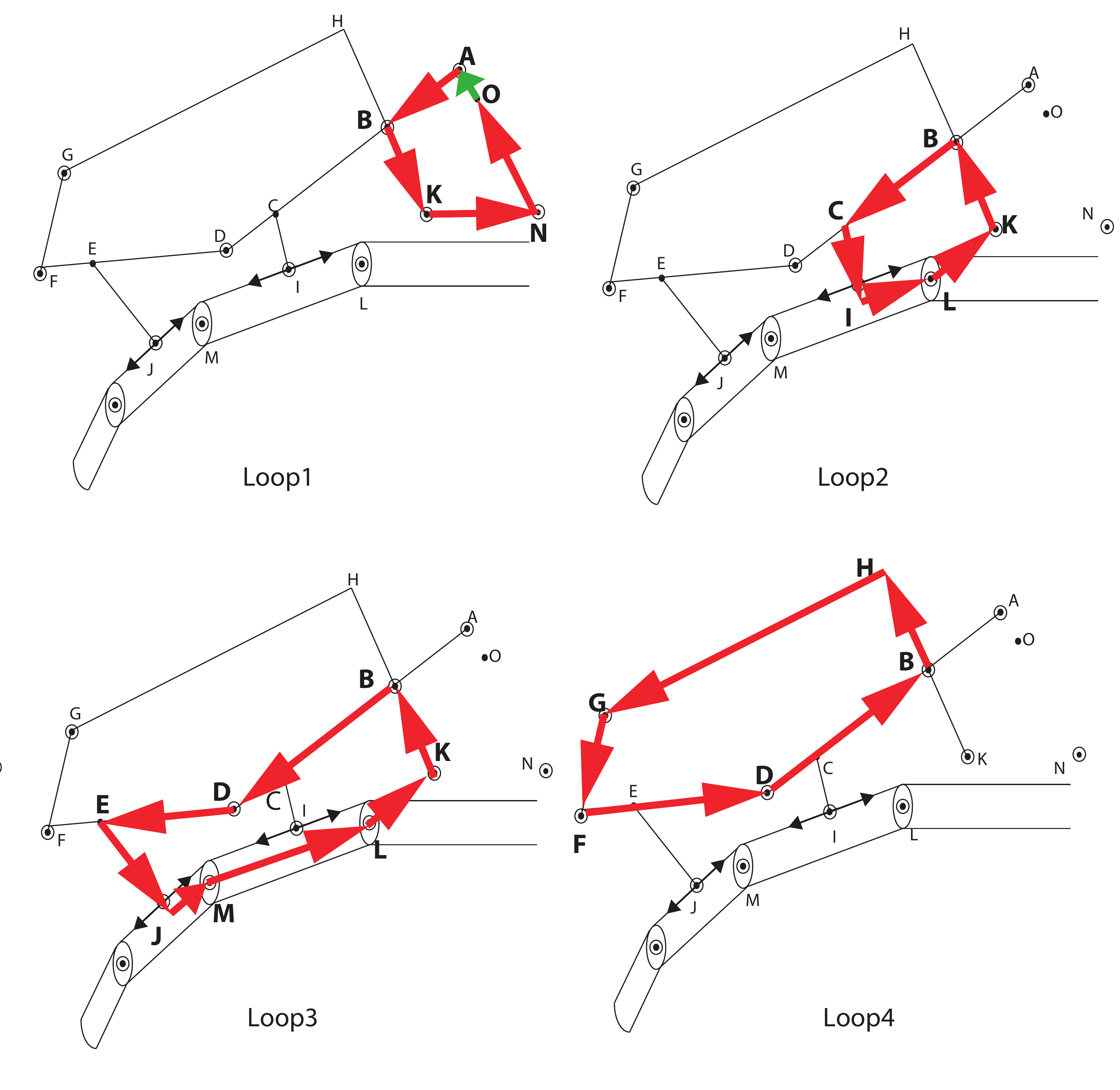}
\caption{Independent kinematic loops to solve numerical inverse kinematics. Red arrows show the vectors defined by passive joints and rigid links, while the green arrow shows the active actuator displacement, which can be controlled.}
\label{fig:loops}
\end{figure}

Table~\ref{tab:variables} shows that the whole mechanism can be defined with vectors using $8$ unknown parameters as \{$l_x$, $c_1$, $c_2$, $q_B$, $q_D$, $q_G$, $q_K$, $q_N$\} and $2$ known finger pose variables as \{$q_{o1}$, $q_{o2}$\}. Even though there are other passive revolute joints (\{$q_A$, $q_F$, $q_I$, $q_J$\}), their rotations are not required to define the device configuration since they are constrained by other variables. To find a unique solution for these $8$ unknowns, $8$ independent equations are needed: four independent kinematic loops were identified as presented in Figure~\ref{fig:loops} using vector chains along the mechanical links.

$Loop 1$ defines a relation between the actuator displacement and the mechanism itself using a vector loop equation and its corresponding exponential expression as:

\begin{align}
\label{eq:loop1}
\nonumber \vect{r}^{O}_{A} + \vect{r}^{A}_{B} + \vect{r}^{B}_{K} + \vect{r}^{K}_{N} + \vect{r}^{N}_{O} = \vect{0} \\
l_xe^{iq_N} + l_{AB}e^{iq_B} + l_{BK}e^{iq_K} + l_{KN}e^{iq_{KN}} + l_{act}e^{iq_N} = 0.
\end{align}

\noindent In this loop, we can solve the unknown parameters \{$l_x$, $q_B$, $q_K$, $q_N$\}. $Loop 2$ defines the corresponding motion around the MCP joint of user's finger using passive joints affected by the actuator:

\begin{align}
\label{eq:loop2}
\nonumber \vect{r}^{K}_{B} + \vect{r}^{B}_{C} +   \vect{r}^{C}_{I} + \vect{r}^{I}_{L} + \vect{r}^{L}_{K} =  \vect{0}\\
l_{BK}e^{iq_K} + l_{BC}e^{iq_B} + l_{CI}e^{iq_B} + c_1e^{iq_{o1}} + l_{LK}e^{iq_{LK}} = 0.
\end{align}

\noindent The unknown parameters of $Loop 2$ can be listed as \{$q_K$, $q_B$, $c_1$\}, as much as the finger pose parameter \{$q_{o1}$\}. Similarly, the PIP joint rotation of user's finger is defined by $Loop 3$ using passive joints of the mechanism:

\begin{align}
\label{eq:loop3}
\nonumber \vect{r}^{K}_{B} + \vect{r}^{B}_{D} + \vect{r}^{D}_{E} + \vect{r}^{E}_{J} + \vect{r}^{J}_{M} + \vect{r}^{M}_{L} + \vect{r}^{L}_{K} = \vect{0} \\
l_{BK}e^{iq_K} + l_{BD}e^{iq_B} + l_{DE}e^{iq_D} + l_{EJ}e^{iq_D} + c_2e^{iq_{o2}} + l_{ML}e^{iq_{o1}} + l_{LK}e^{iq_{LK}} = 0.
\end{align}

\noindent $Loop 3$ covers the unknown parameters \{$q_D$, $c_2$\} and and finger pose parameters \{$q_{o2}$, $q_{o1}$\}. Finally, $Loop 4$ provides a relation between mechanical passive joints alone using the variables \{$q_K$, $q_B$, $q_G$, $q_D$\} in order to achieve a unique solution for the pose analysis:

\begin{align}
\label{eq:loop4}
\nonumber \vect{r}^{B}_{H} + \vect{r}^{H}_{G} + \vect{r}^{G}_{F} + \vect{r}^{F}_{D} + \vect{r}^{D}_{B} = \vect{0} \\
l_{BH}e^{iq_K} + l_{HG}e^{iq_K} + l_{GF}e^{iq_G} + l_{FD}e^{iq_D} + l_{DB}e^{iq_K} = 0.
\end{align}

These loops were chosen such that each loop ($Loop 1$ - $Loop 4$) has at least one parameter, which is not covered by the other loops. In particular, Table~\ref{tab:unique_inverse} details the input and the output parameters covered by each loop, and unique output parameters covered by each loop. 

\begin{table}[htb]
%\vspace*{-1.5\baselineskip}
\caption{Input and output parameters used for each loop and unique output parameters for inverse kinematics.}
\label{tab:unique_inverse}
\begin{center}
\begin{tabular}{P{0.5in}|| P{1.2in}| P{1.3in}| P{1.1in}}
\hline\hline \\
Loop & Input Parameters & Output Parameters & Unique Output Parameters \\ [2ex] \hline
$Loop1$ & & $l_x, q_N, q_B, q_K$ & $l_x, q_N$ \multirow{6}{*}{} \\ [2ex] \cline{1-4}
$Loop2$ & $q_{o1}$ & $q_B, q_K, c_1$ & $c_1$ \\ [2ex] \cline{1-4}
$Loop3$ & $q_{o1}, q_{o2}$ & $q_D, c_2, q_K$ & $c_2$ \\ [2ex] \cline{1-4}
$Loop4$ & & $q_K, q_G, q_D, q_B$ & $q_G$ \\ [2ex] \cline{1-4}
\hline \hline
\end{tabular}
%\vspace*{-2\baselineskip}
\end{center}
%\label{table_parameters}
\end{table}

\noindent We can assume that the $4$ loops are chosen carefully to be independent from each other, since there is at least one parameter covered only by each loop. Even though the loops above can be chosen in a different manner as well, they were chosen in the simplest way possible.

The vertical and the horizontal ($X$ and $Y$ as depicted in Figure ~\ref{fig:definitions}) components of the vector equations in Equations~(\ref{eq:loop1} - \ref{eq:loop4}) achieve $8$ nonlinear equations. Due to the nonlinearity of the system, the numerical method finds a unique configuration for actuator displacement $l_x$ as well as passive joints [$c_1$, $c_2$, $q_B$, $q_D$, $q_G$, $q_K$, $q_N$] for given finger pose $q_{o1}$ and $q_{o2}$. %These loop equations can also be used to calculate the Jacobian of the system.

The inverse kinematics analysis outputs were compared to the CAD simulation results for the index finger exoskeleton. We present the index finger results only, since the mechanical design is identical for the other fingers as well (except the thumb). %In particular, a sinusoidal input of joint angle rotations  Figure~\ref{fig:CadComp_qo} were supplied to both CAD design of the mechanism in CREO CAD program and the kinematics analysis simulation that was run in MATLAB.

\begin{figure}[htb]
\centering
\includegraphics[width=0.8\textwidth]{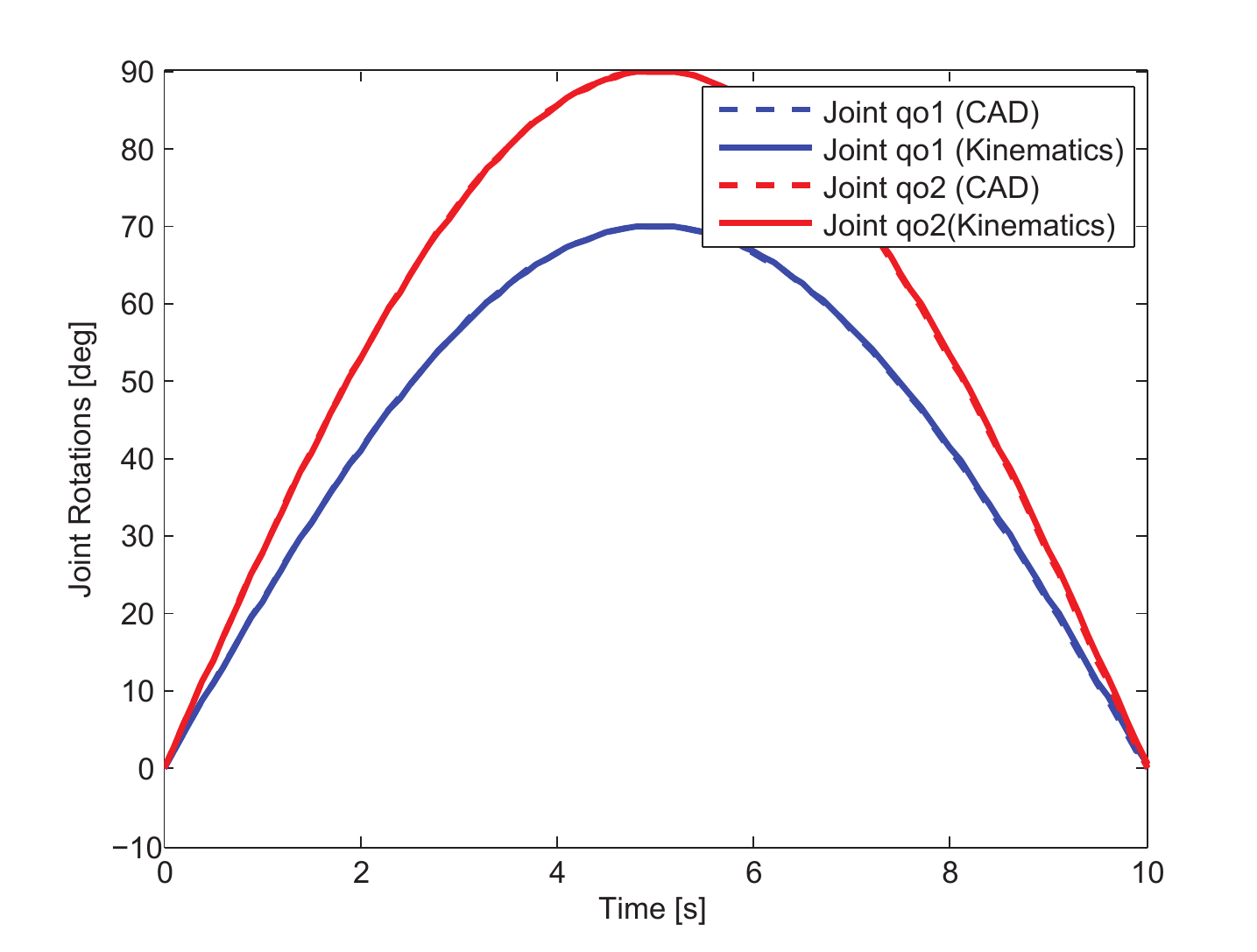}
\caption{Finger joint rotations defined as input for the CAD simulation and the kinematics analysis to calculate inverse kinematics.}
\label{fig:CadComp_qo}
\end{figure}

\begin{figure}[htb]
\centering
\includegraphics[width=1\textwidth]{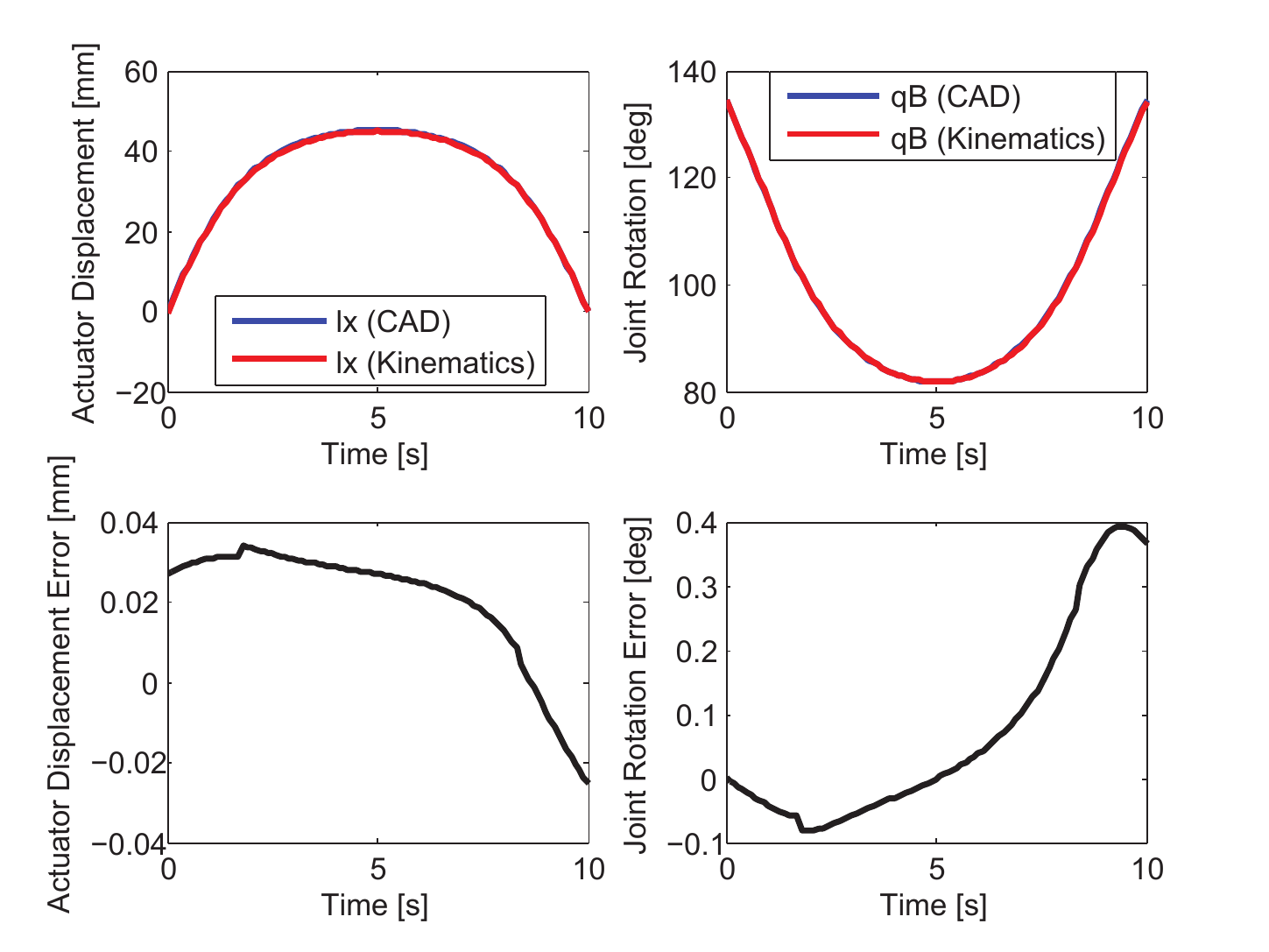}
\caption{Comparison between the simulations of the CAD model and the inverse kinematics analysis through actuator displacement $l_x$ and an additional passive joint $q_B$.}
\label{fig:CadComp_joints}
\end{figure}

The finger joint rotations for MCP and PIP joints as shown in Figure~\ref{fig:CadComp_qo} was supplied to both the mechanical simulation in the CAD software, and the kinematics simulation in MATLAB. Figure~\ref{fig:CadComp_joints} presents the resulting actuator displacement $l_x$ and joint rotation $q_B$. Firstly, the actuator displacement $l_x$ was compared between CAD simulation and kinematics analysis, since it is the only controllable state of the system. We chose passive joint $q_B$ to be an observable state, so $q_B$ is the second joint compared between the two simulations. Even though any of other passive joint could be chosen for confirmation, joint $q_B$ will be used for further analysis for reasons to be justified later. The errors between simulation tools for both states prove that inverse kinematics calculations are sufficiently accurate define the mechanical device.

\newpage
\subsection{Forward Kinematics}\label{sec:forward}

Even though inverse kinematics gives sufficient information about the mechanism behavior, physical limits, actuator requirements and feasible workspace for finger joints; it is useless for real time implementation of the device. The kinematics analysis will be useful to predict finger pose using mechanical pose and measurements. In particular, each finger component of the hand exoskeleton is actuated by a linear actuator and its linear displacement is the only available measurement tool during operation.  

%Embraced underactuation concept dictates the single actuator to control two finger joints, defining an extra mobility between them. Therefore, the measurement coming from actuator displacement is not sufficient to obtain a unique solution for forward kinematics, which is important to achieve to obtain finger pose estimation simultaneously during rehabilitative or haptic applications. %However, the finger pose during the operation is crucial to track the patient's medical improvements, or to provide haptic implementation of the device using virtual environment.

A unique solution for the forward kinematics can be reached simply by providing an additional sensor to the mechanism. Such addition does not affect the underactuation since it does not provide any energy to the overall system, but contributes to estimate the coupled mobility between finger joints that is left uncontrolled. The proposed kinematics have multiple passive joints and choosing the appropriate one to measure during operation should consider the following issues:

\begin{itemize}
\item It should be away from the passive joints at the connection tools near finger phalanges, not to interfere with user and other mechanical components.
\item It should be close to the exoskeleton base to shorten the cables as much as possible.
\item Its displacement during operation should be wide to increase the accuracy of pose estimation.
\end{itemize}

These considerations highlighted the joint along point $B$ depicted on Figure~\ref{fig:definitions} as a second measurement tool for forward kinematics and pose estimation. The wide workspace of joint $B$ can easily be obtained in Figure~\ref{fig:underact} and Figure~\ref{fig:underact2}. The pin connecting two links along joint $B$ allows a simple assembly of a rotational potentiometer while its place minimizes the complexity of cable transmission.

Using such an additional sensory measurement, forward kinematics of the mechanism can be calculated uniquely. Numerical forward kinematics approach use the same vector definitions in Figure~\ref{fig:loops} using $2$ input parameters as \{$q_B$, $l_x$\}, and $8$ unknown parameters as \{$q_o1$, $q_{o2}$, $c_1$, $c_2$, $q_D$, $q_G$, $q_K$, $q_N$\} as redefined for forward kinematics in Table~\ref{tab:variables_forw}. Similar to inverse kinematics, additional passive joints \{$q_A$, $q_F$, $q_I$, $q_J$\}, are ignored, since their movements are constrained by other parameters. Similarly, $8$ independent loop equations are obtained using $X$ and $Y$ components of $4$ loop equations as described in Figure ~\ref{fig:loops} in order to find a unique solution for these $8$ unknowns. The forward kinematics loop equations are the same as inverse kinematics equations, Equation~(\ref{eq:loop1}) - Equation~(\ref{eq:loop4}) having different input and output variables.

\begin{table}[!htb]
%\vspace*{-1.5\baselineskip}
\caption{Variable and constants for the vectors to solve forward kinematics.}
\label{tab:variables_forw} % \multirow{6}{*}{}
\begin{center}
\begin{tabular}{P{0.5in}|| P{0.8in}| P{0.8in}| P{0.8in}}
\hline\hline \\
Vector & Unknown Parameters & Constants & Sensor Parameters \\ [2ex] \hline
$\vect{r}^{O}_{A}$ & \multirow{2}{*}{$q_N$} & & \multirow{2}{*}{$l_x$} \\ [2ex] \cline{1-1}\cline{3-3}
$\vect{r}^{O}_{N}$ & & $l_{act}$ & \\ [2ex] \cline{1-1}\cline{2-4}
$\vect{r}^{A}_{D}$ &\multirow{2}{*}{} & $l_{AD}$ & \multirow{2}{*}{$q_B$} \\ [2ex] \cline{1-1}\cline{3-3}
$\vect{r}^{C}_{I}$ & & $l_{CI}$ &\\ [2ex] \cline{1-1} \cline{2-4}
$\vect{r}^{G}_{K}$ & $q_K$ & $l_{GK}$ & \\ [2ex] \cline{1-1}\cline{2-3}
$\vect{r}^{K}_{N}$ & & $l_{KN}$, $q_{KN}$ & \\ [2ex] \cline{1-1} \cline{2-3}
$\vect{r}^{I}_{L}$ & \multirow{2}{*}{$c_1, q_{o1}$} & &\multirow{2}{*}{}\\ [2ex] \cline{1-1} \cline{3-3}
$\vect{r}^{M}_{I}$ & & $l_{ML}$ & \\ [2ex] \cline{1-1} \cline{2-3}
$\vect{r}^{L}_{K}$ & & $l_{LK}$, $q_{LK}$ &\\[2ex] \cline{1-1} \cline{2-3}
$\vect{r}^{D}_{F}$ & \multirow{2}{*}{$q_D$} & $l_{DF}$ & \\ [2ex] \cline{1-1} \cline{3-3}
$\vect{r}^{E}_{J}$ & & $l_{EJ}$ & \\ [2ex] \cline{1-1} \cline{2-3}
$\vect{r}^{J}_{M}$ & $c_2, q_{o2}$ & & \\ [2ex] \cline{1-1} \cline{2-3}
$\vect{r}^{G}_{F}$ & $q_G$ & $l_{GF}$ & \\ [2ex]
\hline \hline
\end{tabular}
%\vspace*{-2\baselineskip}
\end{center}
%\label{table_parameters}
\end{table}

Even though the vectors used to define the mechanical links are the same for the forward and the inverse problems, changing input and output variables for each problem changes the whole analysis. Not only the analysis results, but also the independency of the defined loops vary by changing the variable definitions. Table~\ref{tab:unique_forward} redefines input and output parameters that are covered by each loop and signifies unique output parameters that are not covered by other loops.

\begin{table}[!htb]
%\vspace*{-1.5\baselineskip}
\caption{Analyzing input and output parameters used for each loop and unique output parameters for forward kinematics}
\label{tab:unique_forward}
\begin{center}
\begin{tabular}{P{0.5in}|| P{1.2in}| P{1.3in}| P{1.1in}}
\hline\hline \\
Loop & Input Parameters & Output Parameters & Unique Output Parameter \\ [2ex] \hline
$Loop1$ & $l_x, q_B$ & $q_N, q_K$ & $q_N$ \multirow{6}{*}{} \\ [2ex] \cline{1-4}
$Loop2$ & $q_B$ & $q_{o1}, c_1, q_K$ & $c_1$ \\ [2ex] \cline{1-4}
$Loop3$ & $q_B$ & $q_{o1}, q_{o2}, c_2, q_D, q_K$ & $c_2, q_{o2}$ \\ [2ex] \cline{1-4}
$Loop4$ & $q_B$ & $q_K, q_G, q_D$ & $q_G$ \\ [2ex] \cline{1-4}
\hline \hline
\end{tabular}
%\vspace*{-2\baselineskip}
\end{center}
%\label{table_parameters}
\end{table}

Similarly, the existence of unique parameters for each loop signifies the independency of those loops, such that the analysis might be trusted. The accuracy of numerical forward kinematics will be performed together with analytical approach in Section~\ref{sec:forward_analytical}.

\begin{table}[!htb]
%\vspace*{-1.5\baselineskip}
\caption{Variable and constants for the vectors to solve forward kinematics for calibration.}
\label{tab:variables_calib} % \multirow{6}{*}{}
\begin{center}
\begin{tabular}{P{0.5in}|| P{0.8in}| P{0.8in}| P{0.8in}}
\hline\hline \\
Vector & Unknown Parameters & Constants & Sensor Parameters \\ [2ex] \hline
$\vect{r}^{O}_{A}$ & \multirow{2}{*}{$q_N$} & & \multirow{2}{*}{$l_x$} \\ [2ex] \cline{1-1} \cline{3-3}
$\vect{r}^{O}_{N}$ & & $l_{act}$ & \\ [2ex] \cline{1-1} \cline{2-4}
$\vect{r}^{A}_{D}$ &\multirow{2}{*}{} & $l_{AD}$ & \multirow{2}{*}{$q_B$} \\ [2ex] \cline{1-1} \cline{3-3}
$\vect{r}^{C}_{I}$ & & $l_{CI}$ &\\ [2ex] \cline{1-1} \cline{2-4}
$\vect{r}^{G}_{K}$ & $q_K$ & $l_{GK}$ & \\ [2ex] \cline{1-1} \cline{2-3}
$\vect{r}^{K}_{N}$ & & $l_{KN}$, $q_{KN}$ & \\ [2ex] \cline{1-1} \cline{2-3}
$\vect{r}^{I}_{L}$ & $c_1, q_{o1}$ & &\multirow{2}{*}{}\\ [2ex] \cline{1-1} \cline{2-3}
$\vect{r}^{M}_{I}$ & $c_1, q_{o1}, l_{LM}$ & & \\ [2ex] \cline{1-1} \cline{2-3}
$\vect{r}^{L}_{K}$ & & $l_{LK}$, $q_{LK}$ &\\[2ex] \cline{1-1} \cline{2-3}
$\vect{r}^{D}_{F}$ & \multirow{2}{*}{$q_D$} & $l_{DF}$ & \\ [2ex] \cline{1-1} \cline{3-3}
$\vect{r}^{E}_{J}$ & & $l_{EJ}$ & \\ [2ex] \cline{1-1} \cline{2-4}
$\vect{r}^{J}_{M}$ & $q_{o2}$ & & $c_2$\\ [2ex] \cline{1-1} \cline{2-4}
$\vect{r}^{G}_{F}$ & $q_G$ & $l_{GF}$ & \\ [2ex]
\hline \hline
\end{tabular}
%\vspace*{-2\baselineskip}
\end{center}
%\label{table_parameters}
\end{table}

Alternatively, the proposed kinematics structure can be easily adjusted for alternative needs. For instance, a calibration process can easily be created to estimate the length of proximal finger phalange in a quick way instead of measuring it manually for each user. Such a calibration process can be defined simply by changing input and output parameters as \{$q_B$, $l_x$, $c_2$\}, and unknown parameters as \{$q_{o1}$, $q_{o2}$, $l_{LM}$, $c_1$, $q_D$, $q_G$, $q_K$, $q_N$\}, where finger phalange is depicted as $l_{LM}$. For the calibration process, the lack of measurement can be overcome by asking the user to reach a pose in a manner, forcing the linear displacement along middle finger phalange to its maximum limit. Table~\ref{tab:variables_calib} and Table~\ref{tab:unique_calib} summarizes the parameters and uniqueness of these parameters as before.

\begin{table}[!htb]
%\vspace*{-1.5\baselineskip}
\caption{Analyzing input and output parameters used for each loop and unique output parameters to solve forward kinematics for calibration}
\label{tab:unique_calib}
\begin{center}
\begin{tabular}{P{0.5in}|| P{1.2in}| P{1.3in}| P{1.1in}}
\hline\hline \\
Loop & Input Parameters & Output Parameters & Unique Output Parameter \\ [2ex] \hline
$Loop1$ & $l_x, q_B$ & $q_N, q_K$ & $q_N$ \multirow{6}{*}{} \\ [2ex] \cline{1-4}
$Loop2$ & $q_B$ & $q_{o1}, c_1, q_K$ & $c_1$ \\ [2ex] \cline{1-4}
$Loop3$ & $q_B, c_2$ & $q_{o1}, q_{o2}, q_D, l_{LM}, q_K$ & $l_{LM}, q_{o2}$ \\ [2ex] \cline{1-4}
$Loop4$ & $q_B$ & $q_K, q_G, q_D$ & $q_G$ \\ [2ex] \cline{1-4}
\hline \hline
\end{tabular}
%\vspace*{-2\baselineskip}
\end{center}
%\label{table_parameters}
\end{table}

\newpage
\subsection{Analytical Forward Kinematics}\label{sec:forward_analytical}

\begin{figure}[hbt]
\centering
\includegraphics[width=1\textwidth]{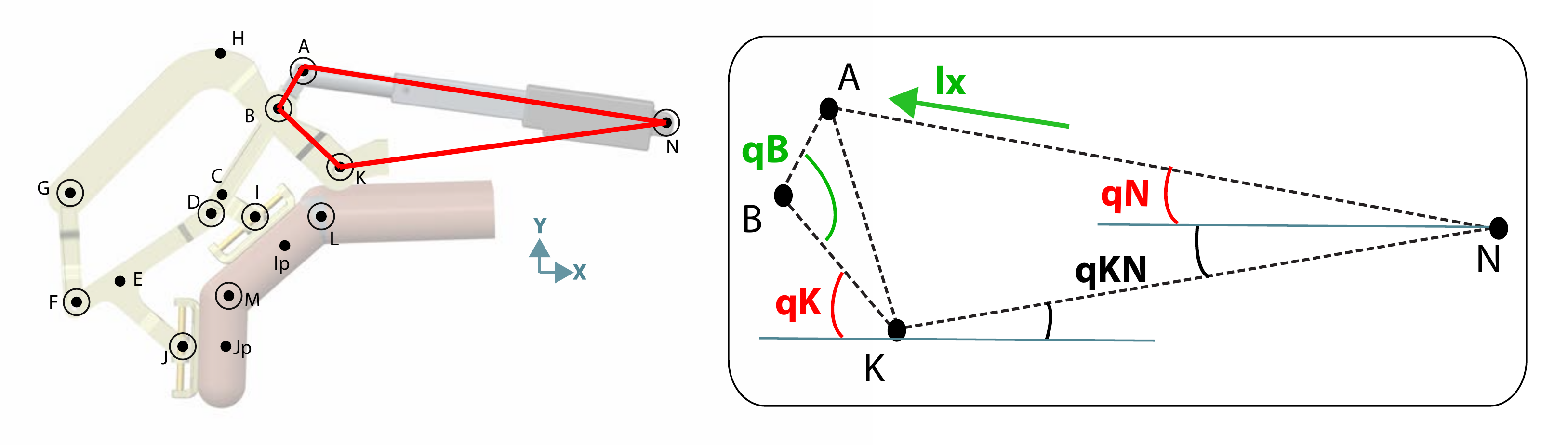}
\caption{First loop for analytical forward kinematics to reach passive joint pose $q_N$ and $q_K$ using actuator displacement $l_x$ and passive joint measurement $q_B$.}
\label{fig:analytical_v1}
%\vspace*{2\baselineskip}
\end{figure}

\begin{figure}[hbt]
\centering
\includegraphics[width=1\textwidth]{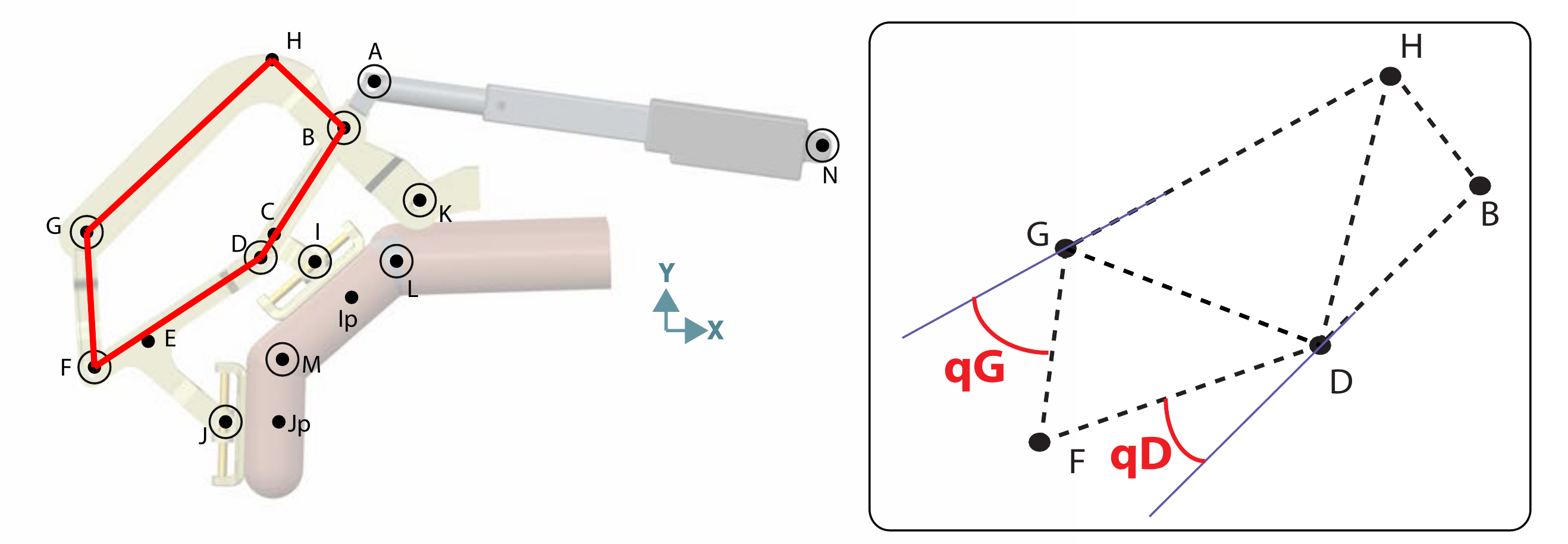}
\caption{Second loop to calculate analytical forward kinematics to reach passive joint pose $q_G$ and $q_D$ using passive joint pose values $q_N$ and $q_K$ obtained in the previous loop.}
\label{fig:analytical_v3}
%\vspace*{2\baselineskip}
\end{figure}

\begin{figure}[hbt]
\centering
\includegraphics[width=1\textwidth]{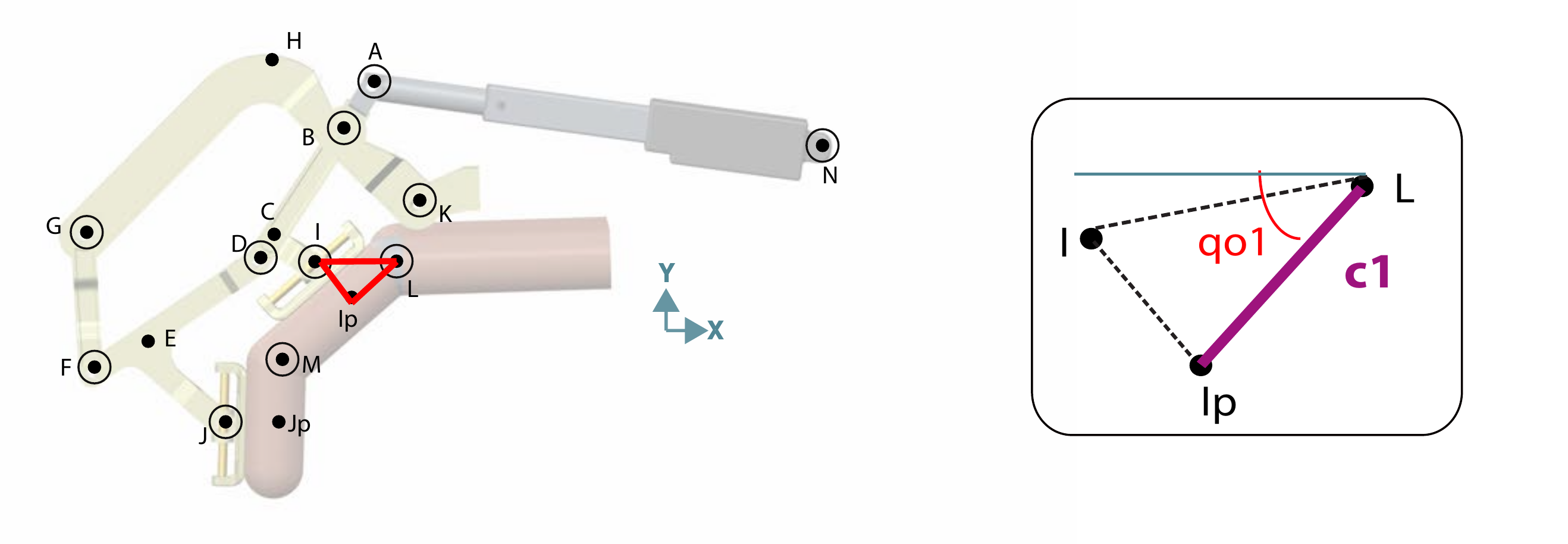}
\caption{Third loop to calculate analytical forward kinematics to reach passive slider displacement $c_1$ and MCP joint rotation $q_{o1}$ using previously obtained $q_K$ and sensory measurement $q_B$.}
\label{fig:analytical_v2}
%\vspace*{2\baselineskip}
\end{figure}

\begin{figure}[hbt]
\centering
\includegraphics[width=1\textwidth]{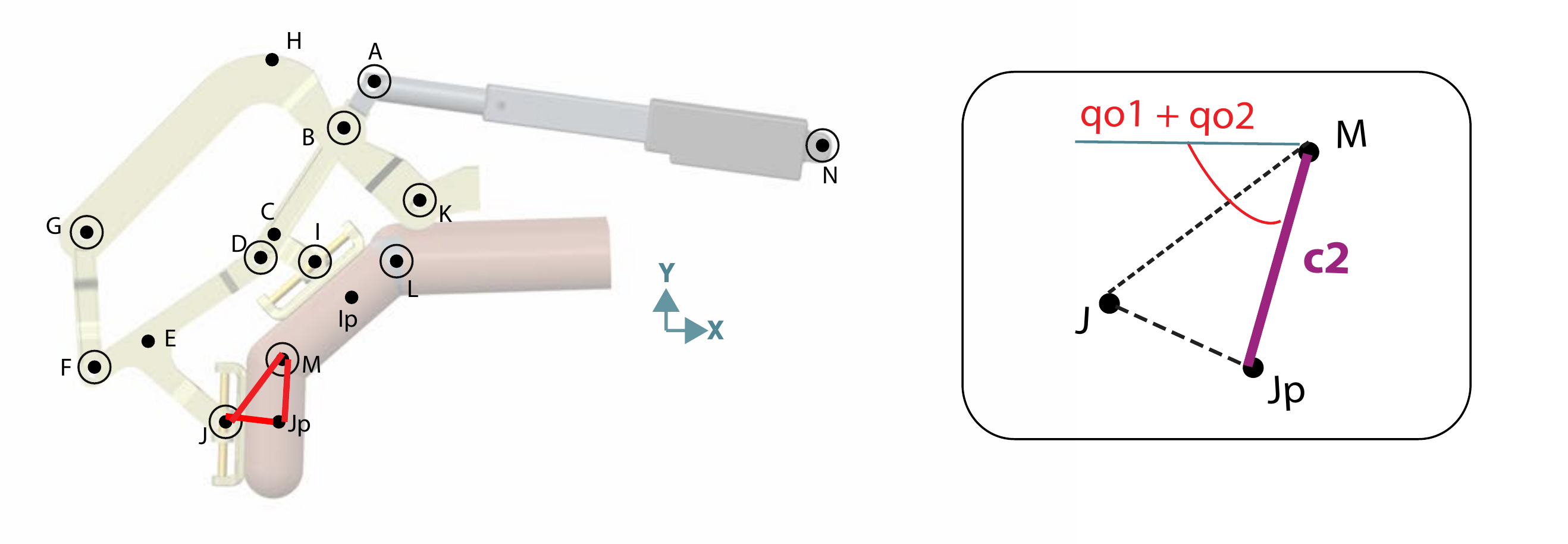}
\caption{Final loop to calculate analytical forward kinematics to reach passive slider displacement $c_2$ and PIP joint rotation $q_{o2}$ using previously obtained $q_D$, $q_K$ and sensory measurement $q_B$.}
\label{fig:analytical_v4}
%\vspace*{2\baselineskip}
\end{figure}

Even though numerical forward kinematics provides an accurate output, the calculational burden of numerical approach might limit the speed of the soon to be implemented control board. For a calibration process, which should be run once for each user and for a short time, such a burden can be ignored. However, pose estimation of finger joints during a complete operation might require a simple, smoother and faster approach. Using the same constants, sensor parameters and unknown parameters as defined in Table~\ref{tab:variables_forw} and independent kinematic loops different than Figure ~\ref{fig:loops}, forward kinematics can be solved analytically thanks to Cosine and Pythagorean Theorems. The new kinematic loops will be shown independently in Figure~\ref{fig:analytical_v1} - Figure~\ref{fig:analytical_v4}.

Considering that input parameters are sensory measurements as $q_B$ and $l_x$, the first loop can be defined as depicted in Figure ~\ref{fig:analytical_v1} in order to obtain $q_N$ and $q_K$ through Equation~(\ref{eq:analytical_qK}) and Equation~(\ref{eq:analytical_qN}).

In particular, constant mechanical link lengths $l_{AB}, l_{BK}$ and measured joints $l_x$ and $q_B$ allows the distance $l_{AK}$ to be known for any orientation using Cosine Theorem. Note that actuator distance $l_x$ is hidden under $l_{AN}$ expression, such that $l_{AN} = l_x + l_{act}$. Once $l_{AK}$ is defined, $q_K$ can be obtained using Equation~(\ref{eq:analytical_qK}).

\begin{align}
\label{eq:analytical_qK}
q_K &= \pi - [acos(\frac{l_{AK}^2 + l_{BK}^2 - l_{AB}^2}{2~l_{AK}~l_{BK}}) + acos(\frac{l_{AK}^2 + l_{KN}^2 - l_{AN}^2}{2~l_{AK}~l_{KN}}) + q_{KN}]
\end{align}

Simultaneously, knowing $l_{AK}$ leads joint rotation $q_N$ to be calculated using Equation~(\ref{eq:analytical_qN}) based on Cosine Theorem for the right-side triangle of the given loop.

\begin{align}
\label{eq:analytical_qN}
q_N &= acos(\frac{l_{AN}^2 + l_{KN}^2 - l_{AK}^2}{2~l_{AN}~l_{KN}}) - q_{KN}
\end{align}

The rest of analytical kinematics method requires the position of each point to be determined independently using previously obtained parameters, sensory parameters, constant parameters and distances between calculated points. Since these distances depend on the orientation of the device, they cannot be classified as constant variables. Nevertheless, obtaining numerical values of unknown parameters turn these distances from being expressions to numerical values. In fact, obtaining $q_N$ and $q_K$ allows the position of points $G$ and $D$ to be calculated, which defines distances $l_{DG}, l_{HD}$ in Figure ~\ref{fig:analytical_v3}

Achieving passive joint rotation $q_G$ using Equation~(\ref{eq:analytical_qG}) and $q_D$ using Equation~(\ref{eq:analytical_qD}) requires the angles within triangles defined by the loop to be calculated using Cosine Theorem.

\begin{align}
\label{eq:analytical_qG}
q_G &= \pi - [acos(\frac{l_{DG}^2 + l_{GH}^2 - l_{HD}^2}{2~l_{DG}~l_{GH}}) + acos(\frac{l_{DG}^2 + l_{GF}^2 - l_{DF}^2}{2~l_{DG}~l_{GF}})]
\end{align}

\begin{align}
\label{eq:analytical_qD}
q_D &= \pi - [acos(\frac{l_{HD}^2 + l_{DG}^2 - l_{GH}^2}{2~l_{HD}~l_{DG}}) + acos(\frac{l_{DG}^2 + l_{DF}^2 - l_{GF}^2}{2~l_{DG}~l_{DF}}) + \nonumber \\
& acos(\frac{l_{BD}^2 + l_{HD}^2 - l_{BH}^2}{2~l_{BD}~l_{HD}})]
\end{align}

So far, analytical kinematic loops are selected similar to $Loop~1$ and $Loop~4$ depicted in Fig~\ref{fig:loops}, which was designed for numerical kinematics approach. Such a similarity is not a surprise, since the previous loops were selected as simple as possible. However, numerical and analytical approaches start to vary in terms of kinematics loops from this point on. Obtaining $q_K$ using Equation~(\ref{eq:analytical_qK}) allows the position of points $L$, $I$ and $Ip$ to be known as well. In fact, point $Ip$ is defined as the projection of point $I$ along the plane where point $L$ is placed and is parallel to finger surface.

Since point $Ip$ is defined as the projection of point $I$, the distance between points $L$ and $Ip$ can be calculated directly using Pythagorean Theorem to define passive linear displacement $c_1$ through Equation~(\ref{eq:analytical_c1}).

\begin{align}
\label{eq:analytical_c1}
c_1 &= \sqrt{l_{LI}^2 - l_{IIp}^2}
\end{align}

Simultaneously, the rotation around the MCP joint $q_{o1}$ can be calculated using Equation~(\ref{eq:analytical_qo1}) simply by combining Pythagorean and Cosine Theorems.

\begin{align}
\label{eq:analytical_qo1}
q_{o1} &= acos(\frac{l_{LI}^2 + c_1^2 - l_{IIp}^2}{2~l_{LI}~c_1}) + atan(\frac{yL - yI}{xL - xI})
\end{align}

Finally, obtaining $q_D$ using Equation~(\ref{eq:analytical_qD}) and $q_{o1}$ using Equation~(\ref{eq:analytical_qo1}) allow the positions of points $J$, $Jp$ and $M$ that are shown in Figure ~\ref{fig:analytical_v4}. In particular, point $J$ is defined thanks to $q_D$, while points $Jp$ and $M$ are formed by $q_{o1}$.

Similar to the previous loop, since point $Jp$ is defined as the projection of point $J$, the distance between points $M$ and $Jp$ can be calculated directly using Pythagorean Theorem to define passive linear displacement $c_2$ using Equation~(\ref{eq:analytical_c2}).

\begin{align}
\label{eq:analytical_c2}
c_2 &= \sqrt{l_{JM}^2 - l_{JJp}^2}
\end{align}

Simultaneously, rotation around PIP joint $q_o2$ can be calculated using Equation~(\ref{eq:analytical_qo1}) simply by combining Pythagorean and Cosine Theorems.

\begin{align}
\label{eq:analytical_qo2}
q_{o2} &= acos(\frac{l_{JM}^2 + c_2^2 - l_{JJp}^2}{2~l_{JM}~c_2}) + atan(\frac{yJ - yM}{xJ - xM}) - q_{o1}
\end{align}

\begin{figure}[hbt]
\centering
\includegraphics[width=0.8\textwidth]{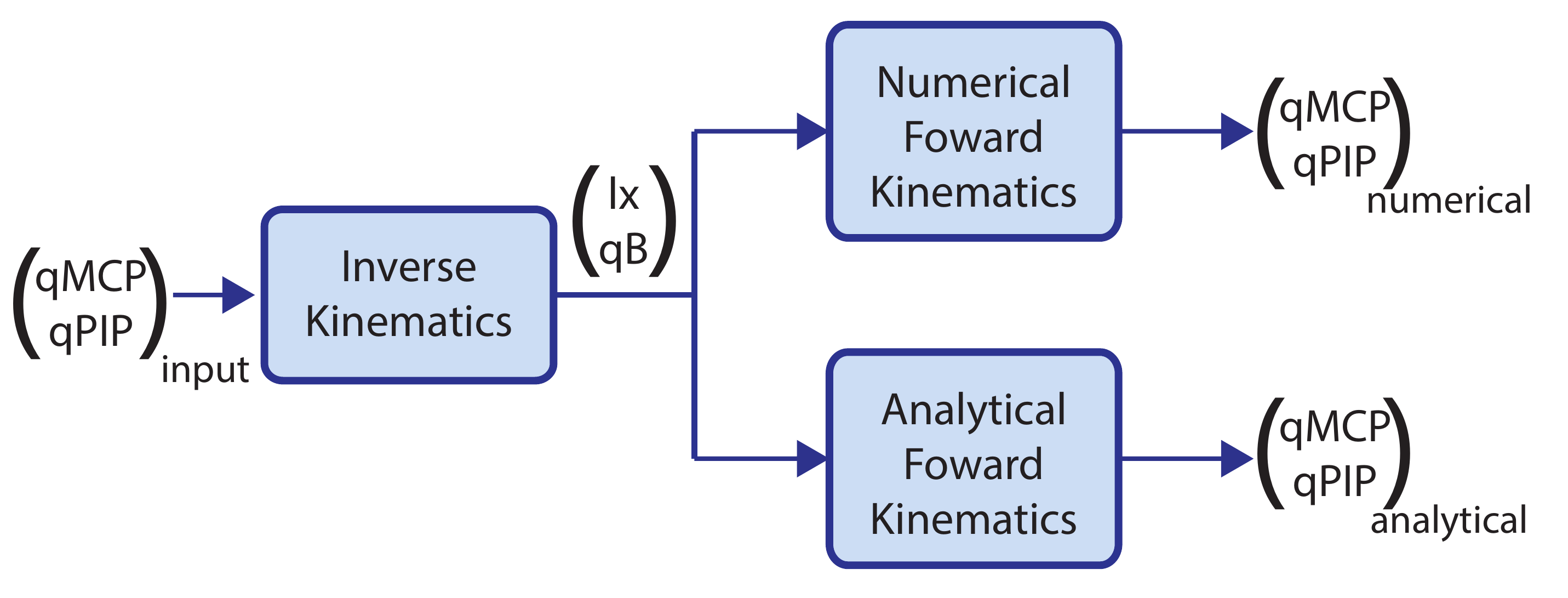}
\caption{Simulation setup to compare numerical and analytical forward kinematic analyses using the same measurement set, which is computed by inverse kinematics for a set of finger joint rotations.}
\label{fig:block_forward}
%\vspace*{2\baselineskip}
\end{figure}

Both numerical and analytical methods to achieve a unique solution for finger pose through forward kinematics should be verified. With this motivation, a simulation setup was designed as can be summarized in Figure~\ref{fig:block_forward}. In particular, inverse kinematics was run to calculate meaningful sensory information of $l_x$ and $q_B$ to reach $70^o$ and $90^o$ DoF along MCP and PIP joints respectively. The finger path was chosen the same as the previous experiment shown in Figure~\ref{fig:CadComp_qo} to achieve compatible simulation results, even though such a verification was performed for many finger curling trajectories. Then, the sensory data obtained from inverse kinematics were used to run numerical and analytical forward kinematics simultaneously to compare their results.

\begin{figure}[hbt]
\centering
\includegraphics[width=1\textwidth]{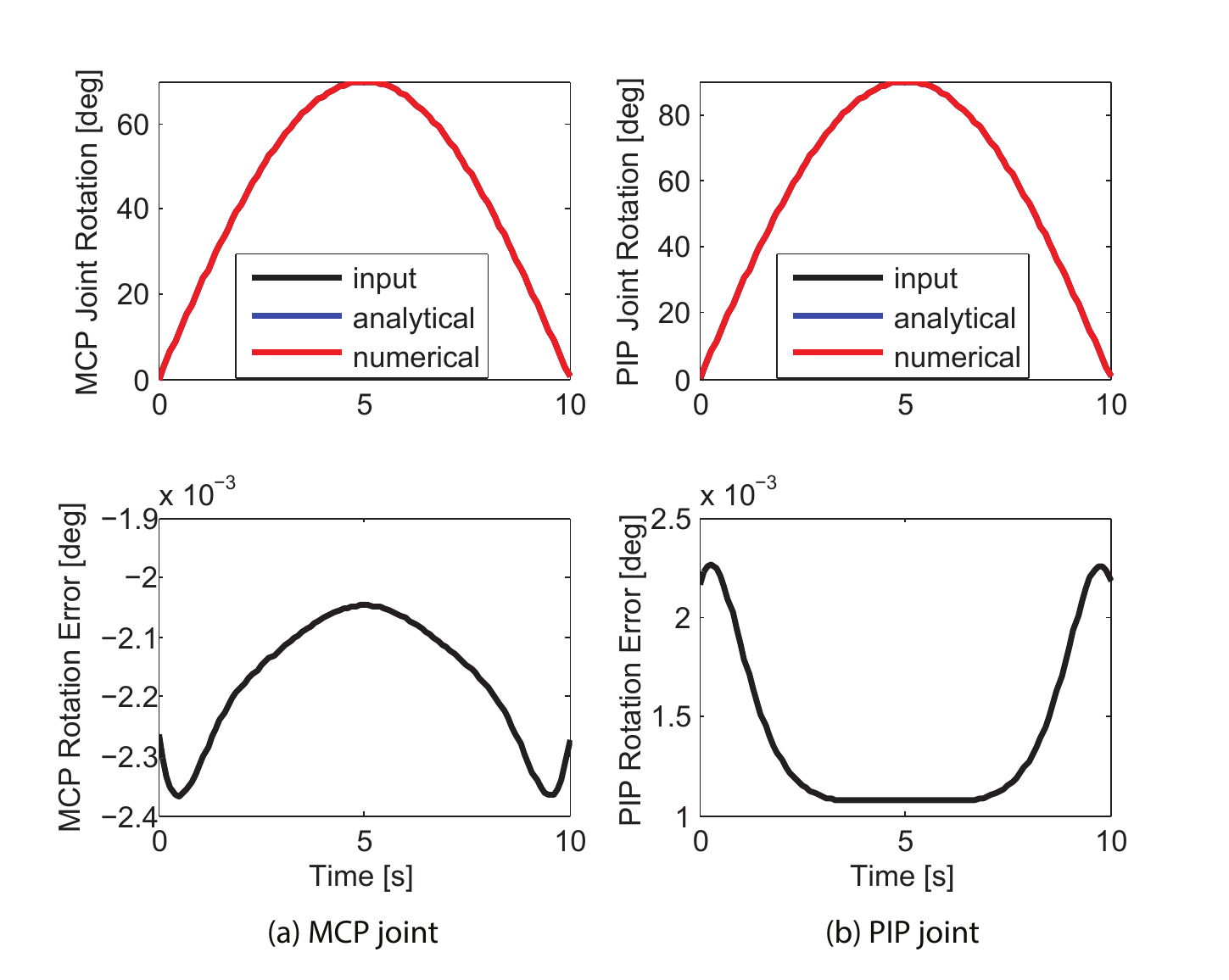}
\caption{Comparison between numerical and analytical forward kinematics in simulation: (a) comparison for MCP joint rotation and (b) comparison for PIP joint rotation.}
\label{fig:numvsan}
%\vspace*{2\baselineskip}
\end{figure}

Figure~\ref{fig:numvsan} shows the MCP and the PIP joint rotations that are given as input, and are calculated from analytical and numerical forward kinematics methods. The error plot shows the error between numerical and analytical methods only, to focus on the variations caused by running different analyses. The obtained error between the two methods are small enough to be neglected and to assume that these two methods provide the same finger pose. The advantages and disadvantages of these techniques will be discussed later while proposing to estimate finger pose during real time tasks.

\newpage
\subsection{Differential Kinematics} \label{sec:jacobian}

Inverse and forward kinematics algorithm provide a relationship between input and output joints in the configuration level, while differential kinematics can be used to set a similar relationship in the velocity level. Such mapping is described by a matrix, based on the instant orientation of the system. %Furthermore, the same matrix might be used to express the statics of the overall system by relating the input and output forces.

Differential kinematics can be calculated simply using the derivative of previously expressed nonlinear equations in Equation~\ref{eq:loop1} - Equation \ref{eq:loop4}. For an underactuated system, where actuator force is transmitted to finger joints by designing more passive joints, differential kinematics also focuses on extracting these passive joints in the velocity level. In order to simplify differential kinematics, finger joint velocities ($\vect{\dot{q}_{fin}}$), measured joint velocities ($\vect{\dot{q}_m}$) and passive joint velocities ($\vect{\dot{q}_p}$) can be defined as presented in Equation~(\ref{eq:joint_def}).

\begin{equation} \label{eq:joint_def}
\vect{\dot{q}_{fin}} =
\left[
\begin{array}{r}
\dot{q}_{o1} \\
\dot{q}_{o2}
\end{array}
\right] \qquad \quad
\vect{\dot{q}_{m}} =
\left[
\begin{array}{r}
\dot{l}_{x} \\
\dot{q}_{B}
\end{array}
\right] \qquad \quad
\vect{\dot{q}_{p}} =
\left[
\begin{array}{r}
\dot{q}_{K} \\
\dot{q}_{D} \\
\dot{q}_{G} \\
\dot{q}_{N} \\
\dot{c}_{1} \\
\dot{c}_{2} \\
\end{array}
\right]
\end{equation}

\noindent Differentiating Equation~\ref{eq:loop1}) - Equation~(\ref{eq:loop4} can be categorized in the matrix form based on finger joint velocities, measured joint velocities and passive joint velocities as Equation~(\ref{eq:matrix_def1}).

\begin{equation} \nonumber \label{eq:matrix_def1}
\left[
\begin{tabular}{r r}
$O_{11}$ & $O_{12}$ \\
$O_{21}$ & $O_{22}$ \\ \specialrule{2.5pt}{1pt}{1pt}
$O_{31}$ & $O_{32}$ \\
$O_{41}$ & $O_{42}$ \\
$O_{51}$ & $O_{52}$ \\
$O_{61}$ & $O_{62}$ \\
$O_{71}$ & $O_{72}$ \\
$O_{81}$ & $O_{82}$
\end{tabular} \right]
\left[
\begin{array}{r}
\dot{q}_{o1} \\
\dot{q}_{o2}
\end{array}
\right] = \left[
\begin{tabular}{c c?c c c c c c}
$R_{11}$ & $R_{12}$ & $R_{13}$ & $R_{14}$ & $R_{15}$ & $R_{16}$ & $R_{17}$ & $R_{18}$ \\
$R_{21}$ & $R_{22}$ & $R_{23}$ & $R_{24}$ & $R_{25}$ & $R_{26}$ & $R_{27}$ & $R_{28}$ \\ \specialrule{2.5pt}{1pt}{1pt}
$C_{31}$ & $C_{32}$ & $C_{33}$ & $C_{34}$ & $C_{35}$ & $C_{36}$ & $C_{37}$ & $C_{38}$ \\
$C_{41}$ & $C_{42}$ & $C_{43}$ & $C_{44}$ & $C_{45}$ & $C_{46}$ & $C_{47}$ & $C_{48}$ \\
$C_{51}$ & $C_{52}$ & $C_{53}$ & $C_{54}$ & $C_{55}$ & $C_{56}$ & $C_{57}$ & $C_{58}$ \\
$C_{61}$ & $C_{62}$ & $C_{63}$ & $C_{64}$ & $C_{65}$ & $C_{66}$ & $C_{67}$ & $C_{68}$ \\
$C_{71}$ & $C_{72}$ & $C_{73}$ & $C_{74}$ & $C_{75}$ & $C_{76}$ & $C_{77}$ & $C_{78}$ \\
$C_{81}$ & $C_{82}$ & $C_{83}$ & $C_{84}$ & $C_{85}$ & $C_{86}$ & $C_{87}$ & $C_{88}$
\end{tabular} \right]
\left[
\begin{array}{r}
\dot{l}_{x} \\
\dot{q}_{B} \\ \specialrule{2.5pt}{1pt}{1pt}
\dot{q}_{K} \\
\dot{q}_{D} \\
\dot{q}_{G} \\
\dot{q}_{N} \\
\dot{c}_{1} \\
\dot{c}_{2} \\
\end{array}
\right]
\end{equation}

Dealing with each joint velocities individually help Jacobian to achieve its components to form a $8~x~8$ matrix and complementary output matrix to form a $8~x~2$ shape. Equation~(\ref{eq:matrix_def1}) can be simplified as Equation~(\ref{eq:matrix_def2}) simply by merging finger joint velocities, measured velocities and passive joint velocities as $\vect{\dot{q}_m}$, $\vect{\dot{q}_{fin}}$ and $\vect{\dot{q}_p}$. The divisions for Jacobian and output matrices are shown with bold lines crossing through Equation~(\ref{eq:matrix_def1}). Table~\ref{tab:jacob_sizes} represent the sizes of simplified matrices to clarify the definitions.

\begin{equation} \label{eq:matrix_def2}
\left[
\begin{array}{r}
\vect{J}_{O_{m}} \\
\vect{J}_{O_{p}}
\end{array} \right]
\vect{\dot{q}_{fin}} = \left[
\begin{array}{rr}
\vect{J}_{R_{m}} & \vect{J}_{R_{p}} \\
\vect{J}_{C_{m}} & \vect{J}_{C_{p}}
\end{array} \right]
\begin{bmatrix}
\vect{\dot{q}_{m}} \\
\vect{\dot{q}_{p}}
\end{bmatrix}
\end{equation}

\begin{table}[h!]
\caption{Matrix sizes of Jacobian matrix components.}
\label{tab:jacob_sizes}
\begin{center}
\begin{tabular}{c || c}
\hline\hline \\
Matrix Name & Size \\ [1ex] \hline \\
$\vect{J}_{O_m}$ & $2 \times 2$ \\ [1ex] \hline \\
$\vect{J}_{O_p}$ & $6 \times 2$ \\ [1ex] \hline \\
$\vect{J}_{R_{m}}$ & $2 \times 2$ \\ [1ex] \hline \\
$\vect{J}_{R_{p}}$ & $2 \times 6$ \\ [1ex] \hline \\
$\vect{J}_{C_{m}}$ & $6 \times 2$ \\ [1ex] \hline \\
$\vect{J}_{C_{p}}$ & $6 \times 6$ \\ [1ex]
\hline \hline
\end{tabular}
\end{center}
\end{table}

The order of equations were chosen such that none of the sub-matrix components are zero matrices, where all the elements are $0$, even though they can have $0$ value in their elements. In particular, matrix components $\vect{J}_{._m}$ indicate coefficients regarding the measured variables while components $\vect{J}_{._p}$ signify passive variables. Similarly, $\vect{J}_{O_.}$, $\vect{J}_{C_.}$ and $\vect{J}_{R_.}$ indicate output coefficients, constraint coefficients and real relationship coefficients.

A Jacobian matrix is needed to form a relationship between finger joint velocities $\dot{q}_{fin}$ and measured joint velocities $\dot{q}_m$, while passive joint velocities $\dot{q}_p$ can be considered as redundant information regarding mechanical behavior. In order to simplify Jacobian for a $2~x~2$ matrix between $\dot{q}_{fin}$ and $\dot{q}_m$, by constraining $\dot{q}_p$ using the bottom raw of Equation~\ref{eq:matrix_def2}:

\begin{equation}
 \vect{J}_{O_p}~\vect{\dot{q}_{fin}} = \vect{J}_{C_m}~\vect{\dot{q}_m} + \vect{J}_{C_p}~\vect{\dot{q}_p},
\end{equation}

\noindent and define $\dot{q}_p$ using $\dot{q}_{fin}$ and $\dot{q}_m$ as in:

\begin{equation}
\vect{\dot{q}_p} = \vect{J}^{-1}_{C_{p}}~[\vect{J}_{O_{p}}~\vect{\dot{q}_{fin}} - \vect{J}_{C_{m}}~\vect{\dot{q}_{m}}]
\end{equation}

The expression of $\dot{q}_p$ can be replaced on the top raw of Equation~(\ref{eq:matrix_def2}):

\begin{equation}
 \vect{J}_{O_m}~\vect{\dot{q}_{fin}} = \vect{J}_{R_m}~\vect{\dot{q}_m} + \vect{J}_{R_p}~\vect{\dot{q}_p} \nonumber
\end{equation}

\begin{equation}
 \vect{J}_{O_m}~\vect{\dot{q}_{fin}} = \vect{J}_{R_m}~\vect{\dot{q}_m} + \vect{J}_{R_p}~[\vect{J}^{-1}_{C_p}~[\vect{J}_{O_p}~\vect{\dot{q}_{fin}} - \vect{J}_{C_m}~\vect{\dot{q}_m}]],
\end{equation}

\noindent which can be simplified into overall Jacobian definition:

\begin{equation}
[\vect{J}_{O_m} - \vect{J}_{R_p}~\vect{J}^{-1}_{C_p}~\vect{J}_{O_p}]~\vect{\dot{q}_{fin}} = [\vect{J}_{R_m} - \vect{J}_{R_p}~\vect{J}^{-1}_{C_p}~\vect{J}_{C_m}]~\vect{\dot{q}_m}. \nonumber
\end{equation}

\begin{align} \label{eq:jacobian}
\nonumber \vect{\dot{q}_{fin}} &= [\vect{J}_{O_m} - \vect{J}_{R_{p}}~\vect{J}^{-1}_{C_{p}}~\vect{J}_{O_p}]^{-1} [\vect{J}_{R_{m}} - \vect{J}_{R_{p}}~\vect{J}^{-1}_{C_{p}}~\vect{J}_{C_{m}}]~\vect{\dot{q}_{m}}
\\
 &= \vect{J}_A~\vect{\dot{q}_{m}}
\end{align}

\begin{figure}[htb]
\centering
\includegraphics[width=0.8\textwidth]{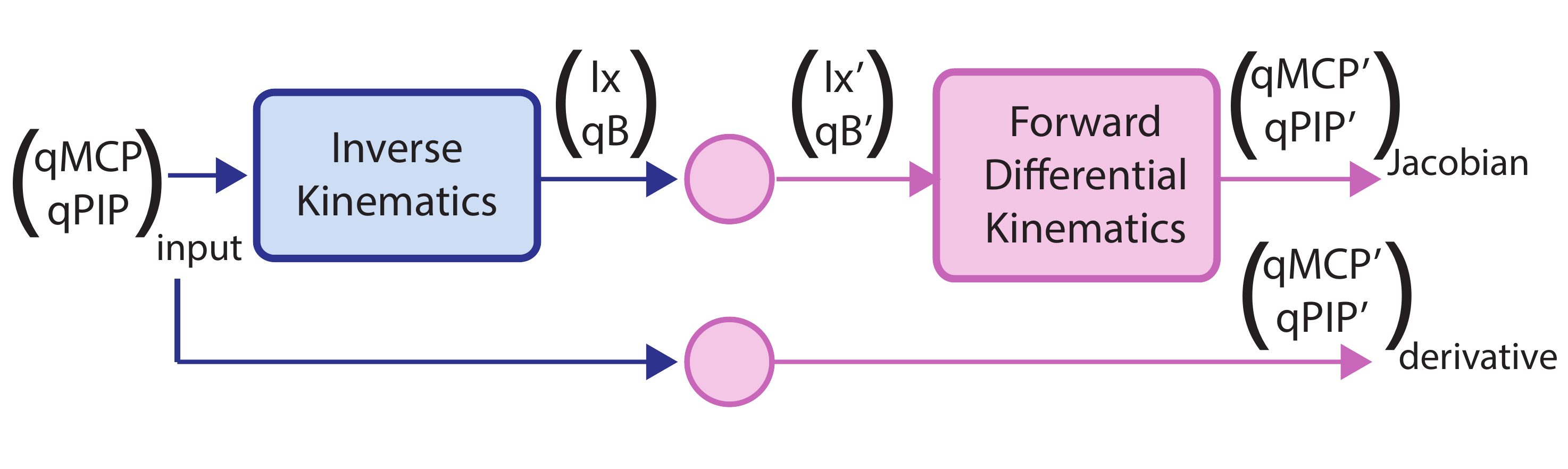}
\caption{Simulation setup to compare differential kinematics to simple derivative of a set of finger joint rotation.}
\label{fig:block_jac}
\end{figure}

The Jacobian obtained by Equation~(\ref{eq:jacobian}) provides a relation between velocities of measured data and finger joint velocities. A simulation setup was designed as in Figure~\ref{fig:block_jac} to compare the outputs of the Jacobian to the derivative of finger joints and verify the efficacy of Jacobian matrix with respect to other analyses.

\begin{figure}[htb]
\centering
\includegraphics[width=1\textwidth]{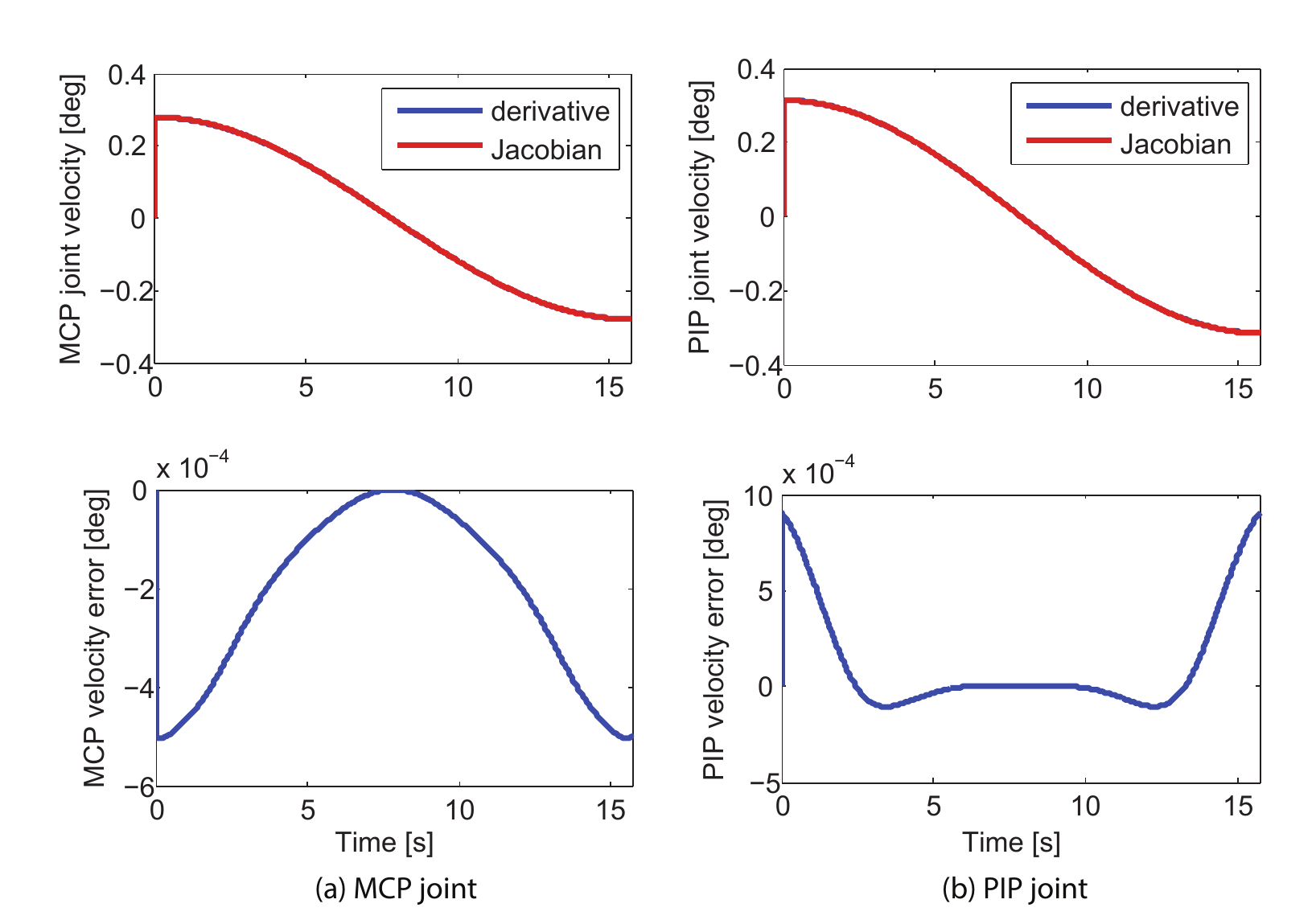}
\caption{Comparison between differential kinematics and simple derivative in simulation: (a) comparison for MCP joint rotation and (b) comparison for PIP joint rotation.}
\label{fig:dervsjac}
\end{figure}

In particular, Jacobian calculation was validated by comparing finger joint velocities. To achieve the corresponding sensory measurements $l_x$ and $q_B$, inverse kinematics were run previously. Jacobian matrix can reach numerical values using the orientation of the device, therefore passive joints are extracted from inverse kinematics as well. As a result, finger joint velocities ($\dot{q}_{o1}, \dot{q}_{o2}$) are compared between Jacobian calculation and derivative of the input. Figure ~\ref{fig:dervsjac} shows the comparison between joint velocities obtained through different methods and the error between them show sufficient evidence to assume their unity.

The same Jacobian matrix provides another relation between forces being applied to finger joints and forces being applied by actuators:

\begin{equation} \label{eq:jacob_tr}
\left[
\begin{array}{r}
F_A \\
\tau_B
\end{array} \right]
= \vect{J}^T_A~
\begin{bmatrix}
\tau_{1} \\
\tau_{2}
\end{bmatrix}.
\end{equation}

\noindent where $\tau_{1}, \tau_{2}$ signify torques around the finger joints MCP and PIP correspondingly, $\tau_B$ signifies output torques around joint $B$, and $F_A$ signifies forces of linear actuator. It is important to know that the device is controlled only by linear actuator while the joint along $B$ is left passive. The force constraint for the given system by the underactuation concept can mathematically be expressed as $\tau_B = 0$. Even though Jacobian matrix might result $\tau_B$ to reach hypothetical non-zero values, that cannot be provided to the system due to the passivity imposed by underactuation.

%\newpage
\subsection{Statics Analysis and Stability of Grasp}\label{sec:statics}

Stability and safety of grasping tasks can be guaranteed only if transmitted forces are applied in the correct direction to provide an interaction between the grasping object and user's finger. Since the proposed underactuated mechanism does not control forces acting on finger phalanges independently, a static analysis is crucial to ensure stability of grasping tasks at any pose of the mechanism.

The static analysis to ensure stability of grasping forces was already introduced previously for a fully mechanical underactuated gripper~\cite{Gosselin2003} through the formulation $\vect{f}_{fin} = {\vect{J}_{T}^{-T}} {\vect{J}_{A}^{-T}} \vects{\tau}_{m}$, where $\vect{f}_{fin}$ is the vector of forces between finger phalanges to the grasping object, $\vects{\tau}_{m}$ is the dual force vector of $\vect{\dot{q}}_m$, $\vect{J}_{T}^{-T}$ is the inverse transpose of Jacobian between angular velocities around finger phalanges and linear velocities at the contact points and ${\vect{J}_{A}^{-T}}$ is the inverse transpose of Jacobian calculated in Equation~(\ref{eq:jacobian}). However, the static analysis in this work aims to control the force transmission on phalanges in terms of finger joint torques $\vects{\tau}_{fin}=[\tau_{1}; \tau_2]$ in order to optimize link lengths of the mechanism. The force transmission is simplified and can be obtained as Equation~(\ref{eq:static}) using the inverse Jacobian transpose (${\vect{J}_{A}^{-T}}$) in Equation~(\ref{eq:jacobian}).

\begin{align}
\label{eq:static}
\vects{\tau}_{fin} = {\vect{J}_{A}^{-T}} \vects{\tau}_{m}
\end{align}

\noindent where $\vects{\tau}_{m}=[F_A; \tau_B]$, and since it can be assumed that $\tau_B=0$, we obtain a direct relationship between actuator force and contact forces. The analysis of sign of $\tau_1$ and $\tau_2$ allows us to study stability of grasp assisted by actuator being in contact with an object.

\newpage

\section{Link Length Optimization} \label{sec:optimization}

The proposed underactuated hand exoskeleton is designed as a linkage based mechanism and the performance of such mechanisms is known to be affected directly by adjusting their link lengths. Even though the first sketch of the proposed kinematics is drawn with random link lengths, the link lengths of each finger components need to be optimized to improve the overall performance and ensure to satisfy natural RoM of MCP and PIP joints. Yet, the mechanical behavior of the device highlights some physical constraints that need to be satisfied:

\begin{itemize}
\item The device is connected to finger phalanges with passive linear sliders. Since these sliders have to be fitted on the finger phalanges, their movements ($c_1$ and $c_2$) have to be limited by average finger phalange lengths.
\item The closing/opening of the hand is performed by transmitting actuator forces to finger joints. To provide a stable grasping, two forces acting on MCP and PIP joints have to be balanced and should be always directed with the same sign, towards opening or closure of the hand.
\item The required actuator displacement to satisfy all combinations of natural RoM for finger joints within his physical stroke limitations.
\end{itemize}

 An optimization procedure was conducted by an extensive search procedure to find a set of link lengths that maximizes the following cost function $p$:

\begin{numcases} { \max~p = \sqrt{\tau_1^2 + \tau_2^2}, \text{ such that: } \label{eq:opt}}
0\leq l_x\leq l_{max} \nonumber \\
0\leq c_1\leq c_{1max} \nonumber \\
0\leq c_2\leq c_{2max} \nonumber \\
0.25\leq \tau_1/\tau_2 \leq 7.5 \nonumber
\end{numcases}

\noindent where $\tau_1$ and $\tau_2$ are MCP and PIP joint torques, $c_{1max}$ and $c_{2max}$ are maximum displacements along passive sliders, which are set as $50~mm$ and $40~mm$, and $l_{max}$ is the maximum stroke capacity of a linear actuator, which is set as $50~mm$ due to the actuator choice. The limitation on $c_{2max}$ might be bigger than the length of middle finger phalange for many users, but the displacement along this joint was found to be crucial to reach natural RoM for PIP joint. Since there is no mechanical equipment on the third finger phalange to cause a mechanical interference, it does not create a practical issue.

%The search space is defined by the constant link length parameters defined in the third column of Table~\ref{table_parameters}.
Before the optimization, a set of initial lengths were selected to define a reasonable range of link lengths belonging to the search space, than the following computational steps were followed:

\begin{algorithm}
\caption{Link Length Optimization Algorithm}\label{alg:stiffnessalg2}
\begin{algorithmic}[1]
\State * Define a feasible and wide link length range
\For $q_{o1} = 0:1:80^o$
\For $q_{o2} = 0:1:90^o$
\State Calculate the movement on the passive prismatic joints $c_1$, $c_2$ and actuator $l_x$
\begin{itemize}
\item Control if $l_x$, $c_1$ and $c_2$ satisfy the physical limits
\item Go to the next set and start from 2 if not satisfied.
\end{itemize}
\State Compute the torques on the finger joints if $1~N.$ force is applied from the actuator using Jacobian transpose
\begin{itemize}
\item Control if the ratio between the torques of two joints satisfy the predefined limits
\item Go to the next set and start from 2 if not satisfied.
\end{itemize}
\State Calculate the optimization objective
$p = \sqrt{\tau_1^2 + \tau_2^2}$
\EndFor
\EndFor
\end{algorithmic}
\end{algorithm}

For the sake of simplifying the search space of the optimization, a sensitivity analysis was conducted to identify the link length variables that do not affect the optimization procedure significantly and so reduce the dimension of the search space.

%\newpage
\subsection{Sensitivity Analysis}

Although numerical or analytical derivatives of the overall cost function $p$ with respect to each search parameter would provide an efficient sensitivity analysis approach, the derivatives are not easy to obtain for complex non linear models. The one-at-a-time (OAT) sensitivity analysis is an alternative method for analyzing the effect of a single parameter on a cost function, keeping other parameters fixed~\cite{Mouida2011}. For this purpose, a sensitivity index ($SI$) is expressed in Equation~(\ref{eq:si}).

\begin{equation}\label{eq:si}
SI = \displaystyle \frac{\frac{S_2-S_1}{S_{av}}}{\frac{E_2-E_1}{E_{av}}}
\end{equation}

\noindent where $SI$ is the sensitivity index of the model output, $E_1$ and $E_2$ are minimum and maximum values of input parameters; $S_1$ and $S_2$ are corresponding output values for $E_1$ and $E_2$; $S_{av}$ and $E_{av}$ are the average values of input and output parameters respectively. This index provides a quantitative relation between model outputs and input variables in terms of sensitivity. Negative $SI$ indicates that the inputs and outputs vary in opposite directions, while positive values signify a change in the same trend. In particular, the aim is to choose the positive, higher than unity $SI$ values, which show that a change in the parameter creates a higher effect in the output.

\begin{figure}[htb]
\centering
% \vspace*{-2\baselineskip}
\resizebox{4in}{!}{\includegraphics{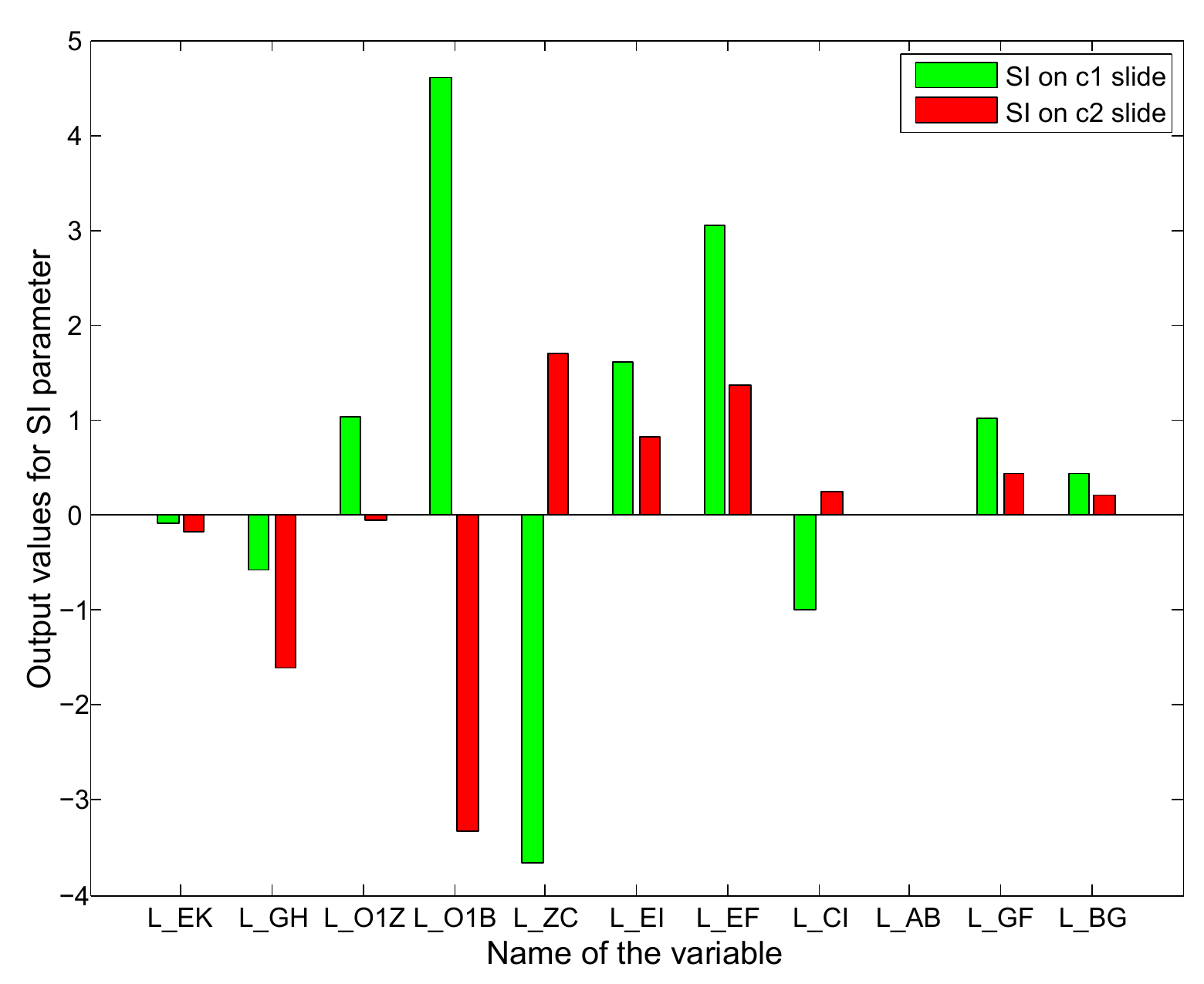}}
\caption{Computation of sensitivity index for first passive slider movement $c_1$ and second passive slider movement $c_2$ independently.}
\label{fig:sensit}
%\vspace*{-1\baselineskip}
\end{figure}

While closing/opening human finger, the most crucial output values were stated as displacements along passive linear sliders ($c_1$ and $c_2$) due to the limitations imposed by finger phalange lengths. Therefore, the impact of each variable on these values was investigated individually. A representative pose of finger was used for this analysis. During the sensitivity analysis, each input variable was changed by $\pm 10\%$ from the initial value, keeping other variables constant. For each set of lengths, the variations of linear displacements $c_1$ and $c_2$ were calculated using the pose analysis discussed in Section~\ref{sec:kinematics}. The sensitivity index was calculated individually for $c_1$ and $c_2$ where the analysis limit is set to $+10\%$ for both cases. In other words, the length parameters, which result in negative or $\leq 0.1$ SI values, are not considered as effective variables over the performance of such linear movement. Figure~\ref{fig:sensit} represents sensitivity analysis results for $c_1$ and $c_2$ displacement individually, calculated as $SI_{c_1}$ and $SI_{c_2}$. Note that the bars with values under $0.001$, for instance, values for $L_{AB}$, cannot be seen in the plot.

It can be easily observed that for some variables, the sign of SI values are different for $c_1$ and $c_2$ displacements. When it happens, it means that increasing a link length affects sliders along first and second finger phalanges in different ways. Equation~(\ref{eq:si2}) combines SI values computed for both sliders. $SI_{g}$ computed from Equation~(\ref{eq:si2}) with positive values indicates similar output effect on both sliders, while the one with negative values indicate different behaviours. Table~\ref{tab:sens} presents output values of $SI_{g}$ from Equation~(\ref{eq:si2}) for each link length variation.

\begin{equation}\label{eq:si2}
SI_{g} = \text{sign}(SI_{c1})\cdot\text{sign}(SI_{c2}) \sqrt{SI_{c1}^2 + SI_{c2}^2}
\end{equation}

\begin{table}[htb]
%\vspace*{-1\baselineskip}
\caption{Results of generic sensitivity index.}
\label{tab:sens}
\begin{center}
\begin{tabular}{c c||c c}
\hline \hline \\
variable & value [mm] & variable & value [mm] \\ [1ex] \hline \hline \\
$SI_{EJ}$ & 0.1939 & $SI_{ED}$ & 3.3437 \\ [1ex]
$SI_{CI}$ & 1.7077 & $SI_{GF}$ & -1.0275 \\ [1ex]
$SI_{KH}$ & -1.0298 & $SI_{AB}$ & -2.1864e-09 \\ [1ex]
$SI_{KB}$ & -5.6804 & $SI_{CD}$ & 1.1021 \\ [1ex]
$SI_{GH}$ & -4.0324 & $SI_{BC}$ & 0.4877 \\ [1ex]
$SI_{EF}$ & 1.8146 & & \\
\hline \hline
%\hline
\end{tabular}
\end{center}
%\vspace*{-2\baselineskip}
\end{table}

From the sensitivity analysis, the most efficient variables are obtained as $L_{EJ}$, $L_{CI}$, $L_{EF}$, $L_{ED}$, $L_{CD}$ and $L_{BC}$ as can be seen in Table~\ref{tab:sens}. It is important to note that, the negative values represent different impacts on $c_1$ and $c_2$ slider displacements as a result for increased link length. These lengths are excluded from parameter search space since minimizing two of the slide displacements simultaneously is not feasible. These constant values of $6$ variables for index, middle, ring and little fingers were set constant as reported in Table~\ref{tab:constants_ind} - Table \ref{tab:constants_lit} during optimization.

\begin{table}[!htb]
%\vspace*{-1\baselineskip}
\caption{Constant variables for the index finger.}
\label{tab:constants_ind}
\begin{center}
\begin{tabular}{c c||c c}
\hline \hline \\
variable & range [mm] & variable & range [mm] \\ [1ex]
\hline \hline \\
$L_{KH}$ & 72 & $L_{KB}$ & 35 \\ [1ex]
$L_{GH}$ & 86 & $L_{AB}$ & 18 \\ [1ex]
$L_{GF}$ & 46 & & \\
\hline \hline
\end{tabular}
\end{center}
%\vspace*{-2\baselineskip}
\end{table}

\begin{table}[htb]
\caption{Constant variables for the middle finger.}
\label{tab:constants_mid}
\begin{center}
\begin{tabular}{c c||c c}
\hline \hline \\
variable & value [mm] & variable & value [mm] \\ [1ex]
\hline \hline \\
$L_{KH}$ & 72 & $L_{KB}$ & 35 \\ [1ex]
$L_{GH}$ & 86 & $L_{AB}$ & 18 \\ [1ex]
$L_{GF}$ & 46 \\ \hline \hline
\end{tabular}
\end{center}
\end{table}

\begin{table}[htb]
\caption{Constant variables for the ring finger.}
\label{tab:constants_ring}
\begin{center}
\begin{tabular}{c c||c c}
\hline \hline \\
variable & value [mm] & variable & value [mm] \\ [1ex]
\hline \hline \\
$L_{KH}$ & 72 & $L_{KB}$ & 35 \\ [1ex]
$L_{GH}$ & 86 & $L_{AB}$ & 18 \\[1ex]
$L_{GF}$ & 46 & & \\ \hline \hline
\end{tabular}
\end{center}
\end{table}

\begin{table}[htb]
\caption{Constant variables for the little finger.}
\label{tab:constants_lit}
\begin{center}
\begin{tabular}{c c||c c}
\hline \hline \\
variable & value [mm] & variable & value [mm] \\ [1ex]
\hline \hline \\
$L_{KH}$ & 72 & $L_{KB}$ & 35 \\ [1ex]
$L_{GH}$ & 86 & $L_{AB}$ & 18 \\ [1ex]
$L_{GF}$ & 46 & & \\
\hline \hline
\end{tabular}
\end{center}
\end{table}

\newpage
\subsection{Pre-optimization Constraints}

 A preliminary optimization procedure was conducted to select the combinations of link length parameters, satisfying displacement and static constraints among link lengths for index, middle, ring and little fingers as ranged in Table~\ref{tab:ranges_index} - Table~\ref{tab:ranges_little}. The variables, which were chosen after the sensitivity analysis, were iterated with a difference of $1~mm$ within the given range throughout the optimization. For the pre-optimization procedure, MCP and PIP joints are moved in different paths for the corresponding iteration set to check whether linear and static constraints are satisfied. The length combinations are eliminated from the optimization performance if the constraints above are not satisfied.

\begin{table}[htb]
%\vspace*{-1\baselineskip}
\caption{Range of variables for index finger optimization.}
\label{tab:ranges_index}
\begin{center}
\begin{tabular}{c c||c c}
\hline \hline \\ [1ex]
variable & range [mm] & variable & range [mm] \\ [1ex]
\hline \hline \\ [1ex]
$L_{EJ}$ & 30 - 48 & $L_{ED}$ & 35 - 45 \\ [1ex]
$L_{CI}$ & 16 - 20 & $L_{EF}$ & 20 - 35 \\ [1ex]
$L_{CD}$ & 9 - 20 & $L_{BC}$ & 36 - 46 \\ [1ex]
\hline \hline
\end{tabular}
\end{center}
%\vspace*{-2\baselineskip}
\end{table}

\begin{table}[htb]
%\vspace*{-1\baselineskip}
\caption{Range of variables for middle finger optimization. }
\label{tab:ranges_middle}
\begin{center}
\begin{tabular}{c c||c c}
\hline \hline \\ [1ex]
variable & range [mm] & variable & range [mm] \\ [1ex]
\hline \hline \\ [1ex]
$L_{EJ}$ & 30 - 48 & $L_{ED}$ & 40 - 55 \\ [1ex]
$L_{CI}$ & 16 - 20 & $L_{EF}$ & 15 - 30 \\ [1ex]
$L_{CD}$ & 9 - 20 & $L_{BC}$ & 36 - 46 \\ [1ex]
\hline \hline
\end{tabular}
\end{center}
%\vspace*{-2\baselineskip}
\end{table}

\begin{table}[htb]
%\vspace*{-1\baselineskip}
\caption{Range of variables for ring finger optimization.}
\label{tab:ranges_ring}
\begin{center}
\begin{tabular}{c c||c c}
\hline \hline \\ [1ex]
variable & range [mm] & variable & range [mm] \\ [1ex]
\hline \hline \\ [1ex]
$L_{EJ}$ & 30 - 48 & $L_{ED}$ & 35 - 45 \\ [1ex]
$L_{CI}$ & 16 - 20 & $L_{EF}$ & 20 - 35 \\ [1ex]
$L_{CD}$ & 9 - 20 & $L_{BC}$ & 36 - 46 \\ [1ex]
\hline \hline
\end{tabular}
\end{center}
%\vspace*{-2\baselineskip}
\end{table}

\begin{table}[htb]
%\vspace*{-1\baselineskip}
\caption{Range of variables for little finger optimization.}
\label{tab:ranges_little}
\begin{center}
\begin{tabular}{c c||c c}
\hline \hline \\ [1ex]
variable & range [mm] & variable & range [mm] \\ [1ex]
\hline \hline \\ [1ex]
$L_{EJ}$ & 30 - 48 & $L_{ED}$ & 30 - 40 \\ [1ex]
$L_{CI}$ & 16 - 20 & $L_{EF}$ & 20 - 35 \\ [1ex]
$L_{CD}$ & 9 - 20 & $L_{BC}$ & 36 - 46 \\ [1ex]
\hline \hline
\end{tabular}
\end{center}
%\vspace*{-2\baselineskip}
\end{table}

\subsubsection{Linear constraints}

A set of link lengths should not cause displacements of passive linear sliders, $c_1$ and $c_2$, to exceed limitations set for them based on finger phalange lengths. With this motivation, inverse kinematics in Section~\ref{sec:kinematics} should calculate variations of $c_1$ and $c_2$ as different joint trajectories are given as input to the kinematics analysis. A similar linear constraint should be set based on the required displacement for actuator to reach required MCP and PIP joints. The limitation of maximum displacement is set based on the properties of chosen actuator. An iteration is passed to the next step if both linear constraints are satisfied at all times, but is aborted from the overall optimization procedure and a new iteration is initialized with a new set of link lengths. Within the given range, $65~\%$ of the iteration sets were eliminated by the linear constraints of the index finger.

\subsubsection{Static constraints}

If the set of link lengths satisfy previous linear constraints, then the static constraints should be ensured during the iteration. Assuming that actuator applies $1~N.$ linear force, torques around MCP and PIP joints, $\tau_1$ and $\tau_2$, should be computed using Jacobian matrix in different mechanical configurations. A static constraint was set to ensure to have a balanced ratio between torques around MCP and PIP joints $\tau_1/\tau_2$ in different orientations of finger. Such a ratio was suggested to be limited by a minimum value of $0.25$ and a maximum value of $7.5$ without considering contact forces acting on the finger from the environment. Similarly, an iteration is passed to the next step if the static constraint is satisfied at all times, but is aborted from the overall optimization procedure and a new iteration is initialized with a new set of link lengths. Setting the statics constraint eliminates $90~\%$ of the remaining sets of variables for index finger.

%\newpage
\subsection{Optimization}

\begin{figure}[htb]
\centering
\resizebox{4.0in}{!}{\includegraphics{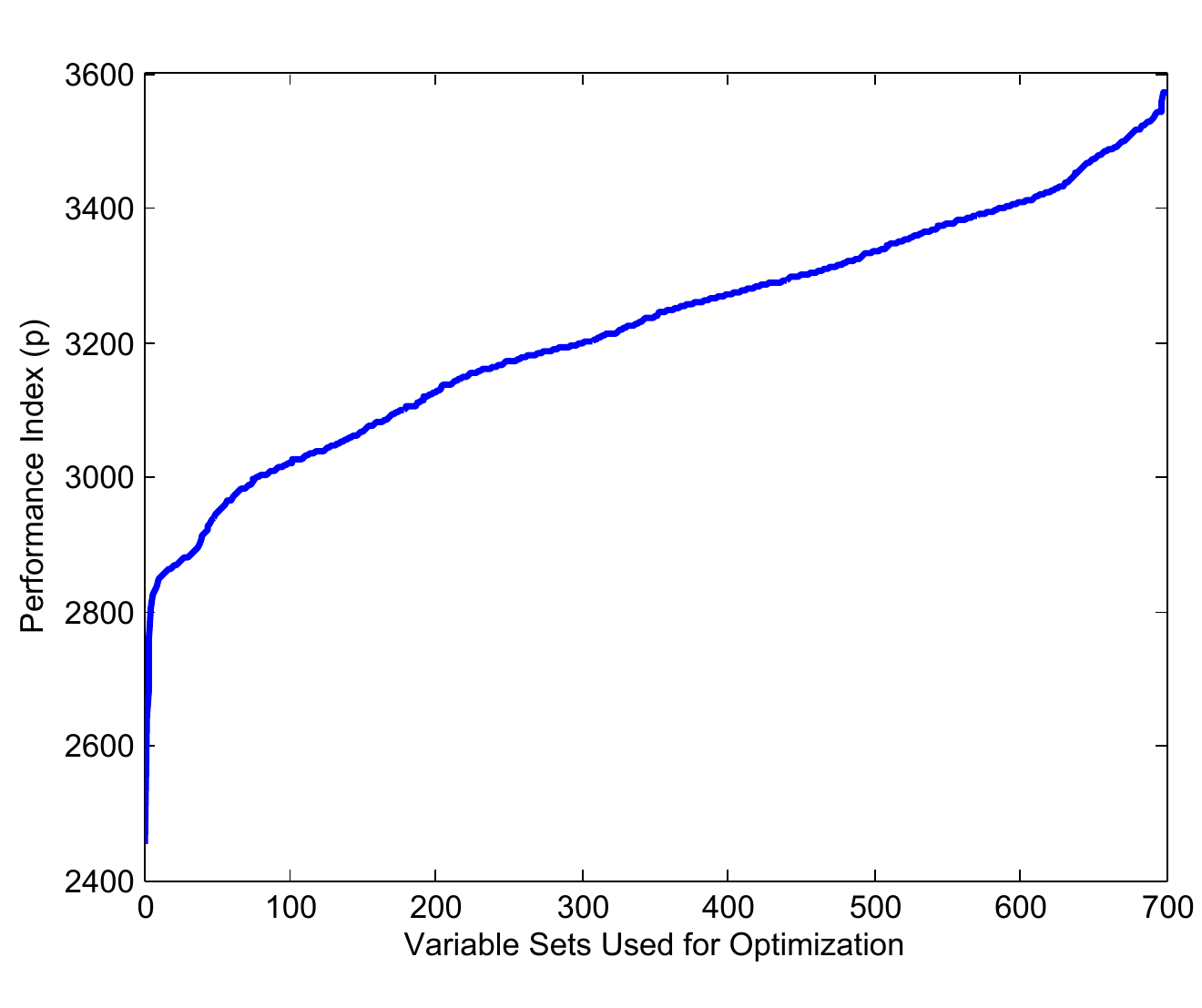}}
\caption{Computation of performance index $p$ for different sets of link lengths, which satisfy the physical linear and statistical constraints, for index finger component optimization.}
\label{fig:performance_plot}
\end{figure}

To enlarge the efficient workspace of finger joints, previous constraints were controlled repeatedly to explore a workspace up to $80^o$ for MCP joint and $90^o$ for PIP joint. After the pre-optimization selection, where physical constraints were ensured to be satisfied, an optimization by exhaustive search was conducted. With this motivation, each successful iteration, which was not interrupted with unsatisfied constraints, was finalized by calculating the performance cost function $p$ such that $p = \sqrt{\tau_1^2 + \tau_2^2}$. In order to provide a better understanding of the difference between the calculated cost function $p$ for each parameter set, Figure~\ref{fig:performance_plot} shows $p$ values over the variable sets during index finger optimization. The $p$ values were ordered to observe a constant increase, revealing how different the performance index can be calculated. It can be observed that the choice of parameters can increase performance by $\%50$. Once the iterations are finalized within the proposed range of variables, the set of link lengths with the highest performance index $p$ is selected as the final set to implement the hand exoskeleton.

Table~\ref{tab:results_index} - Table~\ref{tab:results_little} report the resulting link lengths of the hand exoskeleton components for the index, middle, ring and little fingers independently.Note that the range of variables for the links $L_{GH}$ and $L_{GF}$ could not be enlarged more to avoid any possible mechanical interference.

\begin{table}[htb]
\caption{Results of optimization for index finger.}
\label{tab:results_index}
\begin{center}
\begin{tabular}{c c||c c}
\hline \hline \\ [1ex]
variable & value [mm] & variable & value [mm] \\ [1ex]
\hline \hline \\ [1ex]
$L_{EJ}$ & 39 & $L_{ED}$ & 40 \\ [1ex]
$L_{CI}$ & 16 & $L_{EF}$ & 27 \\ [1ex]
$L_{CD}$ & 9 & $L_{BC}$ & 43 \\ [1ex]
\hline \hline
\end{tabular}
\end{center}
\end{table}

\begin{table}[htb]
\caption{Results of optimization for middle finger.}
\label{tab:results_middle}
\begin{center}
\begin{tabular}{c c||c c}
\hline \hline \\ [1ex]
variable & value [mm] & variable & value [mm] \\ [1ex]
\hline \hline \\ [1ex]
$L_{EJ}$ & 39 & $L_{ED}$ & 52 \\ [1ex]
$L_{CI}$ & 17 & $L_{EF}$ & 21 \\ [1ex]
$L_{CD}$ & 9 & $L_{BC}$ & 41 \\ [1ex]
\end{tabular}
\end{center}
\end{table}

\begin{table}[htb]
\caption{Results of optimization for ring finger.}
\label{tab:results_ring}
\begin{center}
\begin{tabular}{c c||c c}
\hline \hline \\ [1ex]
variable & value [mm] & variable & value [mm] \\ [1ex]
\hline \hline \\ [1ex]
$L_{EJ}$ & 39 & $L_{ED}$ & 38 \\ [1ex]
$L_{CI}$ & 16 & $L_{EF}$ & 29 \\ [1ex]
$L_{CD}$ & 10 & $L_{BC}$ & 42 \\ [1ex]
\hline \hline
\end{tabular}
\end{center}
\end{table}

\begin{table}[htb]
\caption{Results of optimization for little finger.}
\label{tab:results_little}
\begin{center}
\begin{tabular}{c c||c c}
\hline \hline \\ [1ex]
variable & value [mm] & variable & value [mm] \\ [1ex]
\hline \hline \\ [1ex]
$L_{EJ}$ & 42 & $L_{ED}$ & 32 \\ [1ex]
$L_{CI}$ & 16 & $L_{EF}$ & 23 \\[1ex]
$L_{CD}$ & 9 & $L_{BC}$ & 43 \\ [1ex]
\hline \hline
\end{tabular}
\end{center}
\end{table}

\begin{figure}[htb]
\centering
\resizebox{4.5in}{!}{\includegraphics{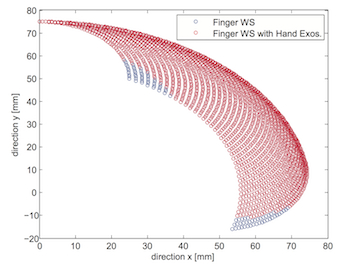}}
\caption{Comparison between anatomical finger workspace and workspace under the assistance of the proposed hand exoskeleton using only the rotation of MCP and PIP joints.}
\label{fig:workspace}
\end{figure}

For a middle-sized hand, the optimized link lengths allow finger joints to reach a workspace stated in Table~\ref{tab:index_rom_exo} for index, middle, ring and little fingers. In fact, Figure~\ref{fig:workspace} represent the index finger workspace of the end of middle phalange to test RoM for MCP and PIP joints only in a natural manner and with physical constraints that are set by the exoskeleton. Even though there are some extreme points that a user cannot reach wearing the exoskeleton, $\%~95$ of the natural workspace of index finger was covered under the exoskeleton assistance.

\begin{table}[htb]
\caption{Ranges of motion for all finger joints with hand exoskeleton}
\label{tab:index_rom_exo}
\begin{center}
\begin{tabular}{c||c|c|c}
\hline \hline \\
Joint & MCP & PIP & DIP \\  [1ex]
RoM & 0 - 80 & 0 - 90 & -- \\  [1ex]
\hline \hline
\end{tabular}
\end{center}
\end{table}

\begin{figure}[htb]
\centering
\subfigure[Full flexion]{\includegraphics[width=0.7\textwidth]{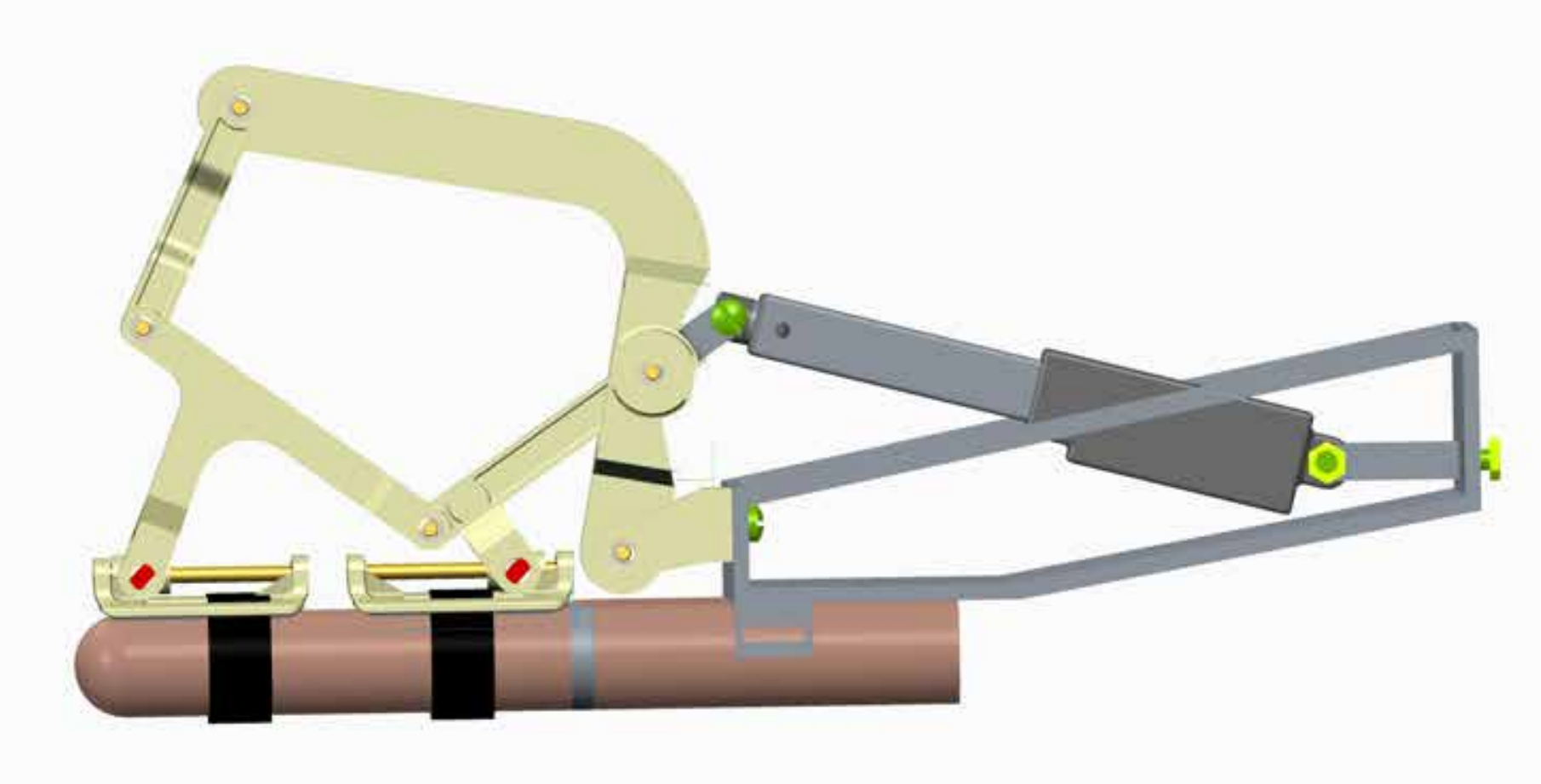}} \\
\subfigure[Full extension]{\includegraphics[width=0.7\textwidth]{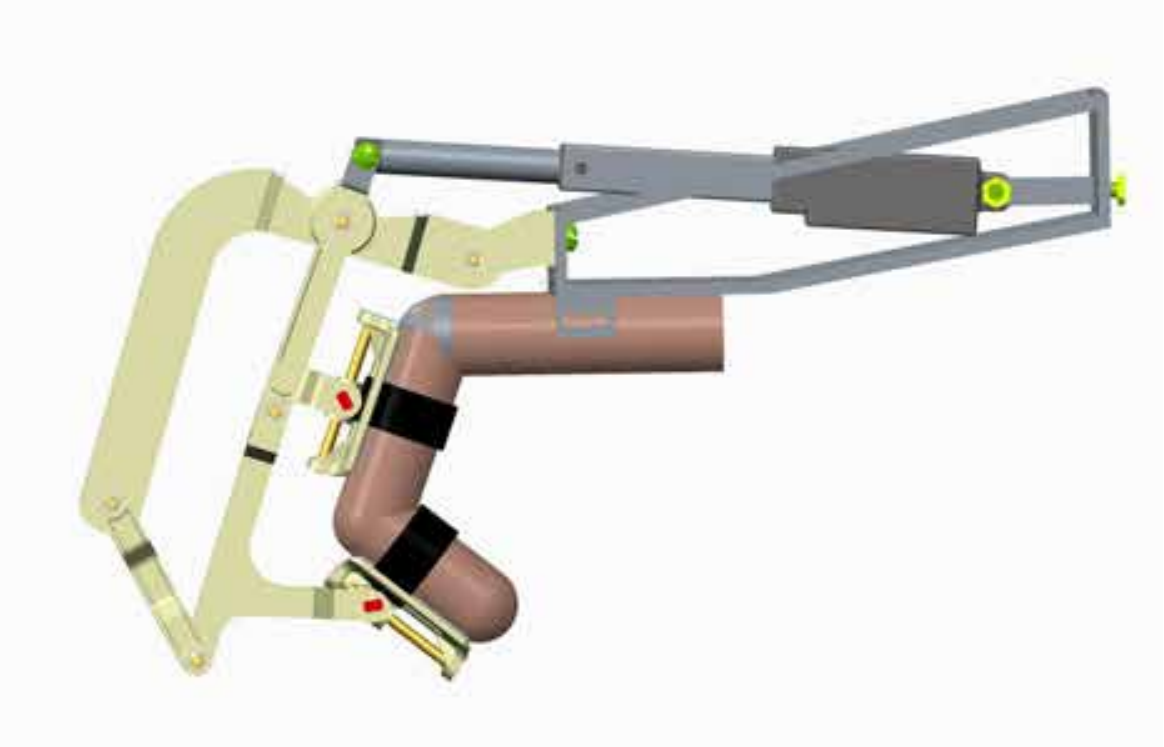}}
\caption{Maximum flexion and extension pose for index finger component of the proposed hand exoskeleton.}
\label{fig:finger_flex_ext}
\end{figure}

\begin{figure}[htb]
\centering
\subfigure[Full flexion]{\includegraphics[width=0.7\textwidth]{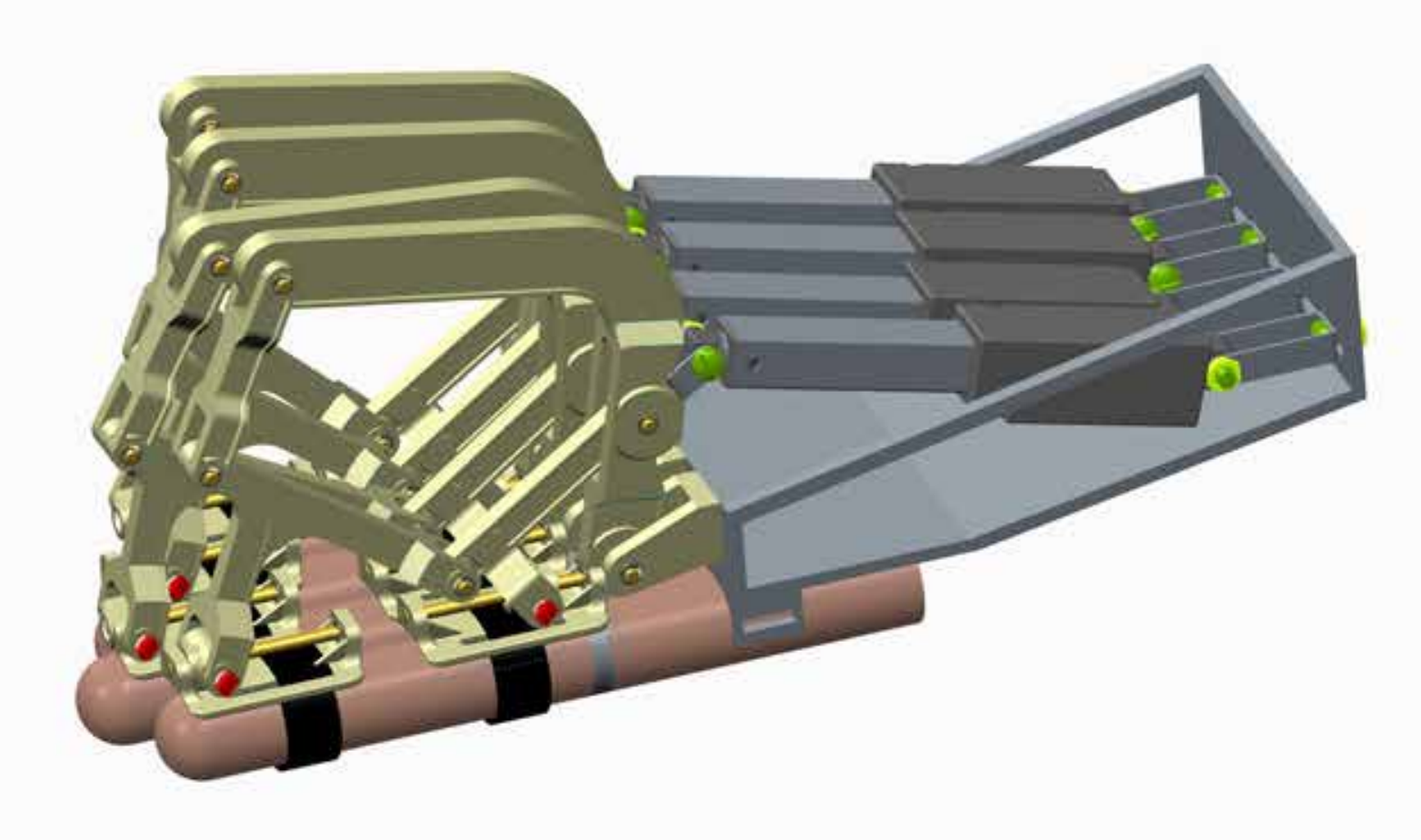}} \\
\subfigure[Full extension]{\includegraphics[width=0.7\textwidth]{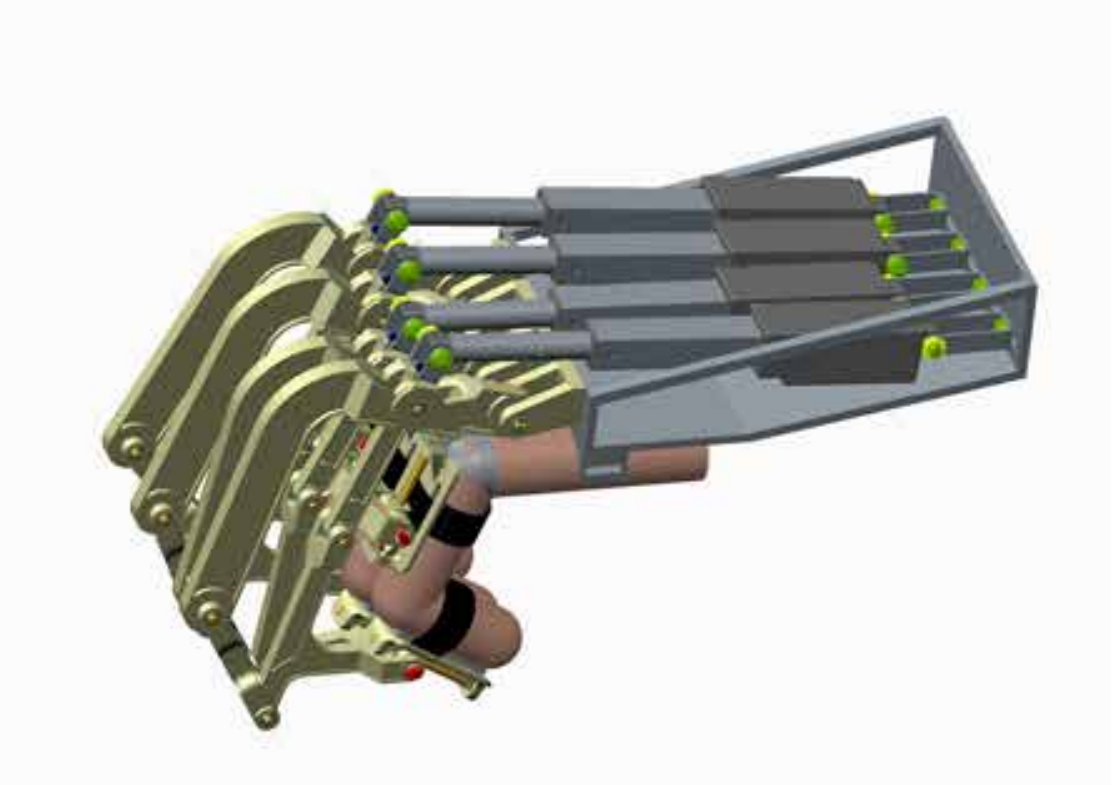}}
\caption{Maximum flexion and extension pose for the proposed hand exoskeleton with index, middle, ring and little finger components.}
\label{fig:hand_flex_ext}
\end{figure}

Figure~\ref{fig:finger_flex_ext} shows the CAD design of index finger component in maximum flexion and extension configurations of a user. Similarly, the overall hand exoskeleton in full flexion and full extension is presented in Figure~\ref{fig:hand_flex_ext} without the thumb. It is useful to highlight that the thumb component for a full hand exoskeleton is left as an independent work due to complexity and diversity of the thumb kinematics compared to rest of the hand.
%
%\begin{figure}[htb]
%  \centering
%  \resizebox{5in}{!}{\includegraphics{finger_ext_flexx}}
%  \caption{Maximum flexion and extension of the index finger with the finger device}
%  \label{fig:optimal_solution}
%\end{figure}

The device is optimized considering the measurements of a middle-sized hand, where index finger has the measurements of $50~mm$ and $30~mm$ for first and second finger phalanges. However, it is observed that the device with selected link lengths satisfy physical constraints and provide an efficient force transmission also for a small-sized hand (index finger of $45~mm$, $27~mm$) and big-sized hand (index finger of $55~mm$, $34~mm$), which cover a wide range of users in the society. Still, it is important to note that assisted RoM of finger joints stated in Table~\ref{tab:index_rom_exo} tend to change for different hand sizes.

\clearpage
\pagebreak

\newpage
\chapter{Implementation of the Hand Exoskeleton} % Main chapter title

\label{sec:Chapter4}% For referencing the chapter elsewhere, use \ref{Chapter1}

\lhead{Chapter 4. \emph{Implementation}} % This is for the header on each page - perhaps a shortened title

After the design, the kinematics calculation and the link length optimization performed in Chapter~\ref{sec:Chapter3}, we will detail the physical implementation and the design of control board and electronic system for the proposed hand exoskeleton. The electronic board mostly focuses on to things: reading sensors (the actuator displacement, additional passive joint rotation, force, pressure, etc.) and drive the actuators based on different control strategies (position control, admittance control, teleoperation, etc.). First, we will detail the design of this electronic board, and then, we will present the 3-D printed hand exoskeleton. Finally, various control algorithms, such as position control with enhanced performance tools, active control through muscular activity using EMG measurements and active control through active backdriveability using additional force sensor will be detailed with feasibility experiment results. The hand exoskeleton for $4$ fingers are manufactured using the kinematics and link lengths presented, but most of the control results are performed only using index finger component for simplicity. Control algorithms that are presented here can easily be utilized for rehabilitation exercise scenarios, as much as assistive applications.

\newpage
\section{Control Board}\label{sec:board}

Designing the hand exoskeleton requires an electronic control hardware to be designed to read sensor measurements coming from position sensors attached to the exoskeleton, to convert analog measurements into digital, to run control algorithms to drive actuators in a stable manner, to use motor drivers to command actuators and to communicate with external environments, such as a host computer, a virtual environment, etc.

\begin{figure}[htb]
\centering
%\vspace*{-2\baselineskip}
\resizebox{6in}{!}{\includegraphics{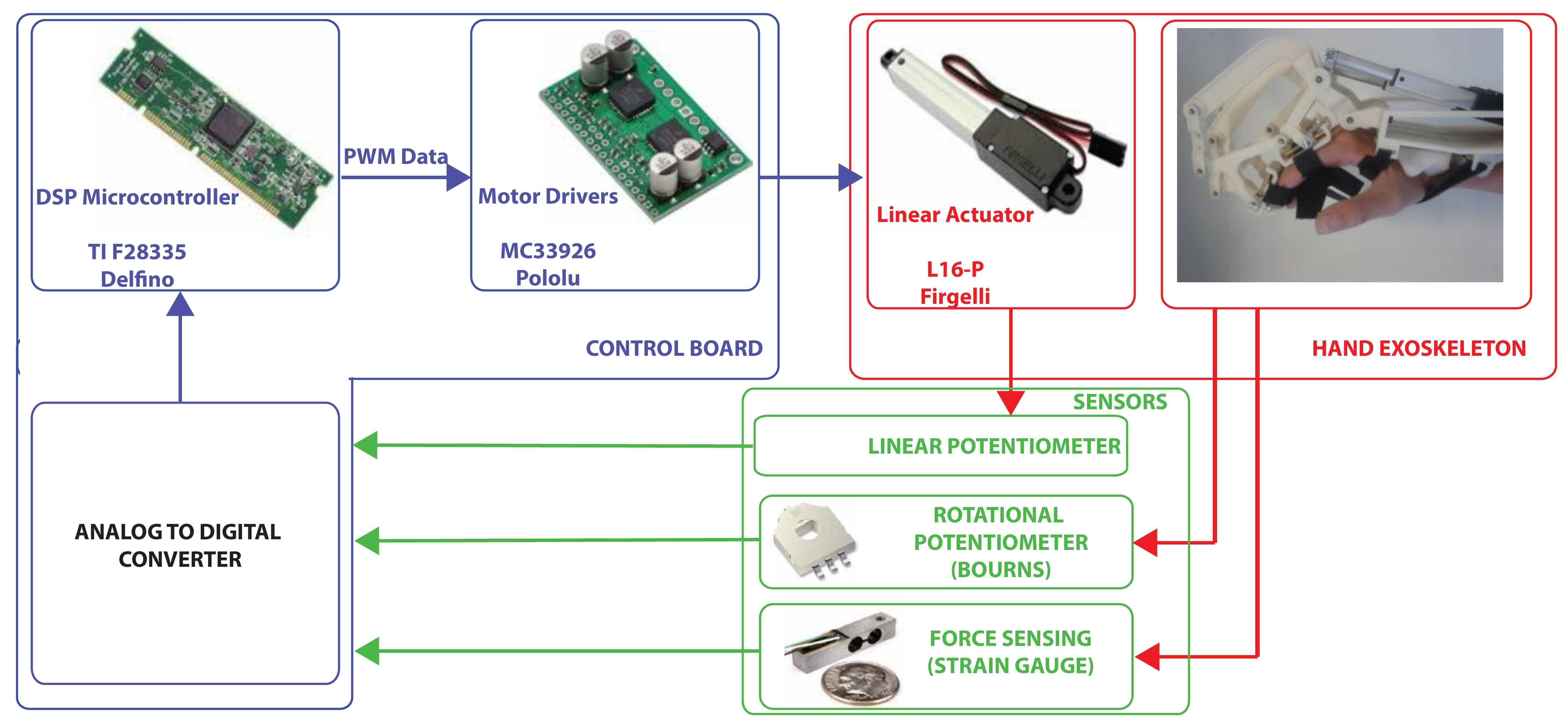}}
\caption{General scheme of electronics components chosen for the actuation and control of the proposed hand exoskeleton: a DSP microcontroller has a control algorithm creating a PWM data to send to the motor driver. This motor driver converts the PWM signal to the current to be given to the linear actuator. As the linear actuator moves, the linear potentiometer inside the actuator and the additional potentiometer measures the movements on the exoskeleton. Furthermore, the force sensor attached on the exoskeleton measures how much force the user applies to open/close his fingers. All of these measurements are converted to digital to be used as feedback by the control algorithm.}
\label{fig:handexoselectronics}
%\vspace*{-1\baselineskip}
\end{figure}

Figure~\ref{fig:handexoselectronics} summarizes the electronics hardware of the proposed exoskeleton. In particular, a DSP microcontroller board (TI F28335 Delfino) was chosen to track sensors, to turn signals to digital and produce Pulse-width Modulation (PWM) signals to run the motors through the Pololu motor driver carrier. The Pololu motor drivers (MC33926) read PWM signals and create corresponding current to drive selected linear actuators (Firgelli, L16-50-P). As the linear actuator pushes the exoskeleton to open/close fingers, its internal potentiometer measures the movement and sends it to the control board. Alternatively, an additional rotational potentiometer will be put over one of the passive mechanical joints to reach a unique finger pose instantaneously and read by the control board through analog pins. Finally, a $1 DoF$ force sensor is used for each finger component to measure user's intentions to open/close his finger and run active backdriveability through control. Similarly, force measurements are read by the control board through analog pins. The blocks represented in %Figure~\ref{fig:handexoselectronics} will be detailed in the following subsections.

\subsection{Actuators}
We know that the actuator directly changes the mechanical design of the proposed exoskeleton. This is why we have to decide the actuator as a first step of the electronics design. Independent finger control can be achieved by implementing a single actuator for each finger component individually. Furthermore, a generic hand exoskeleton should be as portable as possible, such that being attached to a stationary computer through a single cable without turning the overall device into a bulky, heavy design. The portability requires all $4$ linear actuators, without the thumb component, to be placed on top of user's hand. By choosing miniature linear actuators, we can place all actuators in a compact manner, while providing sufficient forces for fingers to perform ADLs or rehabilitation exercises.

Even though there are certain linear actuators with similar stroke capabilities and output forces without a screw mechanism to amplify effective forces. Not having a screw mechanism achieves backdriveability, allowing users to manipulate the actuator by applying forces. However, they can reach high forces using magnetic components, which might be problematic if a certain distance between two actuators cannot be satisfied. This distance requirement would prevent the independent finger implementation in a portable manner, so they are found impractical for this study. Instead, we chose Firgelli actuators, since they are low-cost, practical and efficient in terms of output forces to perform grasping tasks assisted by the hand exoskeleton.

\begin{table}[b!]
\caption{Specifications of Firgelli L16 linear actuator.}
\label{tab:properties_actuator}
\begin{center}
\begin{tabular}{l||c}
\hline \hline \\
{\bf Property} & {\bf Value}\\  [1ex]
\hline \\
Actuator Mass & $\cong$ $56~g.$\\  [1ex]
Actuator Stroke & $50~mm.$ \\  [1ex]
Motor gear ratio & $35:1$ \\  [1ex]
Max. cont force of the motor & $40~N$ \\  [1ex]
Backdrive force & $31~N$ \\  [1ex]
Max. velocity of the output shaft & $32~mm/s$ \\  [1ex]
Positional Error Max & $300~\mu m$ \\  [1ex]
Potentiometer Impedance & $9~k\Omega$ \\  [1ex]
Rated Voltage & $12~V DC$ \\  [1ex]
Maximum Current Consumption & $1~A$ \\  [1ex]
\hline \hline
\end{tabular}
\end{center}
\end{table}

Table~\ref{tab:properties_actuator} details the specific properties of Firgelli $P16-50-22-12$ actuator. It has a rotary DC electromagnetic motor actuating a screw mechanism, defining a gear ratio to amplify the output forces. Such a gear box ratio provides effective output forces with minimized actuator box. The screw mechanism requires high backdrive forces such that intrinsic backdriveability is not allowed when actuators are off. Even though the finger components are manufactured with minimum friction, the backdriveability of the overall system is inhibited. %Nevertheless, the backdriveability can be regained simply by utilizing force sensors and force control techniques.

\subsection{Control Electronics and Motor Drivers}

The chosen Firgelli actuators are driven through Free scale MC33926 H-Bridge IC, mounted on a Pololu Carrier Board. Each driver board drives two DC actuator, so we need $2$ boards to control $4$ actuators, which would simplify the electronics design overall. The control board sends PWM signals to the control inputs of the driver board at $25~KHz$ frequency through PWM peripheral. %Actuators are powered at $12~V$, with maximum current consumption (stall current) of $1~A$.

A microcontroller unit was implemented initially to read actuator displacements and create appropriate PWM signals using control strategies using these measurements. The DSP microcontroller is also used to compute kinematics analyses, communicate with a host PC and read complementary sensors for pose estimation or force control techniques. The Texas Instruments Delfino F28335 DSP microcontroller board was chosen, featuring a core with floating point unit, ADC and PWM peripherals and USB communication with the host PC. It also supports rapid prototyping of the control algorithm through MATLAB Simulink: the Simulink model can access the hardware peripherals and can be compiled, flashed and executed as a stand-alone program on the DSP board.

\begin{figure}[!t]
\centering
%\vspace*{-2\baselineskip}
\resizebox{4in}{!}{\includegraphics{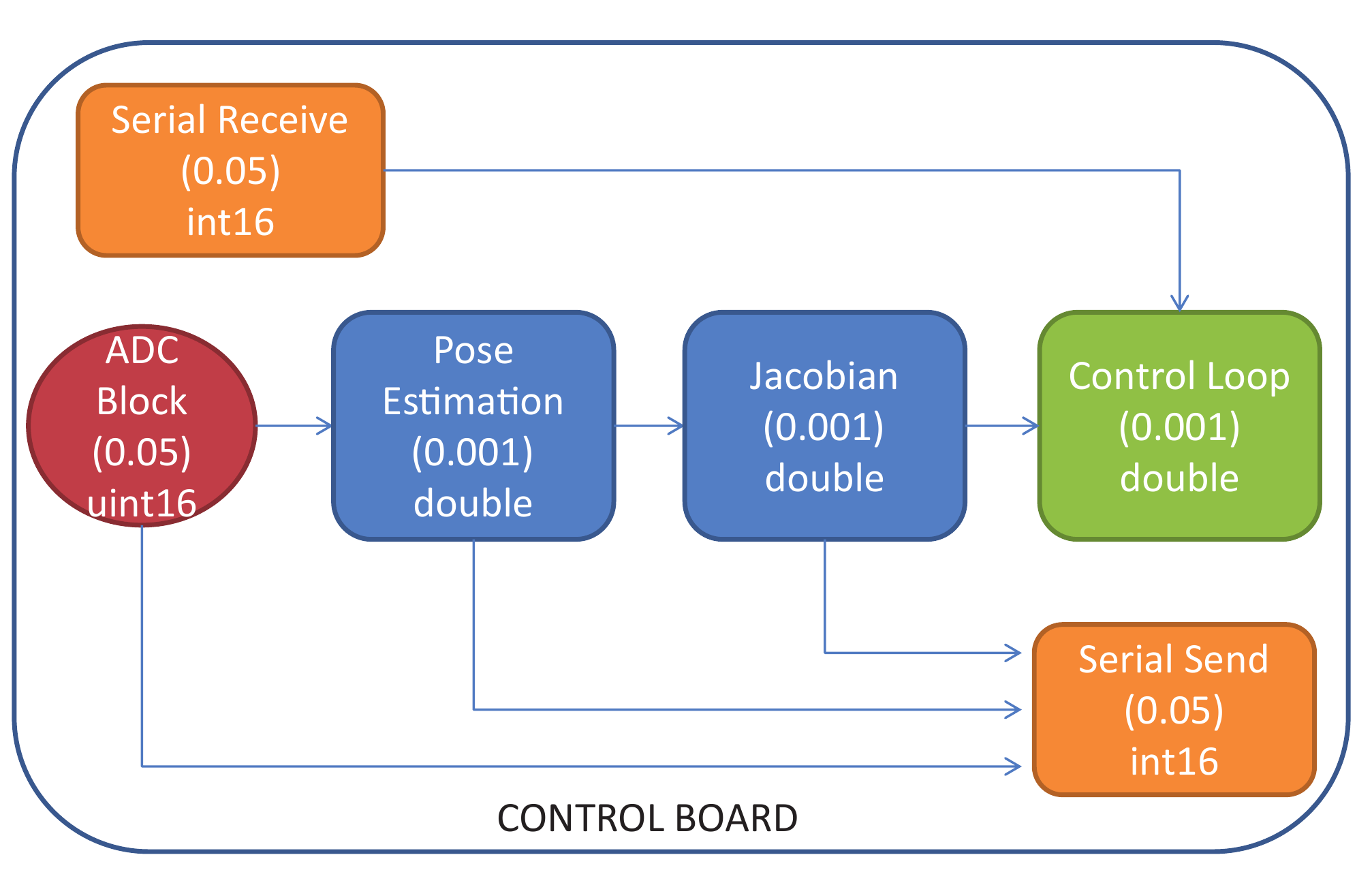}}
\caption{Sample time and data type definitions for different operations performed in DSP control board or communication network with the host computer.}
\label{fig:sample}
%\vspace*{-1\baselineskip}
\end{figure}

Figure~\ref{fig:sample} summarizes the step size and the data type that are used within the control board. Serial send/receive blocks set a USB communication between the board and the host model, and we limit this communication speed to $0.05~s$. Pose estimation, Jacobian calculation and control algorithms are computed at $0.001~s$. Resulting PWM values are fed to actuator with $0.0005~s$ to achieve smoother movements. The data communication is made as $int16$ type and are converted to $double$ right after receiving them or right before sending them.

\subsection{Sensors} \label{sec:sensors}

The most essential sensory measurements of the exoskeleton come from the linear potentiometers attached to Firgelli actuators. These measurements are used for closed-loop control algorithms to form the feedback or to estimate finger pose during operation. Its accuracy and speed are crucial to ensure safety and stability of the overall system. Each actuator potentiometer provides an analog position feedback signal. Minimum and maximum output values of these measurements are set as $0 - 50~mm$ by mechanical properties of the actuator.

In Section~\ref{sec:forward}, we mentioned the need of additional sensor from the mechanism and how we can reach a unique forward kinematics solution by implementing an additional potentiometer to measure $q_B$ (see Figure~\ref{fig:definitions}). A low-cost, lightweight potentiometer (BOURNS 3382G-1-103G) were chosen to be implemented in a simple and lightweight manner. Each potentiometer is DC powered ($5~V$) and the cursor voltage is read through the ADC peripheral of the microcontroller board. The potentiometer was chosen specifically with a hole through property in order to fix it on the mechanism using existing pins. The potentiometer can read measurements between $0 - 330^o$.

Table~\ref{tab:properties_actuator} states the gear box ratio ($35:1$) and high backdrive force ($31~N$) about the chosen Firgelli actuator. This internal friction prevents the user to initiate a physical movement manually. Therefore, achieving active tasks can be ensured only through force control techniques. A force sensor was implemented externally for each finger component to measure interaction forces between the device and the user. $1 DoF$, miniaturized load cell with strain gauges ($S215 Load Cell$ by SMD Sensors) is mounted in series between the actuator and the main body of the exoskeleton for each finger component. The SMD load cell has small dimensions ($26\times6\times6 mm$) and high repeatability ($0.01\%$) with a maximum measurable range of force of $18~N$ (safe overload of $36~N$). Each load cell is connected to a proprietary electronic board ($ZSC31050$ by SMD Sensors) integrating a wheatstone bridge, and a signal amplification and conditioning stage. The analog output signal is regulated between $0-5~V$ and then is read through ADC peripheral of the microcontroller board.

Finally, additional force sensing resistors (FSRs) are occasionally used to measure interaction forces, either between user's fingers and grasped objects, or between device connectors and user's fingers. These sensors help us understand the force transmission between finger phalanges as much as the comparison between resulting forces along finger phalanges and actuator force. FSR $402$ (Interlink Electronics Inc.) sensors have been used, with a circular active area measuring $12,5~mm$ in diameter and a sensitivity range from less $1 - 1000~N$, depending on the electrical setup. The sensors were coupled to a fixed resistor of $2.2~kHz$ according to a voltage divider set up. This configuration allows to measure forces up to $12~N$. 

The FSR sensor was calibrated using precision weights for the chosen fixed resistance values. The output of FSR is dependent on the surface area, so we need to design the contact area carefully. An additional interface should be placed on top of the FSR sensor to minimize the contact dependency, to obtain comparable results within different experimental tasks, and to increase repeatability of applied forces. Within this study, these additional FSR sensors were used in various configurations, so that their placement will be covered in the following sections whenever utilized. Since these sensors are not essential for each task and they are not required for the control loops, their readings are measured through an additional ARDUINO board with serial USB connection to the host computer.

\newpage

\section{Multi-Finger Exoskeleton}\label{sec:hand}

The first prototype of the underactuated hand exoskeleton was implemented with rapid prototyping parts, allowing the device to be low-cost and light-weight. Table~\ref{tab:properties} presents mechanical specifications of the device, actuated by Firgelli $L16$ linear actuator. The link lengths for finger components were found using the optimization algorithm in Section \ref{sec:optimization}.

\begin{table}[htb]
\caption{Specifications of the proposed exoskeleton and its actuators.}
\label{tab:properties}
\begin{center}
\begin{tabular}{l||c}
\hline \hline \\
{\bf Property} & Value \\ [1ex]
\hline \\
Number of Assisted Fingers & 4 \\ [1ex]
Device Mass & $\cong$ 400 g.\\  [1ex]
RoM for MCP & $80^o$  \\  [1ex]
RoM for PIP & $90^o$  \\  [1ex]
Max. torque on MCP & 1485 Nmm\\ [1ex]
Max. torque on PIP & 434 Nmm\\ [1ex]
\hline \hline
\end{tabular}
\end{center}
\end{table}

\begin{figure}[!h]
\centering
\resizebox{3.8in}{!}{\includegraphics{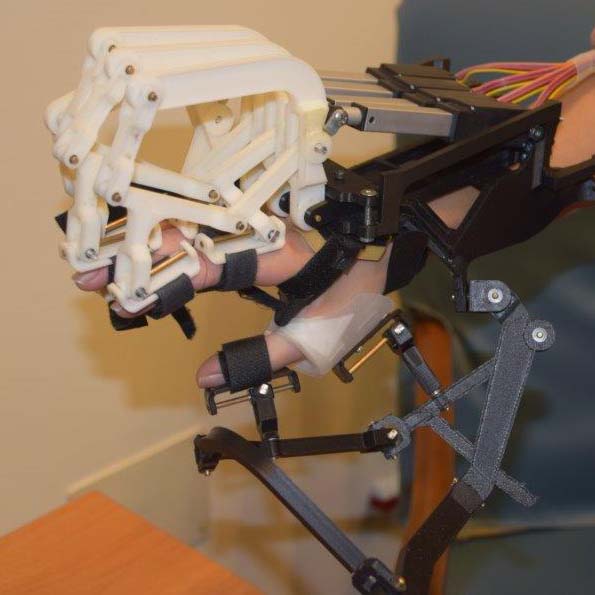}}
\caption{Implementation of hand exoskeleton with 5 finger components, including the thumb, which is the product of an individual research study.}
\label{fig:multiCAD}
\end{figure}

The device is attached to the user from their palm and their wrist, to stabilize the base and from finger phalanges, in order to stabilize finger components. Connecting the device using simple Velcro straps allows the device to be worn in about a minute, without any initial pose requirement of fingers. Actuators are attached on the base of the device in a compact manner. The control board is placed on top of a table, near a host computer, but there is no limitation regarding cable sizes. Figure~\ref{fig:multiCAD} shows the hand exoskeleton that is connected to a user through index, middle, ring and little finger components, whose kinematics and development is detailed in this study and thumb component, whose development is done individually, even though the idea behind its mechanical design is kept the same.

\newpage
\section{Control Algorithms}

The rest of this section details passive and active control algorithms implemented for the assistive and the rehabilitative experiment scenarios. First, a simple position control will be implemented to open/close fingers individually. Underactuation property of the device allows users to perform rehabilitation exercises and real grasping tasks for assistive exercises by adjusting its behavior automatically based on interaction forces, even using a simple, strict position control. The passive position control can be combined with active sensors to measure user's intention. The first active task is to estimate user's intention through EMG sensors, which detect the muscular activity from the healthy or unassisted arm of the user, and open/close fingers based on such estimation as a position reference for the exoskeleton. The second approach is to run a closed-loop force control, by measuring the interaction forces between the user's finger and the device's base, and by driving actuators to make these measurements follow a reference. Such a force control algorithm can be used only as backdriveable task for rehabilitation or assistive tasks, or as stiffness rendering task for rehabilitation and haptic applications.

\subsection{Position Control}\label{sec:position}

\begin{figure}[htb]
\centering
\resizebox{4.2in}{!}{\includegraphics{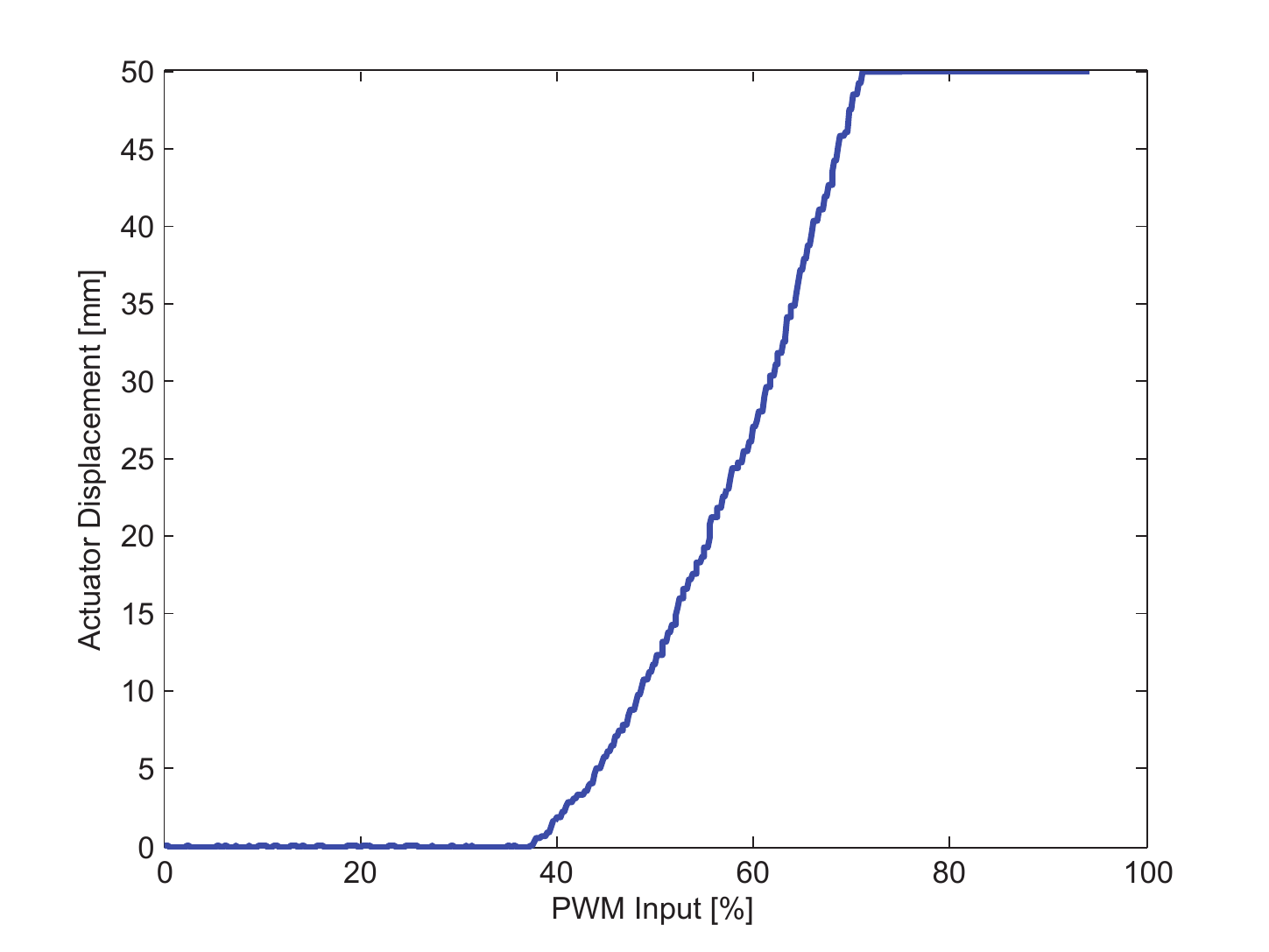}}
\caption{Observed actuator displacement with respect to the PWM percentage input to observe the impact of the high backdrive force of the actuator.}
\label{fig:nonback}
\end{figure}

The mechanical gearbox inside the linear actuator increases the internal friction and requires a backdrive force to overcome before starting the motion. Such a backdrive force not only prevents the actuator to be moved by applying forces to the shaft manually, but also affects the control performance adversely by creating a delay between the PWM signal and the corresponding movement. Figure~\ref{fig:nonback} presents the actuator displacement over the input PWM signal and shows that the PWM input can produce movement after $37.2~\%$, which might be the corresponding PWM input for the backdrive force.

%Nevertheless, the important argument regarding Figure~\ref{fig:nonback} is that actuator displacement cannot be achieved for PWM input values smaller than $37.2~\%$.

\begin{figure}[htb]
\centering
\resizebox{4.2in}{!}{\includegraphics{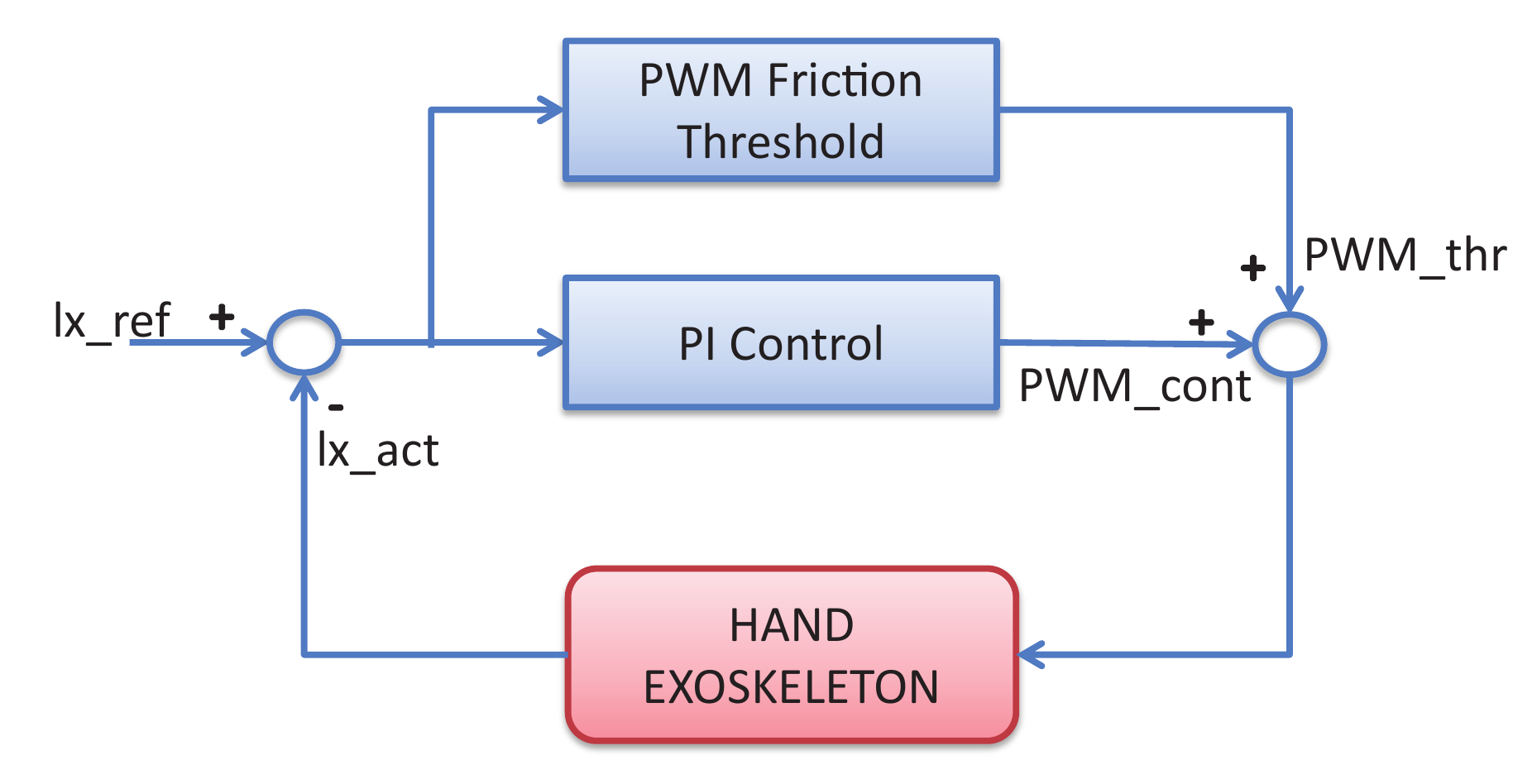}}
\caption{Adjusted position control algorithm with additional PWM supply in the same direction of control algorithm to overcome internal actuator friction.}
\label{fig:pos_alg}
\end{figure}

Actuator displacement cannot be achieved for the PWM input values smaller than $37.2~\%$. From the control point of view, such a limitation can be defined as a dead spot between $\pm 37.2~\%$, which can be compensated by supplying $\pm 40~\%$ to the motor drivers besides the output of the controller algorithm. The sign of such a compensation is adjusted with respect to the sign of the displacement error ($l_{x_{er}} = l_{x_{ref}} - l_{x_{act}}$). PWM input to determine the actuator forces can be calculated as $F_{thr} = sign(l_{x_{er}}) (40)$ allowing the control loop to have a better sensitivity in a stable manner. Meanwhile, the PWM input can be calculated by the controller strategy as $F_{cont} = K_P l_{x_{er}} + K_I \int l_{x_{er}}$, where $K_P$ and $K_I$ are proportional and integral control gains. In the end, the PWM signals being supplied to the motor drivers are calculated as the sum of both components as $F_{PWM} = F_{thr} + F_{cont}$. Figure~\ref{fig:pos_alg} presents the algorithm to compensate the friction PWM and the simple PI Control scheme.

The exoskeleton is controlled by a simple PI control algorithm to minimize the steady state error without calculating the derivative of real time measured sensors. The lack of derivative component allows the overall mechanical system to use simpler and cheaper encoders. The integral component allows the controller to eliminate the steady state error, while affecting the response time of the overall system. However, since the required system speed is much slower than the controller speed itself, response time drawback is not a critical issue.

\begin{figure}[htb]
\centering
\subfigure[Position control with step input.]{\includegraphics[width=0.7\textwidth]{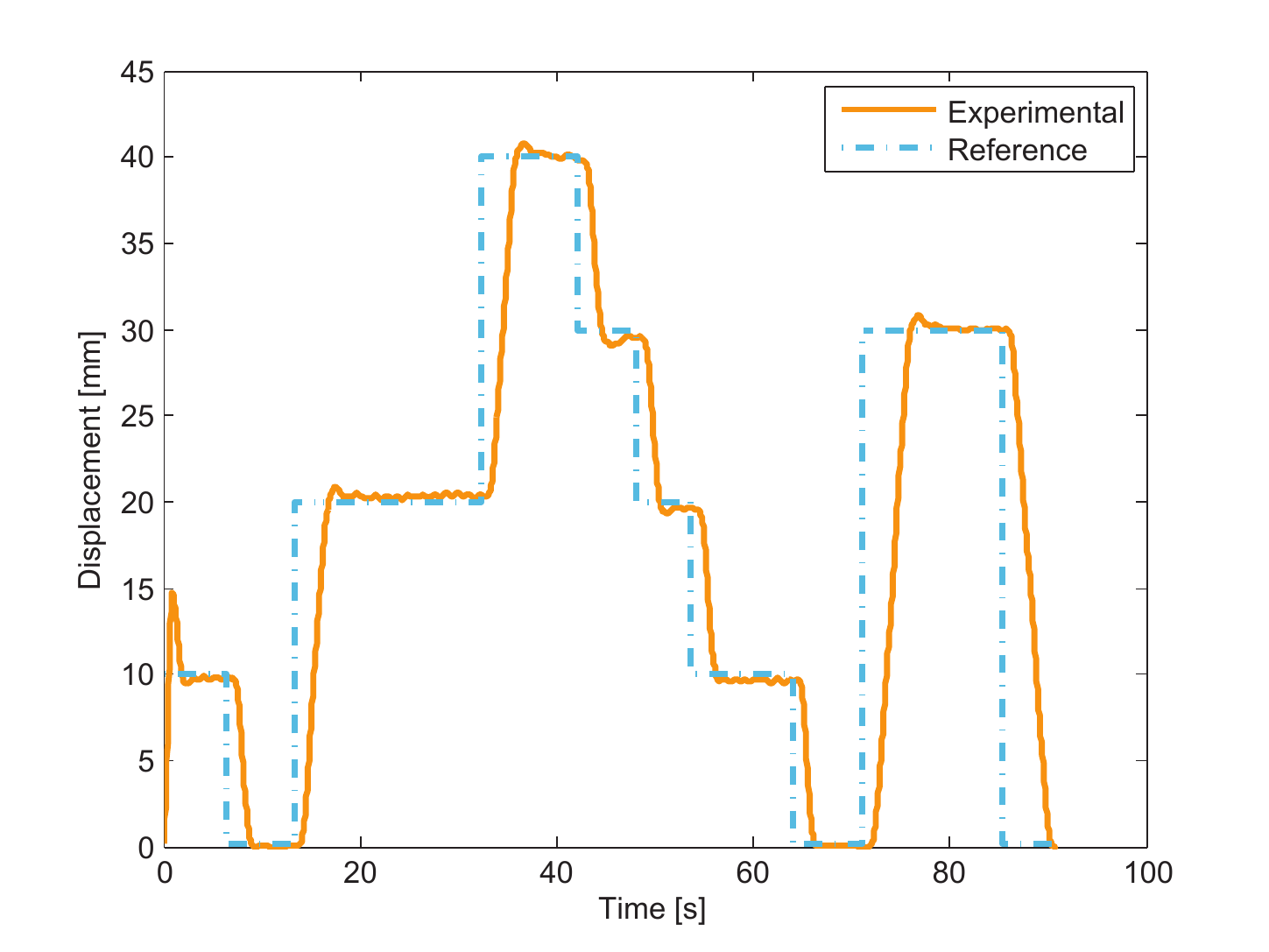}\label{fig:pos_discrete}}\\
\subfigure[Position control with ramp input.]{\includegraphics[width=0.7\textwidth]{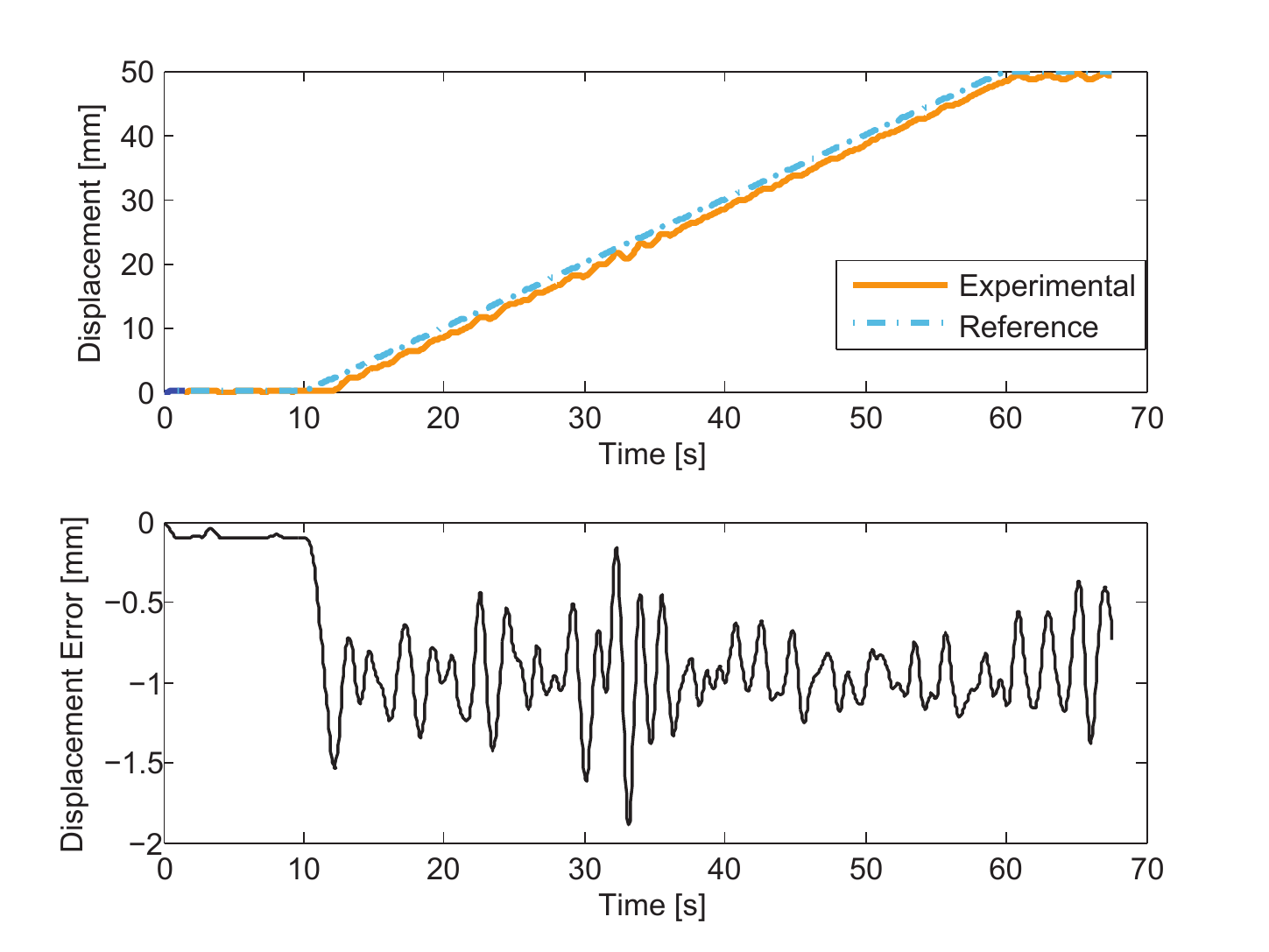}\label{fig:pos_ramp}}
\caption{Simple PI control results with the step and the ramp input references.}
\label{fig:pos_cont}
\end{figure}

Figure~\ref{fig:pos_cont} shows the results of position control tasks. Figure~\ref{fig:pos_discrete} shows the experiment result, where the displacement reference was given randomly and discretely to observe the settling time of the hand exoskeleton with the control scheme presented in Figure~\ref{fig:pos_alg}. From the figure, it can be observed that the finger component of the exoskeleton can open/close fingers in a few seconds easily. Both the response time and the steady state error can be observed to allow the control algorithm to run in an efficient and stable manner.

On the other hand, a ramp reference can show the following abilities of the chosen control gains. Figure~\ref{fig:pos_ramp} shows that the selected controller gains allow the actuator to follow a given continuous trajectory successfully. The displacement error seems to be within $2~mm$ for this experiment, This error range is quite acceptable for assistive or rehabilitative exercises, because they focus on the overall grasping tasks, and do not need precise control.

Besides simple PI control, further control algorithms can be used to open/close fingers in a passive manner. Nevertheless, having adjustable mechanism for various tasks in the environment based on interaction forces does not necessarily require further, more complex controllers to be developed, as long as stability of the controller and safety of the user are guaranteed.

%\newpage

\subsubsection{Temperature filter}\label{sec:temp}

The underactuated hand exoskeleton is designed predicting all the possible physical interference during operation. Mechanical precautions are performed to eliminate such interference, but due to different hand sizes and utilization of Velcro straps, we might still need further measures to minimize their negative impacts. For instance, Velcro straps, which are used to connect the device with user's fingers, might create some clutter and get stuck with each other. Especially while controlling finger components individually, such clutter might create a contradiction between the actuator movement and its reference. On such an occasion, insistent current fed to the actuators might harm the mechanical components, specifically the actuator itself. Such an incident can be prevented by limiting the current supply to the actuator through the PWM values at the controller, simply by lowering the PWM values based on time, preventing actuators to heat and giving enough time for the actuators to cool down.

\begin{figure}[htb]
\centering
\resizebox{5.2in}{!}{\includegraphics{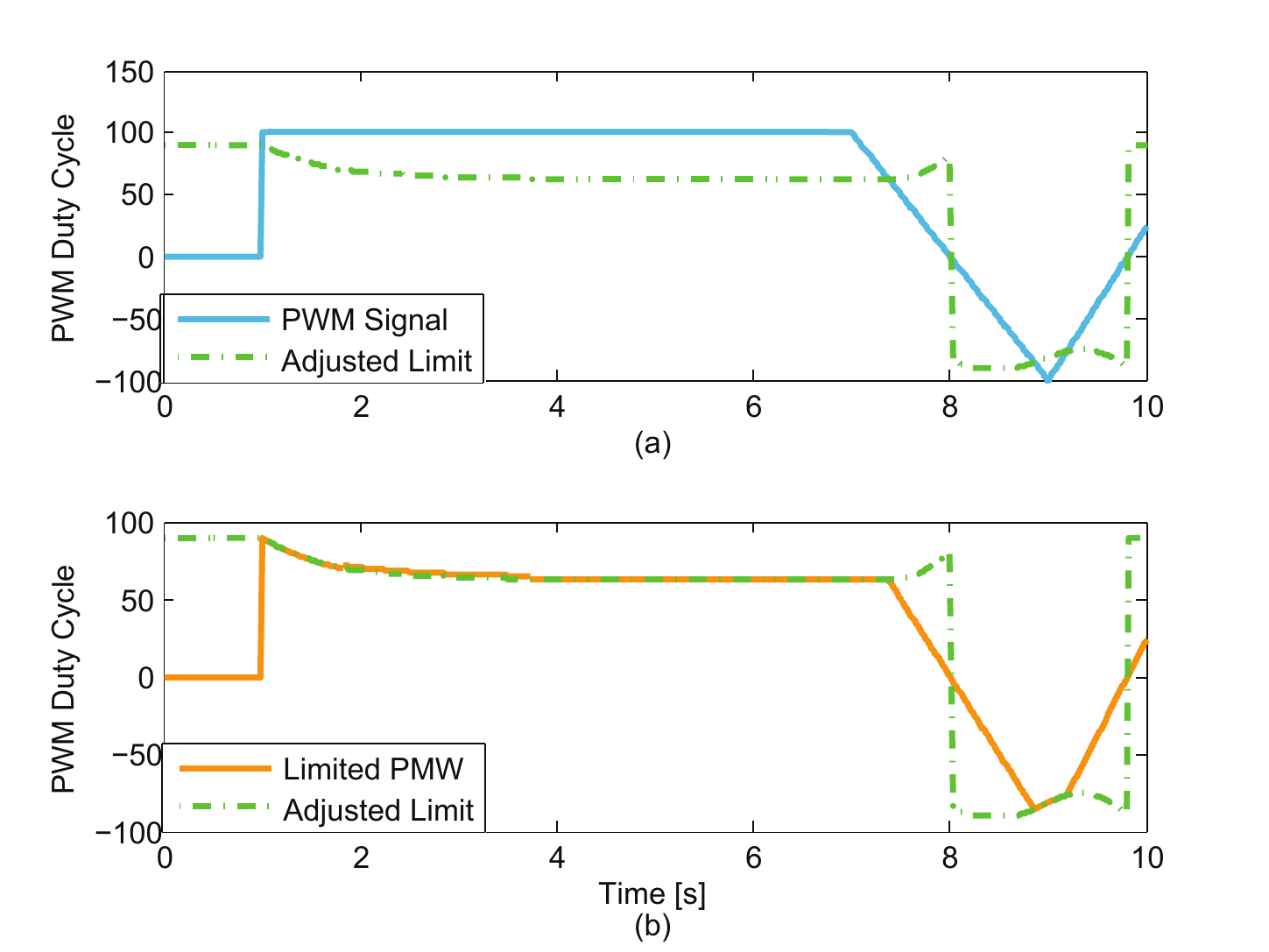}}
\caption{Simulation plot of how the temperature filter works: (a) adjusted limitation set for the input PWM signal based on the PWM signal values over time, and (b) output PWM signal to be supplied for actuator compared to the PWM limitation.}
\label{fig:filter}
\end{figure}

\begin{figure}[htb]
\centering
\resizebox{5.2in}{!}{\includegraphics{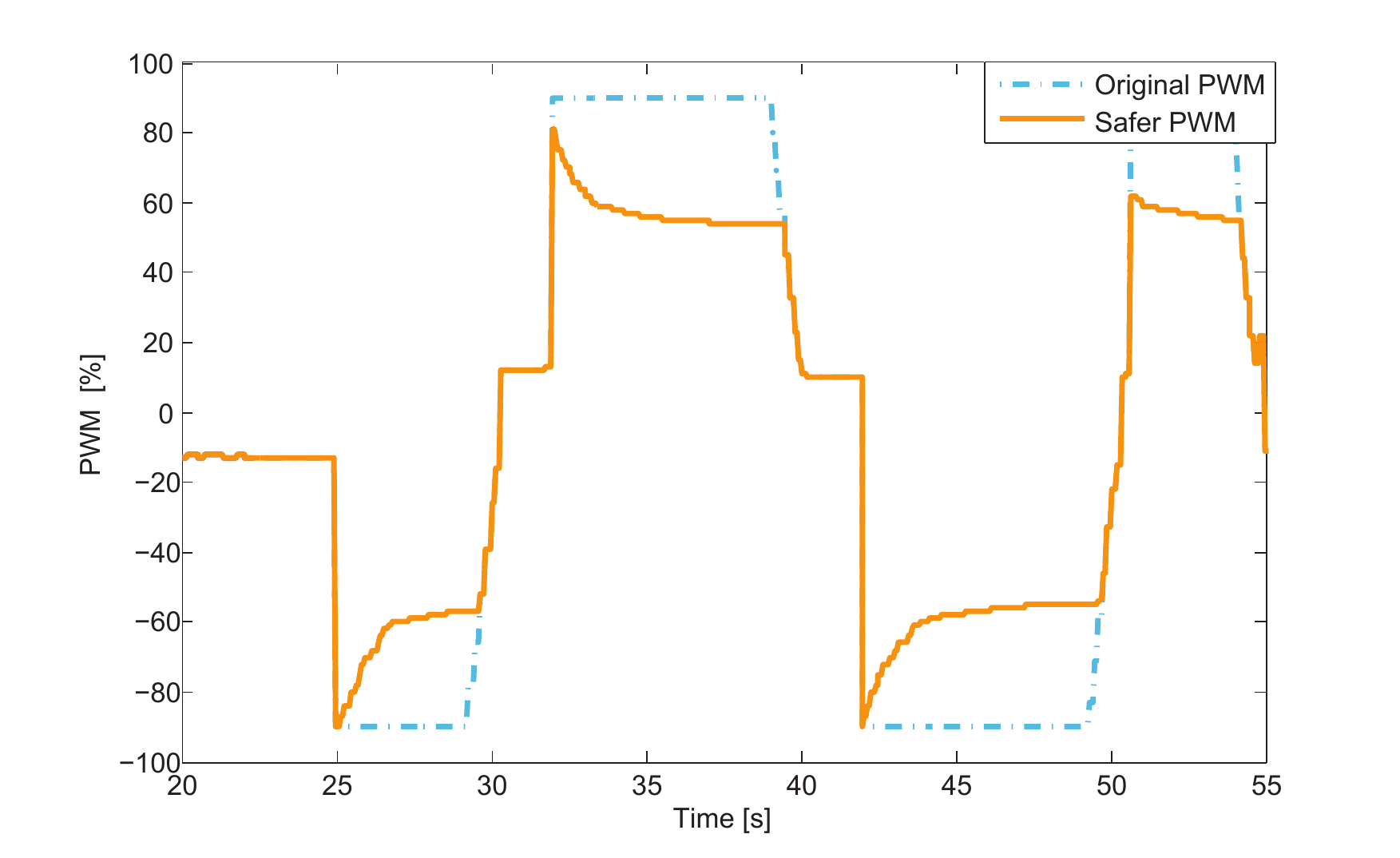}}
\caption{Real time implementation of the temperature filter being applied to the PWM signals generated during a PI control for the chosen actuator.}
\label{fig:pwm}
\end{figure}

Figure~\ref{fig:filter} shows the implementation of the proposed filter. A maximum and minimum values of the PWM duty cycle was defined, such that the controller can deliver PWM signals within those limits to the motor drivers. When the original PWM value is below the minimum limit, the PWM values are directly sent to the motor drivers to run the motors, while the adjusted limit is fixed to the maximum limit. As the PWM value reaches the maximum limit, and continue exceeding the maximum value; the adjusted limit starts to get decreased over time until reaching the minimum limit, as the PWM values supplied to the motor drivers are saturated through these adjusted limits. Whenever the original PWM values start to reach lower values than the minimum limit, the adjusted limit starts to increase until the maximum limit. In the meantime, the motor drivers are supplied the original PWM values. Such a temperature limit should be performed in a bidirectional manner. The minimum and the maximum PWM limits for this study were selected experimentally as $90~\%$ and $60~\%$, since the PWM values lower than $60~\%$ slows down the speed significantly. Doing so, even if the actuator gets stuck, the control unit of the hand exoskeleton will force actuators to perform the desired task using $60~\%$ of duty cycle. This might extend the life span of the actuators. Similarly, Figure~\ref{fig:pwm} shows the implementation of the proposed temperature filter for the hand exoskeleton with a simple PI control.

Implementing a temperature filter prevents the actuators to be exposed for continuous, excessive amounts of current and be burnt out during operation. However, they limit the operational speed significantly. Such a drawback might be insignificant for grasping tasks or repetitive rehabilitation tasks, since the actuator speed can be limited and modified intentionally, with no restriction. However, the speed limit cannot be tolerated for active tasks, where the device transparency is crucial. Even though properties of the filter can be adjusted to improve transparency, it should be eliminated for most of the haptic tasks. Such an elimination does not threaten the safety of the electronics components of the system, because allowing each finger component freely over control reduces the possibility for a mechanical clutter significantly.

%\newpage

\subsubsection{Grasping forces}\label{sec:grasping}

\begin{figure}[htb]
\centering
  %\vspace*{-3\baselineskip}
\resizebox{5in}{!}{\includegraphics{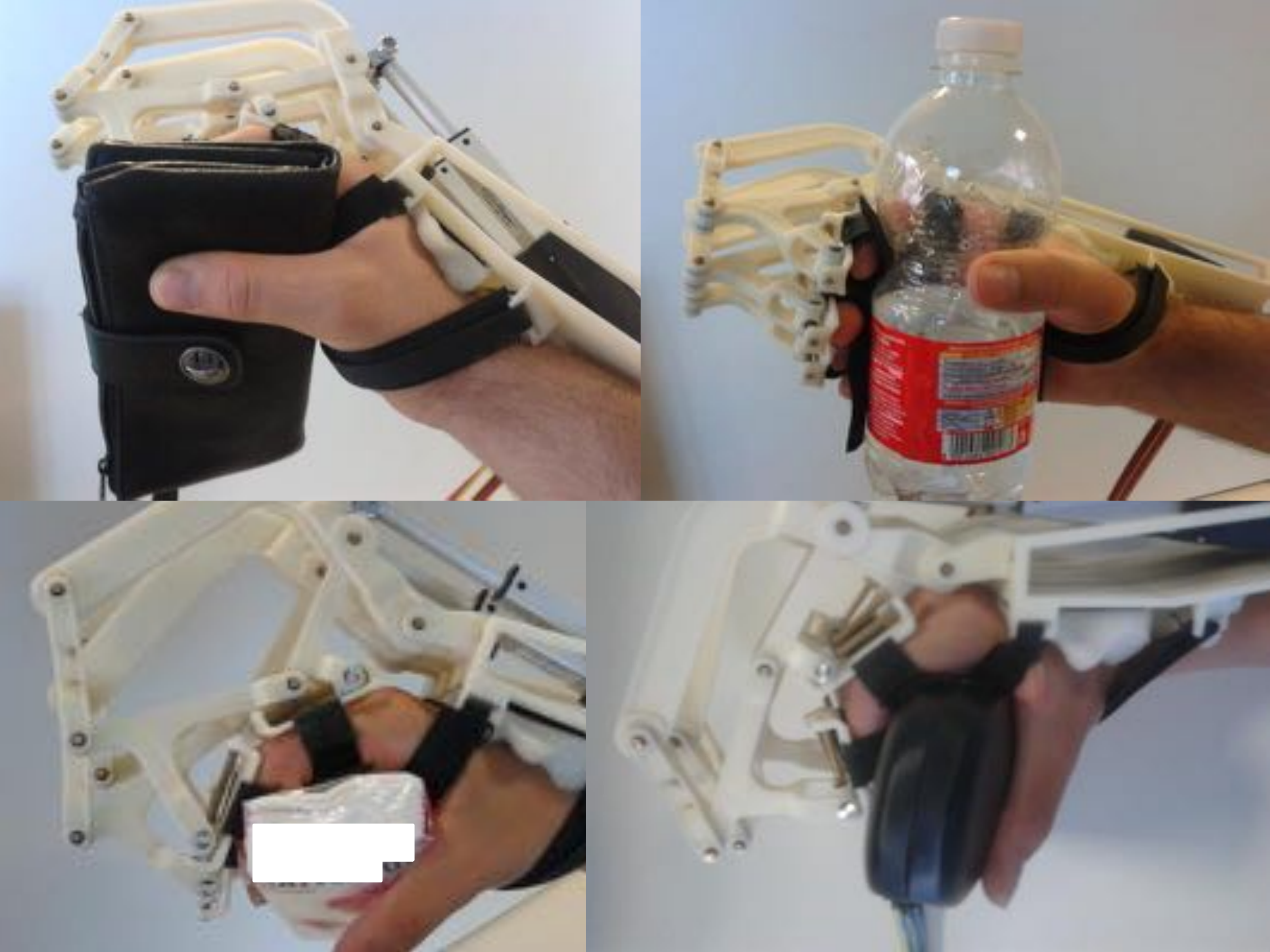}}
\caption{Real time implementation of grasping tasks of objects with different sizes and shapes while the proposed hand exoskeleton assists two subjects with different hand sizes.}
\label{fig:grasping_hand}
\end{figure}

The proposed hand exoskeleton is designed to embrace the underactuation concept, so that the device can adjust its operation mechanically. This adjustability needs to be tested as the exoskeleton assists a user to grasp various objects with different sizes and shapes. The mechanism was controlled by the simple PI control that was detailed previously. Figure~\ref{fig:grasping_hand} shows that the user can grasp objects with different sizes and shapes with no preliminary knowledge of the object. Note that the first two tasks have been completed by a male user and the last two tasks have been completed by a female user. Adaptation of the device for two different hand sizes automatically can be observed.

%The usability of the device has been tested through the grasping tasks of various objects and the force transmission on the finger phalanges to perform the grasping.

\begin{figure}[!b]
\centering
\subfigure[FSR sensors attached on a water bottle.]{\includegraphics[width=0.75\textwidth]{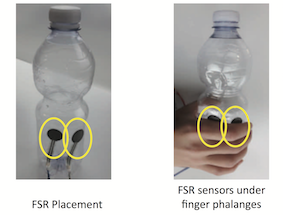}\label{fig:bottle}}\\
\subfigure[FSR sensors attached on a mug.]{\includegraphics[width=0.75\textwidth]{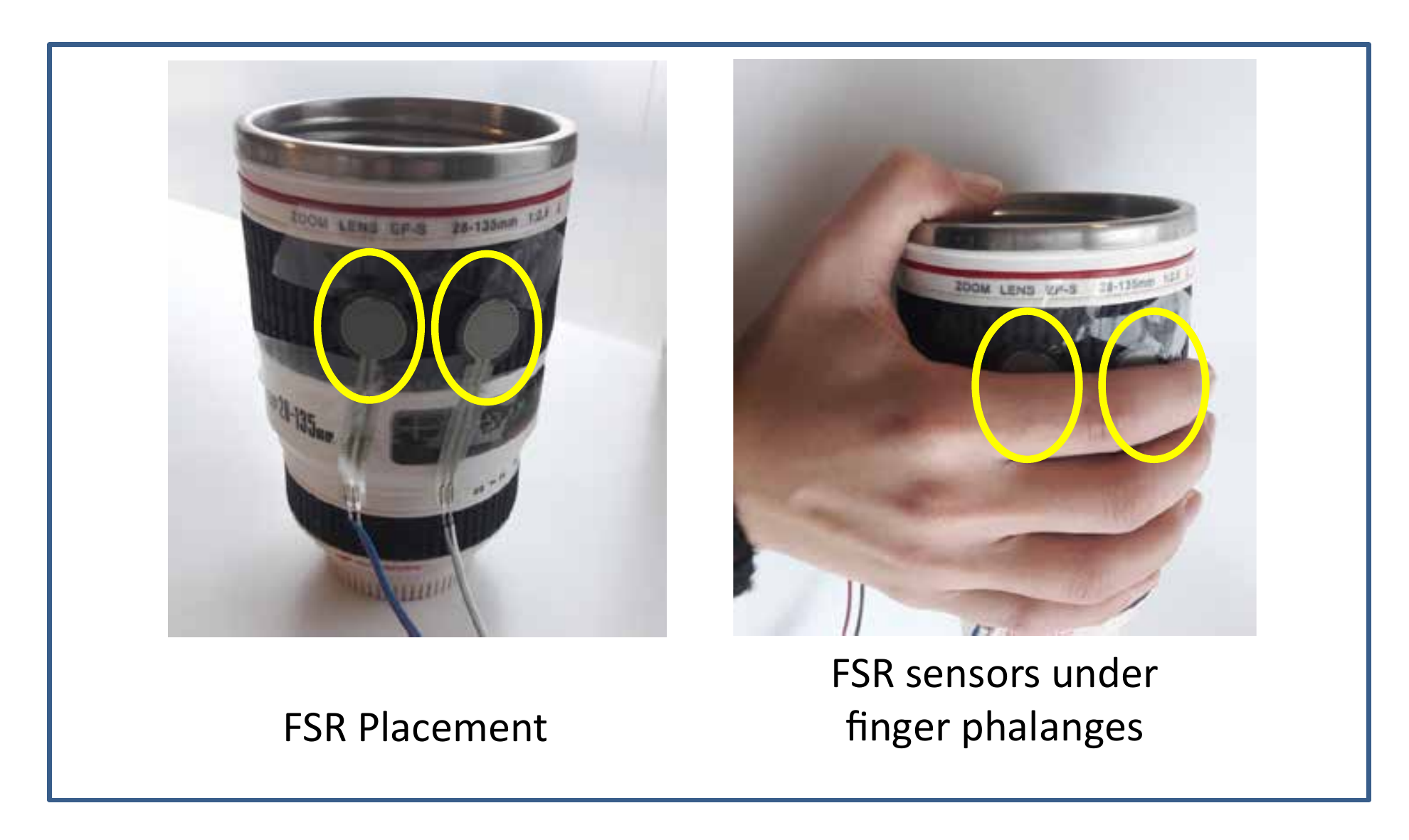}\label{fig:lens}}
\caption{Placement of the FSR sensors to measure the interaction forces between user's finger phalanges and cylindrical objects during grasping: (a) a water bottle and (b) a mug.}
\label{fig:fsron}
\end{figure}

The feasibility of the grasping tasks should be validated by measuring the interaction forces between user's finger phalanges and objects. Two different objects with different dimensions were selected to compare the interaction forces along finger phalanges. To receive efficient measurements during the interaction with the object, the FSR sensors (see Subsection~\ref{sec:sensors}) were inserted on the object as in Figure~\ref{fig:fsron}. Doing so, the FSR sensors can interact with the proximal and the intermediate phalanges of the index finger when the object is grasped. For this verification, we selected cylindrical objects to mount the FSR sensors easily on them. 

\begin{figure}[htb]
\centering
\resizebox{5in}{!}{\includegraphics{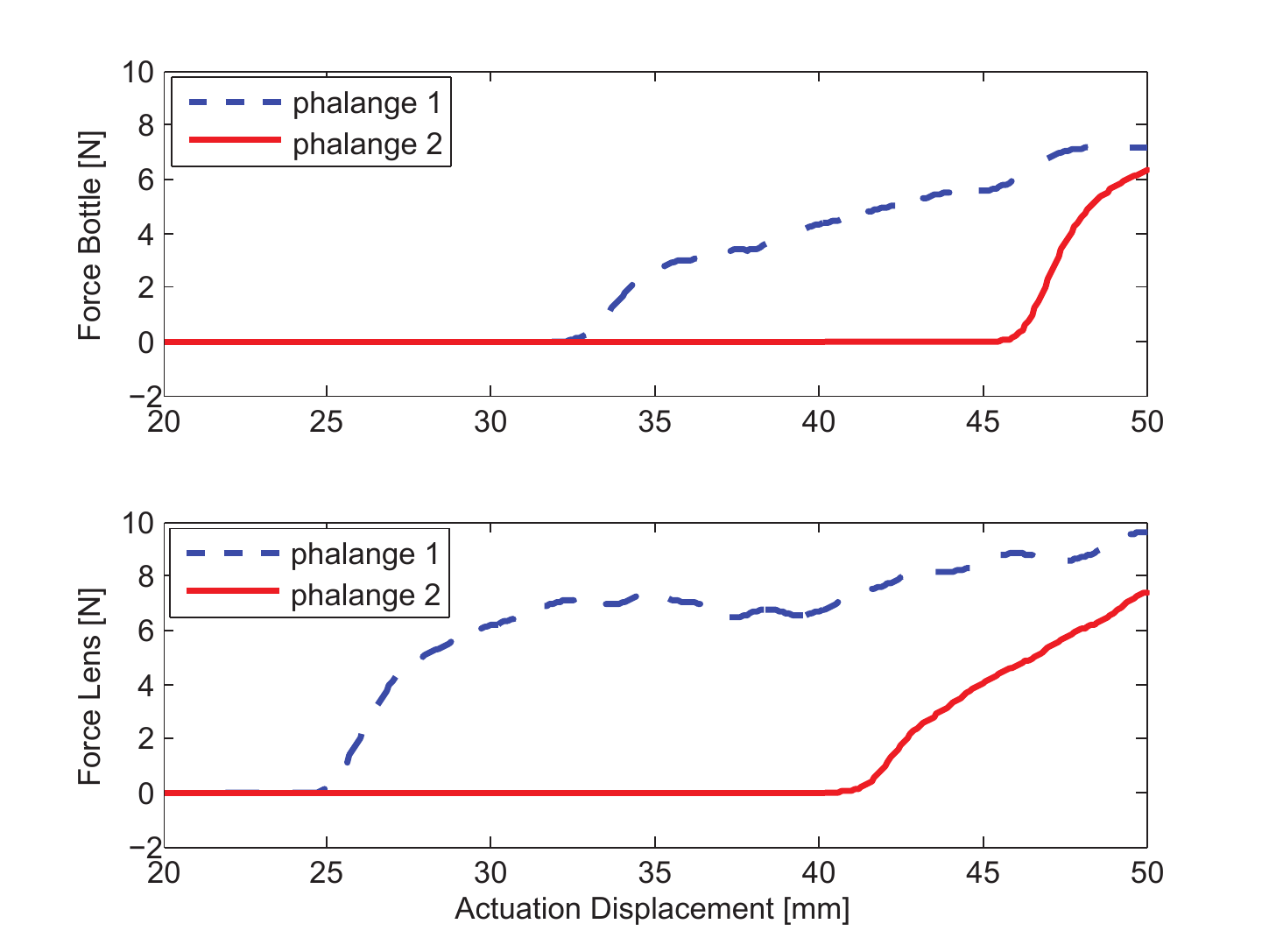}}
\caption{Interaction forces during grasping between user's finger phalanges and FSR sensors attached on the object in Figure~\ref{fig:bottle} and object in Figure~\ref{fig:lens}.}
\label{fig:grasping_force}
\end{figure}

Figure~\ref{fig:grasping_force} shows the grasping forces over the actuator displacements while the user is grasping two cylindrical objects as in Figure~\ref{fig:fsron}. The hand exoskeleton is controlled with a simple PI control using a ramp reference as detailed in Subsection~\ref{sec:position}.

In the first plot, the actuator activation starts to move the MCP joint of the user's finger, until her proximal phalange reaches the object. Then, the interaction forces start to increase, as the PIP joint starts to be rotated. The PIP joint keeps rotating until the intermediate phalange reaches the object, and the interaction forces for phalange 2 start to increase as well. The actuator poses when the interactions occur and the force measurements are shown to be different for each object, emphasizing different behaviours for grasping. Such behavioral difference can be used to prove that the device was adjusted automatically by the kinematics. Yet, the stable grasping can be achieved for both cases. Even though the grasping forces are shown only for the index finger phalanges, other fingers have the same behavior, since they use the same kinematics.

%\newpage

\subsubsection{Pose analysis} \label{sec:pose_estimation}

%Underactuation property allows the hand exoskeleton to be controlled with a single actuator to open/close fingers by moving finger joints in a coupled manner. The lack of the constant ratio between finger joints prevents the finger pose to be known using only the sensory measurements coming from the actuator. 

\begin{figure}[htb]
\centering
\includegraphics[width=0.7\textwidth]{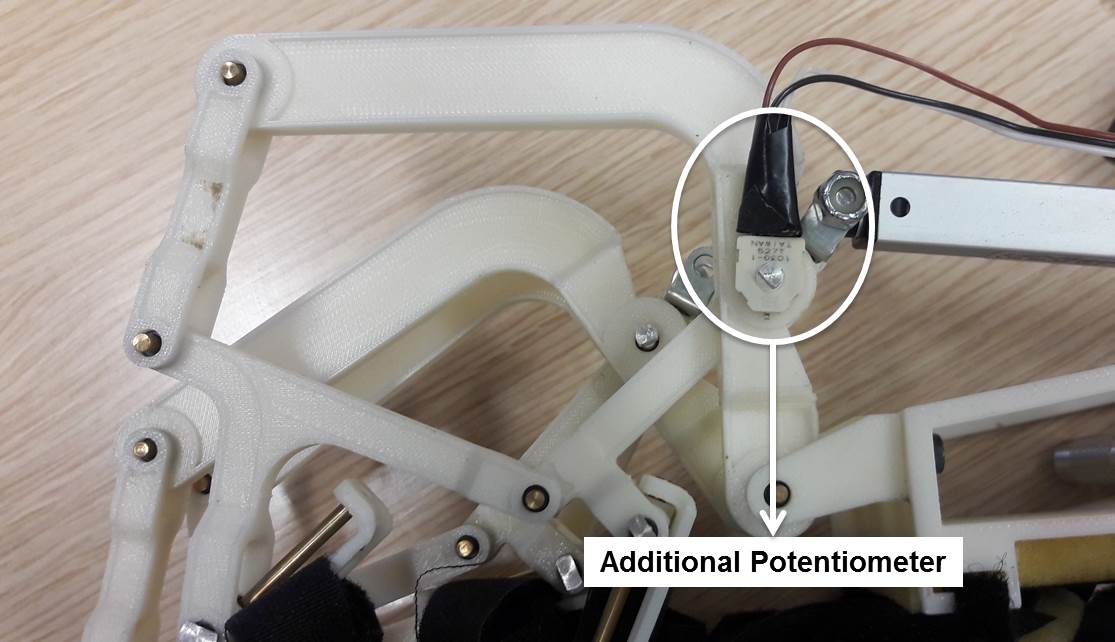}
\caption{An image of the rotational potentiometer for an additional sensory measurement along one of the passive mechanical joints to predict user's pose during operation.}
\label{fig:potentiometer}
\end{figure}

As detailed in Subsection~\ref{sec:forward}, a unique solution for the finger pose can be achieved by placing an additional sensor along one of the passive joints of the mechanical system, and by calculating the forward kinematics of the system. A simple rotational potentiometer can be placed along the mechanical system as represented in Figure~\ref{fig:potentiometer}. The miniaturized sensor size does not cause the system to be bulky and heavy.

Section~\ref{sec:forward} and Section \ref{sec:forward_analytical} solves the forward kinematics problem using numerical and analytical approaches. Even though they give same results in simulations, they have different characteristics in real life. Numerical solution highly depends on initial values for finger pose and mechanical joints, which might cause the finger pose prediction to converge for an unrealistic value.

Another difference between these two approaches is their calculational burden. The analytic solution is composed of single lines of equations, which can be calculated in a fast manner, while the numerical approach might require a couple of iterations to take place. How much time each iteration takes depends on the configuration and initial estimation. For the cases where finger pose is used only for recordings, such a calculational burden can be easily ignored. However, we need the kinematics to be calculated fast and accurately to use the finger pose as a feedback for the control algorithm. 

%The numerical approach can be calculated even in the microcontroller to achieve such fast results, especially for the applications where the finger pose is needed to be used in the control algorithm. 

A final issue is the physical capabilities of the utilized control board. In the proposed electronics design, a single DSP board is used to read all sensory measurements, run kinematics calculations and run a controller to supply motor drivers for $5$ finger components simultaneously. The simplest kinematics approach should be utilized within the control board to calculate the whole kinematics within a single time step, without interruption. 

All these issues address a single solution, where analytical forward kinematics approach should be used to calculate finger pose estimations simultaneously. The finger pose estimation was performed while the index component was controlled by a simple PI controller with a ramp function reference (see Subsection~\ref{sec:position}). During the experiment, the user was asked to move her MCP and PIP joints in different manners:

\begin{itemize}
\item[(a)] keeping the MCP joint stable, and moving only the PIP joint,
\item[(b)] keeping the PIP joint stable, and moving only the MCP joint, and
\item[(c)] moving both the MCP and the PIP joints together freely.
\end{itemize}

\begin{figure}[!b]
\centering
\includegraphics[width=0.9\textwidth]{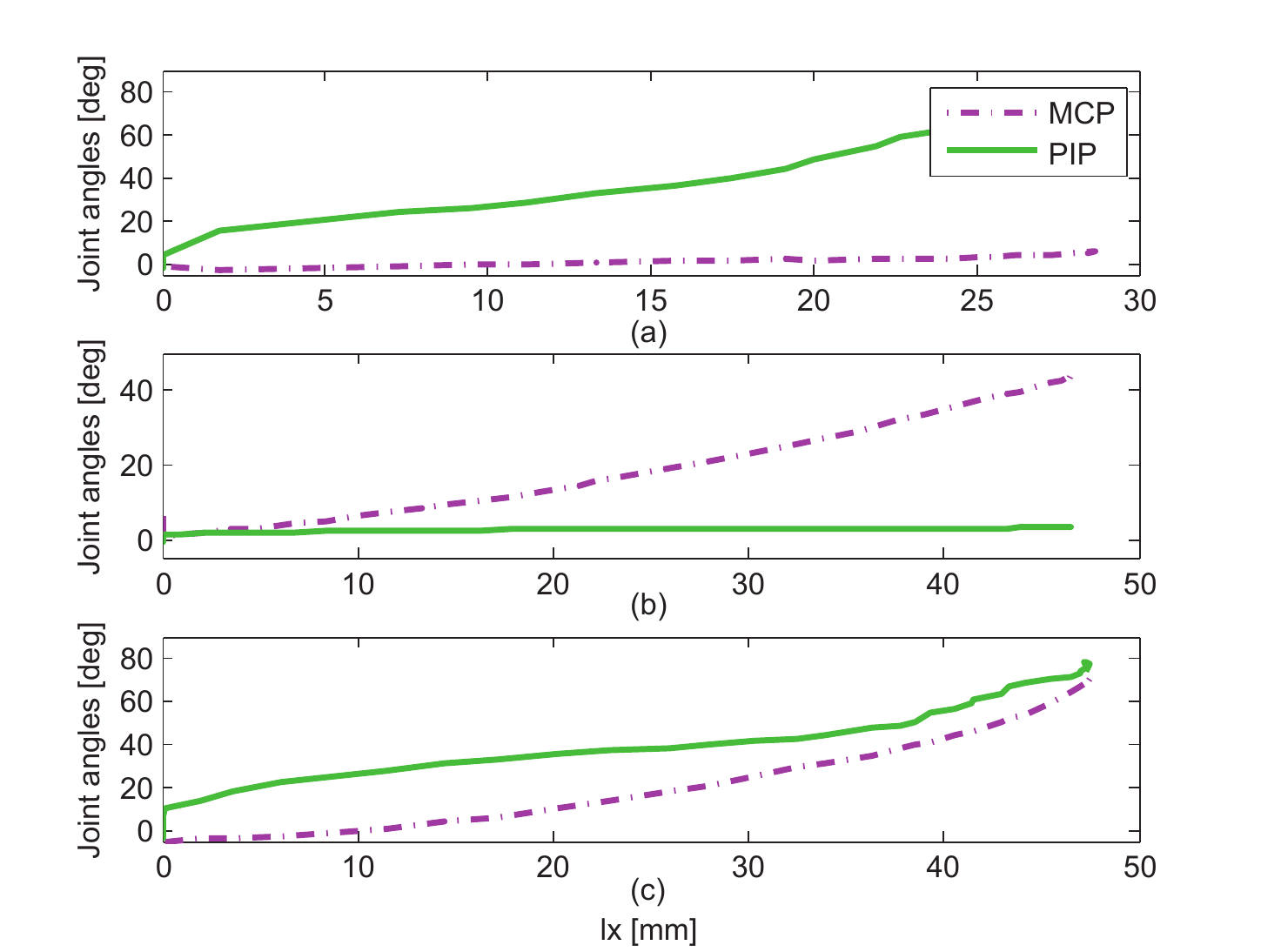}
\caption{Results of pose estimation for different tasks over the actuator displacement $l_x$: (a) moving PIP joint with stable MCP joint, (b) moving MCP joint with stable PIP joint, (c) moving both MCP and PIP joints.}
\label{fig:pose_estimation}
\end{figure}

Physical objects were used to constrain all joint movements during the tasks (a) and (b). Pose estimation results during these three tasks were recorded by the host computer and presented over actuator displacements in Figure~\ref{fig:pose_estimation}. This figure also shows that the same actuator motion can cause the finger move differently, so it presents how the underactuation works and why the additional sensor is necessary for the unique pose estimation.

\begin{figure}[!h]
\centering
\includegraphics[width=1\textwidth]{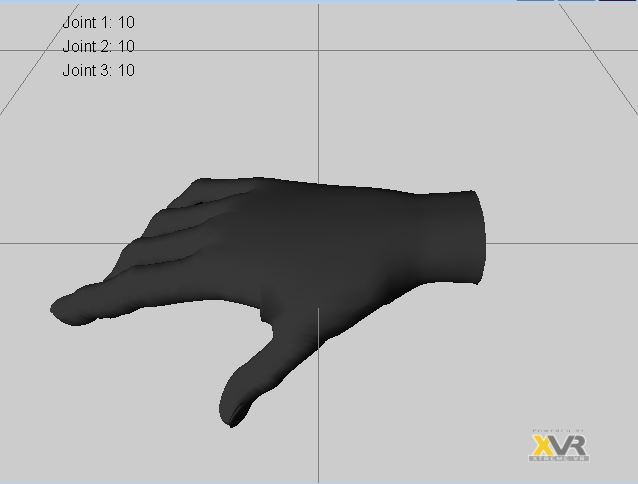}
\caption{Hand mannequin model to visualize finger joints, which are calculated using pose estimation, simultaneously during operation.}
\label{fig:handmodel}
\end{figure}

The actual finger pose can be used for future serious game scenarios, where the exoskeleton and the application tasks are integrated in the virtual environment. A simple snapshot of such a simulation can be observed in Figure~\ref{fig:handmodel}. For such an application, the MCP and PIP joints of user's hand can even be calculated by the host computer independently from the control algorithm, and these joint rotations are mapped to the hand mannequin. The DIP joint is rotated equally as the PIP joint. %Even though such simple simulation would be insufficient for a serious game scenarios or haptic rendering applications, it is useful to give the user a brief image about corresponding motion.

\newpage
\subsubsection{Calibration for finger sizes}

Since the finger phalanges are accepted as a part of the mechanism, the lengths of finger phalanges are considered as a significant parameter for the operation performance as well as the active workspace of finger joints. Even though measuring the lengths of proximal phalanges manually might be sufficient, running a calibration block based on forward kinematics might be more accurate, practical and professional compared to manual methods.

Forward kinematics algorithm can also be calculated, under a certain pose of the user, where passive linear sliders around the middle finger phalange reaches its maximum stroke (Section \ref{sec:forward}). Such a calibration process is needed for each user or patient only once, and this is why it is sufficient to calculate it in the host computer. This actually prevents the necessity to change the built-in code to be used for the actual tasks for each user, increasing the preparation time. Even though the pose estimation was previously calculated through the analytic approach, the calibration can be performed through the numerical kinematics easily. 

\begin{figure}[!htb]
\centering
\includegraphics[width=0.325\textwidth]{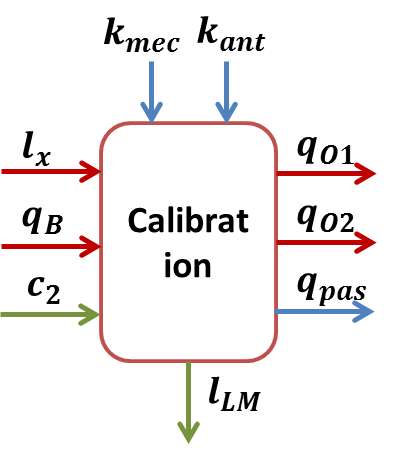}
\caption{A representative scheme shows the pose estimation based calibration to estimate the length of the first finger phalange at a pre-defined pose of user's finger.}
\label{fig:calib}
\end{figure}

When the user reaches a certain pose, a possible singularity might occur when both the MCP and the PIP joints are fully extended and a third measurement estimation can be used without utilizing an additional sensor to run calibration process. Figure~\ref{fig:calib} summarizes the calibration process, where $c_2$ is a constant with its maximum range, $l_{LM}$ represents the length of proximal phalange, $l_x$ and $q_B$ are the sensory measurements.

\begin{figure}[!htb]
\centering
\includegraphics[width=0.75\textwidth]{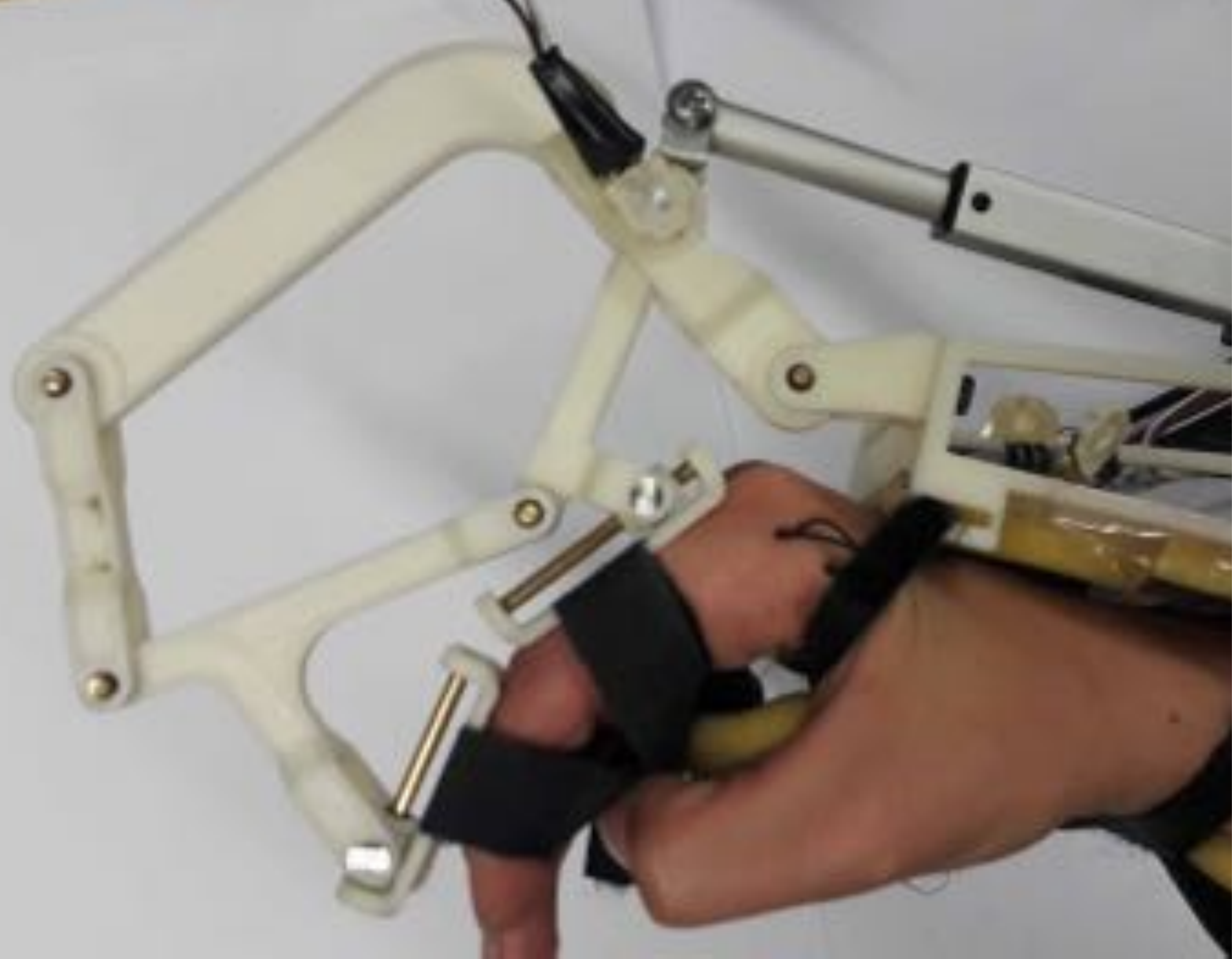}
\caption{Pre-defined pose needed to compute the length of first finger phalange during calibration.}
\label{fig:calib_pose}
\end{figure}

Such a calibration method would work only if user reaches a previously determined pose, to reach a constant value not only at the passive slider, but also at the actuator ($42~mm$). The stable finger pose for the calibration can be observed in Figure~\ref{fig:calib_pose}. With this pose, all controllable joints are known, and the uncontrollable joint $q_B$ is measured. In the end, the numerical forward kinematics can calculate the finger phalange length. 

The calibration process was tested on a single user for the index finger only and her finger length was calculated as $49.19~mm$, while manual measurement was found as $48.32~mm$ with an error rate of $1.77\%$. It is important to note that the initial guess and the finger pose should be close to each other. The initial guesses for passive joints should be determined carefully to increase the analysis performance.

%This process needs to be computed for each patient or user once before the operation under a certain, single mechanical configuration. 

\newpage
\subsection{EMG Control} \label{sec:emg_cont}

For rehabilitation applications, the active participation of patients with disabilities might increase the efficacy of treatment process as much as their interest to the tasks, compared to the passive tasks. One common way of achieving such an active participation is to control the hand exoskeleton, which is worn by the impaired hand, using the muscular activity of the unimpaired hand, which can be measured through Electromyography (EMG) sensors. 

In the literature, most of the studies use the filtered EMG signals to move multiple fingers worn by the exoskeleton together. Wege \textit{et al.}~\cite{Wege2006} created this system, where the EMG signals are used to control fingers independently by choosing the specific muscles to be measured implicitly. In their study the control of little finger component is coupled with ring finger activity, since they share similar muscles to function. 

\begin{figure}[!htb]
\centering
%\vspace*{-.5\baselineskip}
\includegraphics[width=0.7\textwidth]{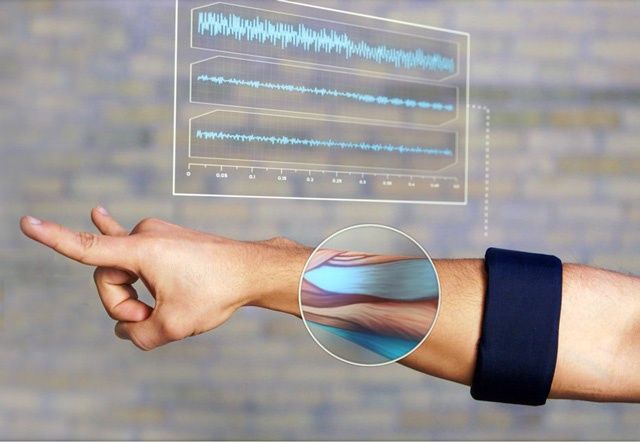}
\caption{Myo bracelet provides the muscle activity of hand gestures in a compact manner~\cite{MyoWebsite}.}
\label{fig:myo_muscle}
% \vspace*{-1.5\baselineskip}
\end{figure}

Usually, the EMG recordings during rehabilitation tasks are collected by attaching individual EMG sensors on each muscle. Even though this attachment approach provides the therapist to reach required muscle activities in a precise manner, placing these sensors requires a longer preparation time. Furthermore, the exact position of each sensor is highly important to obtain meaningful measurements. Figure~\ref{fig:myo_muscle} shows a compact and wearable Myo bracelet~\cite{MyoWebsite}, which can obtain EMG measurements, and address these issues easily. 

\begin{figure}[!htb]
\centering
 %\vspace*{-.5\baselineskip}
\includegraphics[width=0.4\textwidth]{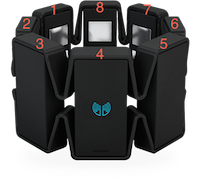}
\caption{Distribution of 8 channel EMG sensors around the Myo bracelet.}
\label{fig:myo_emg}
% \vspace*{-1.5\baselineskip}
\end{figure}

Myo bracelet is a low-cost, compact device mostly developed for gesture tracking, measuring $12$ channels of inertial measurement unit (IMU) and $8$ channels of EMG activities. In this work, only EMG measurements are collected from Myo, while IMU recordings are neglected. $8$ channels of EMG measurements are taken from $8$ EMG sensors placed along the bracelet. Figure~\ref{fig:myo_emg} shows the distribution and placement of these EMG sensors, which allow the muscle activity to be captured all around the arm. Since the quality of gesture estimation depends on the placement of these sensors around the arm, how the device is worn is highly important.

The EMG measurements can easily be extracted from its gesture control software and be used for alternative applications. The device can sends these EMG measurements to the host computer through Bluetooth. Its sensitivity about finger movements, low cost and easy wearability makes Myo sufficiently suitable for teleoperation tasks of the hand exoskeleton.

\begin{figure}[!b]
\centering
\includegraphics[width=1\textwidth]{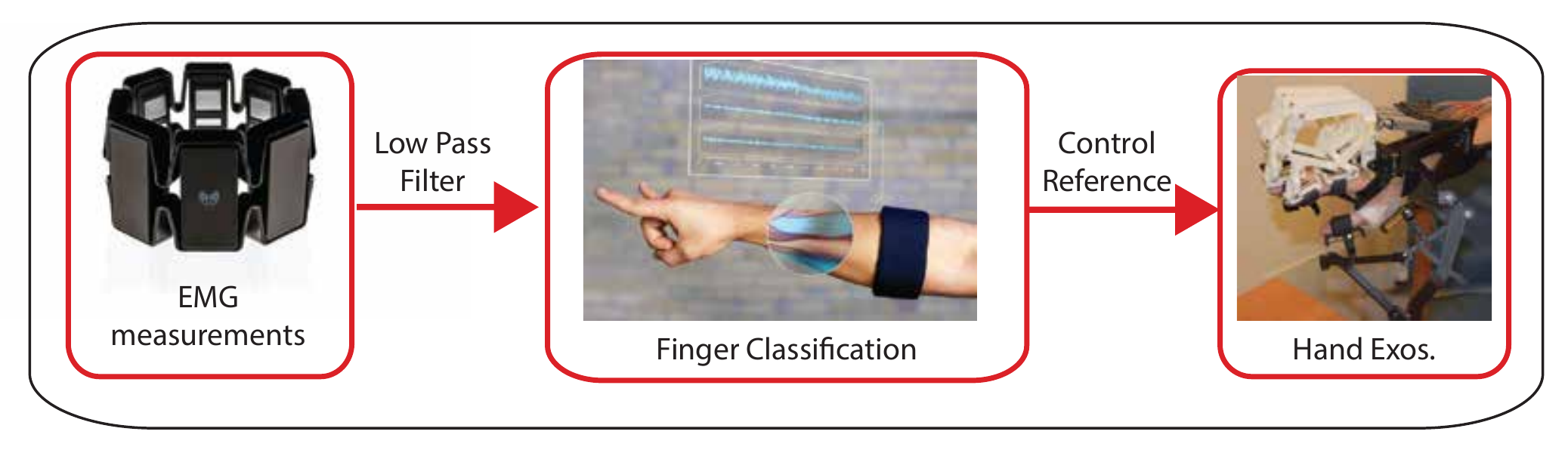}
\caption{A representative scheme to use the EMG measurements as a reference to control fingers independently using a low pass filter and an observation based finger classification process.}
\label{fig:myo_loop}
\end{figure}

Figure~\ref{fig:myo_loop} shows the sequence of the required actions to control the hand exoskeleton based on the EMG measurements from the Myo bracelet. The first process filters the raw measurements. These signals are classified to detect individual finger activities through the intensity of muscle activity using an observation method. The finger activities are normalized to have values between $0$ and $1$, while these normalized values are used as a position reference for actuator displacements between $0$ and $50$. 

The EMG signals extracted from the Myo bracelet are modified to reach more meaningful and useful shape by the classification phase. Since the raw EMG signals are so noisy, a low-pass filter with the frequency of $10~Hz$ is needed to achieve clear signals to be processed. 

\begin{figure}[!b]
\centering
%\vspace*{1\baselineskip}
\subfigure[Index finger calibration]{\includegraphics[width=0.48\textwidth]{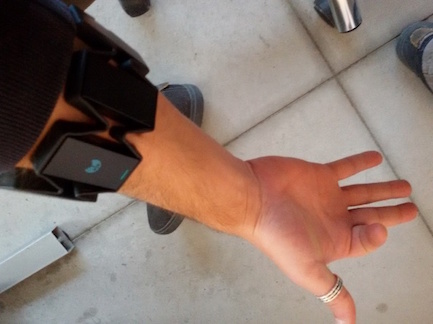}\label{fig:myo_index}} 
\subfigure[Middle finger calibration]{\includegraphics[width=0.48\textwidth]{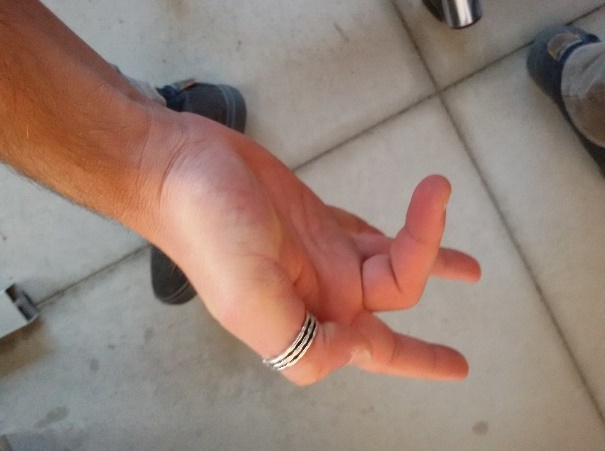}\label{fig:myo_middle}}\\
\subfigure[Ring finger calibration]{\includegraphics[width=0.48\textwidth]{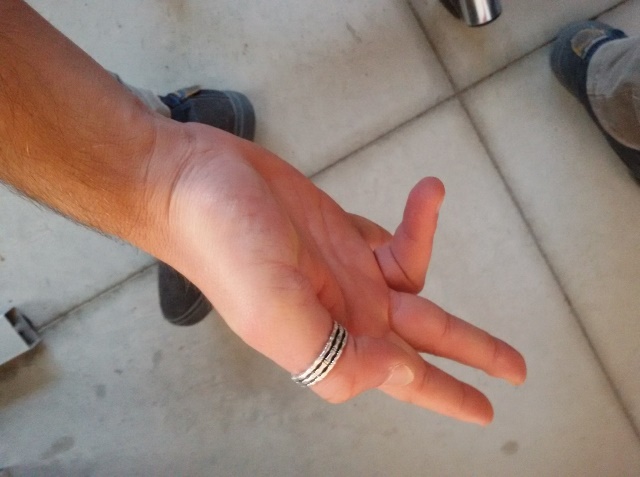}\label{fig:myo_ring}}\\
\caption{Classifying the most effective EMG channels measured by Myo bracelet corresponding to flexion of index, middle and ring fingers independently.}
\label{fig:myo_calibration}
\end{figure}

The filtered EMG signals are observed for each finger movement independently until various weights for the $8$ EMG channels of Myo were found to obtain independent finger movements. We focus on the index, the middle and the ring finger activities, while the little finger is coupled with the ring finger. The ring finger can be moved the same as the ring finger, because they are controlled by the same muscles on the arm. Figure~\ref{fig:myo_calibration} shows the training process to observe the EMG activity based on independent finger movements.

\begin{figure}[!htb]
\centering
\includegraphics[width=0.93\textwidth]{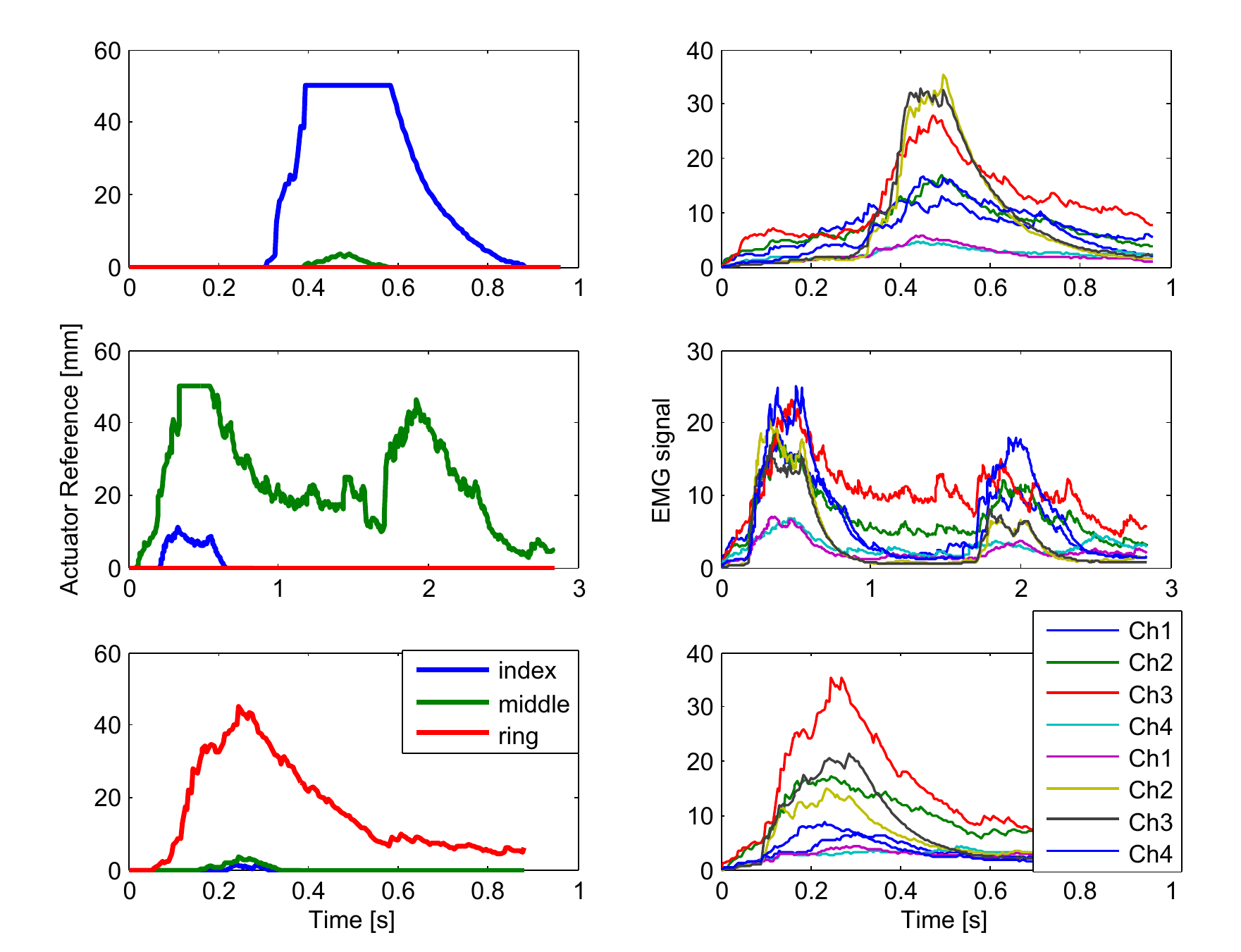}
\caption{The actuator references depicted for independent finger movement through classified EMG activity.}
\label{fig:finger_classification}
\end{figure}

The training process for the finger classification starts by relaxing all fingers to ensure Myo does not detect any activity. Then, each single finger is flexed separately as in Figure~\ref{fig:myo_calibration} while $8$ EMG activities of are observed to detect at least $2$ channels being affected by the movement the most. When these activities are combined, the finger activities are obtained and normalized between $0$ and $1$. Finally, these normalized values are multiplied with the maximum actuator displacement to create the position reference for each actuator. Figure~\ref{fig:finger_classification} shows the actuator references for each finger component using the EMG measurements, while moving the index, the middle and the ring fingers separately.

The Myo bracelet can be used for teleoperation tasks with individual finger control. Even though we can reach independent finger control, we will not design a task with the hand exoskeleton, where we expect users to open/close their fingers to their anatomic limits. We know that we, humans, have fingers, with certain coupling between them and in our daily lives, we tend to move them together in different strategies while performing different tasks. It is extremely useful to be able to detect independent finger movement and control the exoskeleton accordingly, but it will be used only to achieve task adjustability. 

\begin{figure}[!htb]
\centering
 %\vspace*{-.5\baselineskip}
\includegraphics[width=0.7\textwidth]{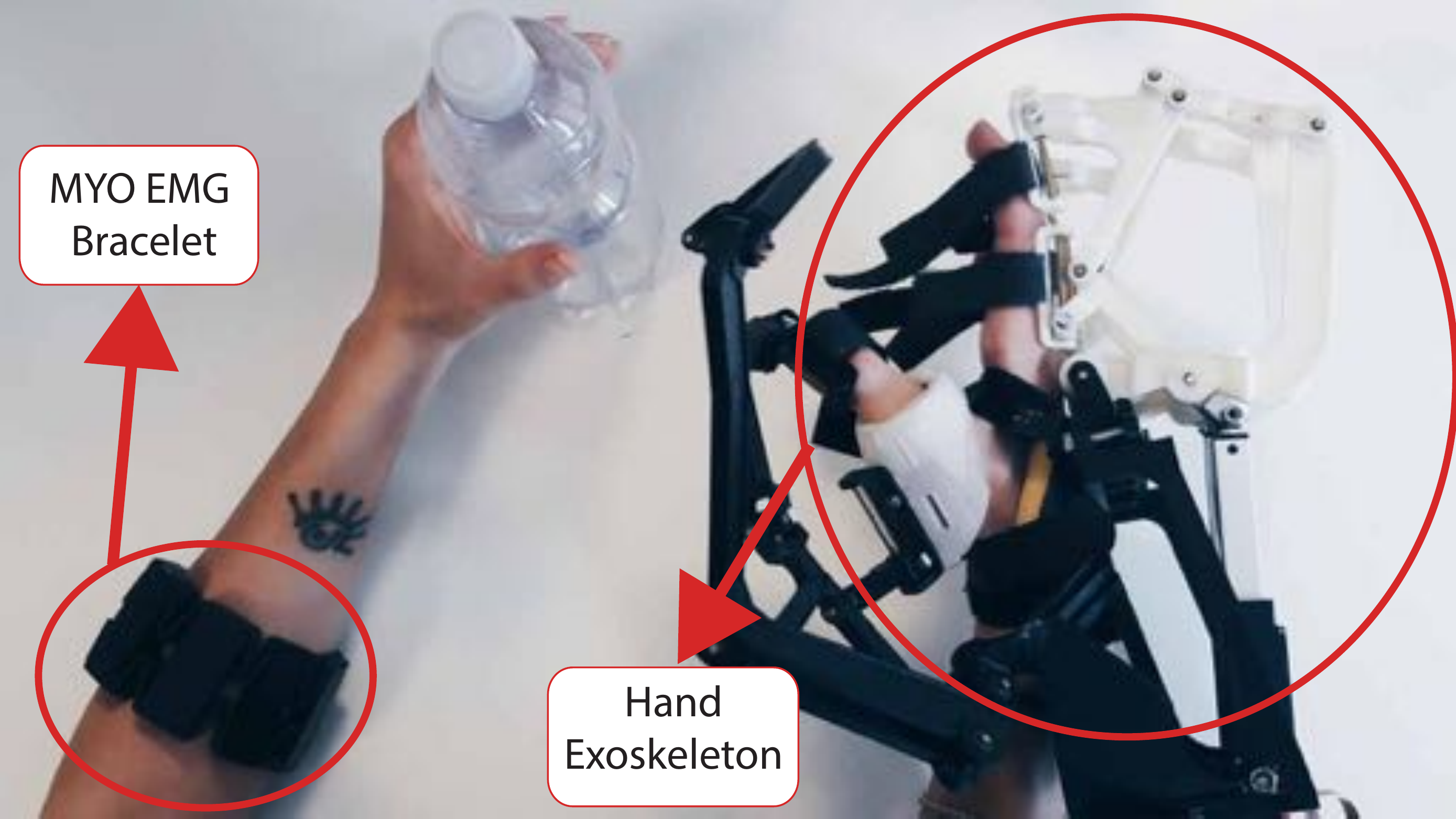}
\caption{The experimental setup for EMG-controlled underactuated hand exoskeleton with MYO bracelet.}
\label{fig:myo_exos}
% \vspace*{-1.5\baselineskip}
\end{figure}

Figure~\ref{fig:myo_exos} shows an example experiment with a Myo bracelet and the hand exoskeleton. The EMG activity occurs only when the fingers are flexed strongly, and disappears when they are relaxed. For such a teleoperation scenario, the left arm with the Myo bracelet is designated as the master, which detects the reference finger movements, while the right arm with the exoskeleton is designates as the slave, which mimics the finger movement of the left arm. The user is asked to 

To obtain tasks that are similar to ADLs, user was asked to push her fingers that are attached to Myo bracelet against an object, a full water bottle in this case, to control her other fingers individually. Using a full water bottle allows EMG activity to be stronger, which would result with a more accurate classification and control performance.

The position reference obtained by classifier in Figure~\ref{fig:finger_classification} are directly given to position control algorithm as detailed previously in Subsection~\ref{sec:position}. Since performance of the position control algorithm was discussed previously, controller performance is not presented again in order to avoid repetitions.

\newpage

\subsection{Force Control} \label{sec:force_cont}

\begin{figure}[!htb]
\centering
\includegraphics[width=0.9\textwidth]{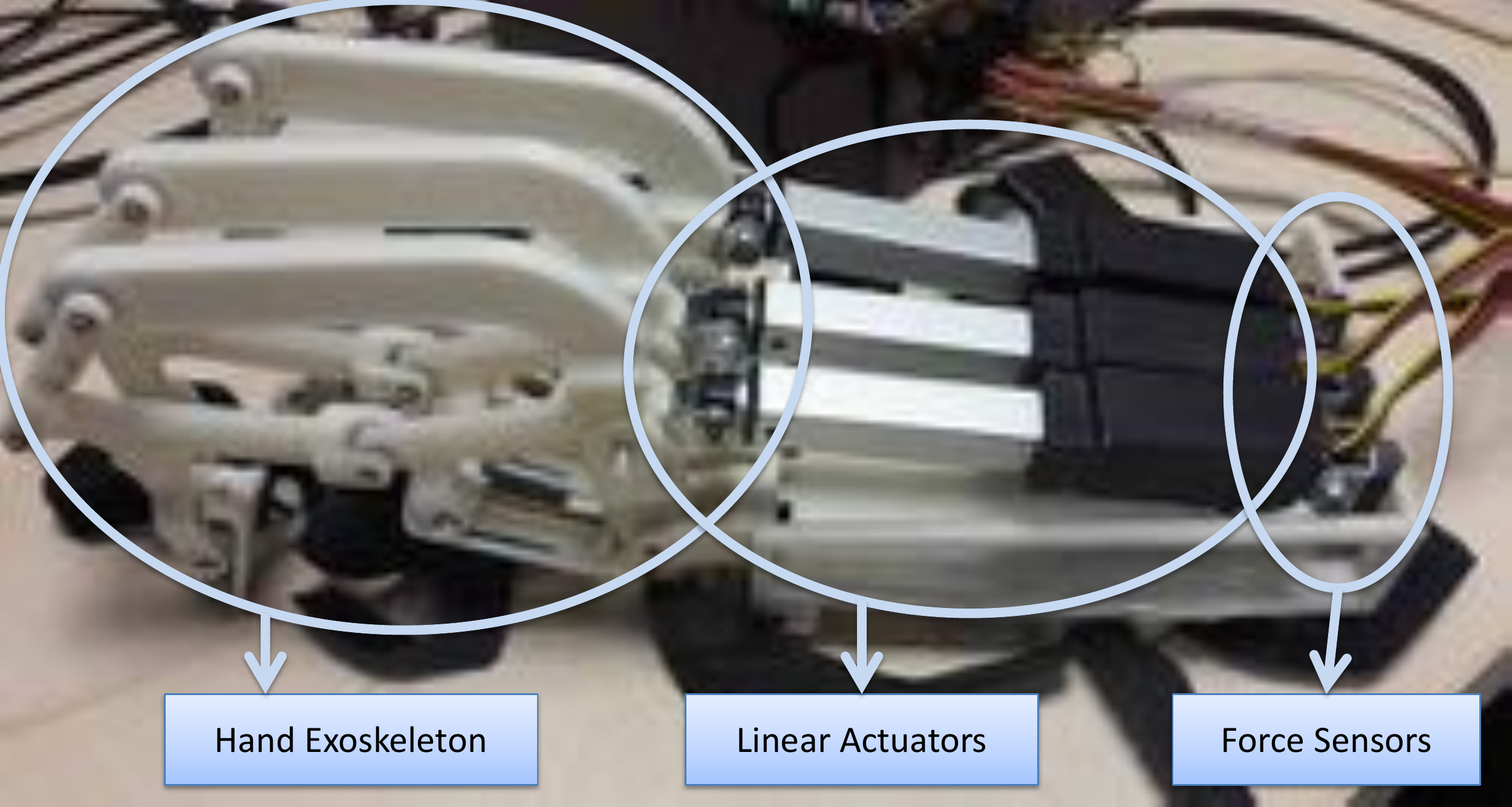}
\caption{Utilization of $1-DoF$ force sensors for each finger component independently to measure user's intention to open/close his finger and move the actuator with high transparency to achieve backdriveability over control.}
\label{fig:setup}
\end{figure}

Firgelli $L16$ actuators were chosen to control the underactuated hand exoskeleton due to its compact and portable design, despite the fact that they are nonbackdriveability. Even though these actuators prevent the proposed hand exoskeleton to reach backdriveability when the actuators are off, it can be achieved actively, by measuring the forces coming from the user and move the actuators accordingly. Figure~\ref{fig:setup} shows the assembly of single $1-DoF$ miniaturized load cell with strain gauges, which are used as force sensors, to the hand exoskeleton. 

\begin{figure}[htb]
\centering
\includegraphics[width=1\textwidth]{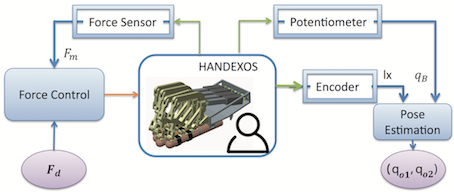}
\caption{A representative scheme of a generic force control algorithm that can be used for active backdriveability when $F_d = 0$ or haptic rendering when $F_d$ is adjusted based on the virtual interactions.}
\label{fig:force_control}
% \vspace*{-1\baselineskip}
\end{figure}

When the user, who is wearing the exoskeleton, wants to move his fingers to open/close them, the force sensor captures these forces, and a force control algorithm in Figure~\ref{fig:force_control} calculates the required PWM supply to drive motors accordingly. As a result, the hand exoskeleton moves the actuators of the finger components, until the user reaches the pose he desires. Depending on the frequency of the control loop, this movement can be achieved almost simultaneously and transparently. The speed of the motion is also adjusted based on the measured forces. 

For active backdriveability, the measured forces $F_m$ created by the user should be the only the reference for the control scheme, so the $F_d$ is set to $0$. Equation~(\ref{eq:backdriveability}) explains the mathematical expression of the force control block represented in Figure~\ref{fig:force_control}. Actuators are controlled using PWM signals $F_{PWM}$ directly related to the measured forces $F_m$, where $K_P$ and $K_I$ are proportional and integral gains of the force controller, since desired force is kept $0$.

\begin{equation} \label{eq:backdriveability}
%\vspace*{-0.5\baselineskip}
F_{PWM} = K_P F_m + K_I \int F_m.
\vspace*{0.3\baselineskip}
\end{equation}

\begin{figure}[htb]
\centering
\resizebox{4.2in}{!}{\includegraphics{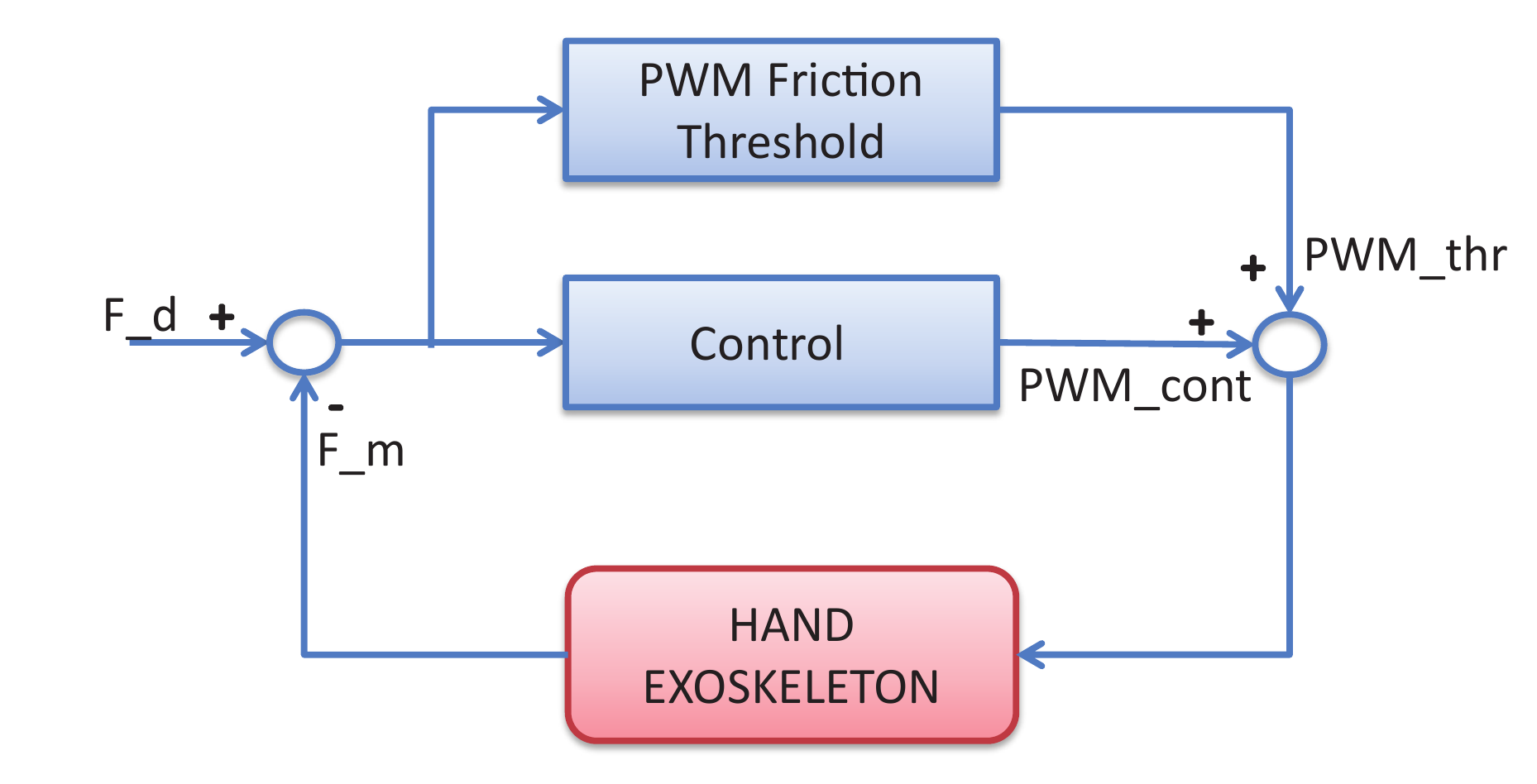}}
\caption{Adjusted force control algorithm with additional PWM supply in the same direction of control algorithm to overcome internal actuator friction.}
\label{fig:for_alg}
\end{figure}

Very similar to the Figure \ref{fig:pos_alg}, the performance of the force control will suffer from the delay between the controller command and the initialization of the actuator movement. Therefore, an additional PWM support is added to the controller output before sending PWM signals to motor drivers.  Figure \ref{fig:for_alg} shows the representative support to overcome backdrive forces during control. 

\begin{figure}[htb]
\centering
\includegraphics[width=0.85\textwidth]{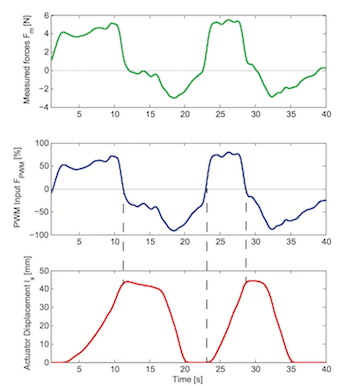}
\caption{Real time implementation of an active backdriveability task using the proposed index finger component of the exoskeleton: (a) measured forces through additional force sensor, (b) PWM input for the given control algorithm in Eqn.~(\ref{eq:backdriveability}) and (c) obtained displacement of the actuator to open/close the finger.}
\label{fig:backdrive_output}
\end{figure}

The PWM signal strictly adjusts the actuator velocity through input current, so that the user can control the movement speed simply by increasing the applied forces. Figure~\ref{fig:backdrive_output} shows the active backdriveability task results. The first plot shows the applied forces by the user, the second plot shows the produced PWM output, and the third plot shows the corresponding actuator displacement. Since the PWM signal controls the actuator velocity, the actuator displacement increases when the PWM value is positive and decreases as the PWM value is negative. Backdrive forces shown in Figure~\ref{fig:nonback} introduces a delay between the second and the third plots. Nevertheless, the relation between the applied forces and the actuator displacement shows that the control algorithm provides sufficient transparency to the device for haptic applications.

\clearpage
\pagebreak

\newpage
\chapter{Haptic Rendering} % Main chapter title

\label{sec:Chapter5}% For referencing the chapter elsewhere, use \ref{Chapter1}

\lhead{Chapter 5. \emph{Haptic Rendering}} % This is for the header on each page - perhaps a shortened title

Haptic rendering focuses on the delivery of virtual object properties to a user through sense of touch using robotic devices. In particular, haptics make it possible to access places and tasks that are too dangerous and too expensive to explore, such as surgical training, video games, simulators, CAD systems for engineers, physical training exercises, etc. With improvements in haptics technology. Haptic tools for these applications have the potential to generate realistic scenarios economically and to create the perception of touch based on the properties of virtual objects, such as shape, size, weight and stiffness.

The perception of touch can be divided into two groups: tactile and kinesthetic. The tactile sensation consists of the perception of pressure, warmth, cold, pain, vibration, etc. The tactile devices attempt to mimic the characteristics of virtual object and local stiffness of the surface, targeting the mecanoreceptors on our skin to detect these properties, without moving the muscles of the user. On the other hand, the kinesthetic system refers to the feedback regarding muscles, tendons and joints for the perception of motion and exchanged forces.

In this chapter, a brief literature survey will be presented for the haptic devices and the approaches to utilize underactuated devices for haptic rendering. We will mostly focus on the virtual grasping tasks, since we are searching for ways to implement haptic scenarios for a hand exoskeleton designed in the previous chapters. The feasibility of force transmission and stiffness rendering tasks will be shown with a simple actuator level rendering task and by simplifying the exoskeleton mobility to a single $DoF$. Once the force transmission of the mechanism and Jacobian is proven, various rendering strategies will be developed based on transmitted forces to fingers and desired pose to reach. Even though these strategy ideas are initially developed to overcome the lack of controllability during haptic tasks, they can still improve the overall performance, even for simple trajectory tracking tasks. These strategies were developed under the assumption of $2~DoF$ mobility guided by anatomical joints and stiff bone structure. Finally, a similar proxy based haptic rendering algorithm will be developed for a haptic simulation of soft finger tissue simply without simplifying the finger mobility to $2~DoF$. Nevertheless, both strategies are generated sharing similar principles.

\newpage
\section{Haptic Devices}

Haptic devices share some characteristics to provide sufficient force feedback through virtual interactions, such as low backdriveable inertia and friction, or minimal motion constraints regarding the device to increase the finger workspace. High control quality can be ensured only by improving the range, the resolution and the bandwidth of position sensing, and force reflection requirements. %Furthermore, the device should be designed in an ergonomic manner to make the user comfortable during operation without any pain, discomfort or attention deficit.

Haptic devices in the literature can be classified based on the number of DoF for motion, which signifies the number of dimensions in an actuated or non-actuated manner to modify the possible movements/forces exchanged between the device and the operator~\cite{Salisbury}. The impedance-type devices in the literature can be listed as $1-DoF$, $2-DoF$, $3-DoF$, $6-DoF$ and $>6-DoF$ devices. 

\begin{itemize}
\item \textbf{$1~DoF$ devices:} Most of the $1~DoF$ devices provide tactile feedback perpendicular to user's fingertip. A set of external sensors are equipped to track the position and the orientation of the fingertip. Once a contact takes place in the virtual environment between the user's avatar and the object, the device applies force and/or vibration to imply the textile, the stiffness or other properties of touch. Depending on the application, these devices can be designed in form of a haptic knob~\cite{Snibbe2001, Beni2014}, a haptic scissor~\cite{Okamura2003}, or a force reflecting gripper~\cite{Barbagli2003}. \\

\item \textbf{$2~DoF$ devices:} Tactile devices can be designed to impose $2~DoF$ feedback in perpendicular and linear tangential directions to the fingertip, instead of $1-DoF$ during virtual grasping tasks~\cite{Minamizawa2007, Minamizawa2008, Solazzi2011, Lecuyer2005}. \\
Alternatively, kinesthetic devices can constrain user's movement to a single plain such that a virtual contact can occur only in $2-D$ environment. Doing so, the perception of contact can be delivered to the user in a sufficient manner by utilizing devices in the form of a simple pantograph~\cite{Ramstein1994} or a force feedback mouse (Logitech, Wingman). \\

\item \textbf{$3~DoF$ devices:} Tactile devices can also impose $3~DoF$ feedback in perpendicular and planar tangential directions to the fingertip~\cite{Solazzi2010, Chinello2012, Benko2016}. Doing so, all interaction forces in the virtual contact can be given to the user as in real life, even though the interaction torques are ignored. \\
From the kinesthetic point of view, haptic devices can support contact appearance in $3-D$ environment. OMEGA~\cite{Grange2001} provides $3~DoF$ feedback from its end-effector, where users move its end-effector and grasp virtual objects. Phantom~\cite{Massie1994} on the other hand provides an active $3~DoF$ feedback to user's finger, while a passive thimble worn at the fingertip provides an additional $3~DoF$ mobility. In particular, passive $DoF$ allow the fingertip orientation to be known, such that the pure force transmission to the fingertip can be used for $6~DoF$ virtual tasks. Nevertheless, contact torques between the user's avatar and the virtual object are ignored.\\

\item \textbf{$6~DoF$ devices:} Tactile devices can also be extended to assist $3~DoF$ contact torques together with contact forces. \\
The same idea goes for the kinesthetic feedback in terms of upgrading $3~DoF$ devices to support also the interaction torques. In particular, $6~DoF$ PHANTOM \cite{Cohen1999}, $6~DoF$ DELTA \cite{Grange2001}, and Freedom 6 \cite{Hayward1998} can measure user's movements and provide the feedback based on the virtual interactions in $6~DoF$. \\

\item \textbf{$>$ $6~DOF$ devices:} A $6~DoF$ manipulation device can be integrated with a single $DoF$ grasping mobility to complete the task~\cite{Lambert2015}. \\
Furthermore, a haptic device can be designed in the joint space rather than the Cartesian space to assist hand or arm movements during virtual operations in the shape of exoskeleton. The device can provide force feedback to the anatomical joints of the user, achieving more than $6~DoF$ overall.
\end{itemize}

Kinesthetic feedback during haptic applications can be given either by controlling the fingertips or the finger joints implicitly. Hand exoskeletons provide force feedback to the finger phalanges to stimulate full grasping tasks~\cite{Aiple2013, Costas2003, Fontana2009, Ryu2008, Feng2011, Jo2014, Wei2017}. Actuating each finger joint independently can be achieved either by aligning actuated mechanical joints with finger joints or through RCM mechanisms. Despite of mechanism choice, utilizing independent actuators for each finger joint increases the overall cost and the complexity. One way to overcome this complexity is to control only user's fingertip by achieving multiple DoF for each finger~\cite{Bergamasco2005, Stergiopoulos2003, Fesharakifard, Gu2016, Gosselin2005, BenTzvi2015, Choi2016, Halabi}. Fingertip devices are attached only to user's fingertip such that the fingertip position and orientation can be measured and resistive kinesthetic feedback is given to the user when any virtual interaction occurs. These devices have simple design and are easy to be worn by user, however the kinesthetic feedback is given only to the fingertip, such that finger phalanges are left free.

During haptic rendering tasks, the desired forces are calculated based on the stiffness value of virtual object, and the distance between the virtual contact and the user's configuration. These forces are mostly delivered to the user by utilizing fully controlled haptic devices, where all mechanical and virtual DoF are controllable individually. Using these devices, all mobility that users perform during virtual tasks can be both tracked and controlled independently. Therefore, the standard haptic rendering algorithm makes the assumption for actuated DoF of device to match the subset of proxy DoF. Unfortunately, this assumption does not hold for many interesting haptic systems. On one hand, computational models and resources grow steadily, and we can currently afford interactive simulation of rich and complex haptic probes~\cite{Perez2016, Wang2014, Xu2016}. On the other hand, novel designs of exoskeletons exploit underactuation to maximize simplicity, wearability, etc.~\cite{Laliberte2002}. In such haptic systems, actuated DoF may not map to a subset of the DoF of the haptic probe. As we show in the paper, naive application of proxy-based rendering to underactuated devices produces undesired results. A formalization and generalization of the approach can be found in~\cite{Otaduy2013}.

The proposed underactuated hand exoskeleton was designed mechanically not only for rehabilitation purposes, but also to respond the requirements of haptic applications. The mechanism was designed in a compact, wearable and portable manner. Then, the properties of pose estimation and force control algorithm allow the device to track user's configuration and provide kinesthetic feedback based on user's activity in the virtual environment. However the underactuation property, which provides many advantages to the device, requires an extension of the conventional haptic rendering algorithm.

\subsection{Underactuated Haptic Devices}

Imposing additional mobility through underactuation might create uncertainties to a haptic system. Nevertheless, their advantages based on portability, affordability and simplicity increase for underactuated devices in the haptic applications as well. The most common way to embrace underactuation in haptic applications is to create haptic devices with $6~DoF$ sensing input (translation and rotation) and $3~DoF$ actuation output (only forces, but not torques). Alternatively, underactuation can be made with $2~DoF$ or $1~DoF$ actuation output and $3-DoF$ sensory input such that the tangential forces are limited to $1$ direction only. The literature involves multiple studies on developing various strategies to handle the underactuation property.

Verner and Okamura~\cite{Verner2009} performed experimental analysis of the performance gain provided by torque feedback over force-only feedback. Lee~\cite{Lee2013} designed a modified penalty-based method that ensures admissible rendered stiffness while correctly balancing directional forces. And Kadlecek \textit{et al.}~\cite{Kadlecek2014} used sensory substitution and pseudo-haptic feedback to simulate torque feedback. Massimino \textit{et al.}~\cite{Massimino1993} studied the capabilities of sensory substitution for force feedback through the tactile and auditory senses, with and without a time delay. In particular, the efficacy of the sensory substitution during the haptic performance has been found to be dependent on the quality of the visual feedback, the characteristics of the force feedback and the complexity of the sensory substitution. Lecuyer \textit{et al.}~\cite{Lecuyer2005} uses an underactuated haptic device, where the device provides $3~DoF$ sensory information and $1~DoF$ actuation. In order to cope with such an underactuation, a new technique is proposed to improve the contact perception during point-based haptic exploration. In particular, the virtual scene is rotated when a contact occurs in order to align the contact normal to the haptic device. However, such technique can be useful when the actuated and the sensed DoF are independent from each other in the end-effector.

Barbagli and Salisbury \cite{Barbagli2003} analyzed underactuated haptic devices in terms of controllability and observability of a virtual environment, and consequently provided guidelines and considerations for the design of a device. However, they did not make an effort at trying to optimize haptic rendering settings for a given haptic device. Verner and Okamura \cite{Verner2006} analyzed passivity in asymmetric devices (i.e., those with a different number of controlled and observed DoF).

The works of Luecke~\cite{Luecke2011} and Meli and Prattichizzo \cite{Meli2014} are probably the closest to our situation with the underactuated hand exoskeleton presented in this thesis. Luecke considered a haptic device whose end-effector maps exactly to a virtual object, but it is underactuated. Haptic rendering is computed by first defining forces for the virtual object, and then finding optimal control parameters that maximize the similarity between those forces and the ones actually displayed by the device. The solution does not consider cases where the virtual object has a higher dimensionality than the end-effector or cases where the haptic device includes non-actuated DoF. Meli and Prattichizzo studied methods to render contact forces through an underactuated device, while maximizing task performance. For each task, they defined an optimality criterion in the selection of underactuated feedback. In contrast to theirs, our work stands on proxy-based rendering, but it can accommodate task-dependent metrics in the computation of the proxy.

Even though there is sufficient effort to improve the use of underactuated devices for haptic rendering, the focused underactuated devices were mostly chosen among Cartesian devices. Independency between DoF of underactuated Cartesian devices might not exist for underactuated exoskeletons, such that some of the proposed techniques cannot be implemented properly. Furthermore, it was observed that most of the optimization methods were shown to be useful under certain assumptions on the designed tasks. Utilizing proxy-based haptic rendering that focuses on optimization algorithms might be useful to generalize the use of underactuation for virtual tasks.

\newpage
\section{Virtual Interaction}

Haptic applications require users to discover a defined virtual environment and make contacts with the virtual objects within this environment. During a virtual grasping task, a haptic device should follow user's intentions with high transparency while providing resistance to user's fingers when encountered with the virtual object. Such an interaction can be designed to be performed in many case, such as exploring with fingertips, grasping, pushing, etc. The hand exoskeleton designed in this study was designed to assist grasping tasks, therefore haptic exercises in this chapter will focus on virtual grasping tasks. Nevertheless, the issues caused by underactuation concept or rendering solutions can be used for any interaction task.

In terms of perceiving force feedback, any virtual task can be divided into two groups: before contact and during contact. Before the contact, the user can move freely without perceiving any force feedback. Once the contact is established between user and virtual object, user starts to perceive resistive forces from the point of connection that signifies the behavior of the task itself. In fact, resistive forces are formed based on the stiffness of the object and forces applied by user to object throughout the task.

%\newpage
\subsection{Actuator Level Stiffness Rendering}

The behavior of force feedback during grasping can easily be imitated by a robotic device. For an underactuated device, the easiest way to implement such a control algorithm is to define a limit for the displacement of the actuator and form the desired force $F_d$ to be used in Figure \ref{fig:force_control} accordingly. Since we are dealing with an underactuated system, the desired force $F_d$ might consist of active and passive components, and to be consistent with the rest of the Chapter, we will call the active desired force $F_a$. Even though defining the stiffness rendering task based on the actuator displacement neglects the pose and orientation of finger joints, it simplifies the control algorithm by ruling out the underactuation.

As mentioned previously, the stiffness rendering task should give the perception of touch only, while allowing the user to explore the environment freely. In order to eliminate forces that might assist the user to reach desired values before getting into the contact, the positive values of the desired force $F_a$ should be kept $0$, while only the negative values should be given to the user as a force feedback related to their touch. In order to achieve such a saturation, the desired displacement $l_{x_{d}}$ is designed to be equal to actual displacement before reaching the virtual limit and to the virtual limit after exceeding it, as summarized in Algorithm~\ref{alg:des_disp}.

\begin{algorithm}
\caption{Setting desired actuator displacement based on virtual limits.}\label{alg:des_disp}
\begin{algorithmic}[1]
\State \textbf{input:} $l_x$, the actuator displacement
\State \textbf{input:} $l_{x_{lim}}$, the virtual displacement limit
\State \textbf{output:} $l_{x_{d}}$, the desired actuator displacement
\If {$l_{x} <= l_{x_{lim}}$}
\State {$l_{x_{d}} = l_{x}$}
\Else
\State {$l_{x} <= l_{x_{lim}}$}
\EndIf
\end{algorithmic}
\end{algorithm}

Equation~(\ref{eq:actuator_stiffness}) presents the mathematical expression of the desired force $F_d$ based on the instantaneous actuator displacement $l_{x}$, the desired actuator displacement $l_{x_{d}}$, which can be set as in using virtual limit for actuator displacement $l_{x_{lim}}$, and the stiffness value of the contact felt by the actuator $K_{ac}$.

\begin{equation} \label{eq:actuator_stiffness}
F_a = K_{ac}(l_{x_{d}} - l_x).
\end{equation}

Once the desired force $F_a$ is calculated, the control algorithm presented in Figure~\ref{fig:force_control} calculates the corresponding PWM signal using Equation~(\ref{eq:stiffness_rend}) and runs the actuator. In Equation~(\ref{eq:stiffness_rend}), $K_w$ signifies the weight acting on the desired actuator force $F_a$, $F_e$ defines the difference between the desired and measured force values ($F_e = F_a - F_m$), and $K_P$ and $K_I$ represent the control coefficients of a simple force PI control.

\begin{equation} \label{eq:stiffness_rend}
F_{PWM} = F_a K_w + K_P F_e + K_I \int F_e.
\end{equation}

\begin{figure}[htb]
\centering
\includegraphics[width=0.8\textwidth]{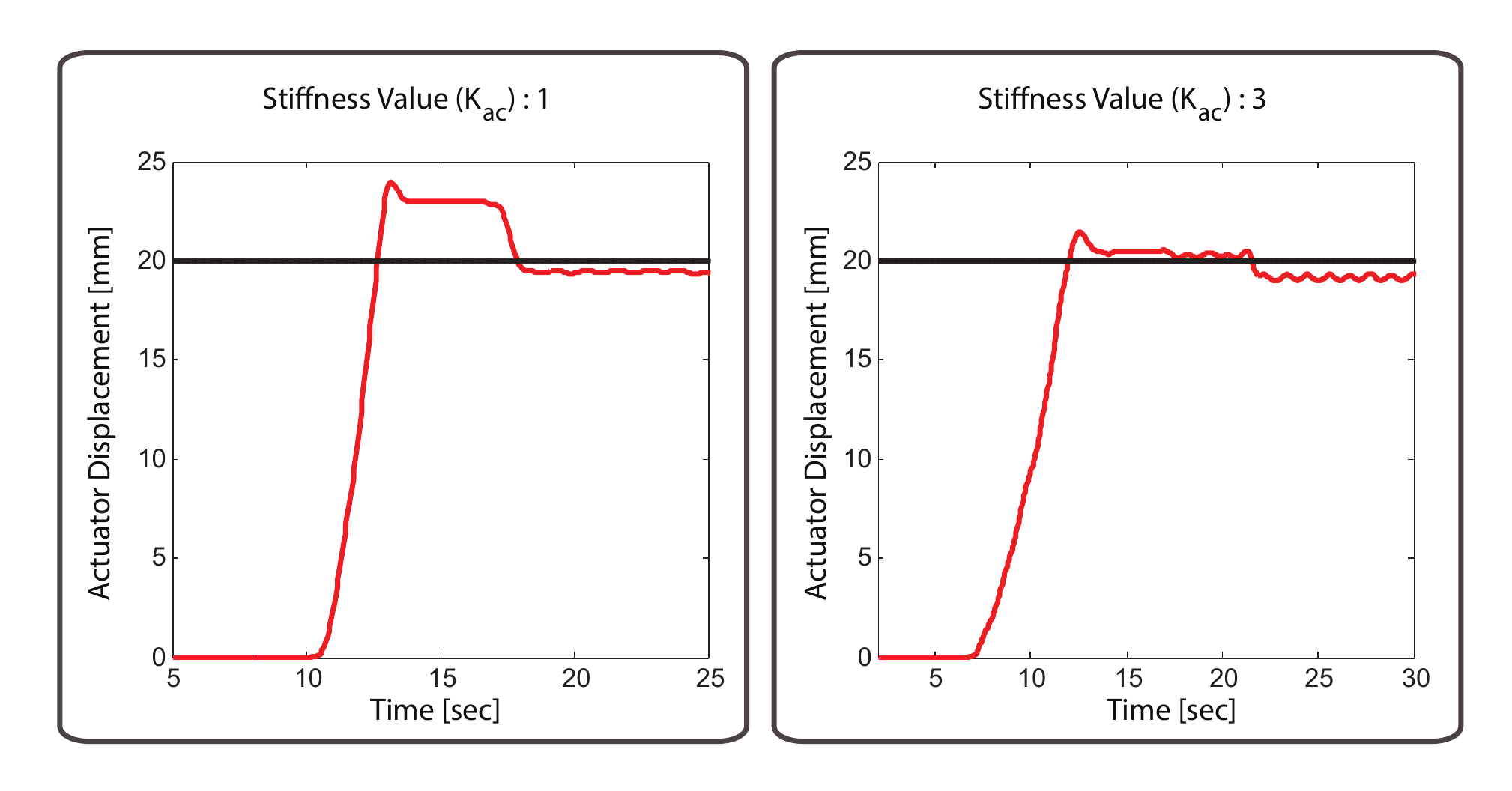}
\caption{Actuator level stiffness rendering while the displacement after the virtual contact are different due to the various stiffness values.}
\label{fig:stiffness_actuator}
\end{figure}

How additional FSR sensors are inserted between the hand exoskeleton and the userIn theory, increasing the stiffness value $K_{ac}$ in Equation~(\ref{eq:actuator_stiffness}) leads to a higher desired force value, which keeps the error between instantaneous displacement and virtual limit to minimum. Conversely, lower stiffness value allows the user to keep up his movement after reaching the contact limit as the force feedback increases, just like squeezing a soft object. Figure~\ref{fig:stiffness_actuator} shows the effect of stiffness values in terms of allowed displacement after the contact limit. Keep in mind that a kinesthetic device, which can control the muscular activity of the user, actually can prevent the user to move after a certain force level is passed unlike tactile devices. In particular, the lower stiffness value allows the user to exceed the given limit more than the high stiffness values, while the user can apply more force to the device due to such ability to move. On the other hand, when the contact is stiffer, the user perceives the existence of the stiff object applying less force to the device.

Performing stiffness rendering algorithm at the actuator level of an underactuated hand exoskeleton is the simplest approach to provide resistive forces to finger opening/closing based on virtual interactions. However, this approach ignores finger joint activities and cannot estimate the actual perception of user, since the resistive actuator forces are delivered to multiple finger joints of the device. Even though for some simple rehabilitation tasks it could be useful to complete the realism of the tasks, defining the contact limit or the stiffness value for the actuator during the haptic applications would be impractical.

\subsection{Joint Level Stiffness Rendering}

\begin{figure}[htb]
\centering
%\vspace*{-0.7\baselineskip}
\includegraphics[width=0.75\textwidth]{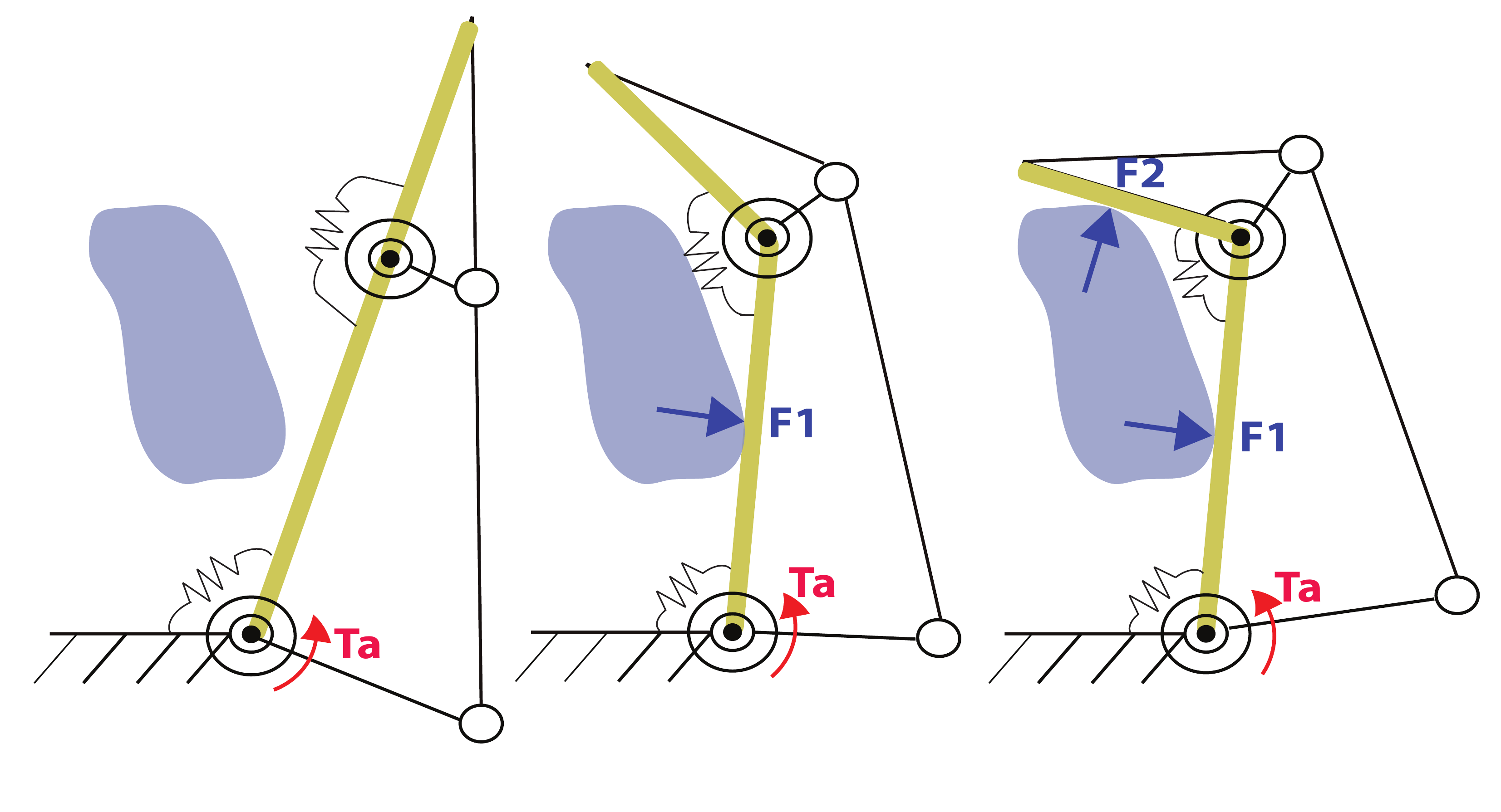}
%\vspace*{-0.7\baselineskip}
\caption{Underactuation concept through a grasping simulation through a $2~DoF$ gripper with a single rotational actuator.}
\label{fig:underactuation}
%\vspace*{-0.7\baselineskip}
\end{figure}

Unlike actuator level stiffness rendering, joint level rendering algorithm calculates the interaction forces based on the activity of each joint independently. For a hand exoskeleton designed to perform virtual grasping tasks in haptic applications, interaction forces should be calculated at finger phalanges, where the interaction occurs. Figure~\ref{fig:underactuation} shows the interaction forces at phalanges during a virtual grasping task while the underactuated hand exoskeleton assists user's fingers. Whenever there is a contact between the virtual object and the user, the actuator should render resistive forces acting on virtual phalanges to allow the perception of touch. Fully controlled exoskeletons impose a kinesthetic feedback by computing the desired torques corresponding to the finger joints, while the proposed underactuated exoskeleton imposes the feedback by computing the desired actuator force to help the user perceive the interaction at both finger phalanges. It is important to emphasize that actuator forces are computed in joint level method based on interaction forces at phalanges, while previous method did not focus on phalange forces at all.

The exoskeleton allows the user to move freely with high transparency until he reaches for the virtual object. When encountered with a virtual object, the interaction forces between the object and the user's phalanges are calculated using stiffness rendering algorithm stated as Equation~(\ref{eq:joint_stiffness}) where $\vect{q}$ and $\vect{q}_d$ represent the instantaneous finger pose and desired pose. The desired pose $\vect{q}_d$ elements are calculated as in Algorithm~\ref{alg:des_joint} based on joint limits for each finger joint.

\begin{algorithm}
\caption{Setting desired joint pose based on virtual limits.}\label{alg:des_joint}
\begin{algorithmic}[1]
\State \textbf{input:} $q$, the joint pose
\State \textbf{input:} $q_{lim}$, the virtual pose limit
\State \textbf{output:} $q_d$, the desired joint pose
\If {$q <= q_{lim}$}
\State {$q_{d} = q$}
\Else
\State {$q_{d} = q_{lim}$}
\EndIf
\end{algorithmic}
\end{algorithm}

\begin{figure}[htb]
\centering
%\vspace*{-0.7\baselineskip}
\includegraphics[width=0.75\textwidth]{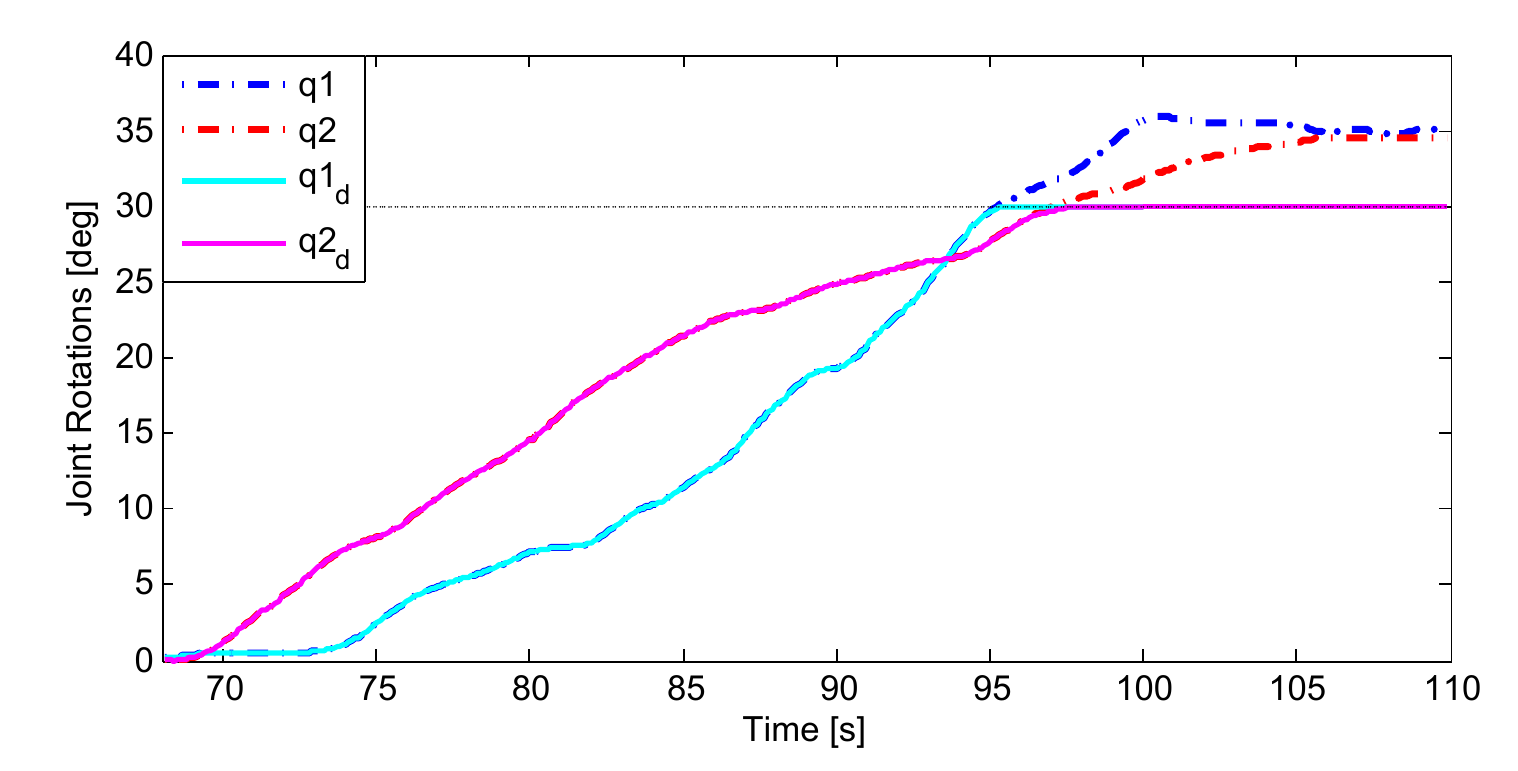}
%\vspace*{-0.7\baselineskip}
\caption{Computing desired pose $\vect{q}_d$ to be used for stiffness rendering algorithm for given actual pose $\vect{q}$ using Algorithm~\ref{alg:des_joint}.}
\label{fig:desiredpose}
%\vspace*{-0.7\baselineskip}
\end{figure}

For a real virtual grasping task, the finger pose limit is determined as the finger pose that the interaction occurs. However, we would like to test the capabilities of the hand exoskeleton before complicating the overall haptic system by introducing a virtual environment. For this reason, we will assume a virtual interaction to occur at a previously defined pose, let's say when both MCP and PIP joints are rotated around $30^o$ each. So. the finger pose limit will be assumed to be $\vect{q}_{lim} = (30, 30)^T$, while the execution of the desired finger pose accordingly can be shown in Figure~\ref{fig:desiredpose}.

Figure~\ref{fig:desiredpose} shows how the user move ($q1$ and $q2$) and how the desired finger pose ($q1_d$ and $q2_d$) change when the user reaches the finger pose limit accordingly. Keep in mind that the desired forces are calculated using the difference between the actual and the desired poses, so the desired forces are zero before the limit. Doing so, any assistance provided by the device until the user reaches the pose limit is prevented, and the device only renders resistive forces once the limit is forces. $\vect{K}_{cont}$ is the diagonal, controllable stiffness matrix, which defines the sensed stiffness regarding the virtual object's properties.

\begin{equation} \label{eq:joint_stiffness}
\vects{\tau} = \vect{K}_{cont}~(\vect{q}_{d} - \vect{q})
\end{equation}

For a fully controlled scenario, the actuator forces can be calculated using Jacobian transpose as  $\vects{F} = \vect{J}^T \vects{\tau}$. Having all components in $\vects{F}$ simply by designing a fully controllable device, the possibility of joint torques $\vects{\tau}$ to be transmitted to finger phalanges can be achieved by implicit joint control. On the other hand, underactuated devices have non-actuated (passive) joints as well as actuated (active) joints to form a square, invertible Jacobian, which can be divided into active and passive components ($\vects{J} = (\vects{J}_a, \vects{J}_p)$). Similar to fully controlled devices, the divided Jacobian can be used to relate actuator forces and desired joint torques using Equation~(\ref{jac_tr_general})

\begin{align}\label{jac_tr_general}
\left[
\begin{array}{r}
\vects{F}_{a} \\
\vects{F}_{p}
\end{array} \right]
&= \left[
\begin{array}{rr}
\vect{J}^T_{a}\\
\vect{J}^T_{p}
\end{array} \right]
\vects{\tau}
\end{align}

\noindent where $\vects{\tau}$ represents the array of end-effector output torques calculated in Equation~(\ref{eq:joint_stiffness}), $\vects{F}_a$ and $\vects{F}_p$ define the arrays for actuated (active) and non-actuated (passive) forces. Passive joints are included for Jacobian formulation since they contribute for motion analysis, even though they do not provide any energy to the system. Such passivity to achieve an underactuation design can easily be transformed into a mathematical constraint in Equation~(\ref{jac_tr_general}) as $\vects{F}_p = \vect{0}$.

As mentioned previously, the underactuated concept performs force transmission based on contact forces simultaneously. In fact, the relationship between joint forces acting by the device to phalanges can be defined simply by detailing the underactuated constraint as:

\begin{equation} \label{eq:constraint}
\vect{0} = \vect{J}^T_{p} \vects{\tau}.
\end{equation}

Since this constraint is implicitly defined by mechanical properties, it cannot be overruled. However, a set of desired forces $\vect{\tau}$ can result the desired passive forces $\vects{F}_p$ to have a non-zero value in order to be transmitted to finger joints, even though mechanically it cannot be performed. The conflict between desired and actual values of passive force $\vects{F}_p$ complicates the stiffness rendering for underactuated devices and need to be solved.

\subsection{Stiffness Rendering for Verification}

\begin{figure}[!t]
\centering
\includegraphics[width=0.8\textwidth]{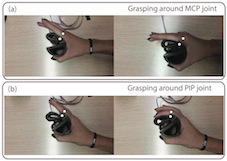}
\caption{Finger motion during the experiment: (a) the user aligns his MCP joint with the mechanical joint of the object to rotate his MCP joint, while his PIP joint remains with minimum movement. (b) the user aligns his PIP joint with the mechanical joint of the object to rotate his PIP joint, while his MCP joint remains with minimum movement. }
\label{fig:GraspingImage}
%\vspace*{-0.5\baselineskip}
\end{figure}

Before getting into the details of underactuation, a simple set of experiments can help us to visualize whether the force transmission through the proposed hand exoskeleton is efficient. In order to do so, a user was asked to perform tasks that require only $1~DoF$ rotation. In fact, the user was asked to rotate only a single joint at a time, while other joint was kept constant, as can be visualized in Figure~\ref{fig:GraspingImage}. It is safe to assume that ensuring the stable joint torque to be constant $0$ throughout the task simplifies the underactuated hand exoskeleton to a simple $1~DoF$ device. The desired torque around active joint should be calculated simply by Equation~(\ref{eq:joint_stiffness}) while Equation~(\ref{jac_tr_general}) shows how to obtain the corresponding desired actuator force $F_a$.

\begin{figure}[htb]
\centering
\includegraphics[width=0.32\textwidth]{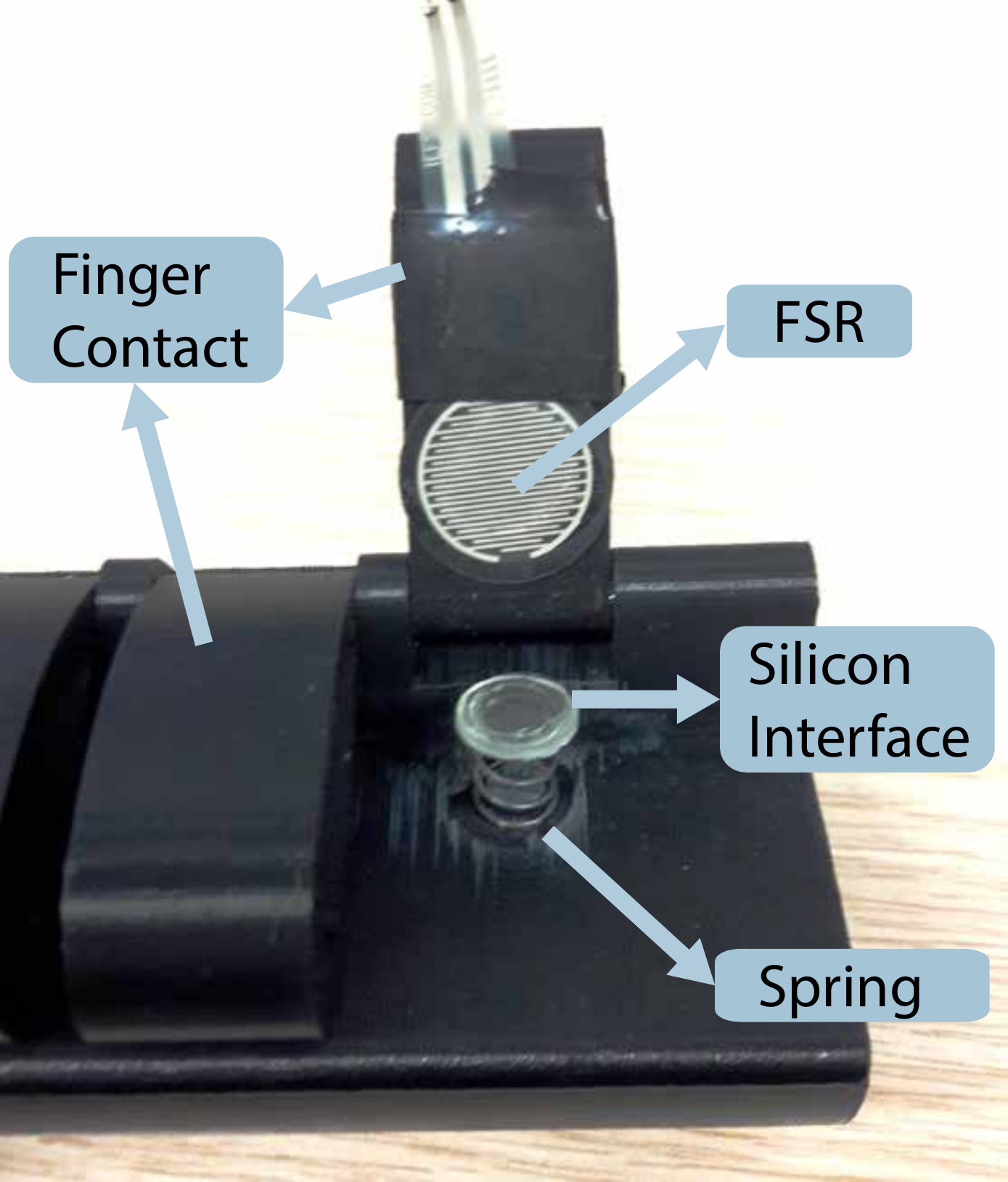}
\caption{FSR Interface to measure forces on the finger phalanges: the finger contact with a mechanical joint allows the user to grasp and squeeze the interface. The FSR sensor measures the pressure applied by the user's finger. The silicon interface aims to imperfection the diversities of how the forces are applied to the FSR sensor. The spring allows the user to squeeze the object, imitating a soft object during haptic rendering task.}
\label{fig:fsr_setup}
%\vspace*{-1\baselineskip}
\end{figure}

Despite of the inconsistency caused by passivity constraint of passive joint, creating a $1~DoF$ movement would provide a more promising approach to begin studying the mechanical behavior with. The feasibility tests on stiffness rendering algorithm and force transmission of the exoskeleton can be shown simply by comparing desired actuator force $F_a$ value based on the active joint torque to external measurements from a customized object and a FSR sensor. In order to minimize the dependency on surface area of the contact with FSR and to obtain comparable, repeatable results during different tasks, a custom object was manufactured as in Figure~\ref{fig:fsr_setup}. The FSR sensor was chosen to provide quick, simple measurements. A spring was attached in front of the FSR to allow user to squeeze the object for force measurements, whose stiffness value is used for the stiffness rendering task $\vect{K}_{cont}$. Finally, a silicon interface was introduced between the spring and the FSR in order to increase repeatability and efficacy of this pressure sensing tool for the setup.

The experiment starts with a calibration process, where the user was asked to align her active joint to mechanical joint and simply rotate her joint until her active phalange reaches the object, as depicted by Figure~\ref{fig:GraspingImage}. Meanwhile, the finger pose of the user was monitored in order to determine the desired joint limits to form $\vect{q}_{d}$ as in Algorithm~\ref{alg:des_joint} for stiffness rendering in Equation~(\ref{eq:stiffness_rend}). The limits for MCP and PIP joints ($q_{{lim}_1}$, $q_{{lim}_2}$) were detected as $15^o$ and $35^o$ by the calibration.

Aligning finger joints with mechanical joints not only increases repeatability of the task, but also allows desired torque values to be converted to forces acting on the contact point, which is the center of the spring shown in Figure~\ref{fig:fsr_setup}. In fact, such conversion can be made easily using the distance between center of FSR to joint of the custom object ($m$) as in $F_{i} = \tau_{i} * m$.

\begin{figure}[!htb]
\centering
%\vspace*{-1.5\baselineskip}
\subfigure[Stiffness rendering task around MCP joint]{\includegraphics[width=0.8\textwidth]{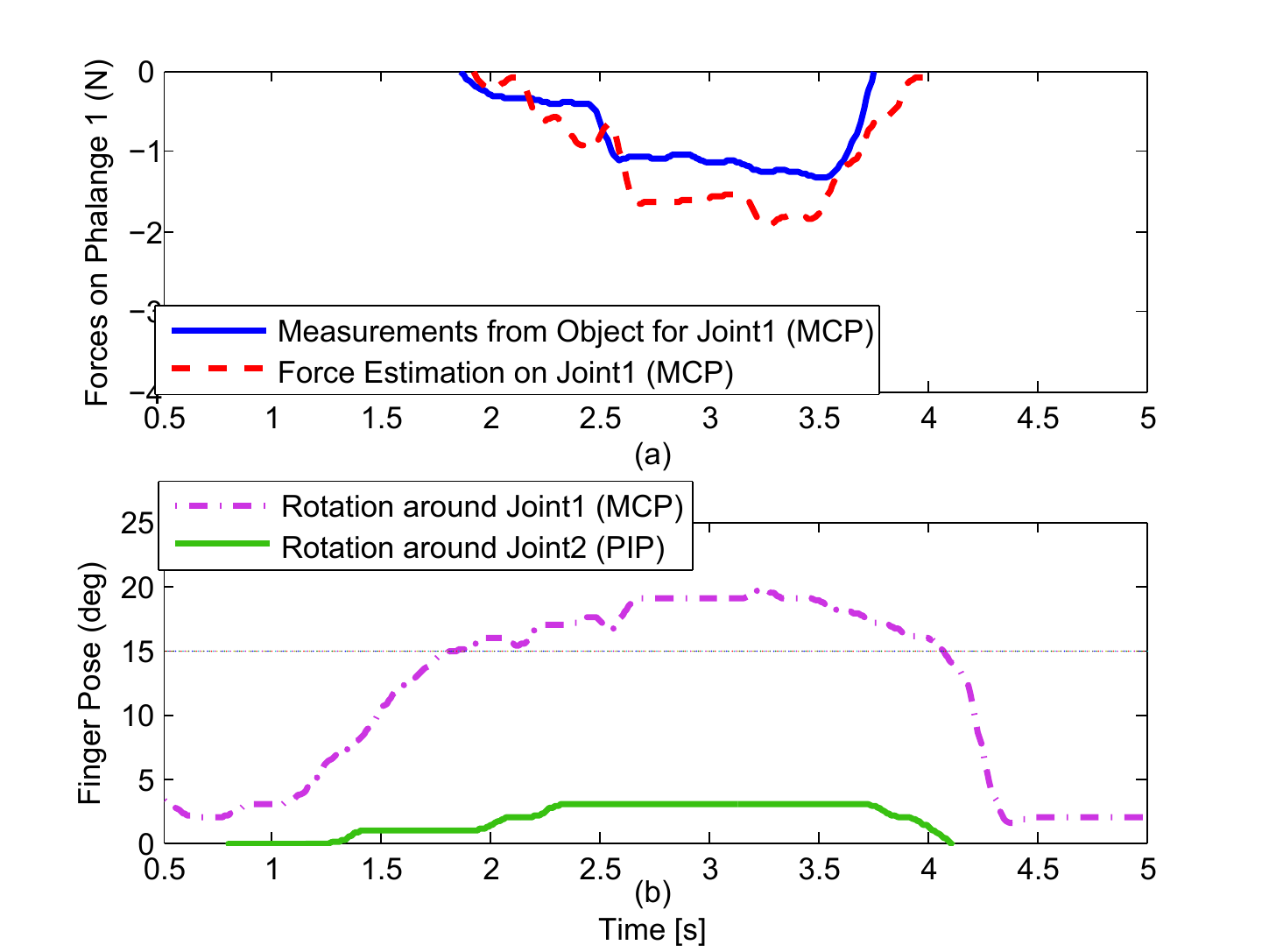}\label{fig:GraspingExp1_NoStiff}}
\subfigure[Stiffness rendering task around PIP joint]{\includegraphics[width=0.8\textwidth]{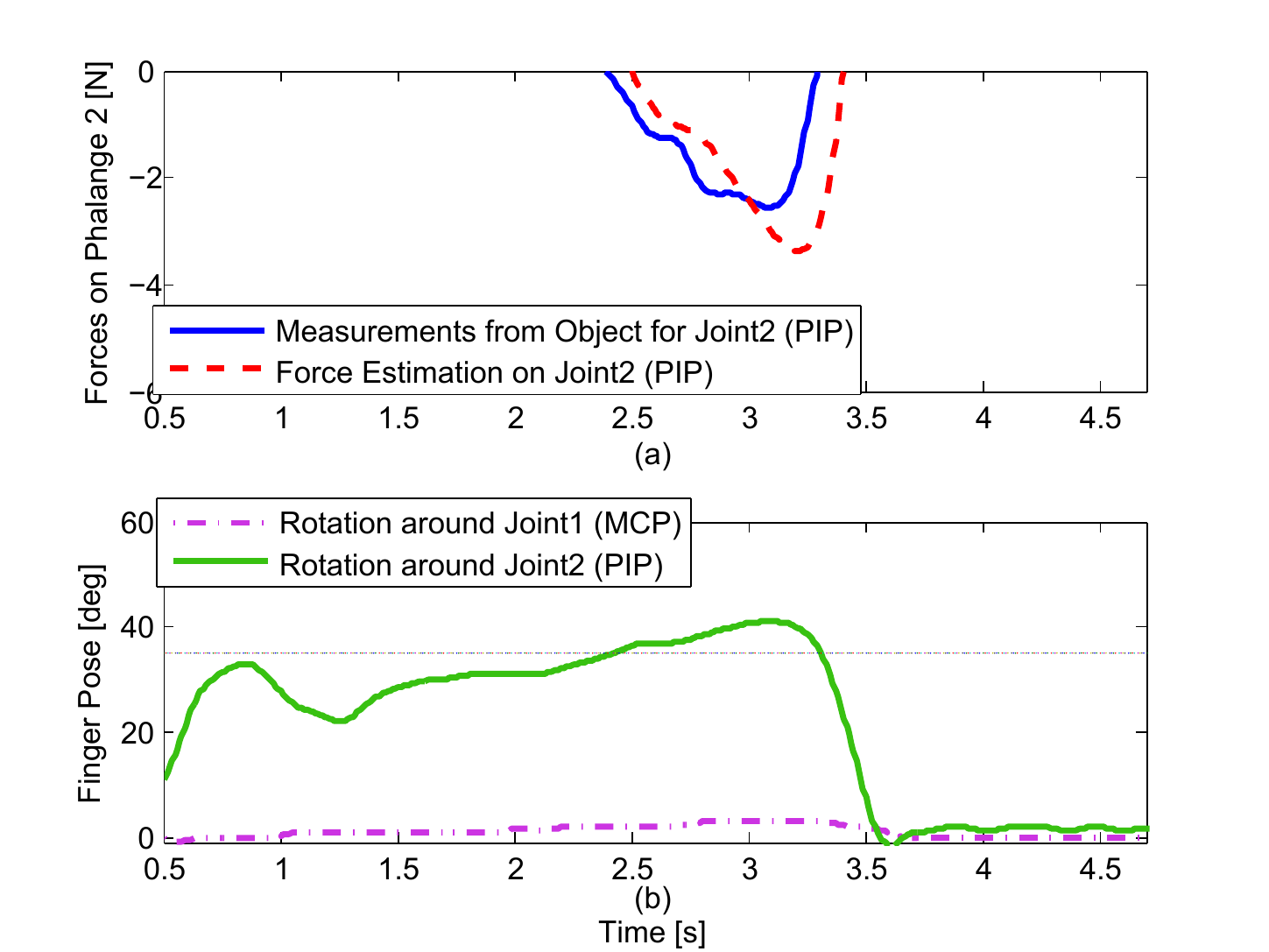}\label{fig:GraspingExp2_NoStiff}}
\caption{Experimental results for force applied by the exoskeleton vs. force measured by FSR and the pose estimation for: (a) stiffness rendering task around the MCP joint; (b) stiffness rendering task around the PIP joint.}	
\label{fig:GraspingExp_NoStiff}
%\vspace*{-0.5\baselineskip}
\end{figure}

\begin{figure}[!htb]
\centering
%\vspace*{-1.5\baselineskip}
\subfigure[Actual impedance for joint MCP]{\includegraphics[width=0.8\textwidth]{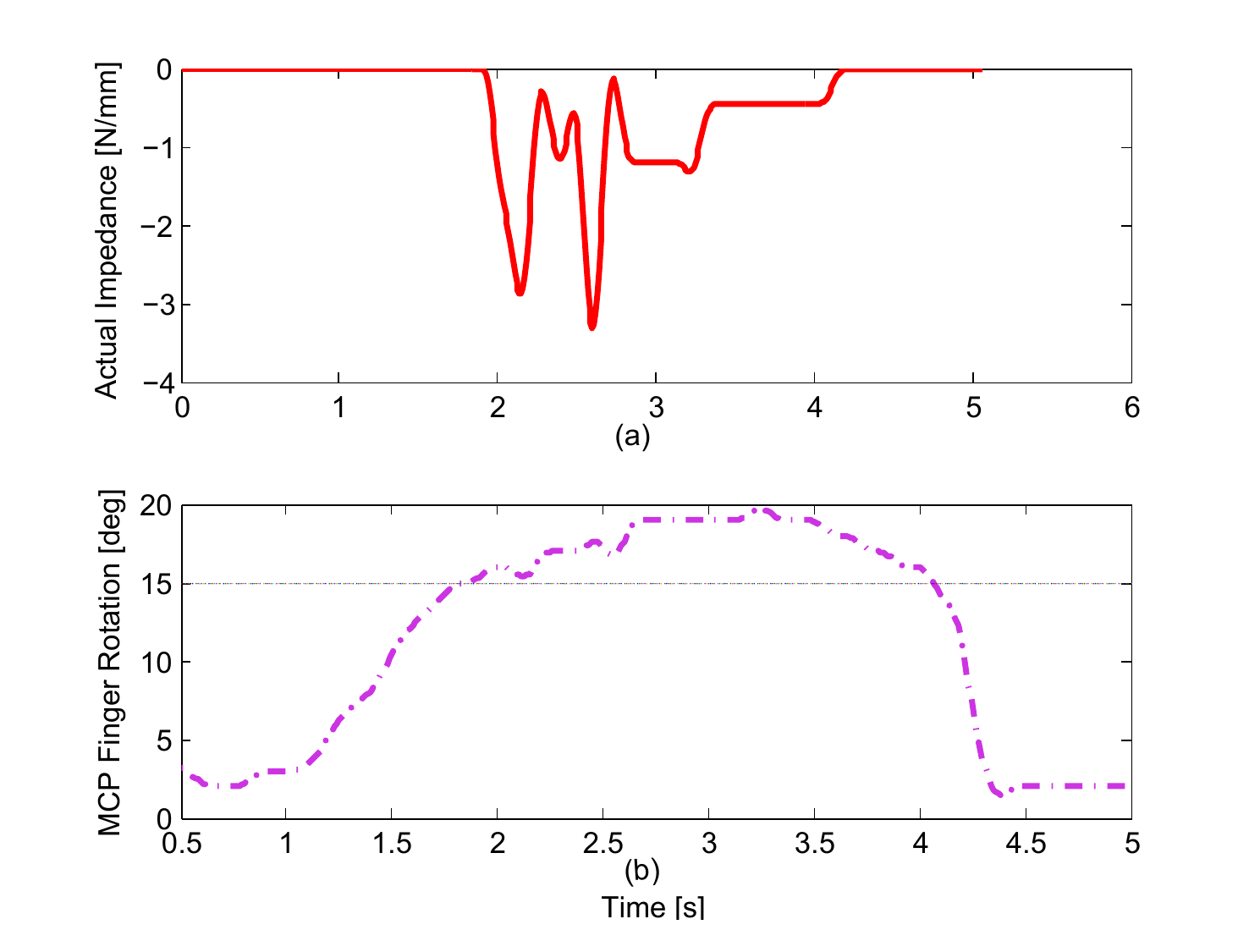}\label{fig:imp1}}
\subfigure[Actual impedance for joint PIP]{\includegraphics[width=0.8\textwidth]{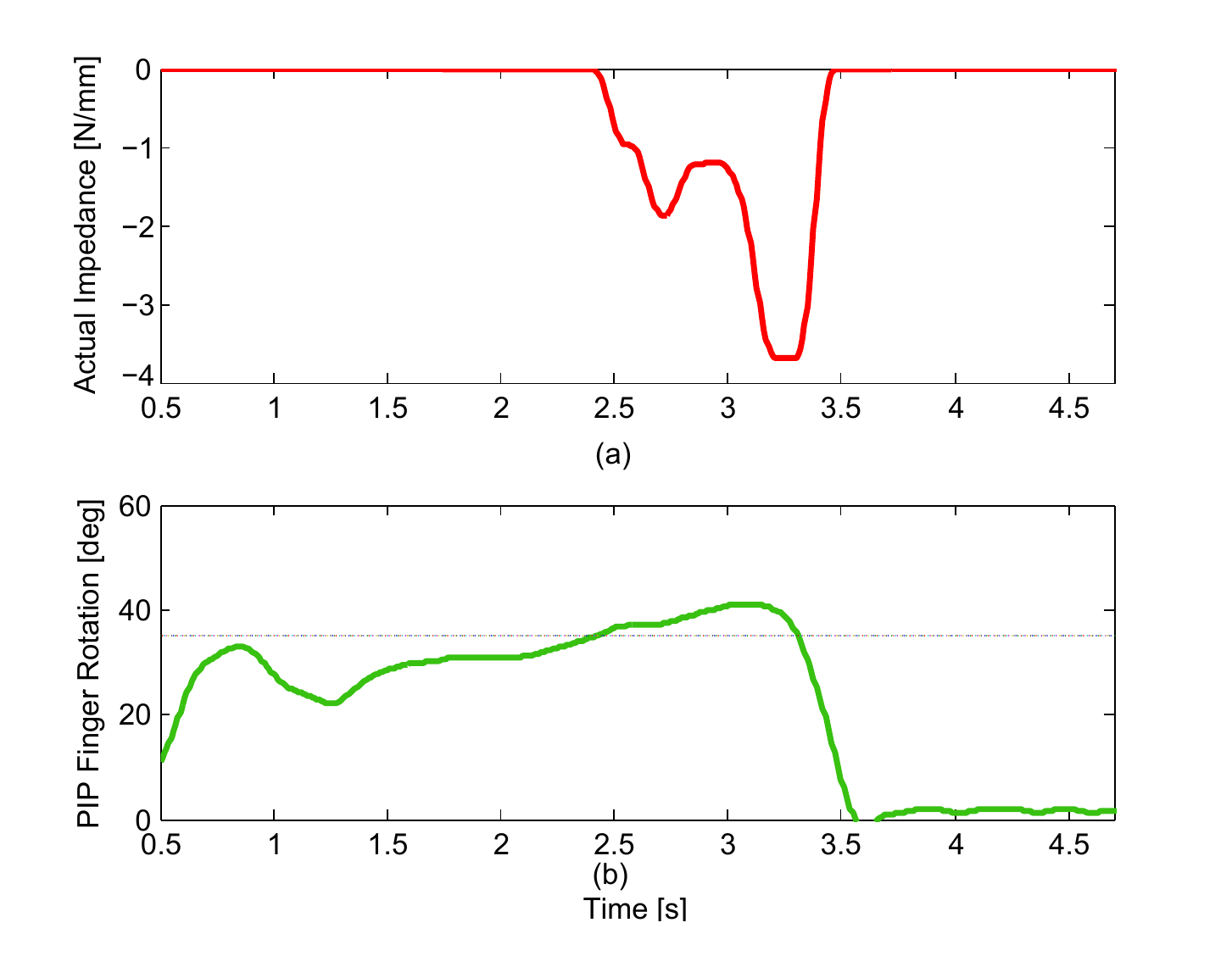}\label{fig:imp2}}
\caption{Sensed impedance calculations in order to study the stability: (a-a) actual impedance value over time for experiment Figure~\ref{fig:GraspingImage} (a), (a-b) rotation around active joint MCP and its virtual limit, (b-a) actual impedance value over time for experiment Figure~\ref{fig:GraspingImage} (b), (b-b) rotation around active joint PIP and its virtual limit.}	
\label{fig:GraspingExp_imp}
%\vspace*{-0.5\baselineskip}
\end{figure}

Meanwhile, stability of the hand exoskeleton can be analyzed simply through displayed impedance, $\Delta F_a / \Delta q$. Since the resistive forces cause the desired impedance to be negative, the negative definition of the impedance values in Figure~\ref{fig:GraspingExp_imp} signify the stability. Having a single active finger joint at a time allows the sensed impedance factor to be considered only for the movable finger joint.

Figure~\ref{fig:GraspingExp_NoStiff} shows the comparison between the FSR measurements and the desired contact forces at the center of spring for rotating MCP and PIP joints individually. The root mean square error between the actual and desired contact forces were calculated to be $0.623$ and $1.74$ for MCP and PIP joints respectively, which can be acceptable by some justifications. First of all, the passive joint is assumed to have $0$ torques around it, however the given setup in Figure~\ref{fig:fsr_setup} does not collect any measurement from the passive finger joint. Furthermore, the FSR sensor was not designed to be suitable for accurate sensing. Nevertheless, observing a similar behavior by the measured and the desired contact forces show sufficient evidence to develop future strategies to generalize the stiffness rendering task with no pre condition about the task or the object. Then, ensuring stability at all times during the experiments as shown in Figure~\ref{fig:GraspingExp_imp} proves the usability of proposed stiffness rendering algorithm for the proposed underactuated hand exoskeleton in a general task.

\newpage
\section{Stiffness Rendering for the Proposed Underactuated Hand Exoskeleton} \label{sec:force_optimization}

%The underactuation concept can be summarized easily as having more mobility than the number of actuation in order to achieve the automatic adjustability for the device based on the contact forces acting to the overall system externally. This extra mobility prevents the independent control of each joint, but allows them to be controlled in a coupled manner. In particular, the underactuation property also prevents the estimation of these degrees of freedoms due to the lack of measurements and the invertibility of the corresponding Jacobian matrix. The issues regarding the measurability can easily be solved by utilizing additional sensors along the mechanical system. However, the lack of controllability remains the same.

Using Jacobian matrix as it is to calculate desired forces corresponding to desired joint torques calculated in Equation~(\ref{eq:joint_stiffness}) creates uncertainties for passive joint forces. The lack of controllability of the passive joint creates uncertainties when a set of desired joint torques require the passive joint to put energy to the system even though its passivity introduces an underactuated constraint as $\vects{F}_p = \vect{0}$.

Such a constraint introduced by underactuation concept prevents all possible sets of finger torques $\vects{\tau}$ obtained by Equation~(\ref{eq:joint_stiffness}) and transmits another set of torques instead of the desired set. It is acceptable not to achieve all possible sets with underactuation due to other advantages of such a lack of controllability, but not knowing the transmitted forces is not very appealing, especially for a device with human interaction. Instead of using the desired set of joint torques and hope for the device transmission to be performed in the best manner, a proxy set that satisfies the underactuation constraint can be found with minimum distance to the desired set. Doing so, even though the original desired set of finger torques cannot be transmitted to finger joints, an optimized set will be ensured to be applied to user through mechanism, solving the issue of uncertainty.

The stiffness rendering algorithm can be improved by finding a proxy set of desired torques or a proxy set of desired pose satisfying this property, such that the underactuated device can turn the desired values to the actual values through force transmission. Implementing such an extra effort based on desired torques might seem more straightforward, since the constraint itself is defined in the force space. However, stiffness rendering tasks are designed on the motion space, simply by forming resistance forces based on user's interactions in the virtual environment. Therefore, another rendering strategy should be defined to merge the same underactuated constrain with joint rotations. All these rendering strategies will be detailed in this section as much as experimental results performed by the underactuated hand exoskeleton. The feasibility of the force transmission will be shown by comparing the theoretical desired force values to the real measurements collected by simple FSR sensors. Then, the stability of the controller will be discussed based on the actual impedance calculations.

Each strategy will be detailed and given theoretical examples to highlight the idea behind it. Then, it is implemented on the hand exoskeleton control algorithm to compare the actual and desired torques acting on the finger joints. To do so, a FSR sensor has been inserted between each finger phalange and passive slider that provides the connection between the user. Furthermore, the stability of the device is investigated by calculating the actual impedance. Finally both strategies are compared to each other in terms of defining the motivations why to implement them.

\subsection{Rendering Strategy based on Joint Torques} \label{sec:opt_torque}

Instead of suffering from such an incompatibility between the desired and actual torques around the finger joints, the constraint imposed by the passivity of sensory joint ($F_p = 0$) can be used to create an applicable ratio between the joint torque values $\tau_{1}$ and $\tau_{2}$. As discussed in Chapter~\ref{sec:Chapter4}, the passive joint is defined as joint $B$, where the additional potentiometer was placed to have an additional sensory measurement for pose estimation. The ratio between the applicable torques can be used to achieve a proxy set of desired torques that the underactuated exoskeleton can actually transmit to the finger joints. Keeping the distance between the proxy set and the desires set of joint torques might provide an efficient stiffness rendering perception while allowing the transmitted torques to be known during operation.

With this motivation, the passivity constraint $F_p = 0$ can be used to achieve a relation, which bounds the desired torques acting on set of MCP and PIP joints $\vects{\tau}$, as in Equation~(\ref{eq:constraint}). In fact, Equation~(\ref{eq:constraint}) defines a relation between torque values around the finger joints for a given orientation. In particular, for the proposed exoskeleton with $2~DoF$ and a single actuator provides a ratio between joint torques at a given instant by the underactuated exoskeleton and the exoskeleton will fail to transmit a set of desired torques if they do not satisfy this ratio.

When a set of desired joint torques ($\vects{\tau}$) is calculated from the stiffness rendering in Equation~(\ref{eq:joint_stiffness}) that do not satisfy the constraint in Equation~(\ref{eq:constraint}), a proxy set of desired finger torques ($\vects{\tau}^* = (\tau^*_{1},~\tau^*_{2})^T$), which is ensured to satisfy the passivity constraint, can be calculated with minimum distance to the desired set $\vects{\tau}$. Instead of $\vects{\tau}$, the proxy set $\vects{\tau}^*$ will be used to generate the force command for control algorithm in Equation~(\ref{eq:stiffness_rend}) to execute the stiffness rendering task. Satisfying the given constraint in Equation~(\ref{eq:constraint}) ensures the underactuated device to transmit the proxy set of torques for user's joints in real life. Even though the finger joints would not achieve the exact force feedback calculated by the stiffness rendering, being able to monitor and control the actual torque values that are being transmitted by the exoskeleton will be ensured.

The stiffness rendering algorithm by finding the proxy set of applicable finger torques $\vects{\tau}^*$ with a minimum error from the set of desired torque calculated by rendering algorithm ($\vects{\tau}$) can be expressed mathematically as in Equation~(\ref{eq:force_opt_math}).

\begin{align} \label{eq:force_opt_math}
F^*_a &= \vect{J}^T_a~\vects{\tau}^*,~\textrm{with}\\
\vects{\tau}^* &= argmin~\frac{1}{2}~\left\| \vect{\tau} - \vect{\tau}^* \right\|^2,~ \textrm{s.t.} ~\vect{J}^T_{p} \vects{\tau}^* = 0, \nonumber
\end{align}

The minimization problem with the underactuated constraint can also be expressed using the pseudo-inverse as in Equation~(\ref{eq:force_opt_math_cl}).

\begin{align} \label{eq:force_opt_math_cl}
\vects{\tau}^* = (\vect{I} - \vect{J}_{p}~(\vect{J}_{p}^T~\vect{J}_{p})^{-1}~\vect{J}_{p}^T)~\vects{\tau}
\end{align}

Using the optimized forces acting on phalanges as in Equation~(\ref{eq:force_opt_math_cl}), the actuator forces can easily be obtained as:

\begin{align} \label{eq:force_opt_math_cl_act}
F^*_a = \vect{J}^T_a~(\vect{I} - \vect{J}_{p}~(\vect{J}_{p}^T~\vect{J}_{p})^{-1}~\vect{J}_{p}^T)~\vect{K}_{cont}~\Delta\vect{q}
\end{align}

\noindent where $\Delta\vect{q} = (\vect{q}_d - \vect{q})$. The sensed impedance of the given strategy $F_a / \Delta\vect{q}$ can be calculated simply as:

\begin{align} \label{eq:force_imped}
\vect{S}_a = - \vect{J}^T_a~(\vect{I} - \vect{J}_p~(\vect{J}^T_p~\vect{J}_p)^{-1}~\vect{J}^T_p)~\vect{K}_{cont}.
\end{align}

\begin{figure}[htb]
\centering
% \vspace*{-2\baselineskip}
\includegraphics[width=0.5\textwidth]{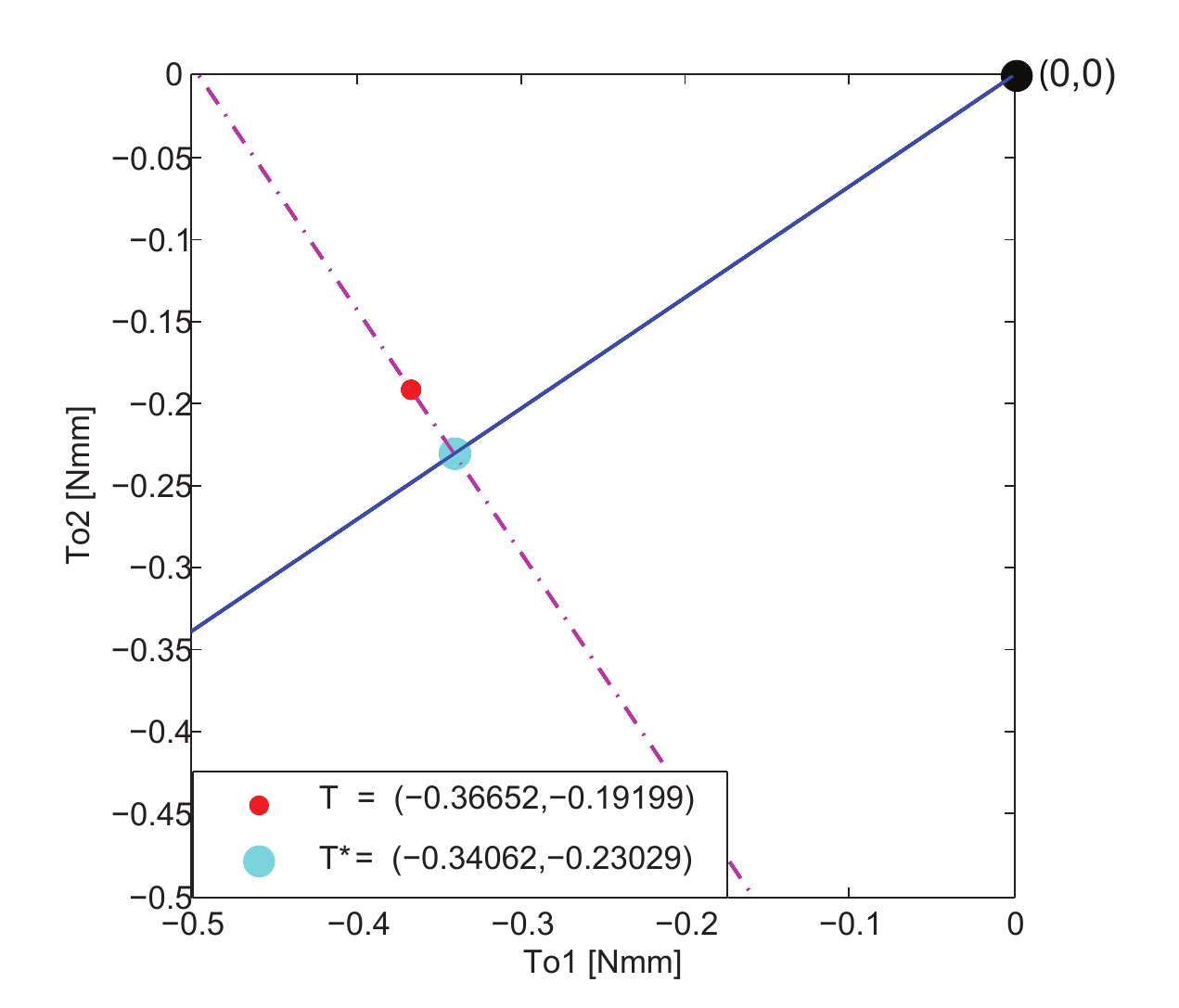}
\caption{Simulation plot for the desired torque optimization based on minimizing the error between the real and the proxy torque sets.}
\label{fig:opt_tor}
%\vspace*{-1\baselineskip}
\end{figure}

The idea behind the given strategy can be shown with a simple simulation setup before the real time implementation. The finger joints were simulated to perform a closing movement for this simulation, while Jacobian matrix ($\vects{J}$) , actuator displacements ($l_x$), finger pose estimation ($\vect{q}$) and rotation around the passive joint ($q_B$) were recorded. Through these recordings, a random finger pose was selected as $\vect{q} = (51^o, ~51^o)^T$ while a desired pose was assumed to be $\vect{q}_d= (30^o,~40^o)^T$. The stiffness matrix around the finger joints based on the virtual object was set as $\vect{K}_{cont} = (1~0;~0~1)$. Under the given assumptions, the desired torque for the given pose was calculated as $\vects{\tau} = (-0.36652,~-0.19199)^T$ using Equation~(\ref{eq:joint_stiffness}). Figure~\ref{fig:opt_tor} provides the scheme of the given example in terms of the desired torques and how to obtain an optimized set of desired torques that the actuator can actually provide.

The proposed exoskeleton was designed with $2~DoF$ mobility with a single actuator, therefore expressing the optimization process with a $2-D$ sketch gives a clear idea of the strategy. The same design property forms a ratio between joint torques ($\tau_2~/~\tau_1$) using the underactuated constraint defined previously in Equation~(\ref{eq:constraint}). This ratio was shown in Figure~\ref{fig:opt_tor} as the blue constant line, passing through the origin to represent the direction of applicable torques to finger joints for the given instant based on the actuator force values. For the chosen finger pose, the Jacobian matrix caused this ratio as $0.6761$, as the slope of the blue line. The red point in Figure~\ref{fig:opt_tor} shows the set of desired torques $\vects{\tau}$ calculated using stiffness rendering algorithm in Equation~(\ref{eq:joint_stiffness}). Since $\vects{\tau}$ does not lay on the blue line, which represents the applicable sets of joint torques, a proxy set of torques $\vects{\tau}^*$ needs to be calculated as shown in Equation~(\ref{eq:force_opt_math_cl}), which was found as $\vects{\tau}^* = (-0.34062,~-0.23029)^T$.

The fact that proxy set of torques $\vects{\tau}^*$ lay on the blue line unlike desired set of torques $\vects{\tau}$ promises these torque values to be transmitted to finger joints. The verification of minimum distance between $\vects{\tau}^*$ and $\vects{\tau}$ was shown simply by drawing a secondary direction line, which was shown as pink dashed line, perpendicular to the blue line. Since both points are connected by the perpendicular line, the minimization issue can be proven. The calculated proxy set of desired torques are mapped to the actuator forces using the Jacobian transpose as $\vect{F^*_{a}} = \vect{J}_a^T * \vects{\tau}^*$.

\begin{figure}[!htb]
\centering
%\vspace*{-1.5\baselineskip}
\subfigure[$Stiff Object$]{\includegraphics[width=0.48\textwidth]{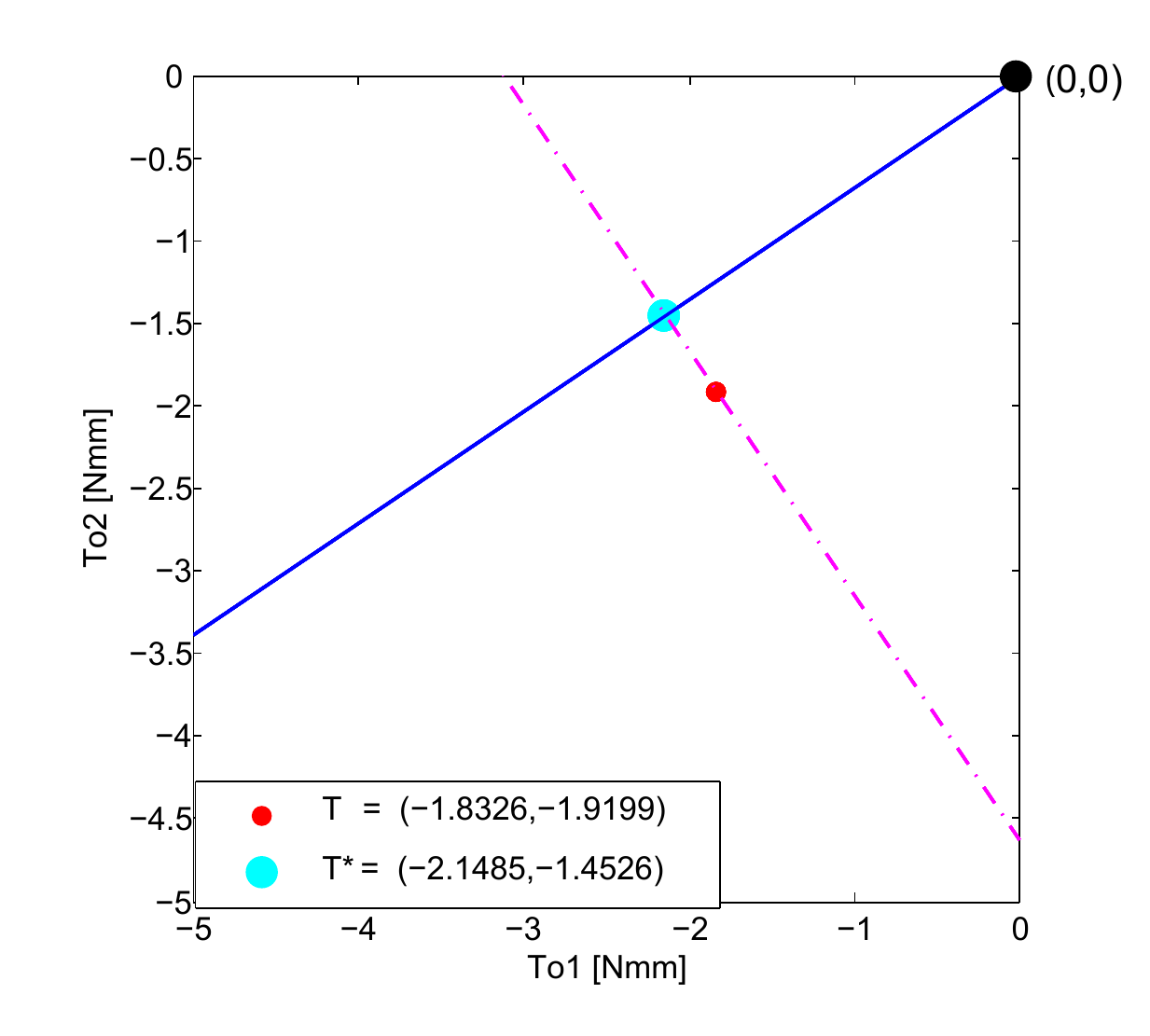}\label{fig:opt_tor_st1}} \subfigure[$Soft Object$]{\includegraphics[width=0.48\textwidth]{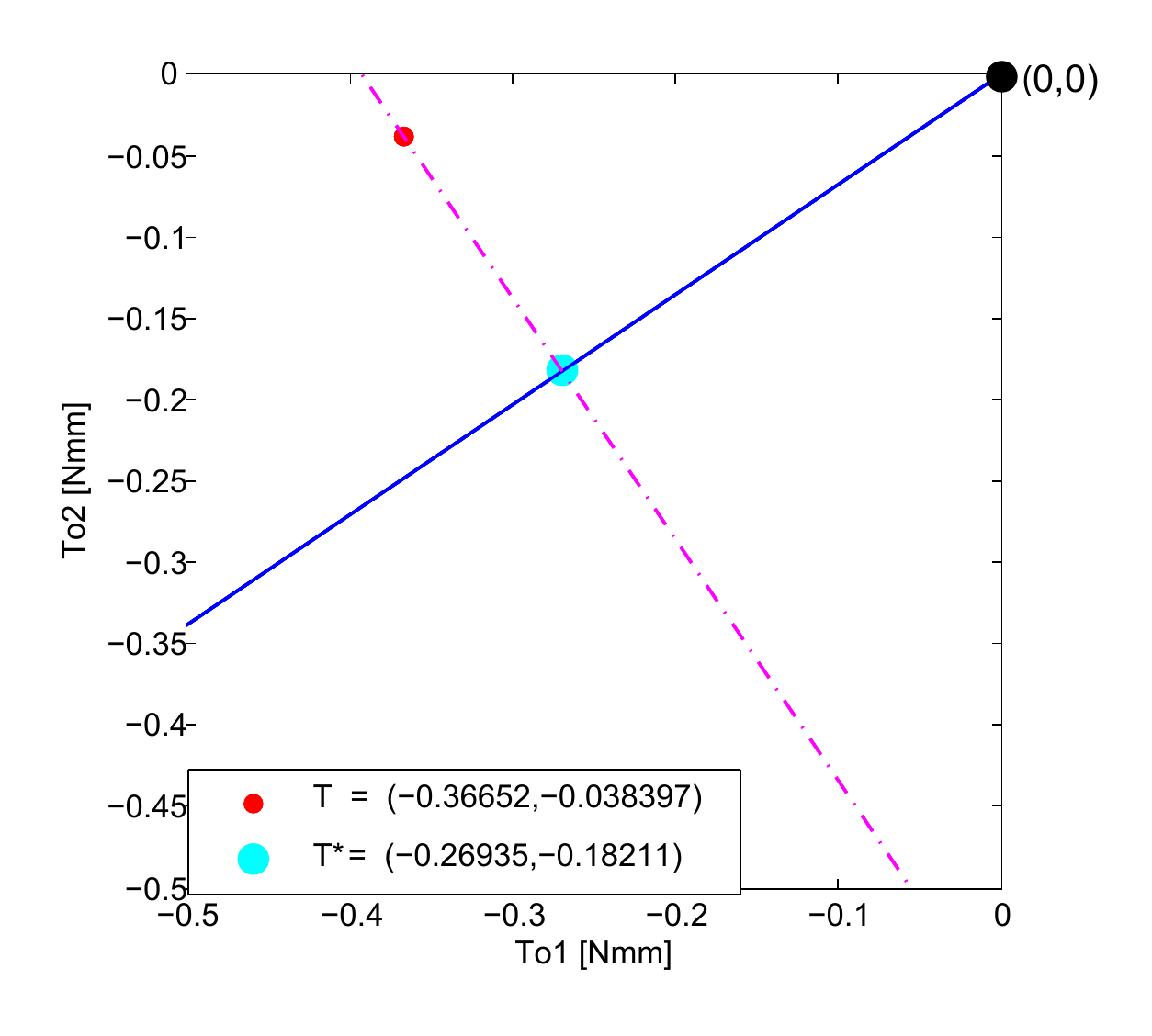}\label{fig:opt_tor_st2}}
\caption{Desired torque optimization for different stiffness assumptions: (a) $K1_{cont} = [5~0;~0~10]$; (b) $K2_{cont} = [1~0;~0~0.2]$.}	
\label{fig:opt_tor_st}
\end{figure}

The impact of stiffness matrix and stiffness ratio between joints can easily be observed by running the same simulation again. Two random stiffness matrices were chosen with different ratios as $\vect{K1}_{cont} = [5~0;~0~10]$ (Figure~\ref{fig:opt_tor_st1}) and $\vect{K2}_{cont} = [1~0;~0~0.2]$ (Figure~\ref{fig:opt_tor_st2}). Changing the stiffness values change the desired torque $\vects{\tau}$ with respect to the blue line depicted by Equation~(\ref{eq:constraint}), which stays constant despite of the stiffness change. Naturally, the proxy sets of desired torques $\vects{\tau}^*$ are found to be different for each experiment, even though the idea remains the same as in Figure~\ref{fig:opt_tor}.

Once the proposed rendering strategy was developed for the simulation, it was implemented on the hand exoskeleton for a real time experiment. For a feasibility experiment, the user was asked to reach an imaginary virtual object with joint limitations randomly selected as $\vect{q}_{lim} = (30^o,~30^o)^T$ and gets in contact with the object using both finger joints. The stiffness of the virtual object was selected in a soft behavior such that the user was allowed to exceed the joint limits after the first contact to complete the second joint contact as well. The desired pose $\vect{q}_{d}$ was computed during operation simultaneously based on Algorithm~\ref{alg:des_joint}. Using the instantaneous pose estimation, the desired torques were calculated using Equation~(\ref{eq:joint_stiffness}), where the stiffness matrix was selected intuitively as $\vects{K}_{cont} = [100~00;~0~100]$.

\begin{figure}[!htb]
\centering
%\vspace*{2.5\baselineskip}
\includegraphics[width=0.9\textwidth]{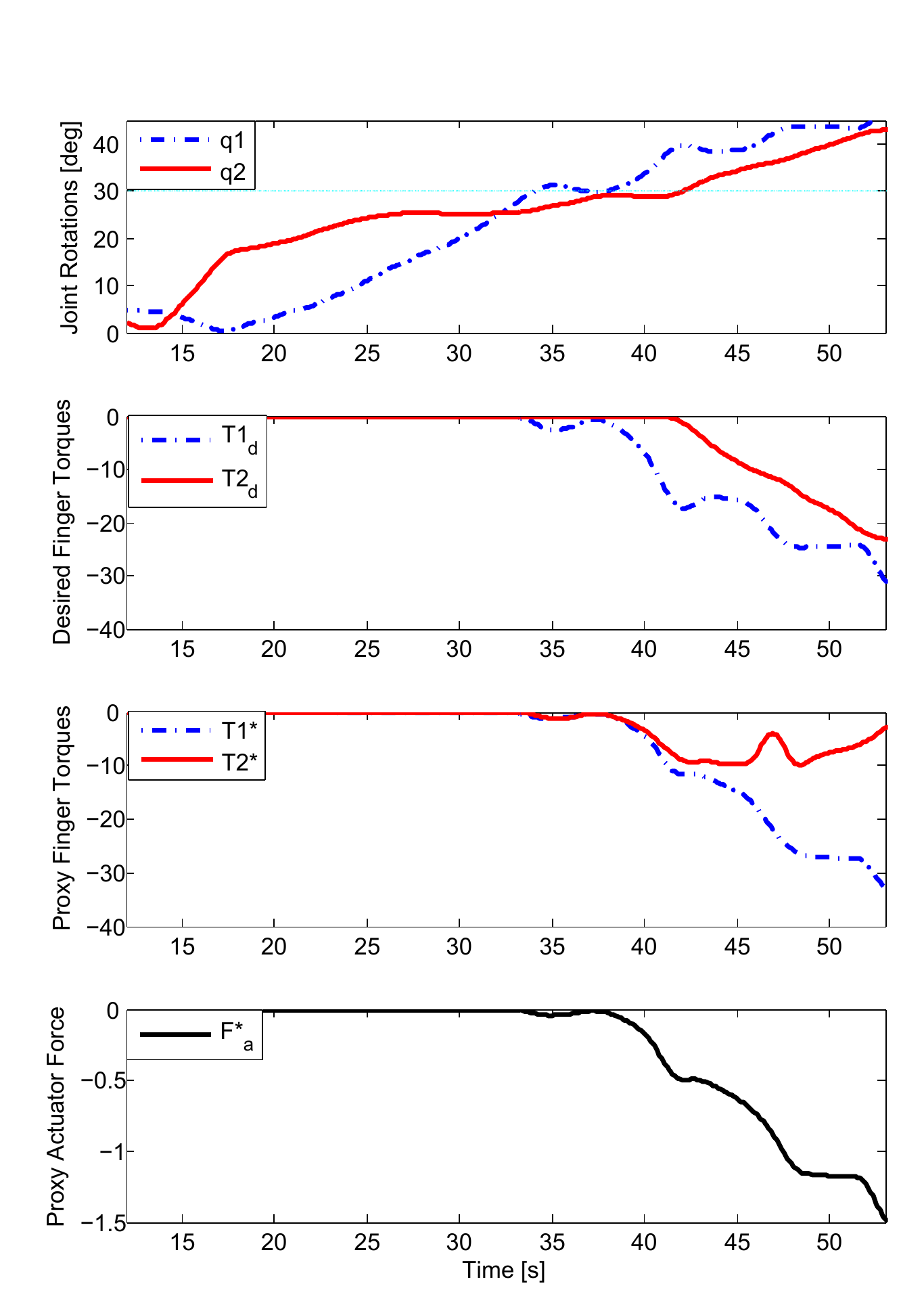}
\caption{Desired torque optimization based on minimizing the error between the real and the proxy torque sets: (a) trajectory of the real time task performed by the user $\vects{q}$ with the virtual limits $\vects{q}_{lim}$ defined for the joints, (b) desired torque values $\vects{\tau}$ calculated for user's trajectory, (c) optimized proxy set of torque values $\vects{\tau}^*$ calculated for user's trajectory and (d) corresponding actuator force for the proxy torque values $F^*_a$.}
\label{fig:opt_tor_real}
%\vspace*{0.5\baselineskip}
\end{figure}
\clearpage

Figure~\ref{fig:opt_tor_st} (a) shows the movement of finger joints within the defined limits regarding the virtual contact with the grasping object. The corresponding desired torques $\vects{\tau}$ are shown in Figure~\ref{fig:opt_tor_st} (b). Meanwhile, the algorithm depicted in Equation~(\ref{eq:force_opt_math}) was being performed to observe the proxy set of desired torques $\vects{\tau}^*$ that can be reached by the hand exoskeleton transmission, which can be observed in Figure~\ref{fig:opt_tor_st} (c). Once the proxy set of desired torques are obtained, the desired forces for the actuator $F_a$ can be calculated through Jacobian transpose through $F_a = \vects{J}_a^T~\vects{\tau}^*$ to be used for the control algorithm presented in Figure~\ref{fig:force_control}.

\begin{figure}[htb]
\centering
% \vspace*{-2\baselineskip}
\includegraphics[width=0.65\textwidth]{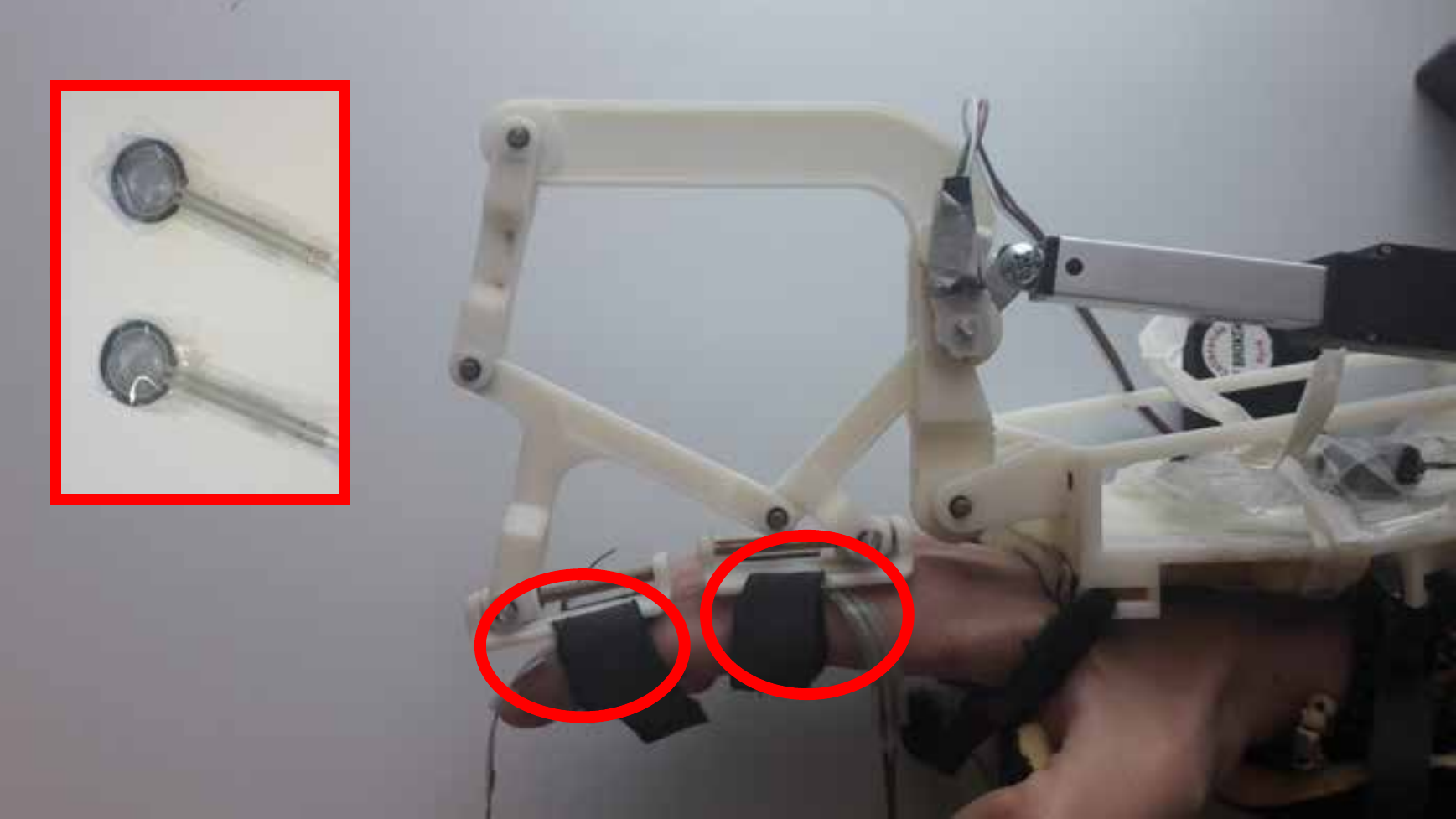}
\caption{How additional FSR sensors are inserted between the hand exoskeleton and the user.}
\label{fig:opt_exos_fsr}
%\vspace*{-1\baselineskip}
\end{figure}

%\newpage
The efficacy of force transmission through proposed underactuated hand exoskeleton was measured by inserting FSR sensors between user's phalanges and passive sliders of the exoskeleton, as shown in Figure~\ref{fig:opt_exos_fsr}. As stated before, a silicon interface was inserted between FSRs and device's sliders to decrease the sensibility of sensors with respect to the form of physical interaction. Even though increasing the distance between user's finger and mechanism would change the kinematics and Jacobian analyses, FSR sensors provide a rough understanding of force transmission compared to desired ones. In order to make a comparison based on finger torques, FSR measurements were converted to joint torques simply as $\tau_{FSR} = F_{FSR} / c_i$, where $c_i$ represents the linear displacement of passive linear slider attached to the corresponding finger phalange.

\begin{figure}[htb]
\centering
% \vspace*{-2\baselineskip}
\includegraphics[width=0.8\textwidth]{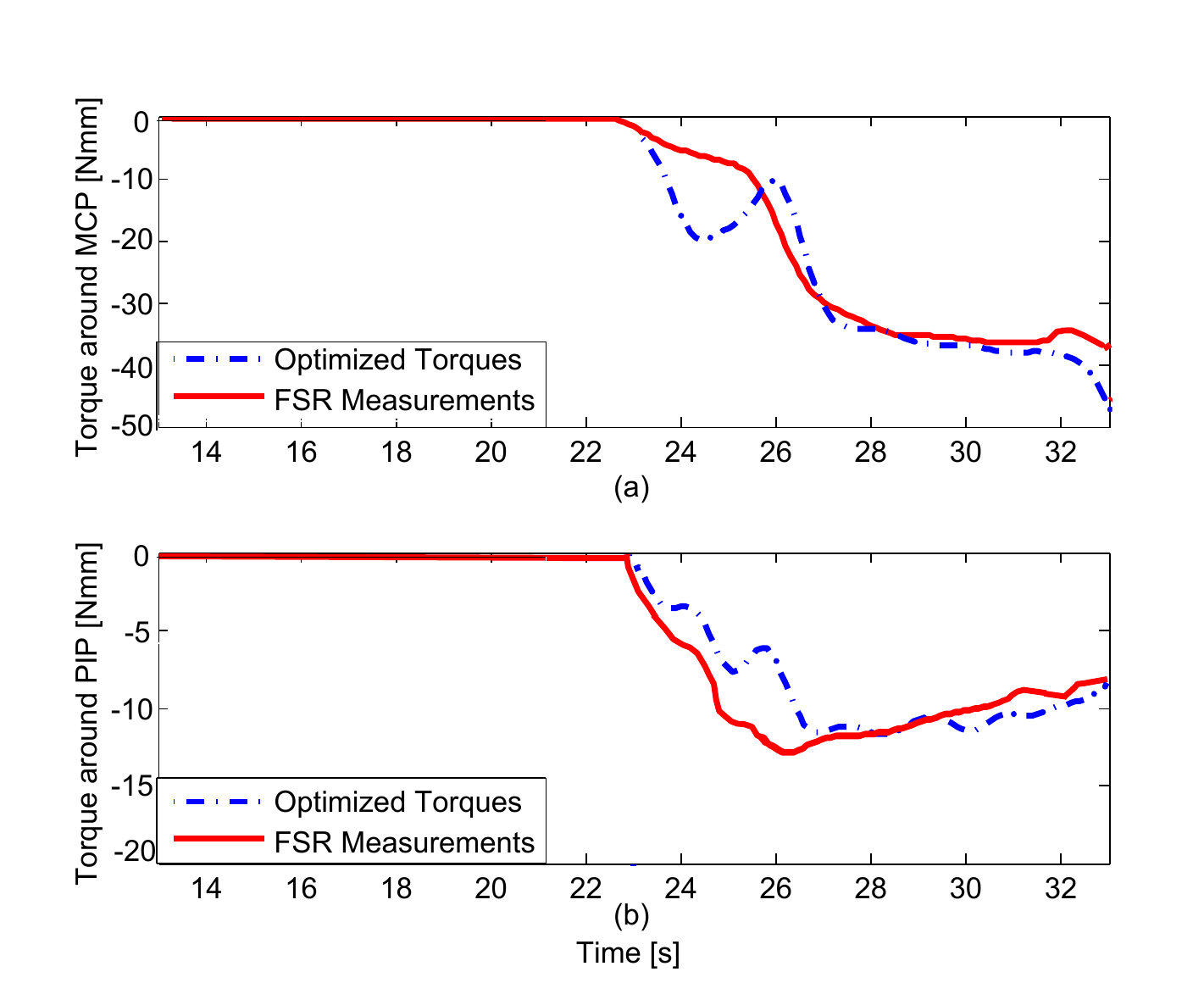}
\caption{Comparison between the calculated proxy set of desired torques and measured torques through FSR measurements.}
\label{fig:opt_tor_fsr}
%\vspace*{-1\baselineskip}
\end{figure}

Figure~\ref{fig:opt_tor_fsr} show sufficient evidence that optimized proxy set of desired torques were applied to user's joints through the device transmission, even though lack of accurate measurements of FSR sensors might create some imperfections during operation.

Comparing actual and desired joint torques as in Figure~\ref{fig:opt_tor_fsr}, one can say that obtaining proxy desired torques can overcome the uncertainties of underactuation concept. Even though proxy set of joint torques cannot assist user's fingers to reach the desired finger pose $\vect{q}_d$, torques being applied to finger joints can be tracked and controlled with this approach for the sake of user's safety and controllability of the device.

\newpage
\subsection{Rendering Strategy based on Finger Pose}
\label{sec:disp_optimization}

Even though optimizing force transmission based on the underactuation constraints by finding a proxy set of applicable torque values, it might be not very efficient in terms of haptics, since the finger pose is not controlled or tracked by the previous strategy. Alternatively, a rendering strategy can be developed to find a proxy set of finger pose $\vects{q}^*$ to replace the desired pose $\vects{q}_d$, which satisfies the underactuated constraint for the instantaneous finger pose $\vects{q}$.

The underactuated grippers were designed simply by introducing more soft or elastic passive joints to perform the force transmission along joints. A similar idea was adopted to create the proposed underactuated hand exoskeleton based on the spring like behavior of user's finger joints anatomically. In particular, anatomical behavior of user's finger joint rotations are modified based on instantaneous stiffness behavior based on user's intention. Using the joint stiffness estimation based on the previous behavior of finger joints, future finger pose might be predicted that will be covered by the finger, assuming the user will continue to move at the same mode. Such a prediction regarding future reachable path by user allows a proxy set of reachable finger pose $\vect{q}^*$ with minimum distance to desired set of joints $\vect{q}_d$ to be found. In practice, running a control board with $1~kHz$ allows these stiffness estimations to be made as fast as the user changes the mode to move. Each stiffness joint value $K_j$ can be calculated as

\begin{align} \label{eq:disp_stiff}
K_j = \frac{\tau_j} {\dot{q}_j}~ \textrm{s.t.}~j = 1,2
\end{align}

\noindent where $K_j$ values are gathered diagonal to form the experimental stiffness matrix $\vect{K}_{stiff}$. For an estimated stiffness matrix of joint stiffness $\vect{K}_{stiff}$, set of joint torques can be calculated using Equation~(\ref{eq:torque_calculation}).

\begin{equation} \label{eq:torque_calculation}
\vects{\tau} = \vect{K}_{stiff}~\Delta\vect{q}
\end{equation}

\noindent where $\Delta\vect{q}$ is the difference between a reachable pose and the initial configuration of finger pose ($\partial \vect{q} = \vect{q}^* - \vect{q}_o$) and is the estimation matrix values. The output forces should calculate to reach actuator forces $F_a$ and passive forces $F_p$ through Equation~(\ref{jac_tr}). Then, the minimization problem is defined as

\begin{align} \label{eq:disp_opt_math}
F^*_a &= \vect{J}^T_a~\vect{K}_{stiff}~(\vect{q}^* - \vect{q}_o),~\textrm{with}\\
\vect{q}^* &= argmin~\frac{1}{2}~\left\| \vect{q}_d - \vect{q}^* \right\|^2,~ \textrm{s.t.} \nonumber \\
& \vect{J}_{p}^T~\vect{K}_{stiff}~(\vect{q}^* - \vect{q}_o) = \vect{0}, \nonumber
\end{align}

Lagrangian optimization method allows the given problem to be expressed in a closed-loop form in Equation~(\ref{eq:disp_opt_math_cl}), where $\vect{K} = \vect{K}_{stiff}~\vect{J}_p$.

\begin{align} \label{eq:disp_opt_math_cl}
\vect{q}^* = \vect{q}_d - \vect{K}~(\vect{K}^T~\vect{K})^{-1}~\vect{K}^T~(\vect{q}_d - \vect{q}_o)
\end{align}

Once the proxy set of joints are achieved through Equation~(\ref{eq:disp_opt_math_cl}), desired actuator force to reach the given pose can be calculated easily as:

\begin{align} \label{eq:disp_opt_math_cl_act}
F^*_a = \vect{J}^T_a \vect{K}_{cont}~(\vect{I} - \vect{K}~(\vect{K}^T~\vect{K})^{-1}~\vect{K}^T)~\Delta\vect{q}
\end{align}

\noindent where $\Delta\vect{q} = (\vect{q}_d - \vect{q}_o)$. The sensed impedance $F_a / \Delta\vect{q}$ of the overall system for configurational space can be depicted as:

\begin{align} \label{eq:disp_imped}
\vect{S}_a = - \vect{J}^T_a \vect{K}_{cont}~(\vect{I} - \vect{K}~(\vect{K}^T~\vect{K})^{-1}~\vect{K}^T).
\end{align}

The joint space based rendering strategy aims to estimate the future joint pose based on the previous behavior of the user, since the underactuation imposes no constraint in the joint space. Due to the lack of underactuated constraint in the joint space, the estimation of reachable finger pose proxy was put into torque form in order to take advantage from the same constraint.

\begin{figure}[htb]
\centering
% \vspace*{-2\baselineskip}
\includegraphics[width=0.6\textwidth]{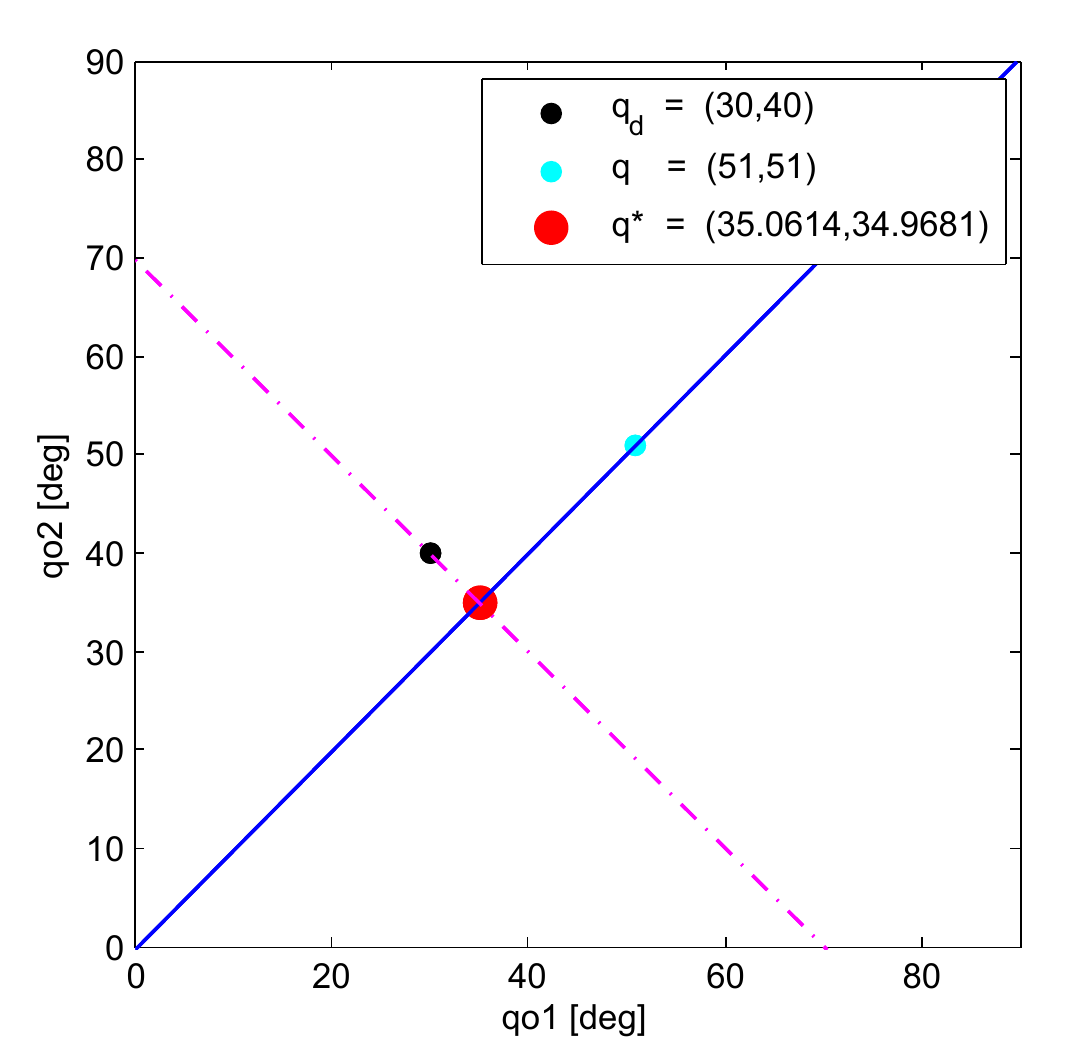}
\caption{Simulation plot for configurational space optimization problem based on minimizing the distance between desired finger pose and proxy finger pose.}
\label{fig:opt_dis}
%\vspace*{-1\baselineskip}
\end{figure}

The feasibility of the optimization algorithm based on desired pose in Equation~(\ref{eq:disp_opt_math_cl_act}) can be shown with the same example as before, where the current finger pose is selected as $\vect{q} = (51^o,~51^o)^T$ while the desired displacement is assumed to be $\vect{q}_d = (30^o,~40^o)^T$. Since the finger joints are moved by a simulation together with accurate relation, the joint stiffness elements were calculated as $1$, which forms the stiffness matrix $\vect{K}_{stiff} = [1~0;~0~1]$. The proxy finger pose$\vect{q}^*$ to replicate desired pose $\vect{q}_d$ was found simply using Equation~(\ref{eq:disp_opt_math_cl}). Such a proxy pose search was shown in Figure~\ref{fig:opt_dis} in a similar way to Figure~\ref{fig:opt_tor} in the operational space, rather than force space.

The constant blue line in Figure~\ref{fig:opt_dis} possible reachable sets of finger pose by the assistance of hand exoskeleton in the next time step for given finger pose $\vects{q}$ using the stiffness matrix $\vects{K}_{stiff}$. Light blue dot shows the actual finger pose $\vects{q}$, while the black dot shows the desired pose $\vects{q}_d$. Since the desired pose $\vects{q}_d$ does not lay on the blue line, it cannot be reached only by actuating the exoskeleton unless the user changes his behavior. That is why a proxy set of finger pose $\vect{q^*}$ is needed to be calculated through optimization in Equation~(\ref{eq:disp_opt_math_cl}), which was found as $(35.0614^o,~34.9681^o)^T$. The verification of reaching the minimum distance between $\vect{q}^*$ and $\vect{q}_d$ was performed simply by plotting an additional line that is perpendicular to the blue line and passes from $\vect{q}_d$. Since the proxy pose $\vect{q}^*$ occurs at the interaction point of both lines, the minimization problem can be verified.

\begin{figure}[!htb]
\centering
\vspace*{-1.5\baselineskip}
\subfigure[MCP active]{\includegraphics[width=0.46\textwidth]{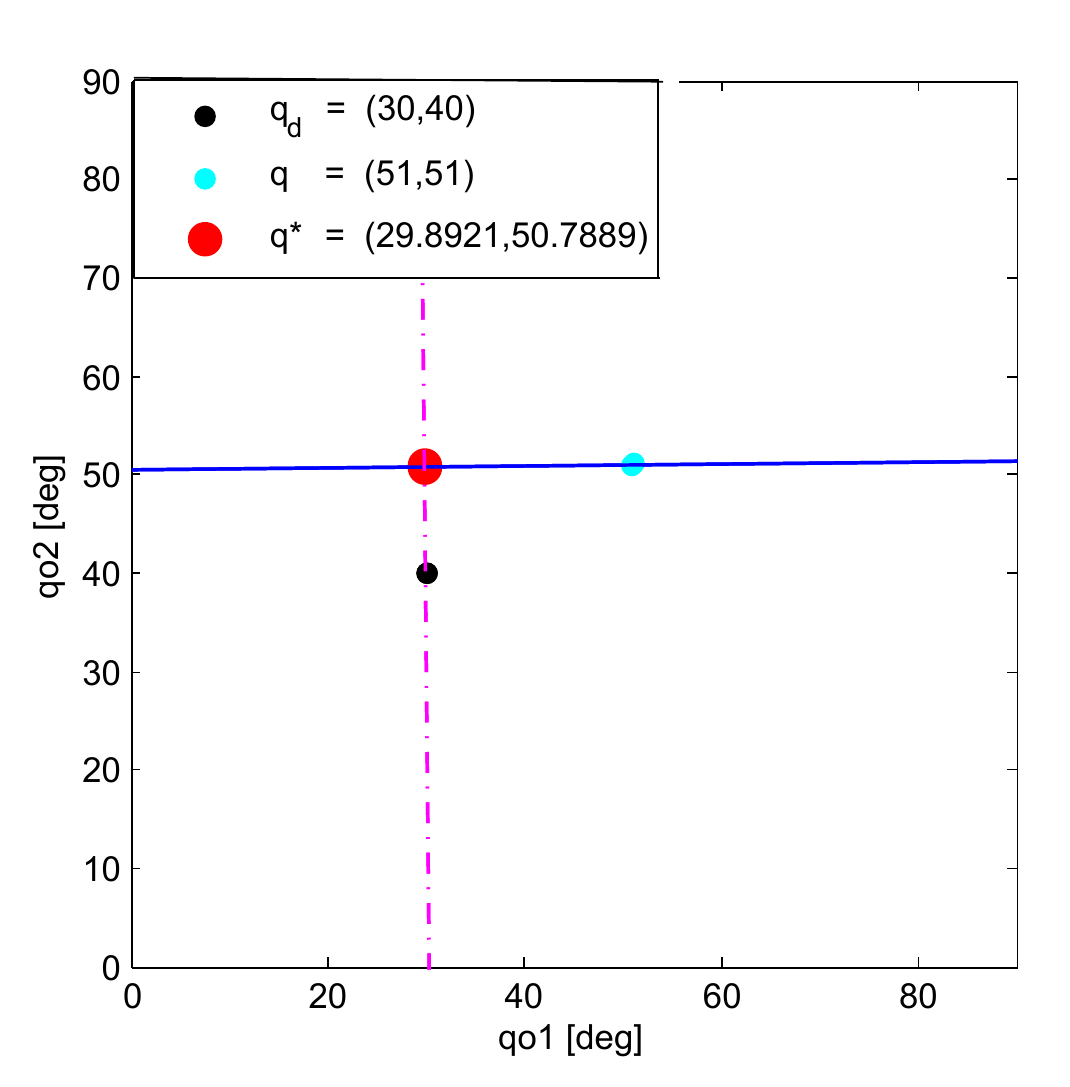}\label{fig:opt_dis_v1}} \subfigure[PIP active]{\includegraphics[width=0.485\textwidth]{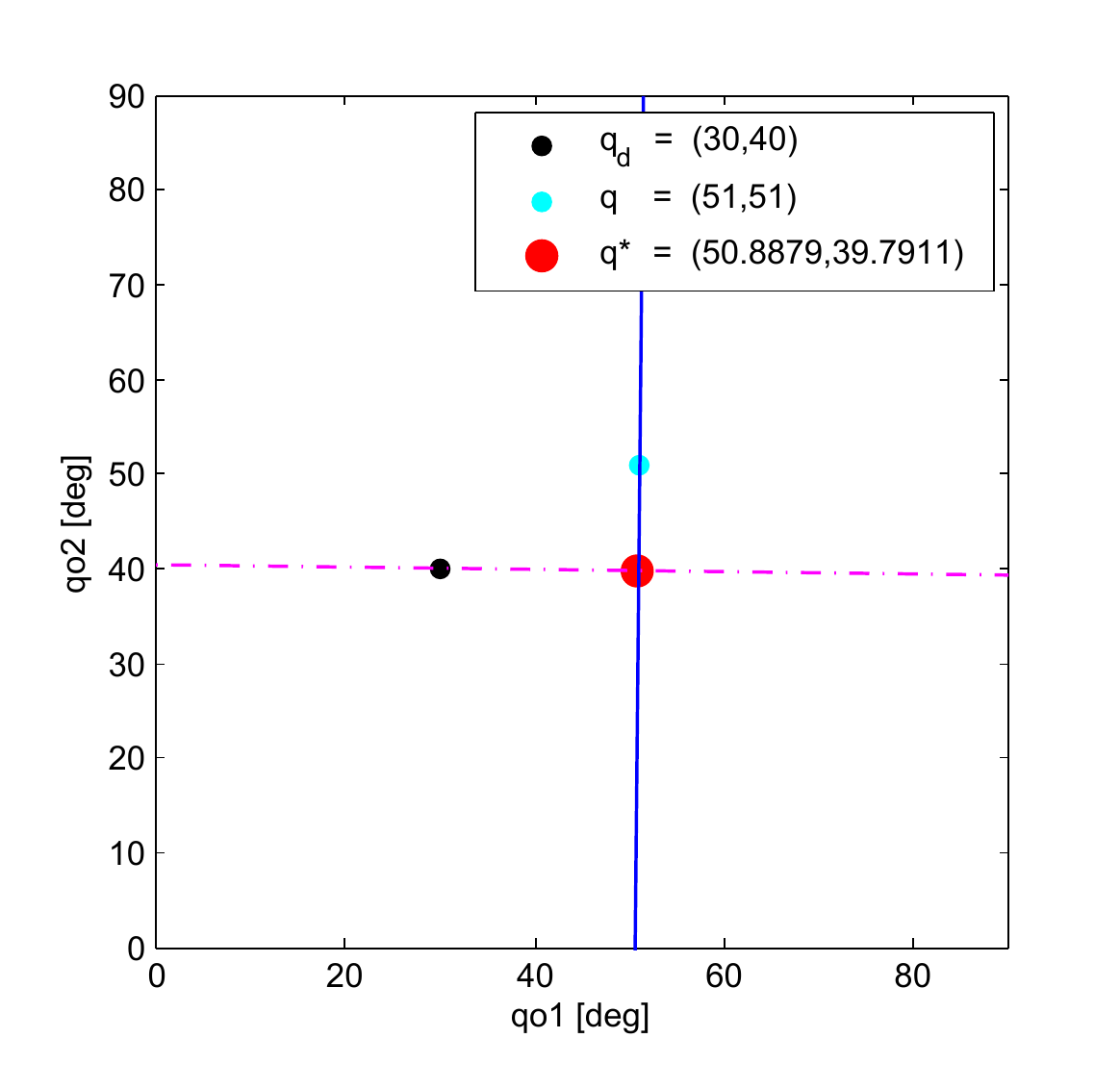}\label{fig:opt_dis_v2}}
\caption{Desired displacement optimization for different previous behavior: (a) only MCP joint was being rotated (ratio = 0.01); (b) only PIP joint was being rotated (ratio = 100).}	
\label{fig:opt_dis_multi}
%\vspace*{-0.5\baselineskip}
\end{figure}

This optimization approach that is based on capturing user's previous behavior can be interpreted with extreme circumstances, as displayed previously in Figure~\ref{fig:GraspingImage}. In particular, changing the finger joint trajectories in the simulation to obtain the ratio between joint stiffness values in $\vects{K}_{stiff}$ matrix as approximately $0.01$ and $100$ define scenarios where only MCP or PIP joint rotates while the other joint remains passive. The same optimization scenario was applied to both simulations and the results are represented in Figure~\ref{fig:opt_dis_multi}. It can be observed that a proxy for desired finger pose can be observed even in extreme situations. %Nevertheless, in order to avoid extreme jumps of the calculated ratio during the experiments with the device, it might be safer to limit these values not to exceed further than $0.01$ and $100$.

\begin{figure}[htb]
\centering
\vspace*{0.1\baselineskip}
\includegraphics[width=0.9\textwidth]{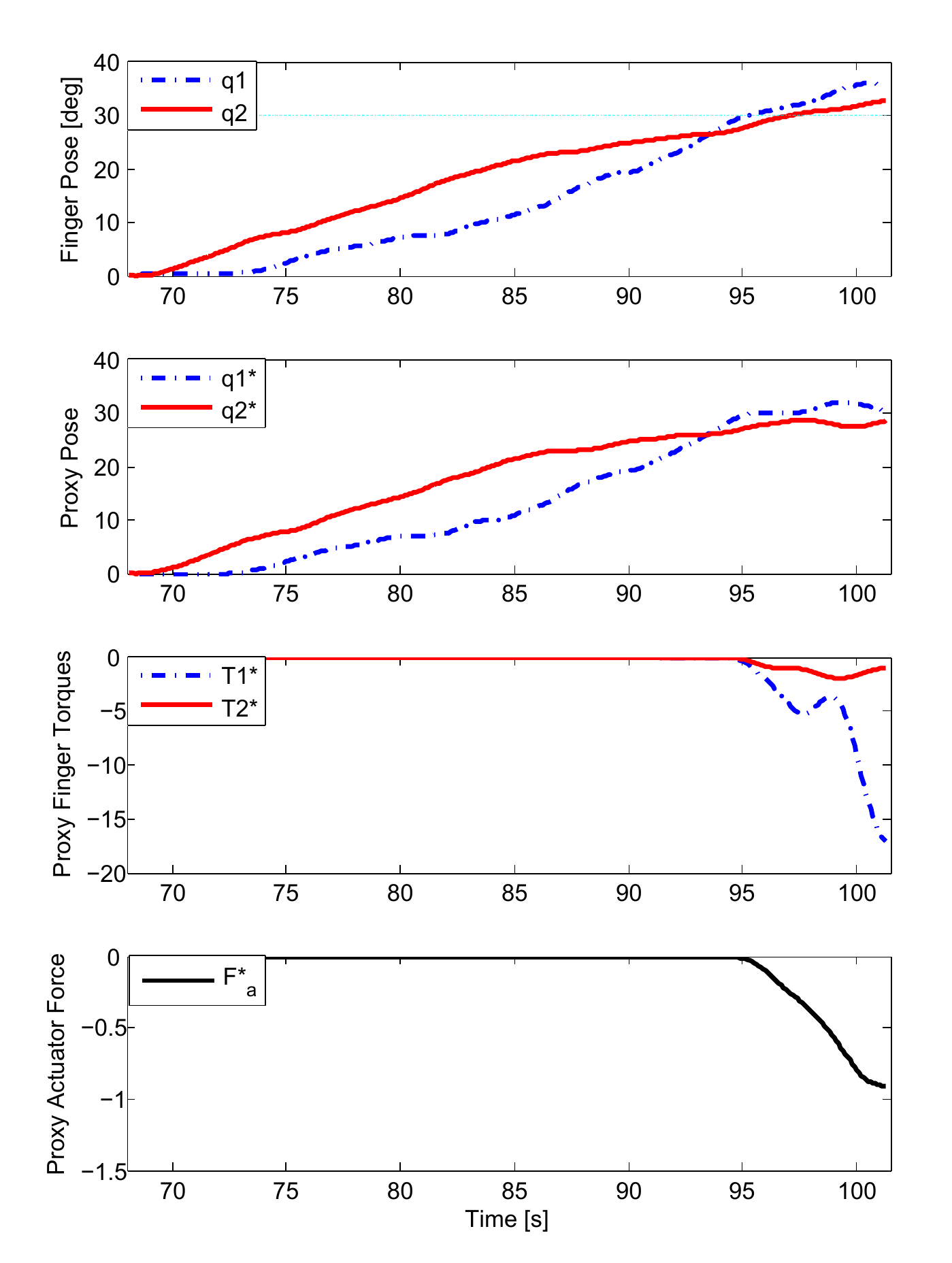}
\caption{Proxy pose optimization based on minimizing the error between the real and the proxy pose sets: (a) trajectory of the real time task performed by the user $\vects{q}$ with the virtual limits $\vects{q}_{lim}$ defined for the joints, (b) optimized proxy set of finger pose $\vects{q}^*$ calculated for user's trajectory, (c) desired finger torques $\vects{\tau}^*$ calculated using proxy set of finger pose $\vects{q}^*$ and (d) corresponding actuator force for the proxy torque values $F^*_a$.}
\label{fig:opt_dis_real}
%\vspace*{0.05\baselineskip}
\end{figure}

The proposed optimization algorithm to find a proxy set of desired finger joints $\vect{q}^*$ was tested using the hand exoskeleton. During the real time experiment, the desired finger pose is randomly selected as $\vect{q}_d = (30^o,~30^o)^T$. The stability of the control algorithm is ensured by running the control algorithm with $1 kHz$ frequency within the control board. Figure~\ref{fig:opt_dis_real} (a) shows the movement of finger joints $\vect{q}$ along with virtual joint limits $\vect{q}_{lim}$. For given instant, Equation~(\ref{eq:disp_opt_math_cl}) calculates a proxy set of finger joints that can be reached based on joint stiffness approximation as Figure~\ref{fig:opt_dis_real} (b). Figure~\ref{fig:opt_dis_real} (c) and Figure~\ref{fig:opt_dis_real} (d) show the corresponding joint torques $\vects{\tau}^*$ and actuator force $F^*_a$ for optimized proxy pose $\vect{q}^*$.

\clearpage

The validity of rendering strategy in the operational space needs to be shown by studying the validation of user's behavior. In particular, the efficacy of the control algorithm under operational space strategy can be studied simply by comparing the ratio between MCP and PIP joints using:

\begin{equation} \label{eq:ratio}
(\frac{q^*(i) - q(i)}{\|q^*(i) - q(i)\|})^T~\frac{q(i+1) - q(i)}{\|q(i+1) - q(i)\|}.
\end{equation}

\begin{figure}[htb]
\centering
%\vspace*{-0.7\baselineskip}
\includegraphics[width=0.9\textwidth]{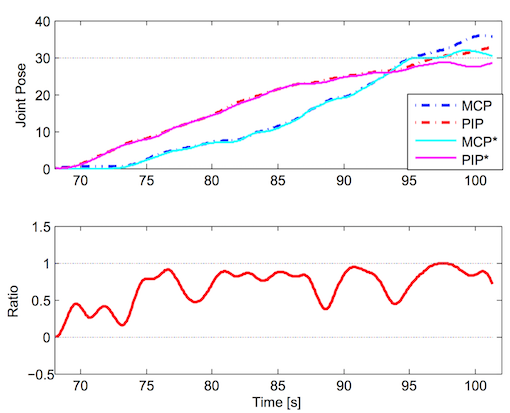}
%\vspace{-.5\baselineskip}
\caption{Validation of user's behavior estimation.}
\label{fig:ratio_pose}
%\vspace{-1\baselineskip}
\end{figure}

\noindent Doing so, the controller will be revealed to move in the estimated behavior and calculated proxy pose using user's previous estimation. Such a multiplication of ratios can be calculated to be similar when the output of Equation~(\ref{eq:ratio}), as depicted in Figure\ref{fig:ratio_pose}.

\begin{figure}[htb]
\centering
%\vspace*{-0.7\baselineskip}
\includegraphics[width=0.9\textwidth]{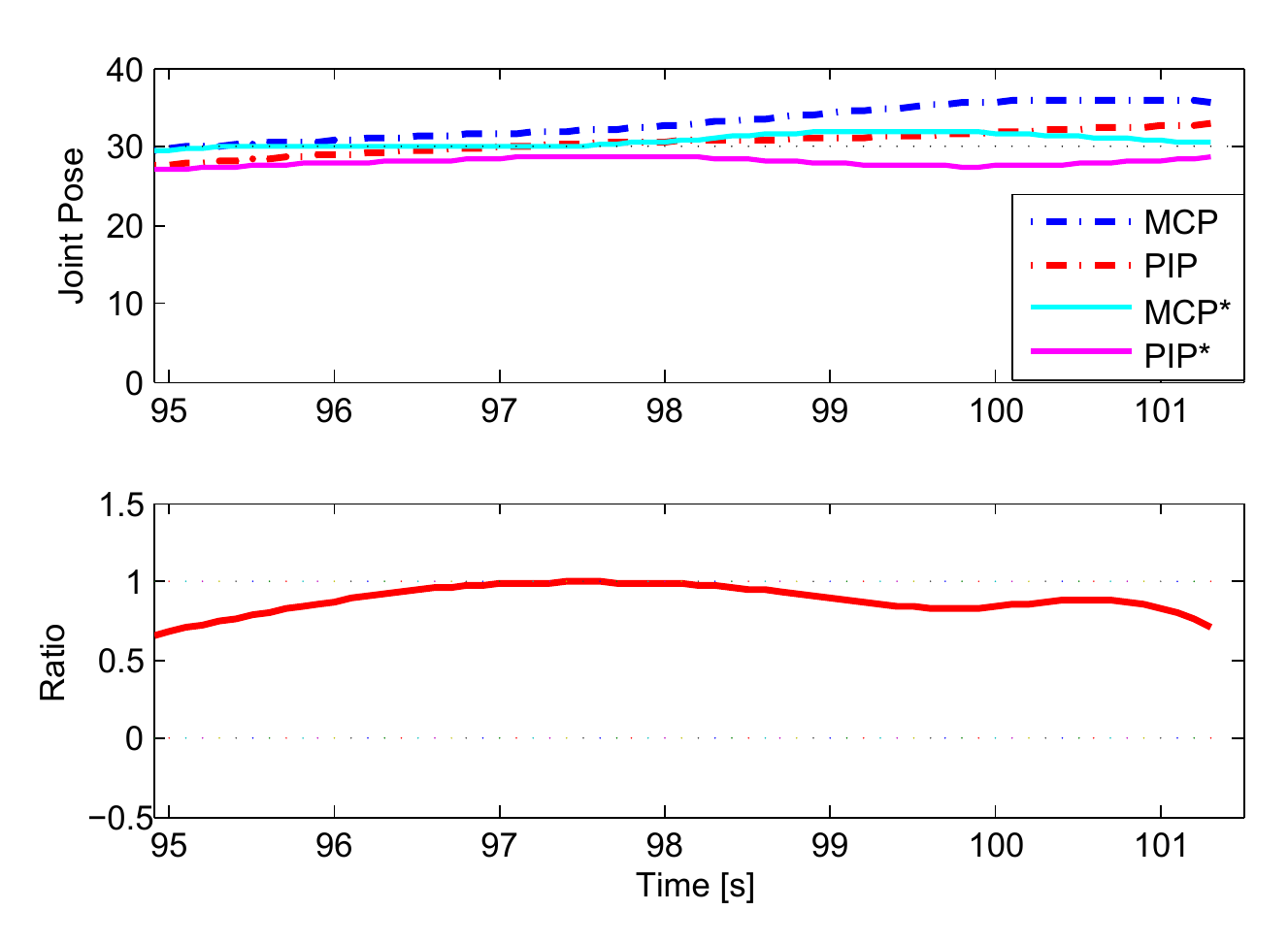}
%\vspace{-.5\baselineskip}
\caption{Validation of user's behavior estimation after user reaches the virtual limit pose.}
\label{fig:ratio_pose_partial}
%\vspace{-1\baselineskip}
\end{figure}

Since the stiffness rendering task, which applies forces to the user in order to impose the proxy finger pose as the reference, applies forces only after user reaches the virtual pose limit, the same ratio between the direction of proxy pose and user's future movement should be focused only after the virtual limit, as in Figure\ref{fig:ratio_pose_partial}.

Having ratio in Equation~(\ref{eq:ratio}) close to value $1$ signifies the closeness between the differences with proxy and actual finger pose and future and actual finger pose. Therefore, the efficacy of the control algorithm can be issued.

\begin{figure}[htb]
\centering
% \vspace*{-2\baselineskip}
\includegraphics[width=0.9\textwidth]{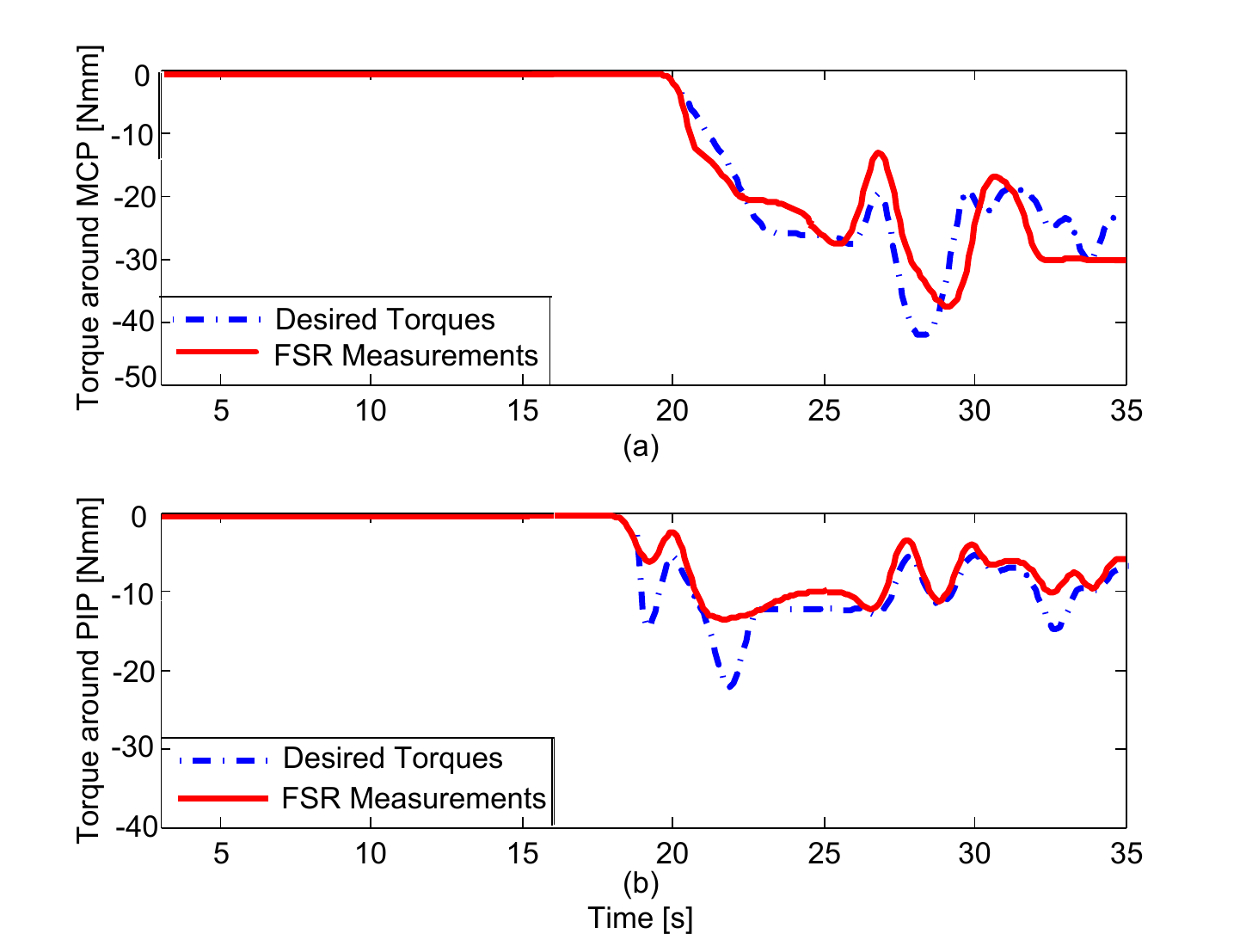}
\caption{Comparison between desired torques and actual torques measured by FSR sensors during rendering strategy based on proxy joint pose.}
\label{fig:opt_dis_fsr}
%\vspace*{-1\baselineskip}
\end{figure}

Similar to the previous strategy, the efficacy of force transmission of the given rendering strategy based on proxy pose should be proven using FSR measurements as depicted in Figure~\ref{fig:opt_exos_fsr}. Figure~\ref{fig:opt_dis_fsr} shows the torques around MCP and PIP joints using FSR measurements and desired torques calculated by the control algorithm during the experiment shown in Figure~\ref{fig:opt_dis_real}.

As expected, despite of the imperfections caused by FSR sensors, Figure~\ref{fig:opt_dis_fsr} shows that measured finger torques are highly correlated with desired ones. Therefore, one can say that even by running an optimization to find an achievable set of desired pose $\vect{q}^*$ the torques that are being applied to finger joints can be estimated. Furthermore, unlike the optimization based on desired torques, the rendering strategy based on proxy finger pose still performs the control algorithm in the configuration space.

\newpage

\subsection{Comparison between Rendering Strategies}

\begin{figure}[htb]
\centering
\vspace*{-.5\baselineskip}
%\resizebox{2.3in}{!}{\includegraphics{HandExos5Fin.jpg}}
\includegraphics[width=0.9\textwidth]{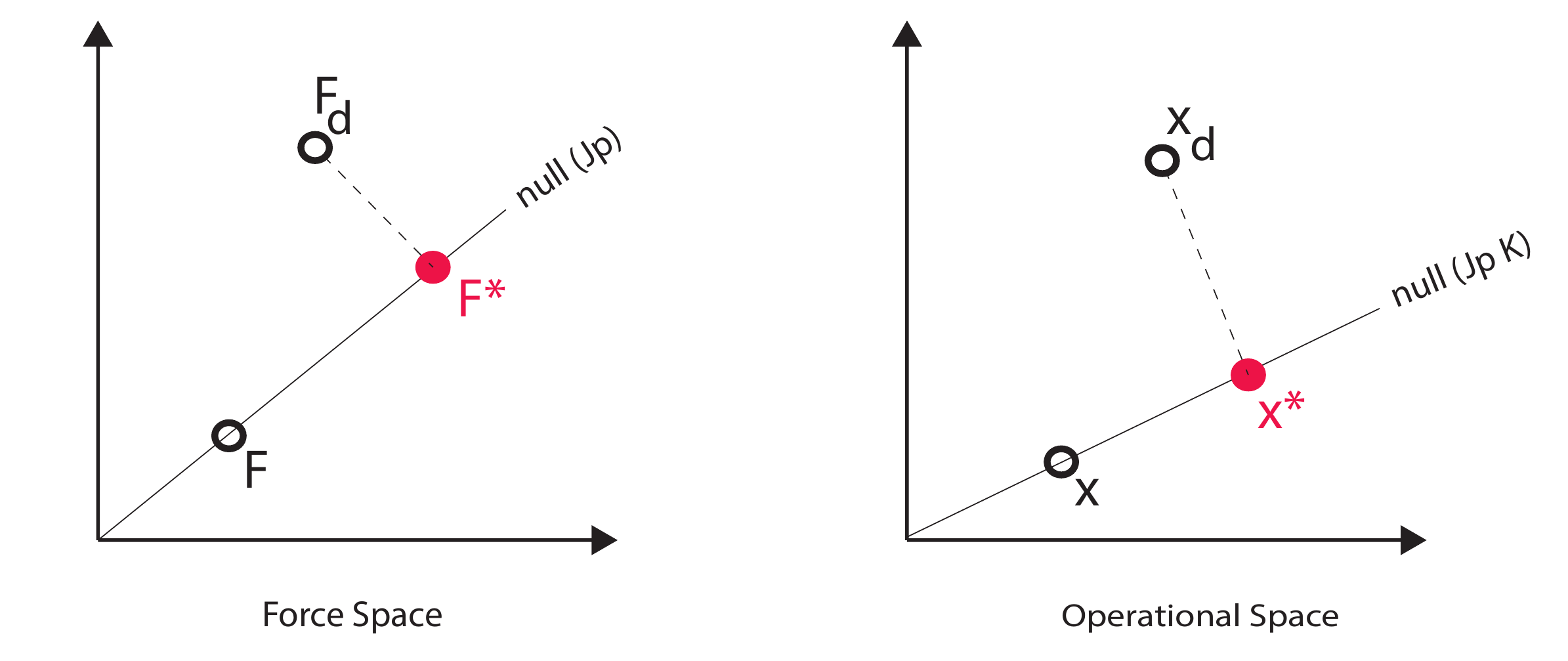}
\caption{Stiffness rendering algorithms for force space and operational space individually.}
\label{fig:optim}
%\vspace*{-0.5\baselineskip}
\end{figure}

Previously two different rendering strategies were proposed to overcome the uncertainties of stiffness rendering tasks for underactuated devices. These two methods were developed simply by computing proxy desired sets that might satisfy the underactuated constraints. The first method aims to find an applicable proxy set in the force space, while the second method reaches a reachable proxy set in the configuration space, as shown in Figure~\ref{fig:optim}.

In particular, the optimization algorithm based on desired torques calculate the set of desired torques $\vects{\tau}_d$ to be rendered based on adjustable stiffness values, which are set based on the properties of the virtual object, and calculates a proxy set of torques $\vects{\tau}^*_d$ that can be applied by the underactuated exoskeleton. Besides the fact that desired finger pose $\vect{q}_d$ cannot be achieved due to the lack of controllability of underactuation concept, resulting finger pose cannot be estimated due to closing the control loop based on proxy torque set. Since the stiffness rendering task is theoretically defined based on joint displacements, the lack of joint estimation to be reached might require further analysis. Nevertheless, for implementations, where the accuracy of rendered forces to finger joints is more crucial rather than the accuracy of imposed finger pose, rendering strategies based on force space might be more useful.

On the other hand, the optimization algorithm based on configuration space calculates the behavior of user's movement and estimates the future trajectory for finger joints. Once the future trajectory is estimated, desired finger pose is optimized to a proxy finger pose $\vect{q}^*$, which satisfies the estimated trajectory and the underactuation constraint. The optimized proxy finger pose is used to calculate corresponding joint torques using the force relation of a spring with estimated stiffness values. The given torques are applied the underactuation constraint, then the rendering stiffness is applied for the calculated torques. Such an optimization method defines the stiffness rendering task assuming the finger joints as imaginary springs. Running the optimization problem based on minimizing the applicable finger pose might improve the performance of underactuated device in the configuration space better than proxy torque strategy.

\begin{figure}[htb]
\centering
%\vspace*{-0.7\baselineskip}
\includegraphics[width=0.95\textwidth]{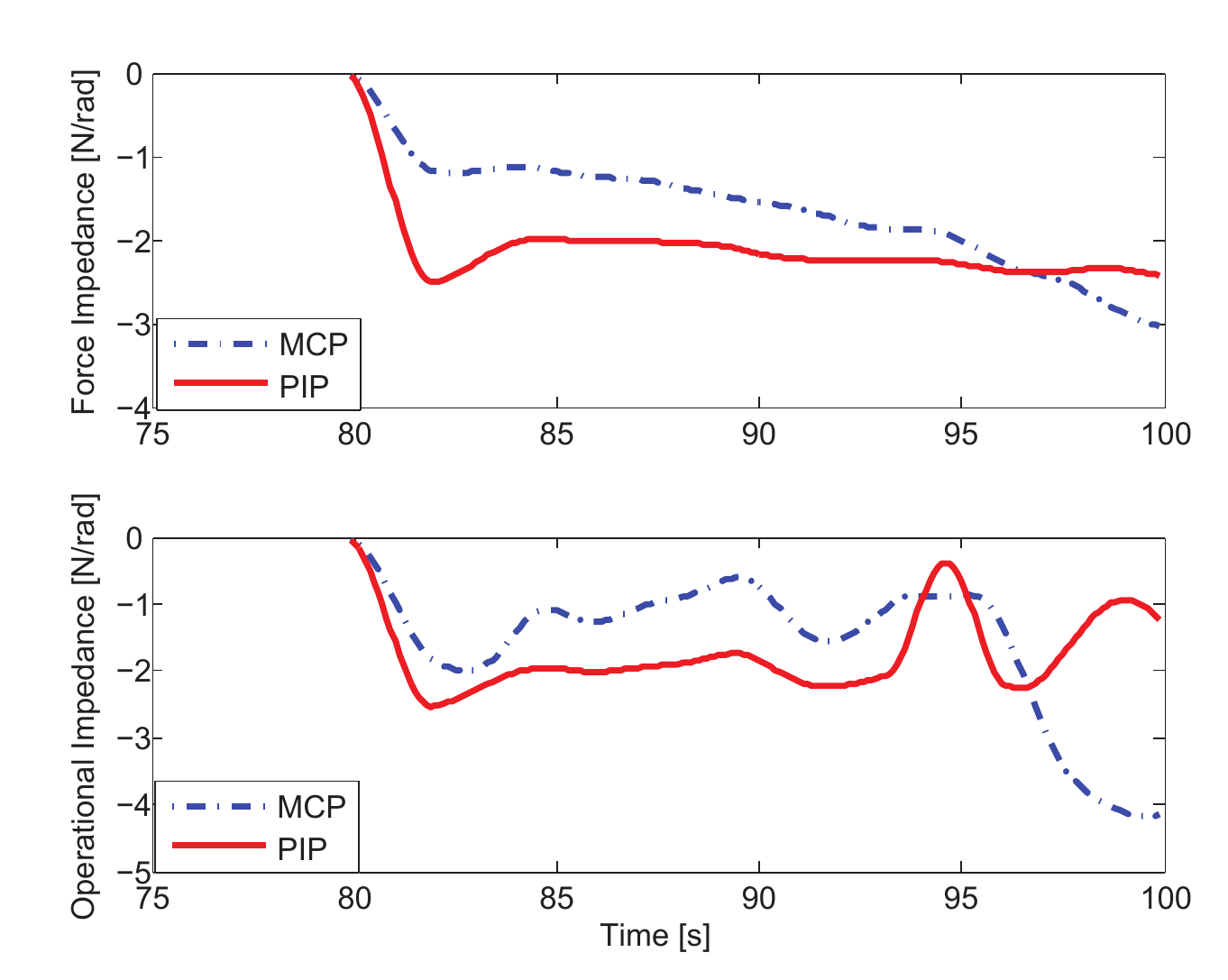}
%\vspace{-.5\baselineskip}
\caption{Sensed impedance $S_a = F^*_a / \vects{q}$ calculated during the experiment in Figure~\ref{fig:opt_dis_real} for (a) proxy torque strategy in an offline manner and (b) proxy pose strategy in an online manner.}
\label{fig:imped}
%\vspace{-1\baselineskip}
\end{figure}

Passivity of the underactuated exoskeleton throughout the operation can be studied for system stability. For both algorithms, the sensed impedance $F^*_a/\vects{q}$ was expressed previously using Equation~(\ref{eq:force_imped}) and Equation~(\ref{eq:disp_imped}). The impedance of both strategies can be compared using the same experiment, where the user perceives force feedback produced by the proxy pose strategy, while sensed impedance for proxy torque strategy was computed in an offline manner. The motivation of such a choice is that force feedback is highly crucial in order to obtain meaningful joint stiffness prediction to form the matrix $\vects{K}_{stiff}$ for proxy pose strategy. The experimental expressions of sensed impedance for proxy torque and proxy pose strategies can be seen in Figure~\ref{fig:imped}, for the experiment shown in Figure~\ref{fig:opt_dis_real}.

Composing Equation~(\ref{eq:disp_imped}) is different than Equation~(\ref{eq:force_imped}) only because the estimation of joint stiffness values are considered based on previous movement of the user. Since the user cannot ensure a smooth behavior while opening/closing fingers in a free environment, such an estimation can cause the impedance plots to be more chattering. Obtaining all negative values for joint stiffness values of both methods signify stability of the system.

\clearpage

\newpage
\section{Proxy based Haptic Rendering}

Once the possibility of developing stiffness rendering strategies for underactuated systems by real time experiments performed by the proposed underactuated hand exoskeleton, the similar idea can be generalized and extended to haptic rendering. In this scenario, the user will be asked to control an avatar in a virtual environment simulation, where the hand exoskeleton will monitor the user's movements and provide force feedback in case of a contact. Even though a different notation will be introduced, that is more favorable for the haptic rendering point of view, the idea behind the soon to be introduced proxy based haptic rendering is the same as the rendering strategy based on finger pose.

The novelty of the proposed proxy-based haptic rendering is to provide fully simulated and constrained by the virtual environment, even with soft objects and to be constrained to the subspace defined by the actuated DoF of the device. Thanks to the subspace proxy, we can compute feedback haptic commands that are optimally constrained to actuated DoF, and we can do this using regular controllers as in the standard rendering method. It admits haptic and virtual configuration spaces of different dimensionality, that are not defined using the physical constraints of fingers as we did previously, and connected through nonlinear mappings.

\subsection{Notation}

Since the following rendering strategies will be proposed for general haptic simulations, countless configurational contact points with virtual objects will be used to compute actuator forces for feedback. With this motivation, a different, generic notation will be used as $Q = Q_a \times Q_n$ the configuration space of the haptic device, with $Q_a$ the actuated configuration space (resulting from actuated DoF), and $Q_n$ the non-actuated configuration space (resulting from non-actuated DoF). $\vect{q} = (\vect{q}_a; \vect{q}_n)^T~\epsilon~Q$ represents a device state, with $\vect{q}_a~\epsilon~Q_a$ the state of actuated DoF, and $\vect{q}_n~\epsilon~Q_n$ the stated of non-actuated DoF. Similarly, we denote as $\chi$ the (unconstrained) configuration space of the virtual object. $\textrm{x}~\epsilon~\chi$ represents a virtual state.

\begin{figure}[hbt]
\centering
\includegraphics[width=1\textwidth]{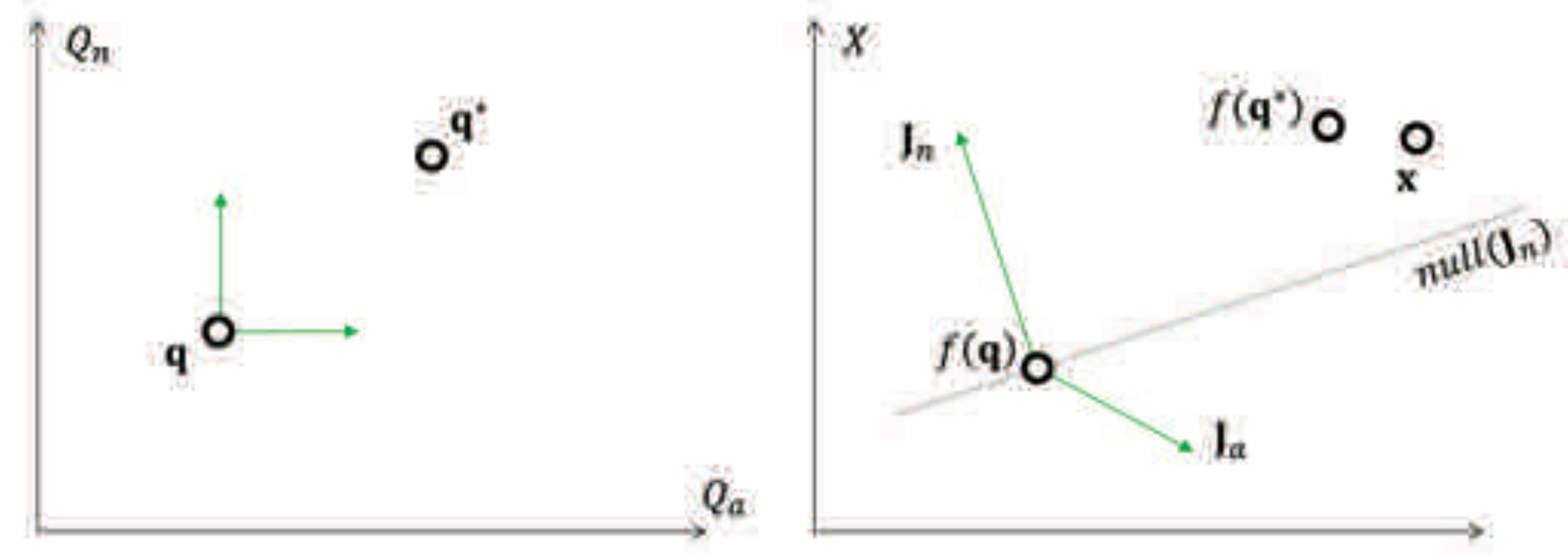}
\caption{Schematic representation of the device configuration space $Q$(left) and the virtual configuration space $\chi$ (right). $\vect{q}$ represents the device state and $f(\vect{q})$ its corresponding configuration in the virtual environment, i.e., the haptic probe; $\textrm{x}$ represents the standard proxy; $\vect{q}^*$ represents the subspace proxy and $f(\vect{q}^*)$ its corresponding configuration in the virtual environment. The images also represent the linear subspaces of actuated and non-actuated motion in the virtual environment, $\vect{J}_a$ and $\vect{J}_n$}
\label{fig:conf}
%\vspace*{-1\baselineskip}
\end{figure}

There is a kinematic mapping $f: Q \rightarrow \chi$ from the configuration space of the device to the configuration space of the virtual object. Then, we can compute a virtual state corresponding to a device state, i.e., $\textrm{x} = f(\vect{q})$. Figure~\ref{fig:conf} shows a conceptual representation of haptic and virtual configuration spaces, very similar to the one in Figure~\ref{fig:optim}, together with their mapping. We denote as $\vect{J} = \partial \textrm{x} / \partial q$ the Jacobian of the mapping $f$, which allows a linearization of the typically nonlinear mapping $f$. We also split the Jacobian into $\vect{J} = (\vect{J}_a; \vect{J}_n)$, with $\vect{J}_a = \partial \textrm{x} / \partial q_a$ and $\vect{J}_n = \partial x / \partial q_n$ with respect to the actuated and non-actuated DoF respectively. Only when the number of DoF of $\chi$ and $Q$ match, $\vect{J}$ is square, and the mapping $f$ may be invertible.

Finally, we denote as $\tau$ a device force vector, and as $\vect{f}$ a generalized virtual force. Since the proposed hand exoskeleton is an impedance-mode rendering, such assumption will be made to proceed the optimization procedure. Nevertheless, all methods could be applied in admittance-mode rendering as well, simply by exchanging force commands with position commands.

\subsection{Review of Proxy-based Haptic Rendering} \label{sec:proxy_based}

This section provides a formal description of the standard proxy-based haptic rendering method as much the discussions regarding the major assumptions and the problems induced when the method is applied to underactuated devices. Furthermore the variation of proxy-based rendering to compute optimal actuator forces will be investigated, within the underactuation constraints. Even though such method typically satisfies passivity, it yields a configuration-dependent rendering impedance.

\subsubsection{Standard rendering algorithm}

Assume $\vect{q}$ and $f(\vect{q})$ define the state of the device and the corresponding position of the haptic probe in the virtual environment respectively. Proxy-based rendering computes a proxy $\textrm{x}$ that minimizes the distance to the probe according to some metric, subject to environment constraints. Then, the method computes a force $\vect{f} = \vect{Z}_x \Delta \textrm{x}$ in the virtual environment, based on some mechanical impedance $\vect{Z}_x$ and the displacement from the probe to the proxy $\Delta \textrm{x} = \textrm{x} - f(\vect{q})$. Next, the force is transformed to the configuration space of the device using the Jacobian transpose approach: $\tau = \vect{J}^T \vect{f}$. Finally, forces are displayed.

The standard proxy-based rendering algorithm makes important assumptions about the configuration spaces $Q$ and $X$: their dimensionality is the same, the Jacobian $\vect{J}$ is hence square, and the mapping $f$ is invertible. This is the case, for example, in typical 3-DoF and 6-DoF haptic rendering systems. In haptic rendering of deformable objects using stylus devices, it is easy to extract a rigid subspace of the full deformable configuration space (using, e.g., rigid modes or a rigid handle), and define this rigid subspace as X for the purpose of applying the proxy-based rendering algorithm.

\subsubsection{Analysis for underactuated systems}

With underactuated devices, even if a full force vector $\tau$ is computed, force can obviously be rendered only on actuated DoF. The effective force resulting from the application of the proxy-based rendering algorithm to an underactuated device is then:

\begin{align} \label{eq:hapt1}
\tau_a &= \vect{J}^T_a \vect{f} \nonumber \\
&= \vect{J}^T_a \vect{Z}_x \Delta \textrm{x},
\end{align}

Simple projection of the forces to the actuated DoF fails to reproduce target forces that lie in the null-space of the actuated DoF. In some popular types of underactuated haptic systems, it is easy to map the actuated DoF to a well defined subspace of the full virtual configuration space, e.g., in systems with 6-DoF input (translation and rotation) and 3-DoF output (force only). In these cases, the Jacobian $\vect{J}$ is block diagonal, and the forces of virtual DoF that map to actuated DoF are matched exactly, while the forces of other virtual DoF are simply zero.

The rendering algorithm can formally be analyzed in terms of the displayed impedance $\partial \tau / \partial \vect{q}$. The rendering method is passive if the displayed impedance is negative definite, i.e., all its eigenvalues are negative. At this point, two main assumptions have been made regarding the following analysis: the local change of the proxy position due to the motion of the device is ignored ($\partial \textrm{x} / \partial \vect{q}$) and also the local change of the Jacobian is ignored ($\partial \vect{J} / \partial \vect{q}$).

Combining the definition $\tau = \vect{S}^T \tau_a$ with Equation~(\ref{eq:hapt1}), the expression shown in Equation~(\ref{eq:hapt2}) is obtained. It is important to emphasize that this expression does not give any guarantee that the displayed impedance is negative definite.

\begin{align} \label{eq:hapt2}
\frac{\partial \tau}{\partial \vect{q}} = - \vect{S}^T \vect{J}^T_a \vect{Z}_x \vect{J}
\end{align}

\subsubsection{Null-space force optimization}

Alternatively, a variation of proxy-based haptic rendering can be studied, which accounts for underactuation prior to transforming the target force $\vect{f}$ to the configuration space of the device. In essence, the method transforms a different force $\vect{f}^*$, as close as possible to the target force, but which yields no forces on non-actuated DoF. The approach can be formulated as a constrained optimization problem:

\begin{align}
\tau_a &= \vect{J}^T_a \vect{f}^* \textrm{with} \\ \nonumber
\vect{f}^* &= \textrm{argmin}\|\vect{f}^* - \vect{f}\|^2, \textrm{s.t.} \vect{J}^T_n \vect{f}^* = 0,
\end{align}

\noindent which was expressed previously as Equation~(\ref{eq:force_opt_math}) in Section~\ref{sec:opt_torque}. The proposed optimization problem can be expressed in a closed-form solution, just as Equation~(\ref{eq:force_opt_math_cl_act}) with the adjusted notation as:

\begin{equation} \label{eq:hapt3}
\tau_a = \vect{J}^T_a (\vect{I} - \vect{J}_n (\vect{J}^T_n \vect{J}_n)^{-1} \vect{J}^T_n) \vect{Z}_x \Delta \textrm{x}.
\end{equation}

The interpretation of the equation above is that the method projects the target forces $\vect{f}$ to the null-space of the nonactuated DoF prior to applying the Jacobian transpose. In this case, instead of Equation~(\ref{eq:force_imped}), the displayed impedance is:

\begin{align} \label{eq:hapt4}
\frac{\partial \tau}{\partial \vect{q}} = - \vect{S}^T \vect{J}^T_a (\vect{I} - \vect{J}_n (\vect{J}^T_n \vect{J}_n)^{-1} \vect{J}^T_n) \vect{Z}_x \vect{J}.
\end{align}

In a simple case where $\vect{Z}_x$ is a uniform stiffness for all DoF of the virtual object, i.e., $\vect{Z}_x = k \vect{I}$, then $\partial \tau_a / \partial \vect{q}_a = 0$ and all eigenvalues of $\frac{\partial \tau_a}{\partial \vect{q}_a} $ are negative, hence passivity is guaranteed. But this is not necessarily the case if the impedance $\vect{Z}_x$ is more complex.

In addition, to ensure stability of the rendering, the stiffness of the displayed impedance must be bounded as a function of the sampling rate. As $\frac{\partial \tau}{\partial \vect{q}}$ depends on the actuated and non-actuated Jacobian matrices $\vect{J}_a$ and $\vect{J}_n$, stability imposes complex nonlinear conditions on the impedance $\vect{Z}_x$. That being said, one can conclude that, with null-space force optimization, maximization of rendering transparency depends in a complex nonlinear way on the mapping from device configuration space to virtual configuration space.

Based on these conclusions, instead of just optimizing rendered forces of the standard proxy-based method, this study seeks a novel rendering method that addresses the challenge of underactuation while remaining passive, and also simplifies maximizing transparency.

\subsection{Rendering for Underactuated Devices}

In this section, a novel haptic rendering method for underactuated devices is proposed. Firstly a nonlinear formulation of a subspace proxy constrained to actuated DoF is presented, then the problem is linearized to yield an efficient rendering method. Finally, the benefits of the method are analyzed compared to the two methods discussed in the previous section.

\subsubsection{Subspace proxy}

The core of the proposed method is simple. The main motivation is to exploit all benefits of the proxy-based rendering method, namely: (i) accurate visual simulation of the virtual object, (ii) simple rendering of forces based on deviations between proxy and probe, and (iii) simple tuning of the rendering impedance.

To achieve this, and to circumvent the dimensionality difference of $Q$ and $\chi$, two different proxies are defined. The classical proxy, $\textrm{x} \epsilon \chi$, is a virtual object that is simulated by minimizing the distance to the haptic probe subject to environment constraints; and a \emph{subspace proxy}, $\vect{q}^*~\epsilon~Q$, is a proxy constrained to the actuated configuration space of the device $Q$. Thanks to the classical proxy $\textrm{x}$, the visual accuracy of the simulation is retained. Thanks to the subspace proxy $\vect{q}^*$, the forces can be computed directly based on the deviation $\Delta \vect{q}_a = \vect{q}^*_a - \vect{q}_a$ in the actuated state of the device. And as a corollary, transparency is easily maximized by tuning the display impedance directly on the actuated DoF.

Given a haptic device state $\vect{q}$, the subspace proxy $\vect{q}^*$ is proposed to be computed by finding the corresponding virtual configuration $f(\vect{q}^*)$ that minimizes the distance to the proxy x. In practice, this is done by solving an optimization problem, as in Section~\ref{sec:disp_optimization}.

Once the subspace proxy is computed, we can also compute the device force (on the actuated DoF) $\tau_a$ based on a rendering impedance $\vect{Z}_q$ and the deviation $\Delta \vect{q}_a = \vect{S} \Delta \vect{q}$ on the actuated DoF alone. Formally, we define our subspace proxy-based haptic rendering as follows:

\begin{align} \label{eq:hapt5}
\tau_a &= \vect{Z}_q \vect{S} (\vect{q}^* - \vect{q}), \textrm{with} \nonumber\\
\vect{q}^* &= \textrm{argmin} \| \textrm{x} - f(\vect{q}^*)\|^2.
\end{align}

The main challenge of this formulation is that the mapping $f$ is nonlinear, and finding $\vect{q}^*$ requires solving a nonlinear
optimization. Therefore, the next step would be to relax this challenge.

\subsubsection{Linearized subspace proxy}

By linearizing the mapping $f$ at the current device state $\vect{q}$, the optimization problem in Equation~(\ref{eq:hapt5}) turns into a simple quadratic
optimization:

\begin{align}\label{eq:hapt9}
\tau_a &= \vect{Z}_q \vect{S} \Delta \vect{q}, \textrm{with} \nonumber \\
\vect{q} &= \textrm{argmin} \|\Delta \textrm{x} - \vect{J} \Delta \vect{q} \|^2,
\end{align}

with the following closed-form solution:

\begin{equation} \label{eq:hapt6}
\tau_a = \vect{Z}_q \vect{S} (\vect{J}^T \vect{J})^{-1} \vect{J}^T \Delta x.
\end{equation}

Please note that the proposed method in Equation~(\ref{eq:hapt9}) and Equation~(\ref{eq:hapt6}) is the same method in Section~\ref{sec:disp_optimization} presented as Equation(\ref{eq:disp_opt_math}) and Equation(\ref{eq:disp_opt_math_cl_act}). In the previous method, simplifying the configurational space into stiff structure of human finger instead of defining a avatar simulation with limitless points gave us the ability to study the previous behavior of user to improve the future estimations. However, such an estimation was eliminated for this analysis.

The interpretation of the equation above is that the method projects the proxy deviation $\Delta \textrm{x}$ onto the device DoF to compute a subspace proxy deviation prior to the force computation. This implies an important conceptual difference with respect to the standard proxy-based rendering method. The standard method transforms forces from the virtual environment to the device, whereas our method transforms displacements and keeps force computation at the device level.

The computational cost of the method is negligible. It requires the solution of a linear system whose size is given by the number of DoF of the device.

%\subsubsection{Analysis}

Similar to Section~\ref{sec:proxy_based}, the displayed impedance is analyzed, which in this case is:

\begin{equation} \label{eq:hapt7}
\frac{\partial \tau}{\partial \vect{q}} = -\vect{S}^T \vect{Z}_q \vect{S}.
\end{equation}

This impedance yields two notable results. First, passivity is easily enforced, simply by ensuring that the rendering impedance $\vect{Z}_q$ is positive definite. Second, transparency is easily maximized, simply by setting the stiffness terms in $\vect{Z}_q$ to the maximum allowed by stability constraints. Unlike the previous methods, the displayed impedance is not affected by configuration-dependent scaling factors.

\subsection{Results}

\begin{figure}[hbt]
\centering
\includegraphics[width=1\textwidth]{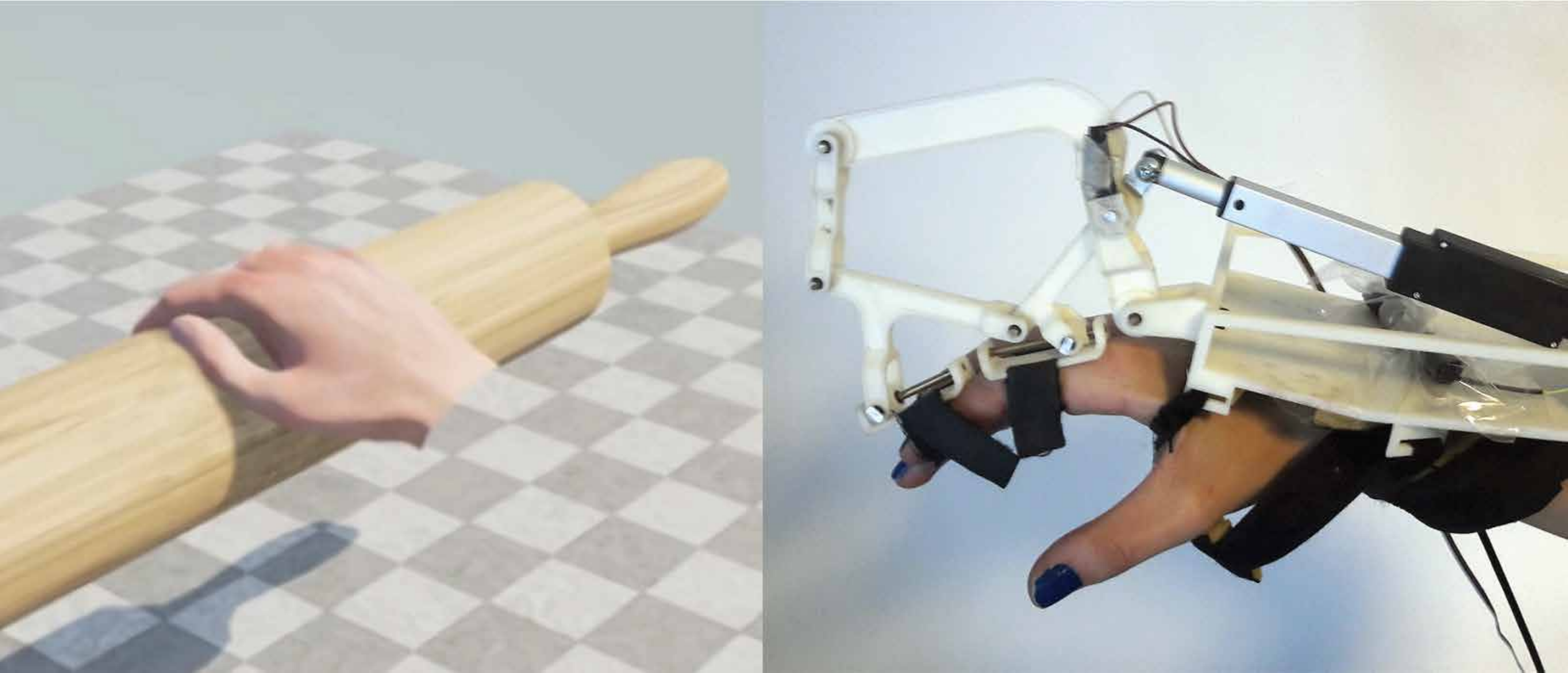}
\caption{Contact with a soft hand being rendered through an underactuated exoskeleton.}
\label{fig:unreal_handexos}
\end{figure}

The feasibility of the proposed rendering algorithm has been expressed in this study through the given optimization formulation. This approach enables simple and elegant formulation and interpretation, and it leaves the mathematical and implementation details to each particular type of device, virtual probe, and objective function. In this section, a particular implementation for a single finger component of the proposed underactuated hand exoskeleton will be described for haptic rendering of soft grasping interactions, as shown in Figure~\ref{fig:unreal_handexos}.

The virtual interactions are rendered in the virtual environment between a soft finger model \cite{Perez2013} and other objects, as shown in Figure~\ref{fig:unreal_handexos}. In particular, the motion of the palm is tracked through a Leap Motion device while the finger tracking is performed by the exoskeleton (see Section~\ref{sec:pose_estimation}). This combined tracking sets the configuration $\vect{q}$ of the device, which we transform into the probe representation of the phalanges $f(\vect{q})$ using the two joints mentioned above. The contact between the soft finger and the grasping objects are modelled by constraining the proxy phalanges. Finally, the proposed rendering algorithm is applied to compute the force command for the finger exoskeleton. The given force command is utilized as a reference for the force control algorithm discussed in Section~\ref{sec:force_cont}.

%\subsubsection{Results}

Several finger trajectories were recorded as well as their associated rendering computations. For the experiments, we have used as impedance a normalized stiffness to factor out the average scale in $\vect{J}_a$, i.e., $\vect{Z}_q = 1$ and

\begin{equation} \label{eq:hapt8}
\vect{Z}_x = \frac{1}{avg(\|\vect{J}_a\|)^2} \vect{I}.
\end{equation}

In order to compare the proposed subspace rendering method to the standard and null-space methods, a force computation and impedance analysis have been carried out in a controlled setting. Given a recorded finger trajectory, the proxy is fixed at $\textrm{x} = 0$, and the output force for the linear actuator as well as the displayed impedance $d\tau_a / d\vect{q}_a$ are computed. Such impedance is computed (a) following the theoretical formulations in Equation~(\ref{eq:hapt2}), Equation~(\ref{eq:hapt4}) and Equation~(\ref{eq:hapt7}) respectively for the three rendering methods, and (b) through finite differences of the applied force and the device motion between frames, i.e., $\Delta\tau_a / \Delta\vect{q}_a$, as performed in the previous sections.

\begin{figure}[h!]
\centering
\includegraphics[width=0.7\textwidth]{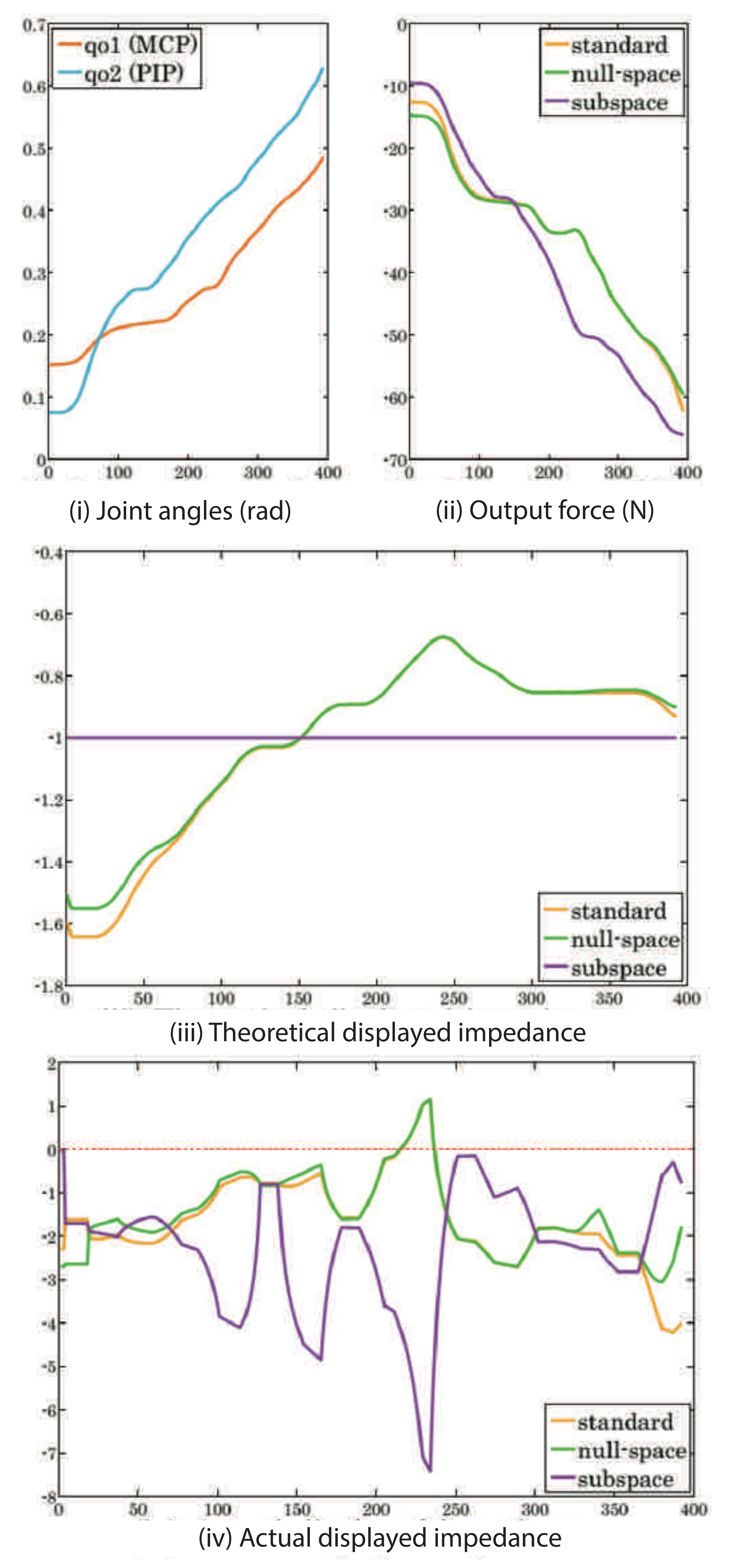}
\caption{Performance comparison of the three rendering methods discussed in the paper (standard, optimized, and subspace). In the test, the proxy is kept still at a zero angle.}
\label{fig:PerformanceComparison}
%\vspace*{-1\baselineskip}
\end{figure}

Figure~\ref{fig:PerformanceComparison} shows the results for a sample finger trajectory. Our subspace-proxy-based method is always passive in the
experiment, according to the theoretical result but also in practice. The standard and null-space methods, on the other hand, are not always passive in practice. This contradicts the theoretical results, due to the missing $\partial\vect{J}/\partial \vect{q}$ term.

\clearpage

Such a novel rendering method opens up multiple avenues for further investigation. First, the proposed rendering algorithm linearizes the mapping from device to virtual workspace, which works well when the deviation between device state and subspace proxy is small. A full nonlinear solve would be more robust under large proxy deviations, but it requires efficient solution methods. Second, in the impedance analysis two important approximations have been made, namely that the proxy remains still and that the Jacobian of the mapping from device to virtual workspace is constant. A passivity controller could be needed to enforce passivity in all cases. And third and most important, the overall quality of haptic rendering can be optimized in a task-specific and device-specific manner by tuning the objective functions that guide the computation of the proxy and the subspace proxy.

\clearpage
\pagebreak

\newpage

\chapter{Conclusion} % Main chapter title

\label{sec:Chapter6} % For referencing the chapter elsewhere, use \ref{Chapter1}

\lhead{Chapter 6. \emph{Conclusion}} % This is for the header on each page - perhaps a shortened title

In this dissertation, a novel underactuated hand exoskeleton was proposed mainly to offer help for post-stroke patients with hand disabilities during physical rehabilitation exercises and daily activities with robotic assistance. Its use can also be extended for haptic applications.

This study began by stating the importance and complexity of the hand in terms of kinematics and crucial functionality during the activities of daily living (ADLs), as much as the importance of providing an efficient treatment process to overcome these physical injuries. The robotic devices can prove efficient, repetitive and accurate physical exercises, therefore the motor recovery and the output of rehabilitation therapy can be improved simply by creating robot assisted rehabilitation scenarios. The kinematics model of the hand was analyzed to understand how the assistance should be provided to the user and possible simplifications due to the focus of motion. A wide range of hand exoskeletons, which focus on recovering the hand functionality, were investigated in details to analyze their characteristics and weaknesses in order to create a generic hand exoskeleton. Furthermore, the devices were categorized based on (i) their active and passive mobility for the hand and for each finger, (ii) their number of connection points between the hand exoskeleton and each finger, (iii) their pose estimation technique throughout the operation, (iv) their actuation type, (v) their placement of each finger component and (vi) their control algorithm. Each category was studied to give kinematics decisions to meet the design requirements of the proposed hand exoskeleton.

The design criteria of the proposed hand exoskeleton were listed based on basic safety precautions for human interaction and specific requirements based on rehabilitation or assistive applications. Using these requirements, the design properties for the proposed hand exoskeleton were listed as:

\begin{itemize}
\item assisting only for $2~DoFs$ finger rotation of each finger for flexion/extension only;
\item reaching independent control of fingers, implementing single actuators for each finger component;
\item achieving $2~DoFs$ flexion/extension through a single actuator;
\item adjusting the grasping tasks for various object with different sizes and shapes based on the contact forces simply by embracing the underactuation concept;
\item adjusting the operation for different hand sizes by completing the linkage based mechanism kinematics around user's fingers;
\item placing the finger components and actuators on the dorsal side of the hand and
\item constraining only perpendicular forces to be transmitted to finger phalanges.
\end{itemize}

The underactuation concept simplifies the mechanical design to control multiple finger joints using a single actuator and aligns the overall operation automatically to different sizes and shapes of the grasping object. Placing the mechanical finger component and its actuator on the dorsal side of the hand increased the portability of the device. The use of miniaturized linear actuators made possible to extend the finger components for the implementation of a complete hand exoskeleton with independent finger control. The chosen linear actuator were selected with internal gearbox to amplify output forces, which are sufficient for power grasping tasks, with minimum magnetic interference with the given placement. However, the overall backdriveability of the finger components were lost due to high backdrive force caused by such a mechanical gearbox. Embracing linkage based mechanism allowed the actuator forces to be distributed to finger phalanges in an efficient and practical manner, improving the wearability of the device considering the disabilities of stroke patients. As a novelty, the proposed hand exoskeleton was designed with passive linear sliders for connecting the mechanism and user's finger phalanges. Doing so, the tangential forces are turned into motion and perpendicular forces are transmitted to user in a more realistic manner.

Using the aforementioned design criteria, the first sketch of the proposed finger component mechanism was drawn. Further analyses were required to analyze the mechanism behavior, the range of motion covered by the exoskeleton and the force transmission performance during operation. The performed analyses can be listed as:

\begin{itemize}
\item numerical inverse kinematics analysis to calculate the required actuator displacement for given finger pose;
\item numerical forward kinematics analysis to estimate the finger pose during operation using the actuator displacement and an additional sensory measurements attached on one of the passive revolute joints;
\item a numerical forward kinematics analysis to be used for calibration, where the lengths of proximal finger phalanges are estimated using the actuator displacement, additional potentiometer measurements and a passive slider displacement known thanks to a pre-defined finger pose;
\item analytical forward kinematics analysis as an alternative to the numerical forward kinematics due to the calculational burden of numerical approach and
\item differential kinematics analysis based on numerical kinematics approach to form a square matrix relating the measurement velocities to finger joint velocities.
\end{itemize}

Inverse kinematics and differential kinematics were analyzed in order to perform a link length optimization to find a set of link lengths that satisfies the physical limits and natural range of motion for finger joints and reaches the highest force transmission along finger phalanges for unitary actuator force. The physical constraints were defined as the desired actuator displacement, the displacements along the passive sliders at the connection with user's finger phalanges and the ratio between forces acting on finger phalanges. All these constraints were ensured to be satisfied for natural range of motion of finger joints. For a set of link lengths that satisfied all these physical constraints, the torque values along finger joints for unitary actuator force were calculated. The set of link lengths with highest force transmission to finger joints were chosen as the optimized set. Once the link lengths were optimized, the hand exoskeleton was manufactured using 3D printer thanks to its light weight and low cost. A control board was designed to control the exoskeleton simply by reading sensory measurements, computing kinematics analysis, running the controller and driving the motor drivers.

Integrating the mechanical device to electronic board, the feasibility of the proposed hand exoskeleton is shown with various control algorithms to define various physical rehabilitation scenarios. In particular:

\begin{itemize}
\item simple position control was implemented with improved performance to provide strict finger opening/closing for users during repetitive rehabilitation exercises;
\item simple position control with improved performance was extended to assist real grasping tasks for objects with different sizes and shapes as the interaction forces are measured between the object and the user's finger phalanges;
\item simple position control was performed with active trajectory adjustment using muscular activity measured from the healthy arm through EMG sensors for teleoperation or active rehabilitation exercises and
\item simple force control algorithm was implemented based on force measurements through 1-DoF strain gauges to achieve backdriveability over control or haptic rendering tasks.
\end{itemize}

Achieving active backdriveability allowed the proposed hand exoskeleton to be used for haptic applications to provide force feedback to users regarding their interactions in the virtual environment. The lack of controllability of the proposed hand exoskeleton and the lack of physical interaction forces acting on finger phalanges created complications for the device to be controlled virtually. Even if the lack of controllability cannot be overcome, such complications were proposed to be estimated simply by:

\begin{itemize}
\item degrading the device's mobility to $1-DoF$ by constraining one finger joint at a time to simplify the grasping task;
\item optimizing desired torque values, which are calculated based on stiffness rendering algorithm, to satisfy the passivity of additional sensory measurement in the force space, which is the underactuation constraint and
\item optimizing the desired finger pose to estimate the reachable pose for finger joints by modelling finger joints as spring and to satisfy the passivity of additional sensory measurement in the force space.
\end{itemize}

In particular, two different rendering strategies were proposed to implement underactuated devices for haptic tasks: (i) based on the force space and (ii) based on the configurational space. The feasibility of both methods were shown through real time experiments with the proposed hand exoskeleton. Firstly, the behavior of the rendering strategies were presented based on finger pose and corresponding actuator forces. Then, the comparison between actual transmitted forces for finger joints and desired finger joints were performed using additional FSR sensors attached between the device and user's finger. These two methods were compared to each other by computing the sensed impedance acting on finger joints during operations using the same task. Even though the feasibility of both strategies were shown in this work, an overall comparison to emphasize their advantages and disadvantages was left as a future work.

Once the optimization based rendering strategies were shown effective for user's finger with $2~DoFs$ stiff finger model, they were extended and generalized for proxy-based haptic rendering using Unreal simulation engine. The generalization of the optimization problem led us to run haptic rendering tasks while grasping soft objects. In these tasks, the user's hand was not modeled as simple $2~DoFs$ chain, but all virtual points consisting user's avatar was considered as an independent $DoF$. The experiment results showed the positive effect of the optimization method compared to the standard one.

The experiments and the analyses that were presented in this thesis show sufficient proof for the feasibility of the proposed hand exoskeleton to be used for physical rehabilitation exercises with post-stroke patients. In particular, obtaining active and passive tasks with improved control techniques in terms of perception is promising for the motor learning of patients with disabilities during grasping tasks. Despite of all that was done so far, there are still some works to be done in the future. These future works can be listed as:

\begin{itemize}
\item improving the ergonomic and comfort of users, especially while wearing the device as preparation period,
\item integrating the hand exoskeleton as a master part for teleoperation tasks to control a slave robotic hand,
\item integrating the hand exoskeleton with a wrist exoskeleton and an arm exoskeleton, which were designed in the same environment, to run some clinical tests with stroke patients,
\item performing a detailed user study to study the perceptional feedback of the haptic rendering simulation integrated with the hand exoskeleton, and
\item comparing both proposed stiffness rendering strategies to each other in a more detailed manner.
\end{itemize}

\clearpage
\pagebreak

\newpage

%----------------------------------------------------------------------------------------
%	THESIS CONTENT - APPENDICES
%----------------------------------------------------------------------------------------

%\addtocontents{toc}{\vspace{2em}} % Add a gap in the Contents, for aesthetics

%\appendix % Cue to tell LaTeX that the following 'chapters' are Appendices

% Include the appendices of the thesis as separate files from the Appendices folder
% Uncomment the lines as you write the Appendices

%\input{Appendices/AppendixA}
%\input{Appendices/AppendixB}
%\input{Appendices/AppendixC}

%\addtocontents{toc}{\vspace{2em}} % Add a gap in the Contents, for aesthetics

%\backmatter

%----------------------------------------------------------------------------------------
%	BIBLIOGRAPHY
%----------------------------------------------------------------------------------------

%\label{Bibliography}

%\lhead{\emph{Bibliography}} % Change the page header to say "Bibliography"

\bibliographystyle{unsrt} % Use the "unsrtnat" BibTeX style for formatting the Bibliography

\bibliography{references}

\end{document}